\documentclass[OPT,biber]{nowfnt} 

\usepackage[utf8]{inputenc}

\usepackage{commands}
\usepackage{hyperref}
\usepackage{url}

\newcommand{\ie}{\emph{i.e.}}

\usepackage{titlesec}

\titleformat{\paragraph}
{\normalfont\normalsize\bfseries}{\theparagraph}{1em}{}
\titlespacing*{\paragraph}
{0pt}{3.25ex plus 1ex minus .2ex}{1.5ex plus .2ex}

\makeatletter
\let\oldparagraph\paragraph
\renewcommand{\paragraph}{%
  \@ifstar{\oldparagraph*}{\oldparagraph}}
\makeatother

\title{Causal Machine Learning}

\subtitle{A Survey and Open Problems}

\maintitleauthorlist{
Jean Kaddour*, Aengus Lynch* \\
AI Centre, Department of Computer Science, UCL \\
jean.kaddour.20@ucl.ac.uk
\and
Qi Liu \\
Department of Computer Science, University of Hong Kong \\
liuqi@cs.hku.hk
\and 
Matt J. Kusner \\
AI Centre, Department of Computer Science, UCL \\
m.kusner@ucl.ac.uk
\and 
Ricardo Silva \\
AI Centre, Department of Statistical Science, UCL \\
ricardo.silva@ucl.ac.uk
}

\issuesetup
{%
 copyrightowner={A.~Gupta and M.~Casey},
 volume        = xx,
 issue         = xx,
 pubyear       = 2024,
 isbn          = xxx-x-xxxxx-xxx-x,
 eisbn         = xxx-x-xxxxx-xxx-x,
 doi           = 10.1561/XXXXXXXXX,
 firstpage     = 1, 
 lastpage      = 18
 }

\usepackage{mwe}

\author[1,*]{Kaddour,Jean}
\author[1,*]{Lynch,Aengus}
\author[2]{Liu,Qi}
\author[1]{Kusner,Matt J.}
\author[3]{Silva,Ricardo}

\affil[1]{AI Centre, Department of Computer Science, UCL}
\affil[2]{Department of Computer Science, University of Hong Kong}
\affil[3]{AI Centre, Department of Statistical Science, UCL}
\affil[*]{equal contribution}

\articledatabox{\nowfntstandardcitation}

\makeatletter
\def\nowfntarxiv@makechapterhead#1{\vbox to 11pc{%
  \vspace*{12\p@}%
  {\parindent \z@ \raggedright \normalfont
    \ifnum \c@secnumdepth >\m@ne
      \if@mainmatter
        {\nowfnt@font@chapternumber\thechapter\par}\addvspace{10pt}\nobreak
        \noindent\hrule\@width1\columnwidth\@height1\p@
        \vskip 14\p@
      \fi
    \fi
    \interlinepenalty\@M
    {\nowfnt@font@chaptertext#1\strut\par}\addvspace{10pt}\nobreak
    \noindent\hrule\@width1\columnwidth\@height1\p@
    \vfill
  }}}
\def\nowfntarxiv@makeschapterhead#1{\vbox to 8pc{%
  \vspace*{24\p@}%
  {\parindent \z@ \raggedright
    \interlinepenalty\@M
    {\nowfnt@font@chaptertext#1\strut\par}\addvspace{8pt}\nobreak
    \noindent\hrule\@width1\columnwidth\@height1\p@
  }\vfill}}
\AtBeginDocument{%
  \renewcommand\nowfnt@bookauthortitlepage{%
    \newpage
    \thispagestyle{\nowfnt@titlepagestyle}%
    \begingroup
    \raggedleft
    {\renewcommand{\baselinestretch}{1.05}%
     \nowfnt@font@title
     \nowfnt@articletitle
     \ifdefempty{\now@subtitle}{}{\par\bigskip\nowfnt@font@subtitle\now@subtitle}%
     \par}%
    \vspace{2.5ex}%
    \noindent\rule{\textwidth}{2pt}%
    \par\vspace{1ex}%
    \ifbool{authortwocolumn}
      {\nowfnt@printauthorsbooktitletwocolumn}
      {\nowfnt@printauthorsbooktitle}%
    \par\vfill
    \endgroup
  }%
  \fancypagestyle{abstractpage}{\fancyhf{}}%
  \raggedbottom
  \tcbset{before skip=4pt, after skip=4pt}%
  \let\@makechapterhead\nowfntarxiv@makechapterhead
  \let\@makeschapterhead\nowfntarxiv@makeschapterhead
}
\makeatother
\begin{document}

\makeabstracttitle

\begin{abstract}
\emph{Causal Machine Learning} (\causalml) is an umbrella term for machine learning methods that formalize the data-generation process as a causal model. This perspective enables one to reason about the effects of changes to this process (interventions) and what would have happened in hindsight (counterfactuals). We categorize work in \causalml into five groups according to the problems they address: (1) causal supervised learning, (2) causal generative modeling, (3) causal explanations, (4) causal fairness, and (5) causal reinforcement learning. We systematically compare approaches in each category and point out open problems. Further, we review field-specific applications in computer vision, natural language processing, and graph representation learning. Finally, we provide an overview of causal benchmarks and a discussion of the state of this nascent field, including recommendations for future work. 
\end{abstract}

\chapter{Introduction}
Today, machine learning (ML) techniques excel at finding associations in independent and identically distributed (i.i.d.) data. A few fundamental principles, including empirical risk minimization, backpropagation, and inductive biases in architecture design, have led to enormous advances for solving problems in fields like computer vision, natural language processing, graph representation learning, and reinforcement learning.

However, new challenges have arisen when deploying these models to real-world settings. These challenges include: (1) large reductions in generalization performance when the data distribution shifts \cite{underspecification}, (2) a lack of fine-grained control of samples from generative models \cite{steerability}, (3) biased predictions reinforcing unfair discrimination of certain sub-populations \cite{mehrabi2021survey, bender2021dangers}, (4) overly abstract and problem-independent notions of interpretability \cite{lipton2018mythos}, and (5) unstable translation of reinforcement learning methods to real-world problems \cite{dulac2019challenges}.

Multiple works have argued that these issues are partly due to the lack of  causal formalisms in modern ML systems \cite{EOCI, goyal2020inductive, pearl2019seven,causal_world,scholkopf2021towards}. Subsequently, there has been a surge of interest in the research community on \emph{causal machine learning} (\causalml), which are machine learning methods that utilize causal knowledge about the to-be-modeled system.\footnote{In parallel, there is a growing trend of using modern machine learning techniques to estimate causal quantities, e.g., treatment effects, which is \emph{not} the focus of this survey, see \Cref{sec:ml_for_causality} for more details on this line of work.} This survey covers how causality can address open ML problems. 

In a nutshell, causal inference provides a language for formalizing structural knowledge about the data-generating process (DGP) via \emph{causal models} \cite{EOCI}. A variety of causal formalisms exist \cite{spirtes2000causation, pearl2009causality, hernan2010causal, dawid2021, bareinboim2022pearl}. In particular, we will focus on \emph{Structural Causal Models} (SCMs) \cite{pearl2009causality} as they encompass the most common tasks found in the \causalml literature within a common framework. With SCMs, one can estimate what will happen to data after changes (called \emph{interventions}) are made to its generating process. Going one step further, they also allow us to model the consequences of changes in hindsight while taking into account what happened (called \emph{counterfactuals}). We introduce these concepts in more detail in \Cref{sec:prelim}, assuming no prior knowledge of causality.

Despite the extensive work on designing various types of \causalml algorithms, a clear categorization of its problems and methodology is lacking. We believe this is partly explained by the fact that \causalml usually involves assumptions about the data unfamiliar to large parts of ML. These assumptions are often tricky to relate across different problem setups, and are in many cases untestable with the data at hand (or even at all testable, regardless of any data collection procedure). This makes it difficult to measure progress and applicability, and motivates this survey.

\textbf{Contributions}
\begin{itemize}[leftmargin=*]
    \item We give a minimal introduction to key concepts in causality that is completely self-contained (\textbf{\Cref{sec:prelim}}). We do not assume any prior knowledge of causality. Throughout, we give examples of how these concepts can be applied to help further ground intuition.
    \item We taxonomize existing \causalml work into \textbf{causal supervised learning} (\Cref{chapter:cil}), \textbf{causal generative modeling} (\Cref{chapter:cgm}), \textbf{causal explanations} (\Cref{chapter:explanations}), \textbf{causal fairness} (\Cref{chapter:fairness}), \textbf{causal reinforcement learning} (\Cref{chapter:rl}). For each problem class, we compare existing methods and address avenues for future work. 
    \item We review data-modality-specific applications in \textbf{computer vision}, \textbf{natural language processing}, and \textbf{graph representation learning} (\Cref{sec:applications}), and \textbf{causal benchmarks} (\Cref{sec:benchmarks}).
    \item We discuss \emph{the good, the bad, and the ugly} (\Cref{sec:gbu}): what benefits \causalml provides compared to non-causal ML methods (the \emph{good}), what challenges the field faces at the time of this writing (the \emph{bad}), and what inevitable price one has to pay for using \causalml techniques (the \emph{ugly}).
\end{itemize}

\newcommand{\causal}{\textcolor{blue}{causal}\xspace}
\newcommand{\spurious}{\textcolor{red}{spurious}\xspace}

\chapter{Causality: A Minimal Introduction} \label{sec:prelim}
This section introduces the reader to the concepts of causality used in \causalml research. We focus on explaining the intuitions behind basic concepts and leave out many formal statements, e.g., proofs. We direct readers interested in such further details to \cite{pearl2009causality,EOCI,brady,statistical_to_causal_learning}.

\notation
\begin{nota}
\gG & \text{ Graph } \\
\de(X_i), \de_i & \text{ Descendants of }X_i \\
\an(X_i), \an_i & \text{ Ancestors of }X_i \\
\pa(X_i), \pa_i & \text{ Causal parents of }X_i \\
\doo(\cdot) & \text{ do-operator } \\
\scm & \text{ Structural Causal Model } \\
\end{nota}

\section{Bayesian Networks} \label{prelim:bayesian_networks}
To reason about the causal effects of some random variables on others, we need to formalize causal relations. The canonical representation of causal relations is a \emph{causal directed acyclic graph} (causal DAG), also called a \emph{causal diagram}. It can encode a priori assumptions about the causal structure of interest (e.g., from expert knowledge).

Before we define causal DAGs, we introduce some terminology to describe a DAG and \emph{Bayesian Networks} (BNs): a probabilistic graphical model representing probabilistic relationships between random variables. From there, we motivate modeling causal relationships, as they complement BNs with the ability to reason about interventions and counterfactuals. 

\subsection{Graphs} 

A \emph{graph} $\gG$ is a collection of \emph{nodes} and \emph{edges} that connect (some of) the nodes. In a \emph{directed} graph, the edges are directed: pointing from one node to another. Visually, arrows indicate this direction; notation-wise, we call nodes connected by one edge \emph{adjacent}. 

A \emph{path} in a graph is any sequence of adjacent nodes, regardless of the direction of the edges that join them. For example, $A \leftarrow B \rightarrow C$ is a path, but not a \emph{directed path}. The latter is a path consisting of directed edges all directed in the same direction, e.g., $A \rightarrow B \rightarrow C$.

A \emph{directed cycle} is a directed path that starts from a node $A$ and ends in $A$. 
A \emph{directed acyclic graph} (DAG) is a directed graph with no directed cycles. In a DAG, edges point from a \emph{parent} node into a \emph{child} node. We denote the parents of a node $X$ with $\pa(X)$; and $X$ an \emph{ancestor} of $Y$ (denoted by $X \in \an(Y)$), and $Y$ a \emph{descendant} of $X$ (denoted by $Y \in \de(X)$) if there is a (possibly empty) directed path that starts at node $X$ and ends at node $Y$. We denote a random node variable as $X$, assume that all distributions possess a mass or density function, and write $p(x)$ to represent such a function. 

We now cover a selection of topics on the uses of graphical concepts to describe factorization constraints in probabilistic models, resulting on which is known as \emph{graphical models}. DAG-based models are a special case, but our focus in this survey. See \cite{lauritzen1996} for a more thorough coverage of classic graphical concepts and their use in encoding conditional independence statements in probability distributions.

\subsection{Graphs as Joint Distribution Factorizations} \label{prelim:bn_example}
We are interested in modeling probability distributions over random variables in both probabilistic and causal modeling. One reason why graphs are helpful for that is that they allow one to conveniently express how a joint distribution over a set of random variables factorizes. Specifically, we will show how graphs allow one to encode (conditional) independence relationships. To explain this, we adopt an introductory example given by \citet{pearl2009causality}.

\begin{figure}
    \centering
    \begin{subfigure}[b]{.3\linewidth}
        \centering
        \includegraphics[width=\columnwidth]{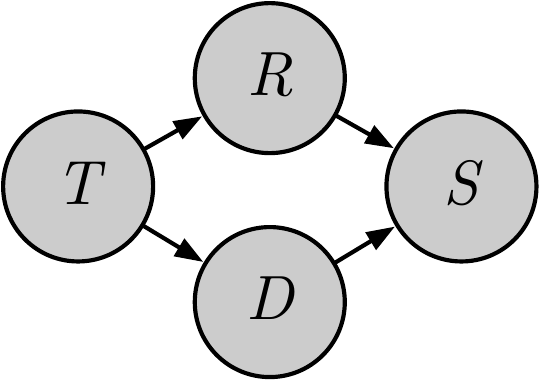}
        (a)
        \label{fig:bn}
    \end{subfigure}
    \begin{subfigure}[b]{.3\linewidth}
        \centering
        \includegraphics[width=\columnwidth]{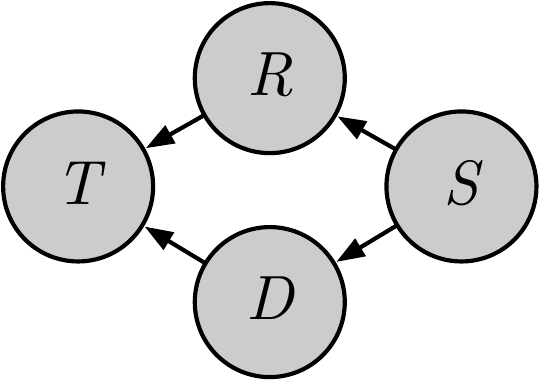}
        (b)
        \label{fig:flippedbn}
    \end{subfigure}
    \begin{subfigure}[b]{.3\linewidth}
        \centering
        \includegraphics[width=\columnwidth]{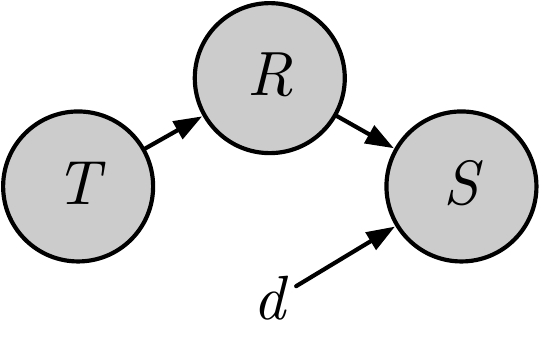}
        (c)
        \label{fig:bn_intervention}
    \end{subfigure}
    \caption{(a), (b) Possible DAGs describing the relationships between the time of the year $T$, the duration of enabled lawn sprinkler $D$, the amount of rain $R$, and the pavement slipperiness $S$. Despite the change of edge directionality, it is possible to verify that the corresponding factorizations impose exactly the same conditional independence constraints \cite{pearl2009causality}. (c) This graph imagines the scenario where there is no random variable $D$, but some fixed control signal $d$ that locally determines the conditional distribution $p_d(s~|~r)$ analogously to the conditional distribution $p(s~|~r, d)$ we use in the model in (a). We will later use the notation $p(s~|~r, \doo(d))$ to denote this change.}
\label{fig:bn_variations}
\end{figure}

Imagine that you are a farmer. You want to understand better the relationship between the time of the year $T$, duration of enabled lawn sprinkler $D$, amount of rain $R$, and pavement slipperiness $S$. To be able to conduct controlled simulations, you want to model the joint distribution $p(t, d, r, s)$.

The chain rule (or product rule) of probability always permits us to factorize a joint distribution over $N=4$ variables as a product of $N$ conditional distributions:
\begin{align}
     p(t, d, r, s) = p(t \mid d, r, s)  p(d \mid r, s)   p(r \mid s)  p(s). \label{eq:factorization}
\end{align}
More generally, it states that
\begin{align*}
    p(x_1, \dots, x_N) = p(x_1) \prod_i p(x_i \mid x_1, \dots, x_{i-1}).
\end{align*}
For discrete distributions without further constraints, modeling this naive factorization takes an exponential number of parameters and quickly becomes intractable as $N$ increases. \citet{brady} illustrates this as follows: Suppose that each $x_i$ is binary and let us denote parameters by $\params$. For modeling the conditional distribution $p(x_n \mid x_{n-1}, \dots, x_1)$, we only need to model $\theta_n = p(X_n = 1 \mid x_{n-1}, \dots, x_1)$ because \begin{align*}
    p(X_n = 0 \mid x_{n-1}, \dots, x_1) = 1 - p(X_n = 1 \mid x_{n-1}, \dots, x_1).
\end{align*} However, this implies that we need $2^{n-1}$ parameters to model the full joint distribution. For example, if $n=4$, to model $p(x_4 \mid x_3, x_2, x_1)$, we now need to store
\begin{align*}
    \theta_1 &= p(x_4 \mid X_3 = 0, X_2 = 0, X_1 = 0), \\ 
    \theta_2 &=p(x_4 \mid X_3 = 0, X_2 = 0, X_1 = 1), \;\;\; \cdots \\
    \theta_8 &=p(x_4 \mid X_3 = 1, X_2 = 1, X_1 = 1).
\end{align*}

Fortunately, some of your farmer colleagues have developed a DAG that describes how the variables are related, shown in Fig. \Cref{fig:bn}. This DAG is helpful because we can use it to describe conditional independence relationships. For example, since the time of the year is independent of all the other variables, we only need to model the marginal $p(t)$ instead of $p(t \mid d, r, s)$. Overall, we can simplify the joint distribution such that, in sum, the conditionals depend only on four instead of six random variables (compared to \Cref{eq:factorization}):
\begin{align}
    p(t, d, r, s) = p(t) p(d \mid t) p(r \mid t)  p(s \mid d,r). \label{ex:joint}
\end{align}

This simplification formally relies on the Markov condition. 
\begin{mydef}{Markov Condition \citep{pearl2009causality}}{markov}
Given a graph $\gG$ of nodes $\rmX$ with joint distribution $p(\vx)$, the Markov Condition states that the parents $\pa_i$ of every node $X_i$ make $X_i$ independent of its other non-descendants in $\rmX \setminus \de_i$, i.e., 
\begin{align*}
    p(x_i \mid \pa_i) = p\left(x_i \mid \rmX \setminus \de_i\right).
\end{align*}
This condition immediately implies the following factorization of the joint distribution
\begin{align*}
    p(\vx) = \prod_{i} p\left(x_i \mid \pa_i\right).
\end{align*}
\end{mydef}

This joint factorization is the product of all variables conditioned on their parents in the graph (if any). The core idea behind DAG models, commonly known in the AI literature as \emph{Bayesian Networks}\footnote{More traditionally, a \emph{DAG} is just a graph, a \emph{DAG model} is a probabilistic model following a DAG factorization, and a \emph{Bayesian network} is a discrete DAG model. We adopt a more informal take where these distinctions are not emphasized.}, is to decompose (potentially complex) joint functions $p(\vx)$ into several small conditional factors according to the assumed DAG relations. 


\begin{mlp}{Autoregressive Distribution Factorizations}{df}
\emph{Autoregressive models} are a popular class of DAG models applied to variables indexed by some relational structure, such as spatial or temporal coordinates, following a repeating pattern. For example, in generative modeling, certain methods \citep{van2016pixel,pixel_cnn} model the distribution over an image $\vx$ using a factorization from a certain pixel-by-pixel order, without imposing any independencies:
\begin{align*}
p(\vx)=\prod_{i=1}^{N} p\left(x_{i} \mid \vx_{<i}\right),
\end{align*} where $\vx_{<i}=\left[x_{1}, x_{2}, \ldots, x_{i-1}\right]$ denotes the vector of pixels with index less than $i$. 
The advantage of doing so is, e.g., to make sampling easier, or enable to learn conditional factors in parallel. Similarly, for sequence modeling tasks (e.g., language modeling with recurrent neural networks \citep{rnn}), one often assumes that the joint distribution over all tokens can be decomposed into conditionals that only depend on hidden states $\vh$ produced by a learnable function $f_{\params}$ with parameters $\params$, i.e.,
\begin{align*}
p(\vx) = \prod_{t=1}^{T} p\left(x_{t} \mid \vh_{t-1}\right), \quad \vh_{t} = f_{\params}(x_t, \vh_{t-1}).
\end{align*}
\end{mlp}

\section{Graphical Causal Models}\label{sec:cbn}

In this section, we will get into the details of what we mean by representing interventions and counterfactuals with graphical causal models.

\subsection{Interventions} \label{sec:cbn_intervention}
BNs utilize structural knowledge about the distribution of interest to represent it more efficiently. However, they remain oblivious to \emph{interventions} on the distribution's underlying process, which are essential to understand the effects of changes, as we will see in this subsection. 

As in the previous example, random variable $D$ quantifies the duration of the sprinkler being turned on. Imagine that you are interested in investigating the effect of fixing it to value $D = d$, given that, e.g., there might be a non-linear relationship between the sprinkler and the slipperiness that you seek to understand better. 

Using the Bayesian network toolbox, perhaps the most obvious strategy to pursue is first to infer the conditional distribution 
\begin{align}
    p(t, r, s \mid d) =  p(r \mid t)  p(s \mid r, d),  \label{eq:conditional_distribution}
\end{align} which we obtain by applying Bayes' rule to \Cref{ex:joint}.
Then, if we are only interested in $p(s \mid d)$, we can marginalize the other variables out using the sum rule. Alternatively, we can train an estimator $\hat p(s \mid d)$ for $p(s \mid d)$.

However, what does $p(s \mid d)$ mean? It is the distribution of $S$ given that we \textbf{observe} variable $D$ for a fixed value, e.g., $d=10$ minutes. In other words, we restrict our focus to those observations in which the sprinkler happened to be set to $d$ (in contrast to set to it actively, which we denote as intervention).

Let us think about the implications of examining this \textbf{observational} distribution: 
$p(s \mid d)$ accounts for the probability of observing $d$ depending on the season of the year $T$ via $p(d \mid t)$ and the amount of rain $p(r \mid t)$, as can be seen in \Cref{eq:conditional_distribution}. 

\begin{figure}
    \centering
    \includegraphics[width=0.9\columnwidth]{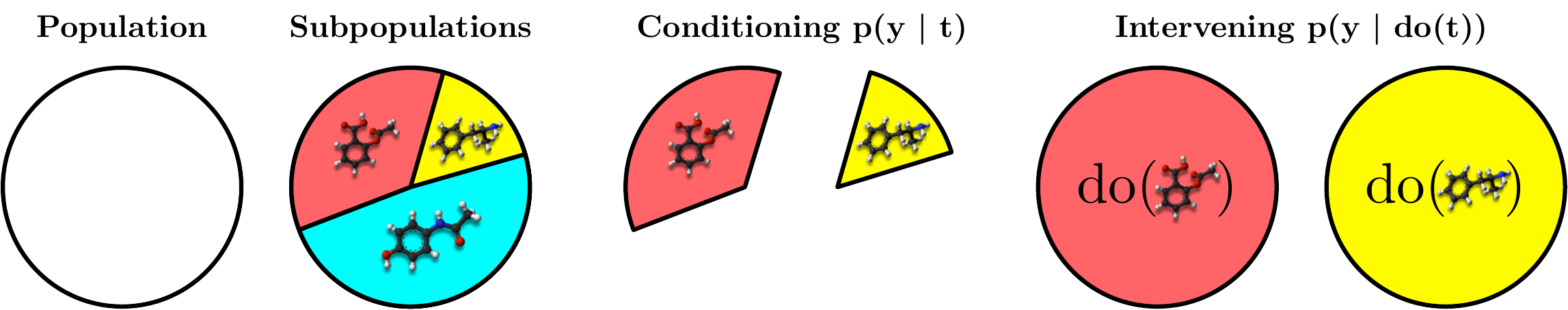}
    \caption{\textbf{Conditioning versus intervening}. Adapted from \cite{brady,SIN}.} 
    \label{fig:cond_vs_intv}
\end{figure}

Now, switching to the causal inference toolbox, our quantity of interest is typically an \emph{intervention}, which we denote by the $\doo$-operator. Following the above example, we are interested in estimating $p\left(s \mid \doo\left(d\right)\right)$. The difference in its meaning is that $p\left(s\mid \doo\left(d\right)\right)$ denotes the distribution of $S$ \textbf{intervened} upon the value of $D$, i.e., if we set it to $d$. This means that we are not considering a specific sub-population for which we observe $D=d$, but we reason about what happens to the (total) population after taking the \textbf{action} $\doo\left(d\right)$.

Returning to our example, \Cref{fig:bn_intervention} shows the new network representation after the action took place. We remove edge $T \to D$ from \Cref{fig:bn} because, after intervening on $D$ by setting it to a constant value\footnote{There also exists the idea of \emph{soft} interventions in which $D$ can still depend on other variables, but for the sake of exposition, we focus on the simpler case of a \emph{hard} intervention here.}, it not longer depends on any other variables. More than that, ``$d$'' here is not a realization of a random variable, but a fixed index denoting a control signal set externally to the joint distribution of the remaining variables.

Formally, there is a small yet significant difference compared to the conditional observational distribution in \Cref{eq:conditional_distribution}: 
\begin{align}
p\left(t, r, s \mid \doo\left(d\right)\right) = p(t)  \underbrace{\cancel{p(d \mid t)}}_{=1}  p(r \mid t)  p(s \mid d, r) \neq p(t,r,s \mid d). \label{eq:interventional_distribution}
\end{align}

We no longer care about the relationship between the year's season and sprinklers before the action because that relationship is no longer in effect while we perform the action. Once we physically turn the sprinkler on, a new mechanism in which the season has no say determines the state of the sprinkler.

\begin{thm}{Truncated Factorization \citep{pearl2009causality}}{truncated} We assume that $p$ satisfies the Markov assumptions in DAG $\gG$. Let $\gS \subset \mathcal X$ be a set of nodes fixed by intervention at level $s$. If entries in $\vx$ are consistent with the corresponding entries in $\mathcal S$ then
\begin{align*}
p\left(x_{1}, \ldots, x_{n} \mid \doo(s)\right) :=\prod_{i \notin \gS} p\left(x_{i} \mid \pa_{i}\right)
\end{align*}
Otherwise, $p\left(x_{1}, \ldots, x_{n} \mid \doo\left(s\right)\right):=0$.
\end{thm}

Hence, to learn a valid \emph{Causal} BN that enables the estimation of the above interventional distributions, we have to make stronger assumptions of how interventions are operationalized in terms of changes to particular factors of the factorization implied by the DAG. Such assumptions rest on causal (and not associational) knowledge about the underlying system, including the physical validity of what an intervention means. A recent detailed discussion of the latter is provided by \cite{jorgensen2025}.

Our takeaway is that Causal BNs differ from regular BNs, e.g. because the conditional independence assumptions in regular BNs do not necessarily imply causality as operationalized by what changes under interventions. Their implied (probabilistic) factorization is testable for any recursive set of independencies and any ordering of the variables, but it does not mean it will remain valid under the control of one or more variables.

To illustrate the difference between the two, recall our initial example shown in \Cref{fig:bn}. Here, we conveniently made independence assumptions that seem causal already. Alternatively, we could have come up with an independence structure shown in \Cref{fig:flippedbn}, where the edges are flipped
\begin{align*}
p(t, d, r, s) = p(t \mid d, r) p(d, r \mid s) p(s).
\end{align*}
Based on common sense, the corresponding DAG does not seem causal. According to commonly known laws of nature, we would not expect the pavement's slipperiness to cause the amount of rain. Yet, for the purpose of \emph{statistical} inference, it is perfectly adequate to factorize the joint distribution that way as we can verify that the implied conditional independencies have not changed \cite{pearl2009causality}.

Another perspective is that causal assumptions postulate how a model behaves under a particular type of \emph{distribution shift}, that which follows an intervention. This means a higher abstraction level for any modeling problem, above common statistical assumptions like the model's flexibility. In pure statistical inference that assumes future test cases will be exchangeable with the training cases (including most ML setups), we often ignore this level and do not impose strong assumptions on the causality between variables. Therefore, we do not yield the ability to predict outcomes under interventions. 

\subsubsection*{Examples of Interventions} \label{example:interventions}
The following examples provide intuition on how conditional and interventional distributions differ. 

\textbf{Espresso Machine \citep{ferenc}.} 
Imagine that $Y$ is the pressure in an espresso machine's boiler, and $X$ is the reading of the built-in barometer. Given a functioning barometer, $p(y \mid x)$ is a unimodal distribution centered around $X$, with randomness due to measurement noise. However, if we forcefully break the barometer and set it to $0$, it will not affect the pressure in the tank. Therefore, $p(y \mid \doo(x)) = p(y) \neq p(y \mid x)$. 

\textbf{Medical Treatment \citep{SIN}.} \label{example:medical_intervention}
Imagine a dataset where each observation $\left(\vx_{i}, \vt_{i}, y_{i}\right) \in \mathcal{D}$ represents a hospital patient's medical history record $\vx_i$, prescribed drug treatment in form of a molecular graph $\vt_i$, and health outcome $y_i$. \Cref{fig:cond_vs_intv} illustrates how in this setup the intervention $p\left(y \mid \doo\left(\vt\right)\right)$ refers to the scenario in which all patients described with medical history features $\vx$ take treatment $\vt$ versus the conditional distribution $p(y \mid \vt)$ restricting our focus to the subpopulation of $\rmX$ that received $\vt$. 

\subsection{Counterfactuals}
Recall our farmer example from \Cref{prelim:bn_example} and imagine that one day, you observe that the slipperiness $S$ was very high, and even worse, a colleague of yours slipped and broke their arm. 
You want to determine under what circumstances $S$ would have been reduced on that day. You come up with a hypothesis that you want to test: \quo{If we had turned the sprinkler off, the slipperiness would have been low.}

Your hypothesis - an \quo{if} statement in which the \quo{if} portion is unrealized - is a \emph{counterfactual} statement. It incorporates the factual data ($D$ was non-zero on that day) and an intervention (setting $D$ to $0$), in which parts of the environment remain unchanged ($T, R$). Thereby, it paves the way to comparing two outcomes under the same conditions, differing only in one aspect.

To formalize a counterfactual statement, we must go beyond just the do-operator. For example, suppose that on that day, we observed slipperiness $s$, time of year $t$, and rain amount $r$. Then, simply writing $p(s \mid \doo(d), s, t, r)$ leads to a clash between the hypothetical slipperiness and the actual slipperiness observed \citep{causal_primer}.

One way to clarify the distinction is to label the two outcomes of interests - the factual and counterfactual - with different subscripts. We denote the actual slipperiness by $s_0$ and the counterfactual slipperiness under the intervention $D=d^{\prime}$ by $S_{d}^{\prime}$, such that our estimand becomes $p(s_{d}^{\prime} \mid s_0, t, r, d)$.

This notation is convenient for expressing our quantity of interest but does not operationalize the estimation of it. We require \emph{Structural Causal Models} (SCMs) to do the latter. We will introduce these in \Cref{sec:scm} and conclude the current one with two more examples of counterfactuals.

\subsubsection*{Examples of Counterfactuals}
Intuitively, counterfactuals are hypothetical retrospective interventions given an observed outcome. They help to explain the data since we can analyze the changes resulting from manipulating each variable. We now describe the difference between interventional and counterfactual distributions.

\textbf{Medical Treatment.} Let us revisit the medical treatment example in \Cref{example:medical_intervention}, where we described $p\left(y \mid \vx,\doo\left(\vt\right)\right)$ as the scenario in which all patients described with medical history features $\vx$ take treatment $\vt$. Perhaps the most obvious counterfactual quantity we can think of in this scenario is $p\left(y_{\vt^{\prime}} \mid \vx,\vt,y \right)$. 

How do $p\left(y \mid \vx,\doo\left(\vt^\prime \right)\right)$ and $p\left(y_{\vt^{\prime}} \mid \vx,\vt,y \right)$ semantically differ? The latter can be interpreted as \emph{imagining} the former \say{after the fact} that $\vx, \vt, y$ occurred. So instead of asking \say{what happens if we give treatment $\vt$ to patient $\vx$?}, the latter is retrospective, asking \say{what would have happened if we had given treatment $\vt^{\prime}$ to patient $\vx$ instead of $\vt$?}.

\textbf{Image Editing.}
Let $\rmX \in \R^D$ be an image. A common assumption in ML is that $\rmX$ is embedded in a lower-dimensional manifold that has semantically meaningful factors of variation \citep{gardner2015deep}. For example, assume we observe an image $\vx$ and infer its (latent) primary object of interest $\vo$ and background features $\vb$. We can then sample images with edited backgrounds in a controlled manner by sampling from the counterfactual distribution $\tilde \vx \sim p(\vx_{\vb^{\prime}} \mid \vx, \vo, \vb)$.

How does $p(\vx \mid \vo, \doo(\vb^\prime))$ differ from $p(\vx_{\vb^{\prime}} \mid \vx, \vo, \vb)$? If we sample images from the former distribution, depending on its variance, we may get a very diverse set of images with object and background features $\vo, \vb$, respectively. If we sample from the latter, we expect sampled images that look identical except for their background.

\subsection{The Ladder of Causation}
\Cref{tab:pearl_ladder} shows the three-layer ladder of causation \citep{pearl2009causality}, summarizing the differences between associational (or \emph{observational}), interventional and counterfactual distributions. It is also called \emph{The Causal Hierarchy}, because questions at level $i \in \{1,2,3\}$ can only be answered if information from level $j \geq i$ is available. Counterfactuals subsume interventional and associational questions, so they sit at the top of the hierarchy. Models that can answer counterfactual questions can also answer questions about observations and interventions, as we will shortly see in \Cref{sec:scm}.

\begin{table}[t]
    \centering
    \begin{minipage}{\textwidth}
        \centering
        \resizebox{\linewidth}{!}{%
            \begin{tabular}{P{3.2cm}|P{2cm}|P{3.5cm}|P{3.5cm}}
                \toprule
                \bf Layer & \bf Activity & \bf Semantics & \bf Example \\ \midrule
                (1) Associational $p(y\mid x)$& Seeing \includegraphics[height=1.5em]{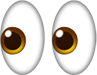}& How would seeing $x$ change my belief in $Y$? & What does a symptom tell us about the disease? \\ \midrule
                (2) Interventional $p(y \mid \doo(x), z)$ & Doing \includegraphics[height=1.5em]{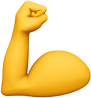} & What happens to $Y$ if I do $x$? & What if I take aspirin, will my headache be cured? \\ \midrule
                (3) Counterfactual $p(y_{x^{\prime}} \mid x, y)$ & Imagining \includegraphics[height=1.5em]{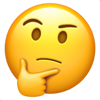} & Was it $x$ that caused $Y$? & Was it the aspirin that stopped my headache? \\ \bottomrule
            \end{tabular}
    }
    \caption{\textbf{The Ladder of Causation \citep{pearl2009causality}} (also called \emph{The Causal Hierarchy} \citep{bareinboim2022pearl}). Distributions on one layer virtually underdetermine information at higher layers, e.g., counterfactuals subsume interventional and associational ones.}
    \label{tab:pearl_ladder}
    \end{minipage}
\end{table}

\section{Structural Causal Models} \label{sec:scm}

In \Cref{sec:cbn}, we learned how causal BNs allow us to move from associational distributions in regular BNs to interventional ones. However, we were not able to construct counterfactual distributions with causal BNs. In this section, we learn another causation formalism that permits counterfactual analysis. 

This formalism is a Structural Causal Model (SCM), sometimes called \emph{Structural Equation Model} or \emph{Functional Causal Model} \citep{pearl2009causality}. 
In SCMs, we express causal relationships through deterministic, functional equations. This formalism reflects Laplace's conception of natural laws being deterministic and randomness being a purely epistemic notion \citep{pearl2009causality}. Hence, we introduce stochasticity in SCMs based on the assumption that certain variables in the equations remain unobserved.

\begin{mydef}{Structural Causal Model}{scm} 
An SCM $\scm := \left(\rmS, p\left(\vec \epsilon\right)\right)$ consists of structural assignments $\mS = \{f_i\}_{i=1}^N$,
\begin{align}
    x_i := f_i(\epsilon_i; \pa_i), \label{eq:scm}
\end{align} where $\pa_i$ is the set of parents of $x_i$ (its direct \textit{causes}), and a joint distribution $p(\vec \epsilon) = \prod_{i=1}^N p(\epsilon_i)$ over exogenous noise variables 
(i.e. unaccounted sources of variation). 

For every SCM, we yield a DAG $\gG$ by adding one vertex for each $X_{i}$ and directed edges from each parent in $\pa_{i}$ (the causes) to child $X_{i}$ (the effect).
\end{mydef}

\Cref{eq:scm} means that, in any SCM, we have that each variable $X_i$ is caused by parent variables, and unobserved exogenous ``noise'' variables $\epsilon_i$. To denote that the noise variables may play a major role causally, sometimes we use the more standard random variable notation $U_i$ to represent them. However, exogenous variables are present in every SCM, and thus, we often omit them from causal graphs for brevity.

Since every SCM induces a (causal) graph, it also implies the previously introduced Markov condition in \Cref{def:markov}. We call this the \emph{Causal Markov Condition}, as the DAG contains the causal relationships. 

\begin{thm}{Causal Markov Condition \citep{pearl2009causality}}{cmc}
Every SCM $\scm$ entails a joint density $p_{\scm}(\vx)$ such that each variable $X_i$ is independent of all its non-descendants given its parents $\pa_i$ in $\gG$. 
\end{thm}

\subsection{Interventions} \label{concept:intervention} 
We already learned the idea of estimating an intervention using a causal BN in \Cref{sec:cbn_intervention}. An SCM allows us to predict the effects of interventions, too: we substitute one or multiple of its structural assignments with the intervention's value. 

The SCM view further highlights the difference between interventions $p(y \mid \doo\left(\vx^\prime \right))$ and counterfactuals $p(y_{\vx^\prime} \mid \doo\left(\vx^\prime\right), \vx)$: interventions operate at the population level in the sense that, if we construct the interventional distribution through an SCM, its exogenous noise terms still consist of the prior distribution $p(\vec \epsilon)$, and not from the posterior $p(\vec \epsilon \mid \vx)$, which incorporates our knowledge of what already happened. 

The solution to a counterfactual query is an individual-level answer to a \say{what would have happened if} question where all the exogenous variable sources had been controlled, leaving information to pass through a (modified) set of structural equations. Information about the exogenous variables can be obtained from observable variables already realized.

\subsection{Counterfactual Inference}\label{concept:counterfactuals} 
To compute counterfactuals, we can manipulate an existing SCM and turn it into a counterfactual one. To do so, we estimate the exogenous noise terms $p(\vec \epsilon \mid \vx)$ given the observed datum $\vx$.

\begin{mydef}{Counterfactual SCM \citep{EOCI}}{cf}
Consider an SCM $\scm = \left(\rmS, p\left(\vec \epsilon\right)\right)$ over nodes $\rmX$. Given observations $\vx$, we define a counterfactual SCM by replacing the prior distribution of noise variables $p(\vec \epsilon)$ with the posterior $p(\vec \epsilon \mid \vx)$: 
\begin{align*}
\scm_{\vx}:=\left(\rmS, p(\boldsymbol{\epsilon} \mid \vx) \right).
\end{align*}
\end{mydef}

Given a counterfactual SCM $\scm_{\vx}$, we yield counterfactual distributions by intervening on its structural assignments $\rmS$. To illustrate this, let $\widetilde \rmS$ denote the modified structural assignments with intervention $\doo(x_i = \tilde x_i)$. Then, we denote the modified, counterfactual SCM as $\widetilde{\scm} := \scm_{\vx, \doo(\tilde x_i)}$. Finally, $\widetilde{\scm}$ yields the counterfactual distribution $p_{\widetilde{\scm}}(\vx)$.

We summarize this \emph{counterfactual inference} procedure in the following.  
 
\begin{mydef}{Counterfactual Inference \citep{pearl2009causality}}{cfi}
We infer counterfactual queries through a three-step procedure:
\begin{enumerate}[leftmargin=*]
    \item \textbf{Abduction:} Infer $p(\vec \epsilon \mid \vx)$, \ie, the exogenous ``state of the world'' $\vec \epsilon$, that is compatible with observations $\vx$.
    \item \textbf{Action:} Replace all equations  corresponding to the intervention $\doo(\tilde{x}_i)$, resulting in a modified SCM $\widetilde{\scm} := \left({\mS}_{\doo(\tilde{x}_i)}, p\left(\vec \epsilon \mid \vx\right)\right)$.
    \item \textbf{Prediction:} Use the modified model to compute $p_{\widetilde{\scm}}(\vx)$.
\end{enumerate} 
\end{mydef}

\subsection{Independent Mechanisms}
A nice property of SCMs is the \emph{Principle of Independent Mechanisms} (also called \emph{autonomy} or \emph{modularity}). It is analogous to the truncated factorization property previously discussed in the context of Causal Bayesian Networks (\Cref{thm:truncated}). Its basic premise is that interventions are local, and intervening on a variable $X_i$ only changes the causal mechanism for $X_i$, leaving the other mechanisms invariant. This principle allows us to encode many different interventional distributions in a single graph \citep{pearl2009causality}. 

\begin{mydef}{Principle of Independent Mechanisms \citep{pearl2009causality,brady}}{pim}
An SCM consists of autonomous modules $p(x_i \mid \pa_i)$,
\begin{align*}
    p(\vx)=p\left(x_{1}, \ldots, x_{D}\right)=\prod_{i=1}^{D} p\left(x_{i} \mid \pa_i\right).
\end{align*}
This principle implies that if we intervene on a subset of nodes $S \subseteq \{1,\dots,D\}$, then for all $i$, we have that
\begin{enumerate}[leftmargin=*]
\item If $i \not \in S$, then $p(x_i \mid \pa_i)$ remains unchanged.
\item If $i \in S$, then $p(x_i \mid \pa_i) = 1$ if $x_i$ is the value that $X_i$ was set to by the intervention; otherwise $p(x_i \mid \pa_i) = 0.$
\end{enumerate}
\end{mydef}

\section{Causal Representation Learning} \label{sec:crl}

The goal of representation learning is to retrieve low-dimensional representations $\rmZ$ that summarize our high-dimensional data $\rmX$, where $\text{dim}(\rmZ) \ll \text{dim}(\rmX)$.
The learned representations then facilitate solving downstream tasks, as the features of interest (e.g., an object in the image) are typically not given explicitly in the granular input data (e.g., pixels). However, these representations often rely on spurious associations and yield entangled dimensions that are hard to interpret \citep{papyan2020prevalence,geirhos2020shortcut,wang2021desiderata}. 

In contrast, \emph{Causal representation learning} (CRL) assumes that an SCM over high-level causal variables generates the data $\rmX$. Representations $\rmZ$ correspond to instances of these typically latent causal variables. 

With access to the SCM, we can estimate the data distribution after interventions on these variables or infer counterfactuals for specific data points. Unfortunately, learning the entire SCM is difficult without extensive supervision or domain knowledge. This task consists of three components, summarized in \Cref{def:causal_rep_learning}.

\begin{mydef}{Causal Representation Learning \citep{scholkopf2021towards}}{causal_rep_learning}
In \emph{causal representation learning}, we aim to learn a set of causal variables $\rmZ$ that generate our data $\rmX$, s.t. we have access to the following: 

\begin{enumerate}[leftmargin=*]
    \item \emph{Causal Feature Learning}: an injective mapping $g: \mathcal{Z} \rightarrow \mathcal{X}$ s.t. $\rmX = g(\rmZ)$ 
    
    \item \emph{Causal Graph Discovery}: a causal graph $\mathcal{G}_{\rmZ}$ among the causal variables $\rmZ$
    
    \item \emph{Causal Mechanism Learning}: the generating mechanisms $p_{\mathcal{G}_{\rmZ}}(z_i \mid \pa(z_i))$ for $i = 1,..,\text{dim}(\rmZ)$
\end{enumerate}

where $\pa(Z_i) \!\subset\! \{ Z_{j} \}_{j \neq i} \cup \epsilon_i$ and $\epsilon_i$ is the exogenous parent of $Z_i$.
\end{mydef}

\section{Spurious Relationships due to Confounding}
\label{sec:confounding}
Recall the medical treatment example from \Cref{example:medical_intervention}, where we have a dataset where each observation $\left(\vx_{i}, \vt_{i}, y_{i}\right) \in \mathcal{D}$ represents a hospital patient's medical history record $\vx_i$, prescribed drug treatment $\vt_i$, and health outcome $y_i$. In real-world scenarios, the patient's pre-treatment health conditions $\vx_{i}$ influence both the doctor's treatment prescription and outcome, thereby $\rmX$ \emph{confounds} the effect of the treatment $\rmT$ on the outcome $Y$ (and we call $\rmX$ a \emph{confounder} or \emph{confounding variable}). Moreover, we say that $\rmT$ and $Y$ are confounded (by $\rmX)$, or \emph{spuriously associated}.

\begin{figure}
    \centering  \hspace*{\fill}
    \begin{subfigure}[b]{\gw}
        \centering 
        \includegraphics[width=\textwidth]{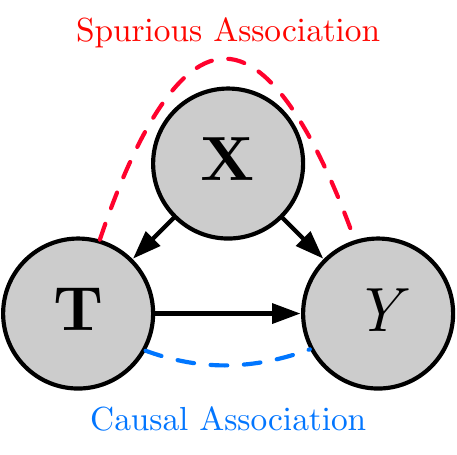}
        \caption{\small $p(y \mid \vx, \vt)$}
        \label{fig:spurious_cond}
    \end{subfigure} \hfill
    \begin{subfigure}[b]{\gw}
        \centering
        \includegraphics[width=\textwidth]{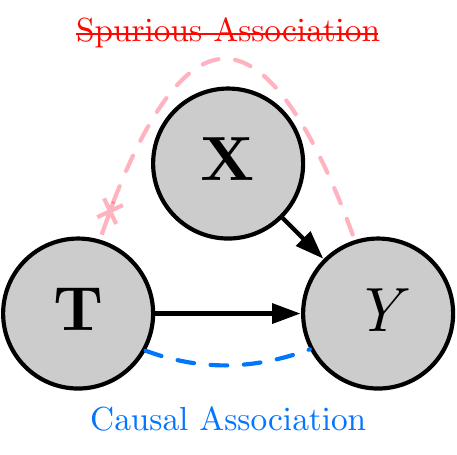}
        \caption{\small $p\left(y \mid \vx, \doo\left(\vt\right)\right)$}
        \label{fig:spurious_intv}
    \end{subfigure}\hspace*{\fill}
    \caption{\textbf{Spurious Relationship} between $\rmT$ and $Y$ due to $\rmX$ being an observed confounder. We adapt the Figures from \cite{brady}.}
    \label{fig:spurious}
\end{figure}

\Cref{fig:spurious} visualizes how associations flow in the observational $p(y \mid \vx, \vt)$ and interventional distribution $p(y \mid \vx, \doo(\vt))$. In \Cref{fig:spurious_cond}, we see that $p(y \mid \vx, \vt)$ entails both \causal and \spurious associations from $\rmT$ to Y, while $p(y \mid \vx, \doo(\vt))$ isolates the \causal association from $\rmT$ to $Y$, as shown in \Cref{fig:spurious_intv}. The \causal effect flows along directed paths, while \spurious associations flow along all \emph{unblocked} paths. To determine whether a path is unblocked, certain criteria need to be checked, which we do not cover here, but one can find in \cite{pearl2009causality}.

Here, we want to highlight the \spurious association between $\rmT$ and $Y$: Imagine a doctor whose policy is to give expensive treatments to very ill patients with low recovery chances and cheap treatments to very healthy patients with high recovery chances. $Y$ is a scalar that denotes the post-treatment health outcome; the higher, the better. Assuming that cheap and expensive treatments are equally effective, cheap and costly treatments are nonetheless positively and negatively correlated with health outcomes, respectively. This correlation is \spurious because it is due to the doctor's policy, based on the patient's pre-treatment health conditions $\rmX$, and not the treatment's actual \causal effect on the outcome. 

When do spurious relationships become problematic? A simplified answer is whenever the confounder is unobserved (also called \emph{hidden} confounding). The reason is that without further knowledge about the data-generating process, a sophisticated ML model will likely rely on spurious associations in the training dataset, which may not occur anymore when the model is in production. This reliance is a feature, not a bug, though: it would be wasteful if the model would not utilize spurious associations if we do not enforce it to avoid them. \Cref{mlp:spurious} illustrates how hidden confounding may harm classification models in a computer vision context. We explore remedies for this issue in \Cref{chapter:cil}.

\begin{figure}
    \centering
    \begin{subfigure}[b]{0.7\textwidth}\centering
    \includegraphics[width=\columnwidth]{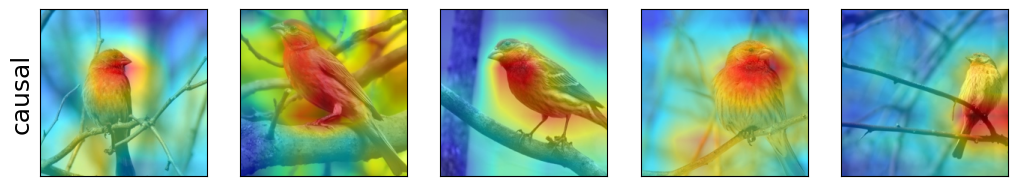}
    \includegraphics[width=\columnwidth]{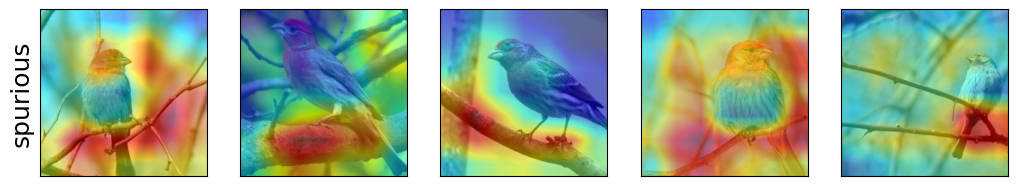}
    \end{subfigure}\begin{subfigure}[b]{0.3\textwidth}\centering
    \includegraphics[width=0.8\columnwidth]{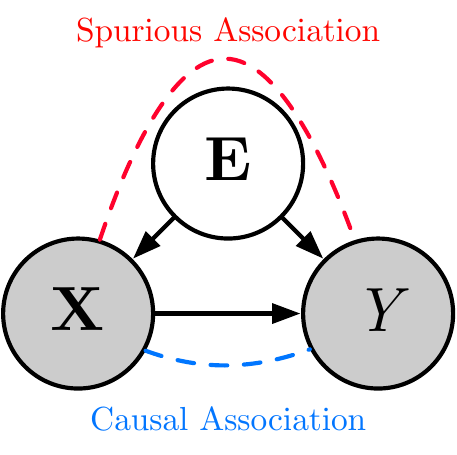}
    \end{subfigure}
    \caption{\textbf{Spurious relationships due to hidden confounding in ImageNet \citep{singla2022salient}.} The hidden confounder animal environment $\rmE$ caused images of birds to include trees and boughs. The heatmaps highlight causal and spurious associations between images $\rmX$ and bird labels $Y$.}
    \label{fig:spurious_imagenet}
\end{figure}

\begin{mlp}{Spurious Relationships in ImageNet  \citep{singla2022salient}}{spurious}In ML benchmark datasets, unobserved confounding may occur too. Consider the image classification scenario with image $\rmX$, a label $Y$, and an unobserved variable $\rmE$ that dictates the background of the classification object of interest, e.g., the environment of animals. It causes pictures of birds to include trees and boughs, as illustrated by \citet{singla2022salient} in \Cref{fig:spurious_imagenet}. The authors find that classifiers trained on the ImageNet dataset rely on confounded relationships to classify objects, i.e., the same object in a different background environment is more likely to be misclassified.
\end{mlp}

\section{Causal Estimand Identification}\label{sec:identifiability}
So far, we have discussed the semantic difference between observational and interventional distributions and the reasons for computing the latter. When is it feasible to estimate the latter? \emph{Identification} of causal estimands refers to the process of moving from the causal estimand (e.g., $p\left(y\mid \doo\left(\vx\right)\right)$) to an equivalent statistical estimand (e.g., $p\left(y\mid \vx\right)$), which we can then estimate from data \citep{brady}.

We call a causal estimand \emph{identifiable} if it is possible to compute it from a purely statistical quantity. If it is not identifiable, then regardless of how much data we have, we will not be able to isolate the causal association of interest in our data. 

In the absence of hidden confounders and as long as we know the causal graph, the causal estimand is identifiable. For example, suppose our estimand of interest is the \emph{average treatment effect} $p(y \mid \doo(\vt))$ in the scenario of \Cref{fig:spurious_cond}, i.e., the causal effect of treatment $\vt$ averaged over all possible patient features $\rmX$. Here, the \emph{backdoor criterion} is satisfied, which makes $\rmX$ a \emph{valid adjustment set} \citep{pearl2009causality}. The mathematical procedure outputting a statistical estimand is called \emph{backdoor adjustment}.

Generally speaking, given any causal DAG (potentially including unobserved confounders), graphical tests allow us to determine the identifiability of specific causal estimands \citep{complete_identification,JMLR:v18:16-319}. Besides the backdoor criterion, a \emph{frontdoor} criterion exists. We do not cover these criteria here; for this survey, it is sufficient to know about their existence, and that finding them can be automated \citep{backdoor_discovery}. We direct readers interested in learning more about them to excellent explanations in \cite{pearl2009causality, EOCI, brady,statistical_to_causal_learning,hernan2010causal}.

\section{Causal Influence} \label{sec:causal_influence}
Besides interventional and counterfactual queries, another common quantity of interest in causal inference is one variable's causal influence on another. For example, in \Cref{chapter:explanations}, given an input vector $\vx$ and a black-box ML model $f_{\params}(\cdot)$ with parameters $\params$, we will look at quantifying the causal influence an input feature $x_i$ has on the model prediction $\hat y = f_{\params}(\vx)$. Another example will occur in \Cref{rl:ca}, where we measure the social influences of agents in multi-agent systems, e.g., how the action one agent takes influences the subsequent action of another agent. 

\citet{causal_influence} postulate a set of natural, intuitive requirements that a measure of causal influence should satisfy. Then, they evaluate various information-theoretic measures on whether they meet these desiderata. Finally, they conclude that the KL-divergence is a suitable measure. 

\begin{mydef}{Causal Influence (KL-divergence) \citep{causal_influence}}{causal_influence}
Given an SCM $\scm$, the causal influence $\scm_{k \rightarrow l}$ from node $k$ to node $l$ is
\begin{align*}
\scm_{k \rightarrow l} &=\sum_{\pa_{l}} \KL \left[p_{\scm}\left(x_{l} \mid \pa_{l} \right) \| p_{\widetilde \scm}\left(x_{l} \mid \pa_{l} \backslash x_{k}\right)\right] p_{\scm}\left(\pa_{l}\right) \nonumber \\
&=\KL \left[p_{\scm}\left(x_{l} \mid \pa_{l}\right) \| p_{\widetilde \scm}\left(x_{l} \mid \pa_{l} \backslash x_{k}\right)\right],
\end{align*} where $\KL(\cdot \| \cdot)$ is the KL-divergence and $p_{\widetilde \scm}$ is the interventional distribution after removing the edge from node $k$ to node $l$. 
\end{mydef}


\newcommand{\ptest}{p_{\operatorname{te}}}
\newcommand{\ptrain}{p_{\operatorname{tr}}}


\chapter{Causal Supervised Learning}
\label{chapter:cil}

\textbf{Important Note on Structural Causal Model (SCM) Assumptions.} 
Throughout this chapter, we discuss a broad spectrum of methods that draw upon causal reasoning or attempt to learn ``causal'' or ``invariant'' representations for supervised learning. However, it is crucial to emphasize that \emph{truly recovering or identifying the full structure of an underlying SCM from observational data alone} requires strong assumptions—some of which may be untestable in principle. For instance, standard assumptions such as no hidden confounding, the Markov property, and faithfulness might be difficult or impossible to verify purely from data. Moreover, claims of “learning the SCM” typically assume that interventions and counterfactual reasoning are valid under these assumptions. In practice, access to multiple environments, additional labels or domain information, and domain knowledge can mitigate but not fully eliminate this need. The methods below are best viewed as ways to exploit partial causal invariances or structure for robustness, not as guarantees of learning an entire causal graph.

The goal of supervised learning is to learn the conditional distribution $p(y \mid \vx)$ by training on data of the form $\gD = \{(\vx_i,y_i)\}_{i=1}^N$, where $\rmX$ and $Y$ denote covariates and label, respectively. One of the most fundamental principles in supervised learning is to assume that our data  $\gD$ is \emph{independent and identically distributed} (i.i.d.). This assumption has strong implications. On the one hand, it allows us to split a set of observations into train, validation, and test dataset, opening up an easy way to perform model training, selection, and evaluation.On the one hand, it allows us to split a set of observations into train, validation, and test datasets, opening up a straightforward way to perform model training, selection, and evaluation. On the other hand, it implies that the test distribution (and future inputs) follow the same distribution as the training set—i.e., i.i.d. Put differently, the i.i.d.\ assumption says that "the past is indicative of the future." These spurious correlations can manifest in multiple ways \citep{lynch2023spawrious}. The simplest form is one-to-one (O2O) correlations, where individual features spuriously correlate with specific classes. More complex many-to-many (M2M) correlations, where groups of features correlate with groups of classes, are also common in real-world data. These spurious correlations can manifest in multiple ways \citep{lynch2023spawrious}. The simplest form is one-to-one (O2O) correlations, where individual features spuriously correlate with specific classes - for example, when each animal species is predominantly photographed in a particular habitat. More complex are many-to-many (M2M) correlations, where groups of features correlate with groups of classes \citep{lynch2023spawrious}. Consider wildlife data collected across seasons: during summer, certain species groups might be photographed primarily in one set of habitats (e.g., forests and tundra), while in winter these groups migrate, leading to entirely different habitat associations. Such M2M correlations pose a particular challenge as they cannot be decomposed into simpler O2O relationships. 

From a causal perspective, an alternative to the i.i.d.\ assumption is to assume that data are sampled from interventional distributions governed by an SCM. For a dataset generated across a set of environments $\gE$, $\left\{\left(\vx_i^e, y_i^e\right)_{i=1}^N \right\}_{e \in \gE}$, we view each environment $e \in \gE$ as sampled from a separate interventional distribution.

\begin{figure}[t]
\centering
\hspace*{\fill}
\begin{minipage}[b]{.45\linewidth}
  \centering
  \centerline{\includegraphics[width=4.0cm]{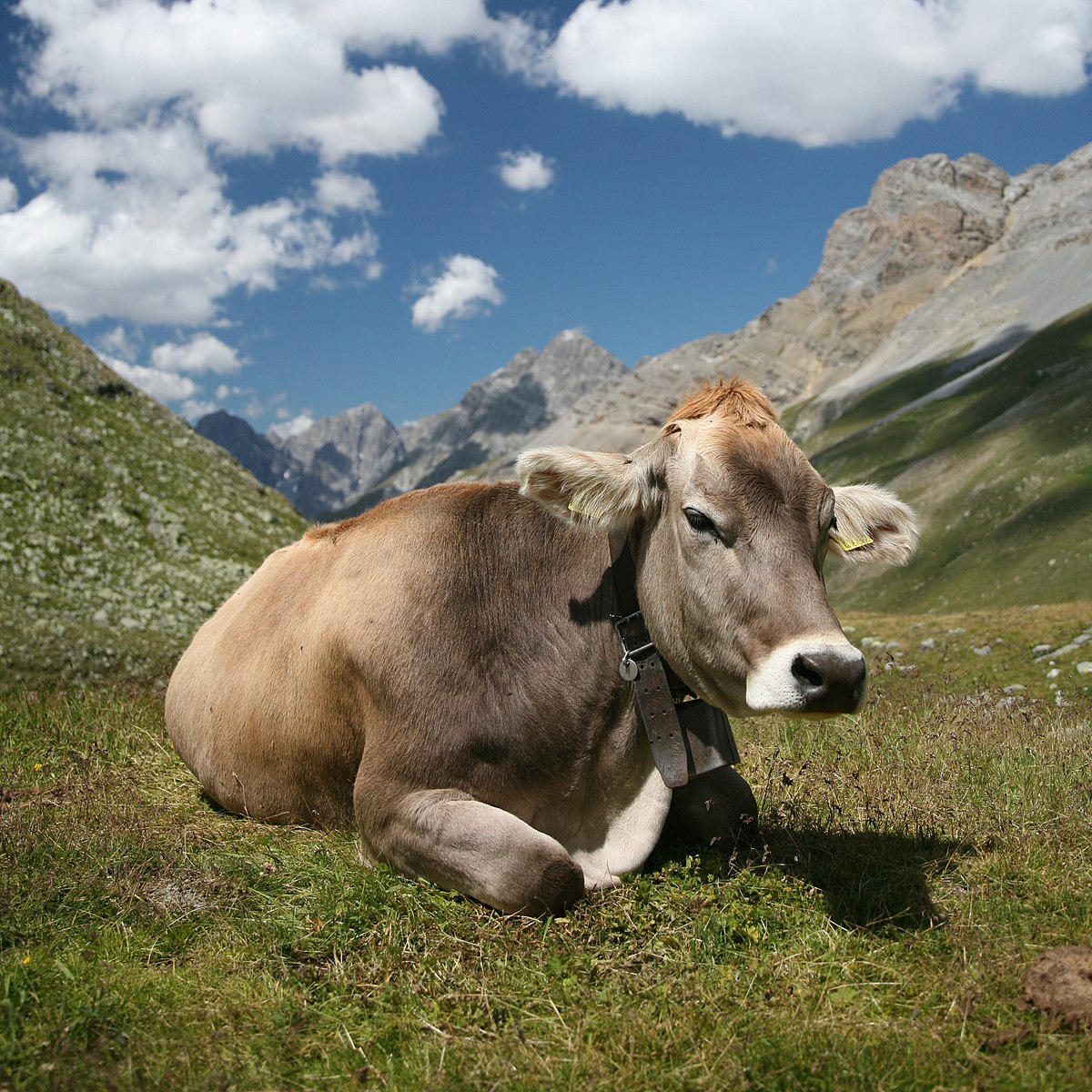}}
{\scriptsize (A) {\bf Cow: 0.99},
Pasture: 0.99,
Grass: 0.99,
No Person: 0.98,
Mammal: 0.98}
  \medskip
\end{minipage}
\hfill
\begin{minipage}[b]{0.45\linewidth}
  \centering
  \centerline{\includegraphics[width=4.0cm]{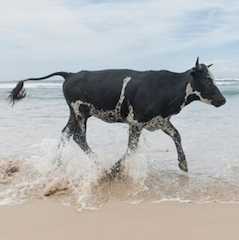}}
{\scriptsize (B) No Person: 0.99,
Water: 0.98,
Beach: 0.97,
Outdoors: 0.97,
Seashore: 0.97}
  \medskip
\end{minipage}
\hspace*{\fill}

\caption{{\bf Image classifiers are prone to spurious relationships in the dataset \citep{beery2018recognition}.} Cows in `common' contexts (e.g., Alpine pastures) are detected and classified correctly (A), while cows in uncommon contexts (beach, waves, and boat) are not detected (B). In such an instance, we can view (B) as a sample from a distribution with intervention on the background. We show the top five labels and confidence produced by ClarifAI.com.}
\label{fig:spurious_associations}
\end{figure}

With no surprise, the validity of this assumption has been challenged \citep{dulac2019challenges, underspecification}; it has been famously called \say{the big lie in machine learning} \citep{IRM}. Whenever we deploy our models in the real world, we have little to no control over the distribution we observe; e.g., variables can change in frequency (see \Cref{fig:spurious_associations}), and novel feature combinations can occur that are not contained in the training set. No matter how many samples we draw from the training distribution, there are changes that occur in the test distribution that cannot be predicted from the training data alone. Put simply, if the i.i.d. assumption breaks down, models relying on it will perform poorly \citep{shen2021towards}. 

Let us consider the example in \Cref{fig:spurious_associations}, inspired by \Cref{mlp:spurious}. In the training dataset, pictures of cows typically exhibit alpine pasture backgrounds, a spurious association caused by the cow's natural habitat. If we train a model under the i.i.d. assumption, the model will rely on this spurious association for future predictions: this is a feature, not a bug. Naturally, problems will arise when we use the model in test settings where cow pictures do not include grass backgrounds. Then, our model is at risk of misclassifying cows. This example illustrates how the i.i.d. assumption breaks down whenever the test distribution differs from the training distribution. As an alternative to the i.i.d. assumption, we can assume that our data is sampled from interventional distributions governed by an SCM. For a given dataset generated across a set of environments $\gE$, $\left\{\left(\vx_i^e, y_i^e\right)_{i=1}^N \right\}_{e \in \gE}$, we view each environment $e \in \gE$ as being sampled from a separate interventional distribution. Regarding the example in \Cref{fig:spurious_associations}, we view the test dataset as being generated by interventions on the latent variables for background features. However, claims about such interventions rest on the assumption that the underlying SCM is valid, which in turn depends on strong assumptions about the data-generating process (e.g., no hidden confounding). These assumptions are difficult to verify and should be explicitly stated when interpreting results. 

How can we estimate $p(y \mid \vx)$ in a principled manner? In the following sections, we will discuss two classes of methods that aim to learn domain-robust, transferable \emph{features} or \emph{mechanisms}: in \emph{invariant feature learning}, we learn a content representation $\rmC$ of the causal parents of Y, $\pa(Y)$, such that $Y \sim p(y \mid \vc)$ across all environments. In \emph{invariant mechanism learning}, we identify a set of mappings $\gF$ that allow us to predict $Y$ from $\rmX$ across a range of interventional distributions.

\notation

\begin{nota}
\rmX & \text{ Observed covariates} \\
\phi(\rmX) & \text{ Embedding obtained by feature map }\phi(.) \\ 
D & \text{ $\text{dim}(\rmX)$ } \\
Y & \text{ Prediction target (label) } \\
\rmS & \text{ Spurious (or style) variables for prediction} \\
\rmC & \text{  Content variables for prediction }(\pa(Y)) \\
\ex & \text{ Independent exogenous causal parents } \\
E & \text{ Environment index } \\
\gE & \text{ Set of environments } \\
\rmU & \text{ Unobserved confounders } \\
\rmA & \text{ Protected attributes} \\
\end{nota}

\section{Invariant Feature Learning}
\label{sec:ifl}

\begin{table}[ht]
    \centering
        \begin{minipage}{\textwidth}
        \centering
        \resizebox{\linewidth}{!}{%
    \begin{tabular}{|P{2.6cm}|P{2.6cm}|P{6cm}|P{2.7cm}|}
        \toprule
        \bf Available resources & \bf Method & \bf Key Idea & \bf Ref. \\
        \midrule
        \multirow{9}{*}{\shortstack{Content-\\Invariant\\Transformations}}
        & Deconfound data & Perform data augmentations and train on the augmented data & \Cref{sec:da_methods}  \\ \cline{2-4}
        & Deconfound SSL feature maps & Deconfound feature maps obtained from a pre-trained model  & \Cref{prediction:causal_transportability} \\ \cline{2-4}
        & Deconfound model during training & Regularize the model to enforce invariances & \Cref{sec:ssl} \\ \cline{2-4}
        & Deconfound predictions post-hoc & Deconfound predictions by identifying and subtracting the confounder bias & \Cref{sec:cf_regularization} \\
        \midrule
        \multirow{5}{*}{\shortstack{Multiple\\Environments}}
        & Feature Maps with Invariant Classifiers & Enforce the feature map to admit an invariant classifier & \Cref{concept:irm}  \\ \cline{2-4}
        & Object Matching & Match content features across environments & \Cref{sec:causal_matching} \\ \cline{2-4}
        & SCD with priors & Place prior on the generative variables and maximize ELBO & \Cref{sc_latent_variables} \\ 
        \midrule
        & Compositional Recognition & Learn invariant features by condition on all causal parents & \Cref{sec:comp_zero_shot} \\        
            \bottomrule
        
    \end{tabular}
    }
    \caption{\textbf{Overview of Invariant Feature Learning (IFL) Methods.}} 
    \label{tab:overview_ifl}
    \end{minipage}
\end{table}

Invariant feature learning (IFL) is the task of identifying features of our data $\rmX$ that are predictive of $Y$ across a range of environments $\gE$. From a causal perspective, the causal parents $\pa(Y)$ are always predictive of $Y$ under any interventional distribution except where $Y$ itself has been intervened upon. This is because of the  \emph{Principle of Independent Mechanisms} \citep{EOCI} (\Cref{def:pim}). 

\begin{figure}[h]
    \centering
    \includegraphics[width = 8cm]{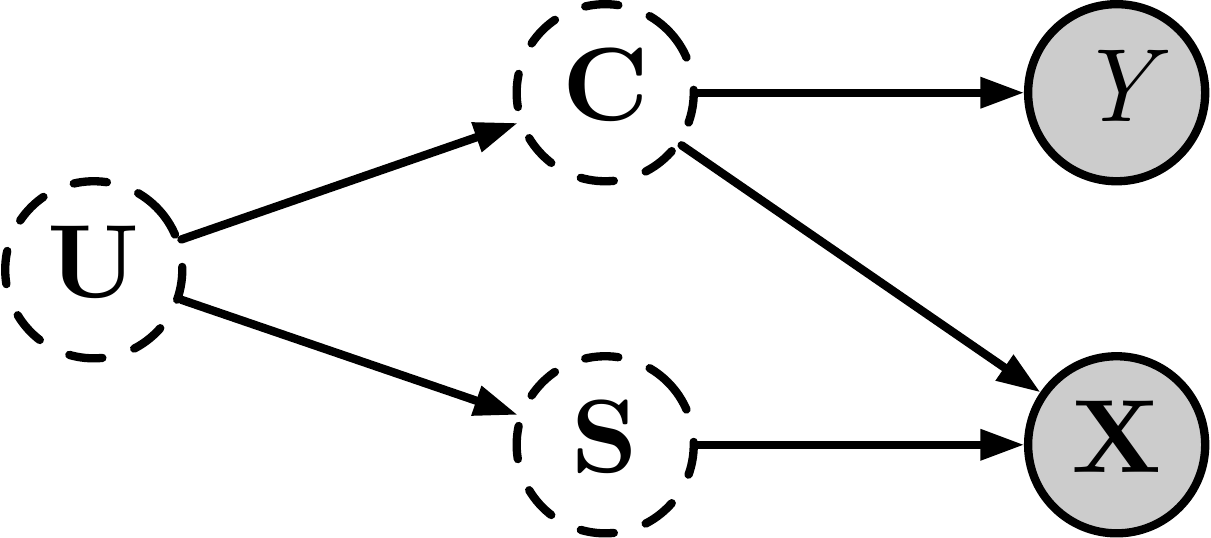}
    \caption{\textbf{General set up of the Style and Content Decomposition}: $\rmS$ and $\rmC$ both generate our data $\rmX$, but only $\rmC$ generates $Y$. The distribution $p(y \mid \vc)$ is assumed to be invariant across environments generated by interventions on $\rmS$. Often we observe spurious associations between $\rmS$ and $\rmC$: this is explained by the unobserved confounder $\rmU$.}
    \label{fig:general_SCD}
\end{figure}

IFL methods often simplify the governing SCM to focus on identifying the causal parents of $Y$. We can abstract a complex SCM into a simple SCM by collecting the causal parents of $Y$ into one variable, while the other variables are collected into another. The most general abstraction we find in the literature is the \emph{Style and Content Decomposition} (SCD) in \Cref{fig:general_SCD} \citep{style_content,DBLP:conf/icml/GongZLTGS16,DBLP:journals/ml/Heinze-DemlM21}. Note that the SCD assumption \emph{itself} requires that no hidden confounders exist among these groups, which can be nontrivial in real applications (e.g., medical imaging with unmeasured confounders).

\begin{mydef}{Style and Content Decomposition  (SCD) \citep{style_content,DBLP:conf/icml/GongZLTGS16}}{scd}
The SCD is a causal graph of a data generating process (DGP) for $\rmX$ and $Y$. We call $\rmS$ the \emph{style} variables and $\rmC$ the \emph{content} variables, and assume both are latent. The content variables group all of the causal parents of $Y$, $\pa(Y)$, while the style variables group the rest of the variables. The generations of $\rmX$ and $Y$ follow the distributions
\begin{align*}
    \rmX \sim p(\vx \mid \vs, \vc), \quad Y \sim p(y \mid \vc).
\end{align*}
\end{mydef}

Assuming that \Cref{def:scd} holds, we define Invariant Feature Learning as follows.
\begin{mydef}{Invariant Feature Learning (IFL) \citep{icp, IRM}}{ifl}
IFL aims to identify the content features $\rmC$ that cause both $\rmX$ and $Y$, i.e., 
\begin{align*} \vc & = \phi(\vx) \; \text{s.t. } Y \sim p(y \mid \vc). \end{align*}
\end{mydef}

\subsubsection*{Content-Invariant Transformations}
\label{sec:data_aug}
This section reviews IFL methods that extract content features by applying content-invariant transformations to the training dataset, feature maps, or predictions.

\subsubsection*{Deconfounding Data}
\label{sec:da_methods}
Data Augmentation (DA) is ubiquitous in modern machine learning pipelines involving high-dimensional datasets. The idea is to identify a set of transformations that can be applied to any sample in our dataset yet will not change its semantics (i.e., its content features). It requires domain knowledge from a practitioner about the types of permissible transformations that avoid diluting the content information. 
Its motivation is to enforce invariances in a model and, therefore, to improve generalization.

This section reviews works that motivate data augmentation through a causal perspective \citep{relic, data_aug_interventions, ssl_data_aug_scholkopf}. This perspective views data augmentations as constructing counterfactuals of instances, where we intervene on style features, akin to the Style and Content Decomposition (\Cref{fig:general_SCD}). 

\begin{cp}{Data Augmentation \citep{data_aug_interventions,relic}}{da}
    Data augmentation can be viewed as a set of interventions on \emph{style} variable in a \emph{style and content decomposition} (see \Cref{fig:general_SCD}). This breaks any spurious associations between style variables and $Y$. 
\end{cp}
The augmentations in this section are hand-crafted. Later in \Cref{chapter:cgm}, we will review how to generate counterfactual data augmentations using causal generative models.

\citet{data_aug_interventions} explain that DA weakens the spurious association between observed domains and task labels. To show that, they introduce the concept of \emph{intervention-augmentation} equivariance, formalizing the relationship between data augmentation and interventions on features caused by the domain. If intervention-augmentation equivariance holds, one can use data augmentation to simulate new environments $e$ using only observational data and remove the spurious association $E \leftarrow \rmU \rightarrow Y$ caused by a hidden confounder $\rmU$. Based on this insight, they derive an algorithm to automatically select appropriate data augmentation techniques from a list of available transformations, in order to achieve better domain generalization.

\citet{cad} propose a method of data augmentation via a human-in-the-loop process: given some documents and their (initial) labels, humans must revise the text with edits sufficient to flip the label, and thus generate a \emph{counterfactual sample}. Notably, the method prohibits edits not sufficient to flip the applicable label. With an augmented dataset of counterfactual samples, they label original samples positive and counterfactual samples negative. Then, they propose to use a contrastive loss \citep{chopra2005learning} to learn useful feature maps.

\citet{gradient_supervision} propose to automatically identify counterfactual samples and suggest a method to exploit them for learning useful feature maps. They propose a regularization term that enforces alignment between local gradients $\nabla_{\vx}f(\vx_i)$ and a ground truth gradient: the ground truth gradient mimics the translation in the input space (i.e. $\vx_j - \vx_i$ for counterfactual samples $\vx_i,\vx_j$) necessary to switch the model output between two contrasting samples and their corresponding labels.

\begin{figure}[h]
    \centering
    \includegraphics[width = \textwidth]{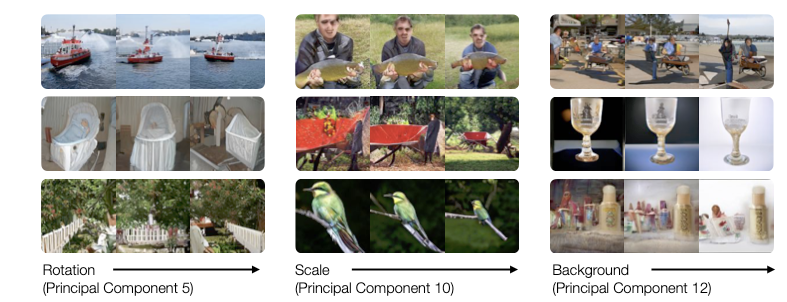}
    \caption{\textbf{\citet{generative_interventions} use \method{GANSpace} \citep{harkonen2020ganspace} to generate a variety of data augmentations}: some of the data augmentations are seen here. The data augmentations are interpreted as interventions on style features that our predictor $p(y \mid \vx)$ must be invariant to.}
    \label{fig:ganspace}
\end{figure}

\citet{generative_interventions} propose a human-in-the-loop strategy to generate effective data augmentations using generative models that mimic interventions on non-causal features, and an objective function to exploit the augmentations. They propose to empirically identify a suite of data augmentations that do not interfere with the object labels according to the method proposed in \method{GANSpace} \citep{harkonen2020ganspace}. They intervene directly on latent factors in the GAN such as those which affect background and angle of view (see \Cref{fig:ganspace}).

\subsubsection*{Deconfounding feature maps}
\label{prediction:causal_transportability}

\begin{figure}[h]
    \centering
    \includegraphics[width = 8cm]{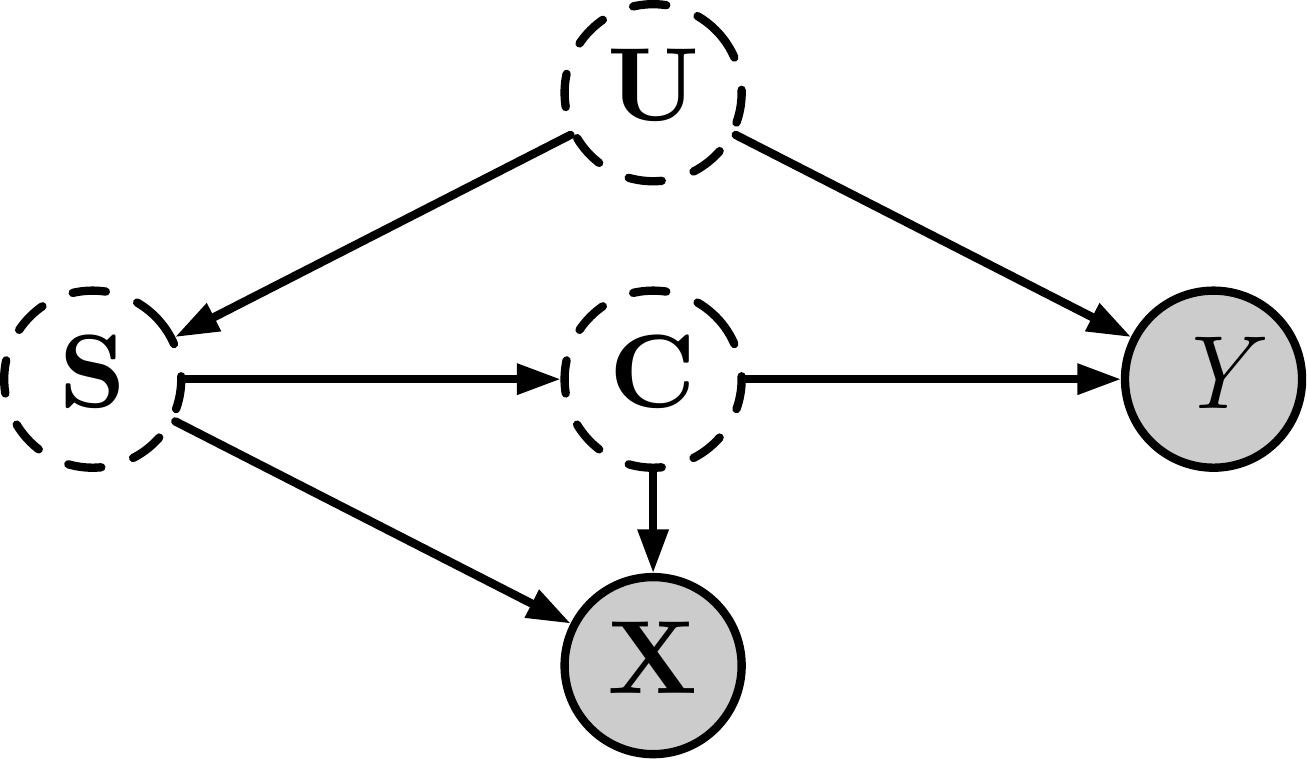}
    \caption{\textbf{Causal Transportability \citep{mao2022causal} identifies causal parents from a set of embeddings}: \citet{mao2022causal} propose to deconfound pre-trained feature map $\phi(\rmX)$ by considering the above DAG. The embeddings are split into disjoint sets to align with variables $\rmS$ and $\rmC$. The hidden variable $\rmU$ confounds $\rmS$ and $Y$.}
    \label{fig:causal_transportability}
\end{figure}

In the supervised learning setting where we also have access to a pre-trained feature map $\phi(.)$, \citet{mao2022causal} propose to deconfound the embeddings for the prediction task. They assume that the embeddings can be split into disjoint sets of style and content variables that generate $\rmX$ and $Y$, and the existence of a confounded directed path $\rmS \rightarrow \rmC \rightarrow Y$ (see \Cref{fig:causal_transportability}). Then, they propose an algorithm to obtain the causal effect $p(y \mid \doo(\vx))$ by learning a predictor that only depends on causal parents of $Y$.

The authors propose to learn $p(y \mid \doo(\vc))$ by marginalising the style variable contributions out from the embedding $\phi(\rmX)$, as shown below,

\begin{align} \label{eq:tran_marg}
    P(y \mid \operatorname{do}(c))=\sum_{\phi} \hat{P}(\phi \mid x) \sum_{x^{\prime}} \hat{P}\left(y \mid \phi, x^{\prime}\right) P\left(x^{\prime}\right)
\end{align}

Here, $\vx'$ denotes data corruptions that aim to remove content features from the sample while maintaining style features. The method includes two algorithms: training and evaluation. In the training phase a predictor $p(y \mid \phi(\vx), \vx')$ is learned, while in the evaluation phase the marginalization of equation \ref{eq:tran_marg} is performed for each prediction.



\begin{algorithm}
    \caption{Causal-Transportability Model Training \citep{mao2022causal}}
    \label{algorithm:CT_train}
    \begin{algorithmic}
        \State \textbf{Input:} Training set $D$ over $\{(\rmX, Y)\}$. 
        \State \textbf{Phase 1:} Compute $\hat{p}(\phi(\rmX) \mid \vx)$ from representation of VAE or pre-trained model.
        \State \textbf{Phase 2:} Sample $\vx_i, \phi(\rmX)_i, y_i \sim \gD^{\prime}:=(\rmX,\phi(\rmX),Y)$
        \For{$i=1,...,K$}
        \State Random sample $\vx_i'$ from the same category as $\vx_i$
        \State Train $\hat{p}(y \mid \vx^{\prime}, \phi(\rmX))$ via minimizing  $\gL_{\text{class}}$
        \EndFor
        \State \textbf{Output:} Model $\hat{p}(\phi(\rmX) \mid \vx)$ and $\hat{p}(y \mid \vx,\phi(\rmX))$
    \end{algorithmic}
\end{algorithm}
\label{alg:caus_trans}

\subsubsection*{Deconfounding models during training} \label{concept:counterfactual_invariance}

\textbf{Counterfactual Invariance.}
\emph{Counterfactual invariance} is a framework for constructing predictors whose predictions are invariant to certain perturbations on $\rmX$ \citep{veitch2021counterfactual}, using practitioner specifications. We specify an additional variable $\rmA$ that captures information that should not influence predictions however $\rmA$ may causally influence the covariates $\rmX$. Our predictor should thus be invariant to changes in $\rmA$.

Consider the image dataset in \Cref{fig:spurious_associations}. We find that our classifier seems unable to identify cows when the background is changed from the mountains to a beach. In this instance, we identify \emph{background} as $\rmA$, since changing our background should not affect the model prediction, but it does have a causal effect on the images $\rmX$. 

Let $\rmX_{\va}$ denote the counterfactual $\rmX$ we would have seen had $\rmA$ been set to $\va$, leaving all else fixed.

\begin{mydef}{Counterfactual Invariance \citep{veitch2021counterfactual}}{cf_inv}
A predictor $f(\cdot)$ is counterfactually invariant to $\rmA$ if for any given sample $\vx$, counterfactual $\vx_{\va} \sim p(\vx_{\va} \mid \vx )$, and $\forall \va \in \gA$, we have $f(\vx) = f(\vx_{\va})$.
\end{mydef}

\begin{figure}[h]
\centering
\hspace*{\fill}
\begin{minipage}[b]{.45\linewidth}
  \centering
  \centerline{\includegraphics[height=4.0cm]{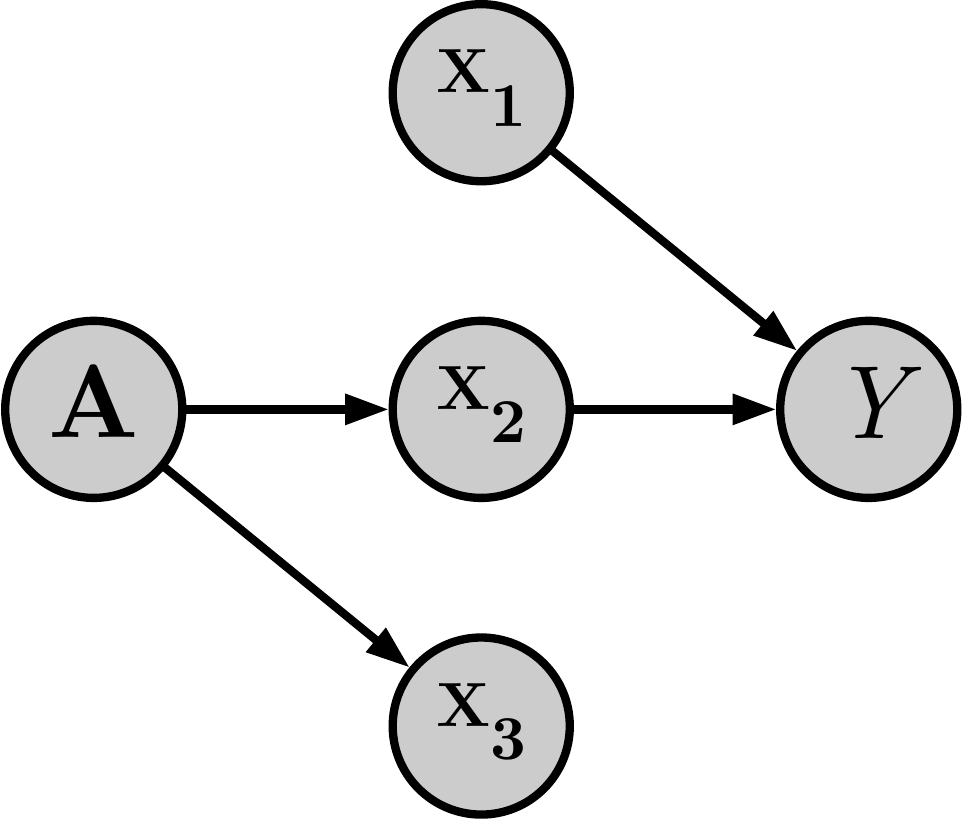}}
{\scriptsize (A) Causal direction
}
  \medskip
\end{minipage}
\hfill
\begin{minipage}[b]{0.45\linewidth}
  \centering
  \centerline{\includegraphics[height=4cm]{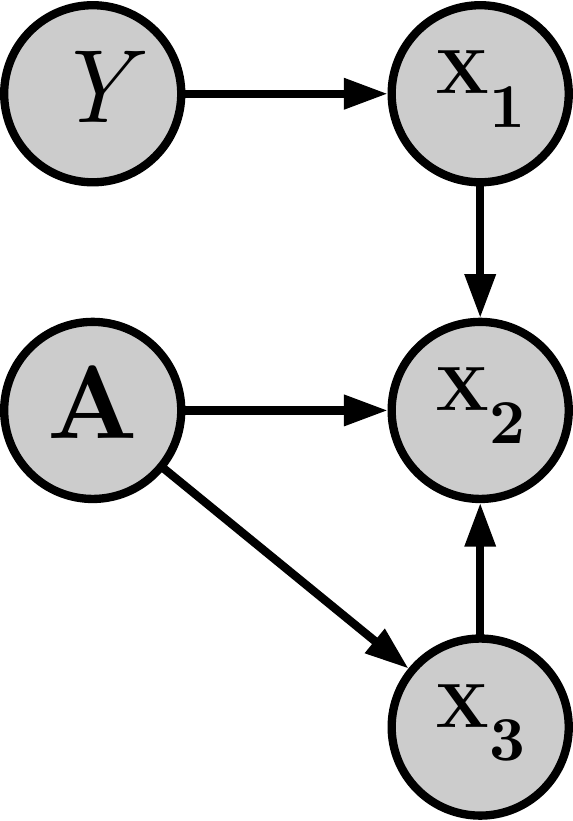}}
{\scriptsize (B) Anti-causal direction}
  \medskip
\end{minipage}
\hspace*{\fill}

\setlength{\belowcaptionskip}{-10pt}
\caption{\textbf{Counterfactual invariance causal graphs \citep{veitch2021counterfactual}}: The counterfactual invariance framework accommodates the above causal models (\Cref{concept:counterfactual_invariance}). In both $\rmX \rightarrow Y$ and $\rmX \leftarrow Y$, $\rmA$ is not a causal parent of $Y$ and we aim to remove any spurious influence $\rmA$ may have on our predictor $p(y \mid \vx)$. The goal is to predict $Y$ using only $\rmX_1$ features.}
\label{fig:cf_inv_graph}
\end{figure}

This framework requires the practitioner to identify i) the causal direction ($\rmX \rightarrow Y$ or $\rmX \leftarrow Y$), ii) and the sensitive attributes $\rmA$ (see \Cref{fig:cf_inv_graph}). With this information the practitioner can appropriately choose between one of two regularizers that enforce a counterfactual invariance signature. 

In the implementation, \citet{veitch2021counterfactual} restricted $\rmA$ to be binary however conceptually, one may extrapolate counterfactual invariance to higher dimensions with appropriate conditioning on $\rmA$. If $\rmA$ is a protected attribute (e.g., sex or race) that we would like to be an invariant input to the predictor, this definition is connected to \emph{counterfactual fairness}, introduced in \Cref{def:cff}.

\textbf{Asymmetry learning.} 
\label{sec:asymmetry_learning}
Similar to the counterfactual invariance framework, \citet{mouli2022asymmetry} present \emph{asymmetry learning} (AL), a framework for encoding transformation invariances into the prediction model. In contrast to data augmentations, AL enforces them in the training objective.

Consider the pendulum experiment included in \cite{mouli2022asymmetry}. The initial potential energy of any pendulum swing system, $\rho_1(.)$ is identified as a property that our prediction model should be invariant to. Define an equivalence relation $\sim_1$ such that for any two samples $\vx^{(1)}, \vx^{(2)}$ which have the same initial potential energy($\rho_1(\vx^{(1)}) = \rho_1(\vx^{(2)})$), we say $\vx^{(1)} \sim_1 \vx^{(2)}$. Each equivalence relation $\sim_i$ induces a set of object transformations $\mathcal{T}_i$ that preserve equivalence class membership. 

The key idea is that these transformations can be composed to model distribution shifts. We can view each $T$ as a concatenation of transformations associated with distinct equivalence relations, $T = t_1 \circ .. \circ t_r$ for arbitrary $r$. A distribution shift can be modeled by the differences between $T^{\mathrm{tr}}$ and $T^{\mathrm{te}}$ in a subset of the transformations: for example, we may have $T^{\mathrm{tr}} = t_1 \circ t_{\text{tr}} \circ t_3$ and $T^{\mathrm{te}} = t_1 \circ t_{\text{te}} \circ t_3$. To generalize to novel test environments, we must specify which equivalence relation-induced transformations will remain invariant. 

\textbf{Self-supervised learning using SCD.} \label{sec:ssl} While the previous methods aimed to improve prediction for one task, the following methods explore how the principles of IFL can be exploited for self-supervised learning (SSL) where representations, or \emph{feature maps}, are learned to assist with a suite of downstream tasks. This strategy has paved the way to utilize a vast amount of unlabeled data in settings for which the availability of labeled data is often limited, e.g., for large-scale language modeling \citep{gpt3}, medical image analysis \citep{chen2019self}, or molecular property prediction \citep{molecule_ssl}. We can divide prevailing approaches primarily into whether they are reconstructive \citep{mae} (learning a distribution over data) or discriminative \citep{simclr, instance_discrimination, byol} (which rely on a handcrafted prediction task, and includes contrastive SSL). 


\citet{ssl_data_aug_scholkopf} prove that data augmentations for contrastive SSL which simulate style interventions in the SCD (\Cref{fig:general_SCD}) can indeed isolate the invariant content features in the learned feature map, and \citet{relic} propose an objective named \method{ReLIC} to do so.

\citet{relic} frame the task of self-supervised learning as proxy task prediction, where they propose a list $\gY$ of proxy labels $y_t$. They formalize the criterion \ref{relic_eq} which enforces invariance to data augmentations and is implemented as a KL-divergence term appended to the objective. 

\begin{align}
    p\left(y_t \mid \vc, \doo\left(\vs^{(1)}\right)\right) = p\left(y_t \mid \vc, \doo\left(\vs^{(2)}\right)\right) \label{relic_eq}
\end{align}

As follow-up work, \citet{relicv2} propose \method{ReLICv2}, which differs from \method{ReLIC} in the selection of appropriate positive and negative pairs and how they combine resulting views of data in the objective function.

\subsubsection*{Deconfounding Predictions Post-Hoc}
\label{sec:cf_regularization}

A variety of methods \citep{counterfactual_analysis, CF_VQA, CF_attention} propose to remove the influence of unobserved confounders $\rmU$ from predictions through \emph{counterfactual regularization}, after model training has ended. This involves estimating the confounding effect of $\rmU$ on the prediction $\tilde{Y}$ and then subtracting it, thus deconfounding the prediction. For a prediction over sample $\vx$, $\vx'$ is generated such that it carries none of the causal information in $\vx$. Then, the prediction is deconfounded using the difference  
\begin{equation} \label{eq:CF_regularization}
    \tilde{Y}_{\text{causal}} = \tilde{Y}_{\vx} - \hat{Y}_{\vx'}.
\end{equation}

\begin{figure}[h]
\centering
\begin{subfigure}[b]{.45\linewidth}
  \centering
  \centerline{\includegraphics[width=4.0cm]{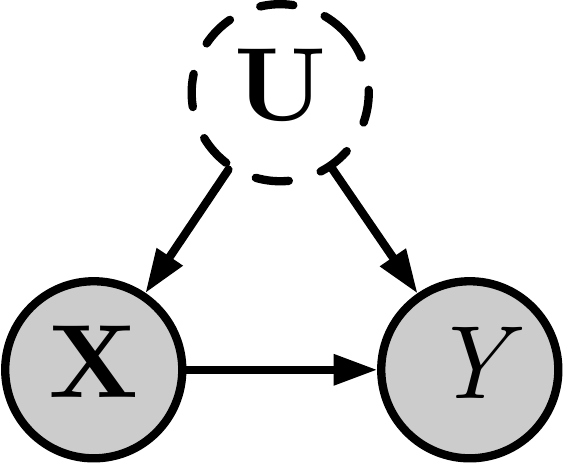}}
\caption{\scriptsize Original causal graph}
    \label{fig:cf_reg_no_int}
\end{subfigure}
\hfill
\begin{subfigure}[b]{0.45\linewidth}
  \centering
  \centerline{\includegraphics[width=4.0cm]{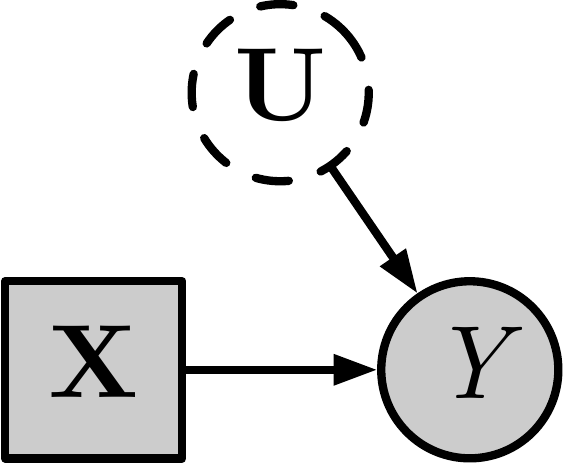}}
\caption{\scriptsize Intervention on $\rmX$}
  \label{fig:cf_reg_int}
\end{subfigure}
\hfill
\setlength{\belowcaptionskip}{-10pt}
\caption{\textbf{Counterfactual Regularization to remove unobserved confounding}: An intervention on $\rmX$ eliminates the influence of $\rmU$, and deconfounds the causal effect $\rmX \rightarrow Y$.}
\label{fig:CF_regularization}
\end{figure}

\citet{counterfactual_analysis} apply an intervention to $\rmX$ by setting it to a zero vector (or random noise), identifying the resulting prediction of $\hat{Y}$, and implementing \Cref{eq:CF_regularization}. The authors propose to use this for domain generalization in trajectory prediction. \citet{CF_attention} propose improving attention mechanisms for visual categorization, and \citet{CF_VQA} propose removing language bias in visual question answering.

\subsection{Multiple Environments}
\label{sec:mult_envs}

Often we collect datasets from multiple environments $\gE$ in the form $\left\{\left(\vx_i^e, y_i^e\right)_{i=1}^N \right\}_{e \in \gE}$. For example, records obtained from hospitals under varying protocols or images of houses collected in different seasons. The WILDS benchmark \citep{WILDSdataset} offers a curated collection of multiple environment data from real-world scenarios. 

As many machine learning algorithms operate on the i.i.d. assumption, practitioners may shuffle the data collected from multiple environments. \citet{IRM} suggest that information about the data generating process is lost when we shuffle: by shuffling the data, we lose any distinction between associations across and within environments. For example, a spurious association exists between grass features and the cow label in \Cref{fig:spurious_associations}, yet this association does not hold across environments if one environment is alpine pastures and another is the beach.

How can we use environment labels effectively? From a causal perspective, we can treat each environment as being a distinct interventional distribution, generated by interventions on style variables in the SCD (\Cref{fig:general_SCD}). A variety of environments should reveal style variables as having unstable correlations with the output, while content variables should remain predictive.

\begin{cp}{Multiple Environments}{multi_env}
    For data in the form $\{\rmX^e, Y^e \}_{e \in \gE}$ where $\gE$ is a set of environments, it is often reasonable to view each environment as being generated by a set of interventions on the \emph{style} variables in the SCD (\Cref{fig:general_SCD}). Of course, this perspective fails when the underlying SCM differs between environments.
\end{cp}

The methods in this section propose identifying the predictive features across all environments and learning an invariant mapping from the invariant features to the output variable $Y$. We interpret the invariant features as content features in the SCD or causal parents of Y.

\subsubsection*{Feature Maps with Invariant Classifiers}
\label{concept:irm}
\citet{icp} introduce the algorithm \emph{Invariant Causal Prediction} (\method{ICP})to find the \emph{causal feature set}, the minimal set of features that are causal predictors of a target variable. They exploit the Principle of Independent Mechanisms (\Cref{def:pim}) to identify an invariant predictor from the causal features to the target variable.

Invariant Risk Minimization (\method{IRM}) \citep{IRM} is an extension of \method{ICP} that learns a mapping from $\rmX$ to a feature set $\phi(\rmX)$ containing only invariantly predictive features, and a linear predictor $w$ on top. It posits the existence of this feature map $\phi(\cdot): \gX \rightarrow \tilde{\gX}$ such that the linear classifier $\hat{w}: \phi(\rmX) \rightarrow \mathcal{Y}$ is optimal in every environment $e \in \gE$. The feature map and classifier then compose to form $f(\boldsymbol{x}) = w \circ \phi(\boldsymbol{x})$. \citet{IRM} define the optimally-invariant classifier as one that minimizes the worst-case environment-specific empirical risk $\gR^e :=\mathbb{E}_{\rmX^{e}, Y^{e}}\left[\ell\left(f\left(\rmX^{e}\right), Y^{e}\right)\right]$ over all environments. 

The authors argue that such a function will only rely on invariant features because non-invariant features will have unstable correlations with the target variables between different environments, and lead to a non-optimal linear predictor in at least one environment. The invariant features $\phi(\rmX)$ can be viewed as the content variables in the SCD (\Cref{fig:general_SCD}), while the linear predictor $w$ maps from content features to the target variable.

\begin{mydef}{Invariant Risk Minimization \citep{IRM}}{irm}
To learn $\phi(\rmX): \rmX \rightarrow \tilde{\mathcal{X}}$ and an invariant classifier function $w_{\beta}: \phi(\rmX) \rightarrow \mathcal{Y}$ parameterized by $\beta$, the \method{IRM} objective is the following constrained optimization problem:
\begin{align}
\begin{split}
    \min_{\phi, \hat{\beta}} \sum_{e \in \mathcal{E}} \mathcal{R}^{e}(\phi, \hat{\beta})  \quad \text { s.t. } \quad \hat{\beta} \in \underset{\beta}{\arg \min } \; \mathcal{R}^{e}(\phi, \beta) \quad \forall e \in \mathcal{E} .\label{eq:irm}
\end{split}
\end{align}
\end{mydef}

In practice, this bilevel program is non-convex and difficult to solve. The authors consider a Lagrangian form as an approximation, whereby the sub-optimality concerning the constraint is expressed as the squared norm of the gradients of each of the inner optimization problems:
\begin{align}
    \min _{\phi, \hat{\beta}}  \sum_{e \in \mathcal{E}}\left[\mathcal{R}^{e}(\phi, \hat{\beta})+\lambda\left\|\nabla_{\hat{\beta}} \mathcal{R}^{e}(\phi, \hat{\beta})\right\|_{2}^{2}\right]\label{eq:irm_lagr}
\end{align}
Assuming the inner optimization problem is convex, achieving feasibility is equivalent to the penalty term being equal to $0$. Thus, \Cref{eq:irm} and \Cref{eq:irm_lagr} are equivalent if we set $\lambda=\infty$.

Unfortunately, both \citet{risks_of_irm} and \citet{kamath2021does} show that \method{IRM} often performs no better than standard \emph{empirical risk minimization} (ERM). \method{IRM} achieves its best results when the underlying causal relations are linear, and there is sufficient heterogeneity observed in the training environments, such that enough \emph{degrees of freedom} are eliminated \citep{risks_of_irm}. However, if either of these conditions is not met, then \method{IRM} can achieve worse results than ERM. \citet{ahuja2020invariant} pose the \method{IRM} objective as finding the Nash equilibrium of an ensemble game among several environments. 

\citet{ahuja2021invariance} point out that while IRM-like approaches provably generalize OOD in linear regression tasks, they do not necessarily in linear classification tasks, which require much stronger restrictions in the form of support overlap assumptions on the distribution shifts. They establish that augmenting the invariance principle with \emph{information bottleneck} \citep{tishby2000information} constraints resolves some of these issues. 

Similar to \method{IRM}, \citet{risk_extrapolation} propose \method{Risk Extrapolation} (\method{REx}), a domain generalization method which uses a weaker form of invariance than \method{ICP}. While \method{IRM} specifically aims for invariant prediction, \method{REx} seeks robustness to the worst risk among a space of affine combinations of the training distributions. The authors prove that variants of \method{REx} can recover the causal mechanism $p(y \mid \pa_y)$ under several assumptions including \emph{homoskedasticity} (see \citep{risk_extrapolation} section 3.2), while also providing some robustness to covariate shifts.

\citet{wang2021desiderata} note that invariant features should be \emph{non-spurious} but also \emph{efficient}. To assess non-spuriousness and efficiency simultaneously, they require the feature map to satisfy a probability of necessity and sufficiency (PNS) condition, which captures the following behavior.

A representation is non-spurious and efficient if the label responds to the feature both ways: non-spuriousness refers to the effectiveness of turning on the label by turning on the feature captured by the representation. Efficiency is the effectiveness of turning off the label by turning off this feature. In other words, if the feature is enabled, the label will be enabled; if it is disabled, the label will be disabled. PNS calculates the probability of this condition. The authors turn this condition into a constrained optimization objective, which they call \method{Causal-Rep}. 

\citet{jiang2022invariant} study anti-causal domain shifts, where $\rmX \leftarrow Y$, instead of $\rmX \rightarrow Y$. Similar to \method{IRM}, the authors propose a training objective for extracting an invariant predictor. Recall that IRM aims to build predictors that only rely on the causal parents of $Y$. If the observed covariates $\rmX$ may be all caused by $Y$, the causal parents of $Y$ are the empty set. In this case, the naive causally invariant predictor is vacuous. \citet{jiang2022invariant} address these scenarios. 

\citet{ISR} introduce \emph{\textbf{I}nvariant-feature \textbf{S}ubspace \textbf{R}ecovery} (ISR), addressing the fact that IRM and its extensions cannot generalize to unseen environments with less than $d_s+1$ training environments, where $d_s$ is the dimension of the spurious-feature subspace. They propose two algorithms, \method{ISR-Mean} and \method{ISR-Cov}. By identifying the subspace spanned by invariant features, \method{ISR-Mean} achieves provable domain generalization with $d_s+1$ training environments. By using the information of second-order moments, \method{ISR-Cov} reduces the required number of training environments to $\gO(1)$. 

\subsubsection*{Object Matching}
\label{sec:causal_matching}

\begin{figure}[t!]
    \centering
    \includegraphics[width=0.7\textwidth]{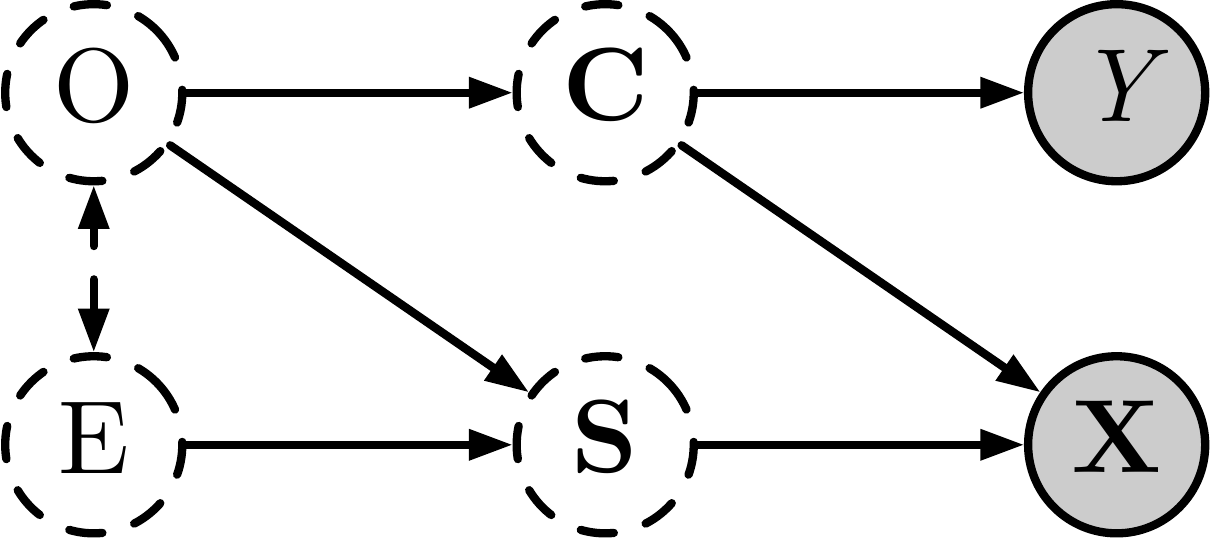}
    \caption{\textbf{Modified Style and Content Decomposition \citep{caus_match}}: When $\rmS$ and $\rmC$ are correlated, we would like to describe the nature of this association using a causal graph. Causal Matching \citep{caus_match} introduces the environment index variable $E$ and object variable $O$ to extend the causal graph. The algorithm aims to identify objects from content features $\rmC$.}
    \label{fig:SC_struc2}
\end{figure}

An alternative to the constrained optimization objectives above is to construct a \emph{matching function} $\Omega$ \citep{caus_match} where similar data points map to 1 and dissimilar points to 0, then learn a feature map $\phi(\rmX)$ where matched data points are close. In the presence of domain knowledge, a practitioner can specify a distinct matching function of their own.

Specifically, the constructed matching function $\Omega: \mathcal{X} \times \mathcal{X} \rightarrow\{0,1\}$ maps pairs of same-label data points from different environments to 1, otherwise pairs are mapped to 0. That is, for any $(\vx_i^{(e)},y)$ and $(\vx_j^{(e')},y)$ where $e \neq e', i \neq j$, we have $\Omega(\vx_i^{(e)}\vx_j^{(e')}) = 1$. The practitioner can choose an alternative criterion for matching data points otherwise.

Given the matching function, \citet{caus_match} propose a 2-stage algorithm, \method{MatchDG}: compute the optimal scaling parameter for the feature map distance function, then learn a predictor whose feature map is constrained such that matched data points are close together and non-matched data points are far apart (i.e. an example of a contrastive loss \citep{chopra2005learning}).

The method is motivated by considering the causal DAG in \Cref{fig:SC_struc2}. The aim is to learn a feature map that aligns with the \emph{object} latent variable $O$ generating data points $(\vx,y)$, given confounding by the environment variable $E$.

\subsubsection*{SCD with priors} 
\label{sc_latent_variables}

The previous methods both proposed learning feature maps that capture content variables and ignore style variables with bi-level optimization objectives and differed in how they constrained the feature map. The following methods instead place a prior distribution $p(\vc, \vs)$ over the content and style variables (\Cref{fig:general_SCD}) and learn a generative model over the joint distribution $p(\vx, y)$, for example through a factorized variational inference objective \citep{original_vae}. 

\citet{LaCIM} propose \method{LaCIM}, which learns separate encoders 
$q_e(\vs, \vc \mid \vx)$ for each environment but shared decoders $p(\vx \mid \vs, \vc)$ and $p(y \mid \vc)$, since $p(\vx, y \mid \vc, \vs) = p(\vx \mid \vs, \vc)(y \mid \vc)$ in the SCD DAG. Given the possibility of confounding between the style and content variables, they also learn environment-specific priors $p_e(\vs, \vc)$. 

Separately, \citet{csg} propose \method{CSG}, which trades off learning environment-specific priors for only requiring one training environment. In contrast to \method{LaCIM}, their best approach assumes independent style and content variables, $\rmC \indep \rmS$.

\citet{lu2022invariant} propose \method{iCaRL}, which performs the same task as \method{LaCIM} but with a 3-stage training procedure instead of the end-to-end approach. Phase 1: learn an environment-conditional encoder $q(\vz \mid \vx, e)$ which maps to disentangled latent variables $\rmZ$ (more on this topic in \Cref{chapter:cgm}). Phase 2: apply the PC algorithm \citep{spirtes2000causation} to learn a set of possible DAGs governing the factorization of the latents $\rmZ$, then identify the causal parents of $Y$, $\rmC \subset \rmZ$. Phase 3: Learn a classifier $w: \gC \rightarrow \gY$ using standard ERM.

\subsubsection*{Compositional Recognition}
\label{sec:comp_zero_shot}

\begin{figure}[h]
    \centering
    \begin{subfigure}[t]{0.6\columnwidth}
        \includegraphics[width=\columnwidth]{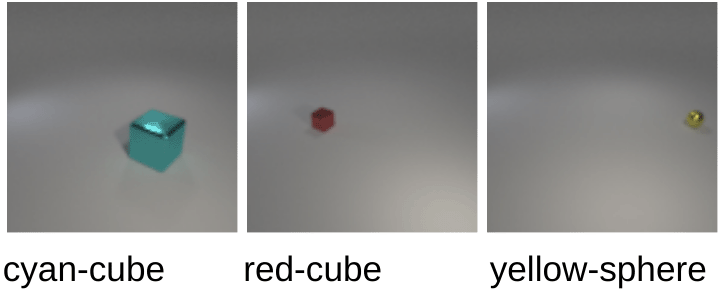}
        \caption{\textbf{Example dataset}}
        \label{fig:comp_images}
    \end{subfigure}  
    \begin{subfigure}[t]{0.3\columnwidth}
    \includegraphics[width=\columnwidth]{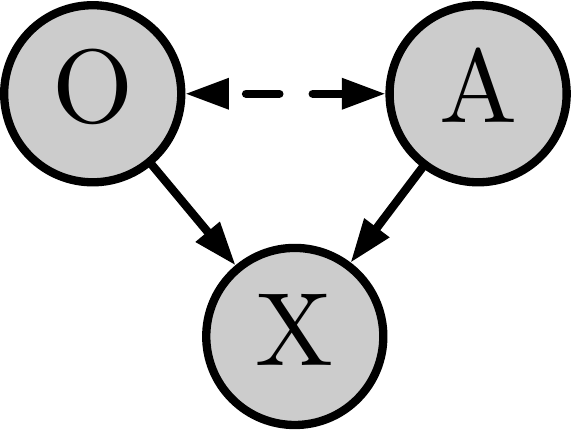}
    \caption{\textbf{Causal DAG}.}
    \end{subfigure}
    \caption{\textbf{Composition of attributes leads to robust generation \citep{atzmon_causal_2020}}: While objects and attributes features are unreliable features for prediction when used by themselves, combinations of such features are assumed to be reliable (\Cref{sec:comp_zero_shot}). Hence, they learn $p(\vx \mid \va, o)$, which will be invariant across interventional distributions. Examples of the dataset required are shown in \Cref{fig:comp_images}, collected from the AO-CLEVr dataset \cite{johnson2017clevr}.}
    \label{fig:cp}
\end{figure}

\emph{Compositional recognition} is a type of zero-shot learning \citep{xian2017zero} where the goal is to identify new combinations of a known set of objects. 
\citet{atzmon_causal_2020} argue that deep discriminative models fail at compositional recognition because of two reasons: (i) distribution-shift and (ii) representation entanglement. Hence, they propose constructing a feature map based on a causal graph that models images as being caused by attributes $\rmA$ and objects $O$, as illustrated in \Cref{fig:cp}(b). Unlike objects or attributes by themselves, \citet{atzmon_causal_2020} argue that combinations of objects and attributes generate the same distribution over images in train and test sets. Hence, they propose considering images of unseen combinations generated by interventions on the attribute and object labels. 
They learn the distribution $p(\vx \mid \va, o)$, which is stable across environments, instead of $p(\vx \mid \va)$ and $p(\vx \mid o)$ which vary. Given $p(\vx \mid \va, o)$, they learn a classifier from this invariant representation.

\section{Invariant Mechanism Learning}
\label{sec:iml}

\begin{table*}[h!]
    \centering
    \begin{minipage}{\textwidth}
        \centering
        \resizebox{\linewidth}{!}{%
    \begin{tabular}{|P{2cm}|P{9cm}|P{1.5cm}|}
    \toprule
     \bf Method & \bf Key Idea & \bf Ref. \\
     \midrule
     Separate Networks & Networks can be combined in novel domains and account for independent interventions & \Cref{sec:inv_mech_sep_nets} \\
     \hline
     Domain Mappings & Mappings from target domain to source domain; each corresponds to an independent intervention & \Cref{sec:inv_mechanisms_mappings} \\
     \bottomrule

    \end{tabular}
    }
    \caption{\textbf{Overview of IML Methods.}}
    \label{tab:invariant_mechanism_learning}
    \end{minipage}
\end{table*}

In the previous section, we exploited the utility of invariant features to remove spurious associations from our models. Now we look at various methods with fundamentally different objectives compared to invariant feature learning. 

Consider how a human may hear someone speaking quietly or loudly but can distinguish between what is said and how loudly it is. If one wants to model what is being said, the content of the speech corresponds to features of the data, while the speech volume corresponds to the mechanism that maps the information to the observer.

\begin{figure}[h]
    \centering
    \includegraphics[width=5cm]{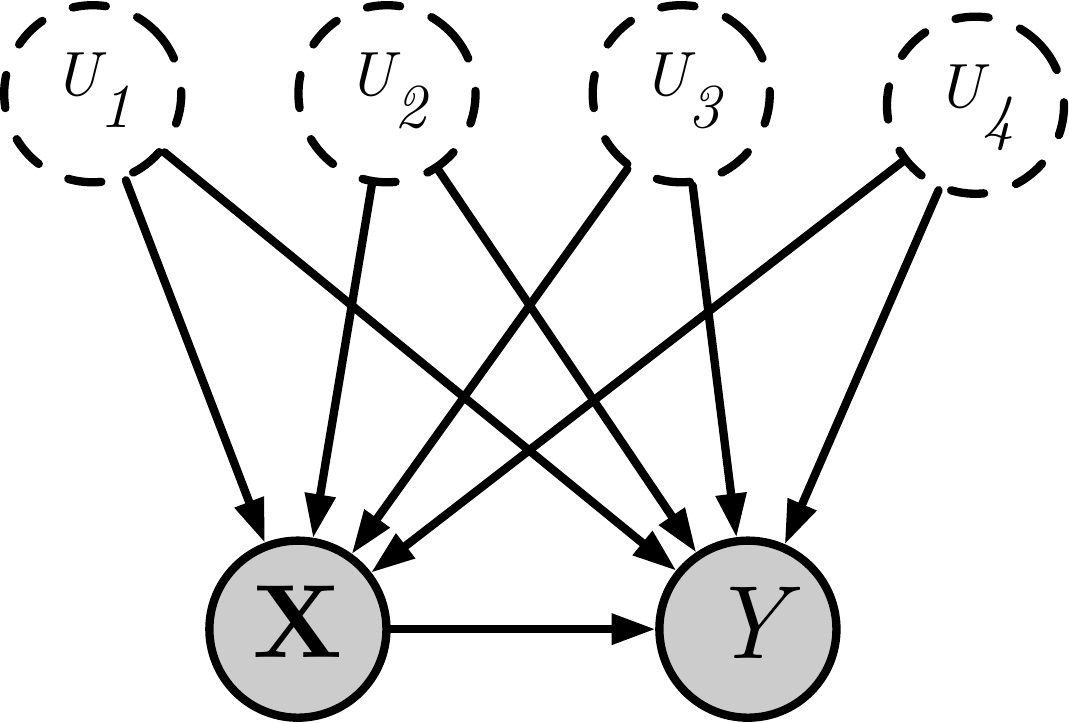}
    \caption{\textbf{Graphical motivation for invariant mechanism learning}: We consider $\rmX \rightarrow Y$ to be confounded by a set of independent confounders $\rmU$, such that an intervention on any $U_i$ generates an interventional distribution.}
    \label{fig:IML_motivation}
\end{figure}

Invariant mechanism learning (IML) aims to identify data-generating mechanisms representing different interventional distributions. For independent latent confounders $\rmU$ of $\rmX$ and $Y$, we view each interventional distribution as being generated by interventions on a subset of the confounders (see \Cref{fig:IML_motivation}). Then, to collect learned mechanisms $\gF$, we employ a subset of the mappings $\gF$ to predict $Y$ from $\rmX$. We emphasize that IML does not learn feature maps of the confounders $\rmU$.

\begin{mydef}{Invariant Mechanism Learning (IML)}{iml}
The task of IML is to identify a set of mappings $\gF$ which capture an interventional distribution generated by independent unobserved confounders $\rmU = \{U_1, .., U_{D} \}$: 
\begin{align*}
    \gF = \big \{ f_i &: \gX \rightarrow \gY \mid Y = f_i(\rmX) \Leftrightarrow Y \sim P\big(Y \mid \rmX, \doo(u_i)\big) \big\}_{i=1}^{D}
\end{align*}
\end{mydef}

\subsection{Independent Invariant Mechanisms as Separate Networks} 
\label{sec:inv_mech_sep_nets}
\citet{learning_icm} propose a method that aims to identify a suite of competing data transformations ('mechanisms') trained to specialize in recovering distinct underlying structures from a sample. \citet{goyal2021recurrent} develop this by applying independent mechanisms to sequential data, specifically video and text. They call their proposed architecture \emph{Recurrent Independent Mechanisms} (\method{RIMs}). 

At each time step, a soft attention layer selects the top $k$ out of $N$ competing mechanisms for input processing. A second attention layer allows for 'sparse communication' between the mechanisms at each time step to aid contextual understanding. The benefits observed over other sequential architectures such as the LSTM (\citet{lstm_original}) and transformers (\citet{attention_is_all_you_need}) are (i) reduced degradation of information over long-dormant phases in the input, and (ii) improved trajectory prediction for multiple objects.

\citet{madan2021fast} use the RIMs proposed by \citet{goyal2021recurrent} for a meta-learning algorithm that quickly adapts the parameters of the modules and slowly adapts the attention mechanism parameters. Their fast and slow learning dynamic demonstrates improvements over LSTM and RIMs in meta-learning benchmarks.

\begin{figure}[h]
    \centering
    \includegraphics[width=4cm]{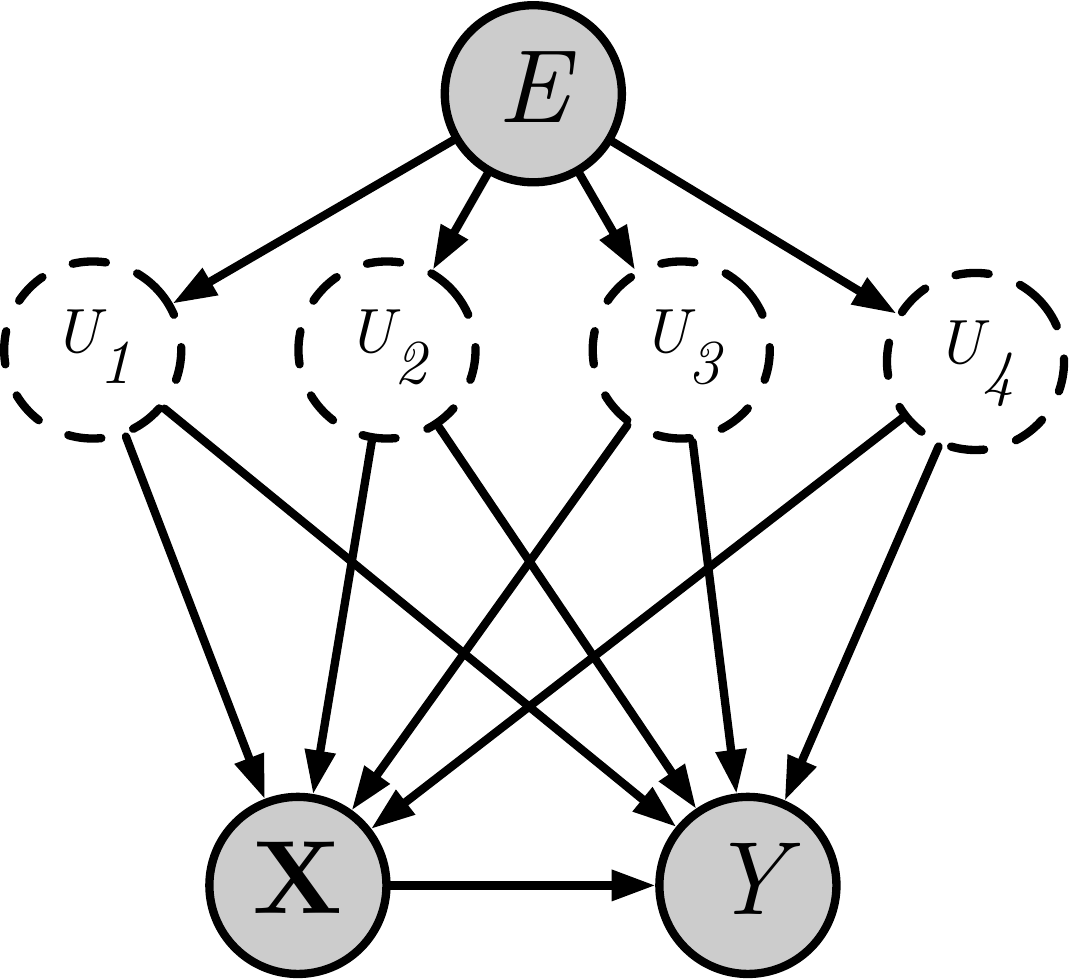}
    \caption{\textbf{Invariant mappings between domains}: \citet{transporting_causal_mechanisms_UDA} propose to learn a set of mappings between different environments generated by the above causal graph (\Cref{sec:inv_mechanisms_mappings}), where $\rmU$ is confounding the causal effect $\rmX \rightarrow Y$. Hence, they propose to learn a set of mappings between the source and test domain that each account for an isolated intervention on independent parts of $\rmC$.}
    \label{fig:UDA_transport}
\end{figure}

\subsection{Invariant Mechanisms as Mappings Between Domains}
\label{sec:inv_mechanisms_mappings}

\citet{transporting_causal_mechanisms_UDA} propose to view domain adaptation as a problem of inferring a set of disentangled causal mechanisms that generate mappings from the source domain to the test domain.


They model the data-generating process using the causal graph in \Cref{fig:UDA_transport}. $E$ represents an environment index and $\rmU$ a set of unobserved domain aware confounding variables on $\rmX$ and $Y$. Given that $\rmU$ are unobserved, the authors propose to learn a set of proxy variables $\hat{\rmU}$ and identify disentangled counterfactual mappings between domains that each correspond to an isolated intervention on $U_i$ while keeping all other $U_{j \neq i}$ constant. Specifically, they learn $\left\{\left(M_{i}, M_{i}^{-1}\right)\right\}_{i=1}^{k}$ in an unsupervised fashion, where $M_{i}: X_{\text{tr}} \rightarrow X_{\text{te}}$ and $M_{i}^{-1}: X_{\text{te}} \rightarrow X_{\text{tr}}$. 

For a given input in the test domain, they propose to map the test domain input to the source domain (i.e., to what the input would have been had it been generated in the source domain) and then predict based on this counterfactual input. 

Similarly, \citet{teshima2020few} propose to identify an invariant mechanism that generates $\rmX$ from a set of independent components $\rmZ$ which vary across domains. The method takes in training data from multiple domains and uses nonlinear ICA to \emph{identify features} $\rmZ$ and an invariant mapping $f:\mathcal{Z} \rightarrow \mathcal{X}$ across domains. The learned mapping $\hat{f}$ identifies the values of $\rmZ$ in the target domain and generates pseudo samples that mimic the target domain. A standard supervised learning algorithm can be trained on this generated data. 

\section{Open Problems}

\subsection{Lack of Targeted Benchmarks for Invariance Learning} 
\label{sec:targeted_benchmarks}

Previous works in invariance learning often evaluate their suggested methods on novel toy experiments instead of a standardized test bed. As a result, practitioners cannot quickly determine the best approach for their problems. We summarize some existing benchmarks that may serve as inspiration for future works.

One common approach to evaluating causal supervised learning models is introducing spurious associations in the training data and evaluating performance in a test domain where the association has changed. For instance, \citet{IRM} introduce the \method{ColoredMNIST} benchmark, which adds colors to the digits and changes the strength of association between color and digit label between test and training data.

\citet{wang2021desiderata} varies spurious associations between face attributes of interest and irrelevant characteristics of the \method{CelebA} dataset \citep{liu2015faceattributes}. For example, the training data contains a spurious association between \emph{black hair} and \emph{necklaces}. Further, they propose to compare the generalization ability under changes to the specified dimension of the representations in their models. 

Recent work has expanded benchmark complexity by developing synthetic datasets that carefully control spurious correlations while maintaining photo-realism \citep{lynch2023spawrious}. These benchmarks often include multiple difficulty levels and explicit training/test distribution shifts, allowing for systematic evaluation of robustness methods. By leveraging modern text-to-image models, researchers can generate large-scale datasets with precise control over spurious associations while maintaining the visual complexity needed to reflect real-world challenges \citep{lynch2023spawrious}.

\subsection{Connections between Invariance Learning, Adversarial Robustness and Meta-learning}
\label{sec:IFL_benefits_AR/ML}

We note two active ML research areas akin to invariance learning: adversarial robustness and meta-learning. We investigate how causal assumptions may benefit these areas, as it is currently underexplored. 

In adversarial robustness, we are interested in learning classifiers that are robust to adversarial perturbations. An adversarial perturbation is an additive random variable $\Delta$ that makes a model fail to classify $\tilde \rmX = \rmX + \Delta$, while the model correctly classifies an image $\rmX$. Typically, $\tilde \rmX$ and $\rmX$ are indistinguishable from the human eye. Naturally, we can see the problem of AR from a causal perspective and interpret a robust model as invariant against perturbations, as we will later see in \Cref{nlp:caus_int_nlp}. 

In meta-learning, we often seek to learn shared structure across tasks and task-specific parameters that can be quickly adapted to unseen tasks \citep{recasting_MAML,mikulik2020meta,PAML}. Here, we can interpret the shared structure as being invariant across tasks. Future work must examine if causal assumptions can be used to learn better task-invariant representations. For example, instead of conditioning on task variables, it could be interesting to explore interventions on them.

\subsection{Exploiting Additional Supervision Signals}
\label{sec:supervison_signals}

The previously discussed methods in invariant feature learning (\Cref{sec:ifl}) require two distinct forms of extra supervision beyond labels $Y$: i) content-invariant transformations (\Cref{sec:data_aug}), or ii) environment indices (\Cref{sec:mult_envs}). These additional supervisory signals serve as means to separate data among different interventions on the spurious features, which helps a model avoid any predictive dependence on them. 

We suggest two directions for future research. First, we can combine these two forms of signals to exploit a combination of domain knowledge of the data-generating process and data collected from multiple environments. One straightforward idea is to imagine data augmentations as themselves generating new environments and thus use existing algorithms to handle data from multiple environments, with data augmentation techniques multiplying the number of environments available.

Second, we can attempt to exploit high-dimensional environment information, as opposed to environment indices $e \in \gE$. For instance, one could complement the dataset collected for the cow images (\Cref{fig:spurious_associations}) with an aerial photo of the pictures' location landscape, which serves as an environment variable that causes the data and label. \citet{PAML} show examples of additional task descriptors, such as one image per robotics environment. 

\subsection{Limitations in Causal Identifiability and Unverifiable Assumptions}
\label{sec:causal_id_limitations}
One recurring challenge is the \emph{gap between idealized causal assumptions and what can be learned (or tested) from finite data.} For instance, many methods presented here rely on the style-and-content decomposition or independence assumptions between different parts of the data-generating process. In practice, these assumptions may fail, or the data may not exhibit enough variety (interventions or natural perturbations) to learn or refute a particular decomposition. Moreover, if there are unobserved confounders, the problem of distinguishing correlations from genuine causal paths may require additional forms of domain knowledge. Future work should make explicit which assumptions are essential and attempt to design benchmarks or protocols that test whether these assumptions hold.

\chapter{Causal Generative Modeling} \label{chapter:cgm}

\textbf{Important Note on Strong Assumptions.}  
Throughout this chapter, we discuss methods that aim to learn or leverage \emph{causal} generative models—enabling interventions or counterfactual modifications to the data. However, \emph{learning the entire structure of an underlying SCM typically depends on strong assumptions} (e.g., no hidden confounders, correct causal ordering, or the correct level of abstraction). These assumptions are often hard to verify empirically. As a result, while causal generative modeling can be powerful for interpretability or robust generation, the term ``causal'' here should be interpreted with the understanding that \emph{some combination of domain knowledge and partial assumptions} is usually needed. Without these, claims of fully identifying causal mechanisms must be treated with caution.

The goal of generative modeling is to produce samples that mimic the characteristics of our training data. The field of \emph{controllable generation} refers to techniques that allow us to enforce a set of attributes that novel samples should satisfy.
Non-causal controllable generation samples from a conditional distribution $p(\vx \mid \va)$ where $\rmA$ is an attribute specification \cite{conditionvae, song2020score}. Such observational distributions restrict us to sample properties as seen in the training data. However, we may want to i) generate samples with combinations of attributes unobserved in our training data and with downstream causal effects automatically incorporated or ii) edit samples using instructions that specify the attributes we want to change while keeping unrelated attributes fixed.

\emph{Causal Generative Modeling} (CGM) offers a causal perspective on controllable generation and sample editing by estimating interventional or counterfactual distributions. For controllable generation, given a causal representation $\rmZ$ of our data, we sample from $p(\vx \mid \doo(\va))$, where $\rmA \subset \rmZ$ is the set of attributes we wish to enforce. If we wish to edit a sample $\vx$ according to attributes $\va$, the target estimand becomes $p(\vx_{\rmA \rightarrow \va} \mid \vx)$.

\begin{figure}
    \centering
    \includegraphics[width=\columnwidth]{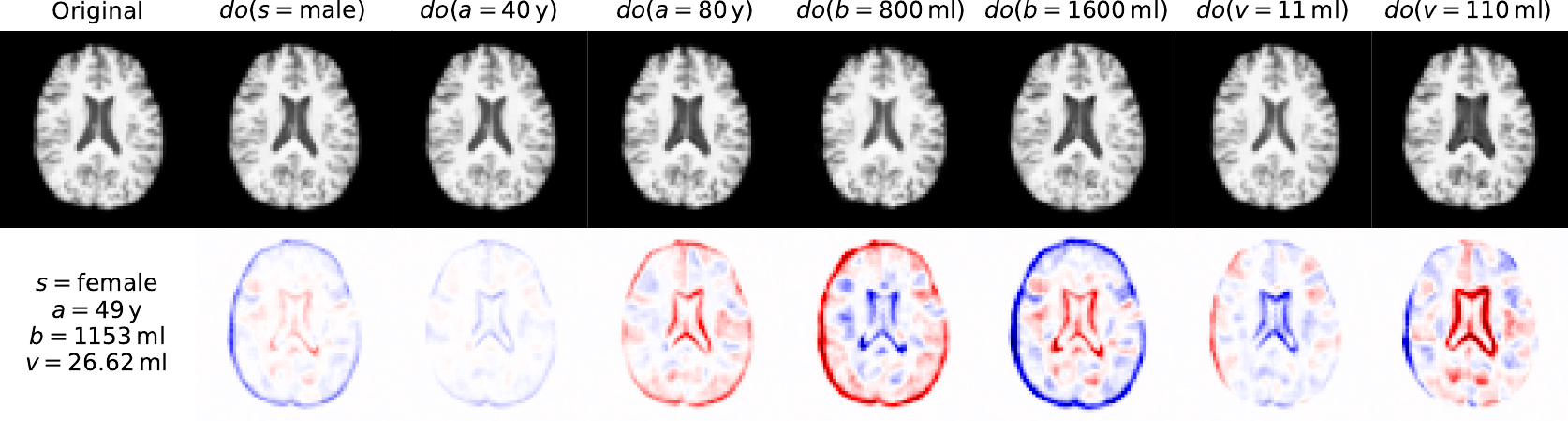}
    \caption{\textbf{Brain image counterfactual samples \citep{pawlowski2020deep}:} One way to study the effects of varying demographics on the brain structure is to generate counterfactual samples, as done here by \method{DeepSCM} (\Cref{crl:deepscm}). The intervention variables are a) age, s) sex, b) brain volume, and v) ventricle volume. \textbf{Top row:} Counterfactual samples with interventions specified for each column. \textbf{Bottom row: Difference maps.} }
    \label{fig:cf_brains}
\end{figure}

One unique use case of CGM is the \textbf{investigation of complex causal mechanisms}. For example, \citet{pawlowski2020deep} imagine a medical imaging setup in which we are interested in how a person's anatomy would change if particular traits were different. By incorporating expert domain knowledge in the form of a causal graph, counterfactual distributions enable us to model the appearance of brain MRI scans under a counterfactual scenario where the person's biological sex is different. Interpreting such counterfactual samples can reveal how demographic attributes (e.g., sex) might physically manifest in the brain—but \emph{only if} the assumed causal structure is accurate enough. \Cref{fig:cf_brains} shows some of the counterfactual samples produced by \citet{pawlowski2020deep}.

Another unique CGM use case is \textbf{counterfactual data augmentation}. Recall the example of training an image classifier in \Cref{fig:spurious_associations}. Given training data on cows commonly found in alpine pastures, the classifier fails to generalize to unfamiliar backgrounds, where cows are observed on the beach, for example. As a result, a classifier may learn a spurious association between grass backgrounds and the cow label in such an instance. One way to mitigate this spurious correlation (``grass''\(\leftrightarrow\)``cow'') is to augment the dataset by generating images of cows in \emph{unfamiliar} backgrounds. For example, \citet{sauer2021counterfactual} propose to learn a generative model that disentangles background and foreground attributes. Then, the model can generate images where cow images are intervened upon to have new backgrounds, such as a beach.

How can we learn such interventional and counterfactual distributions? Given that it is impossible to learn disentangled representations unsupervisedly from data without extensive domain knowledge \cite{locatello2019challenging, khemakhem2020variational}, we explore methods in this section that vary according to the supervision and domain knowledge they require. On the one hand, we discuss techniques demanding some domain knowledge encoded in an underlying causal graph in \emph{structural assignment learning} (\Cref{sec:requires_causal_dag}). On the other hand, methods exist that relax this requirement, which we refer to as \emph{causal disentanglement} (\Cref{sec:does_not_require_causal_dag}).

\notation
\begin{nota}
\rmZ & \text{ Generative variables } \\
K & \text{ $\text{dim}(\rmZ)$ } \\
\rmA & \text{ Attributes / intervention variables} \\
I & \text{ Index set of intervention variables s.t. $\rmZ_{I} = \rmA$ } \\
P & \text{ $\text{dim}(I)$ } \\
\ex & \text{ Independent exogenous causal parents } \\
\rmG & \text{ Graph adjacency matrix } \\
\gF & \text{ Set of structural assignments of an SCM }
\end{nota}

\section{Structural Assignment Learning}
\label{sec:requires_causal_dag}

\begin{table}[h]
    \centering
    \begin{minipage}{\textwidth}
        \centering
        \resizebox{\linewidth}{!}{%
    \begin{tabular}{|P{2.8cm}|P{9cm}|P{1.5cm}|}
        \toprule
        \bf Method & \bf Key Idea & \bf Ref.  \\
        \midrule
        \method{DeepSCM} \cite{pawlowski2020deep} & Learn each structural assignment independently & \Cref{crl:deepscm} \\
        \hline
        \method{VACA} \cite{vaca} & Learn structural assignments jointly with a GNN & \Cref{crl:vaca} \\
        \hline
        \method{DCEVAE} \cite{disCEVAE} & Group structural assignments in the SCM with latent variable modeling & \Cref{sec:dcevae_paragraph} \\
        \hline
        \method{DEAR} \cite{shen2020disentangled} & Learn all structural assignments in the SCM with latent variable modeling & \Cref{sec:dear} \\
        \bottomrule
    \end{tabular}
    }
    \caption{\textbf{Overview of Structural Assignment Learning methods}.}
    \label{tab:SAL_methods}
    \end{minipage}
\end{table}





Here, we typically assume that the full DAG over \(\rmZ\) is known or well-approximated. For instance, \method{DeepSCM} \citep{pawlowski2020deep} uses a small set of variables (often < 5) so that each structural equation \(z_i = f_i(\mathrm{pa}(z_i), \epsilon_i)\) can be learned. In more complex data domains like large-scale medical imaging with dozens of relevant variables, the approach can become intractable unless strong domain knowledge is used to limit the graph. We emphasize that if unobserved confounders exist, the learned structure is only partial. 

Specifically, the practitioner must have access to data of the form $\{(\vx_i, \vz_i) \}_{i=1}^n$, where $\rmZ$ denotes the causal variables of interest. The practitioner also designs a causal DAG $\gG$ of the underlying SCM, with each variable $Z_i \in \rmZ$ represented as an individual node. The methods in the following section aim to learn: i) an invertible mapping $g$, $g:\gX \rightarrow \gZ \times \epsilon$, and ii) a set of structural assignments $\gF = \{ f_i: z_i = f_i(\pa(z_i), \epsilon_i; i = 1,..,K\}$, which combine with $\gG$ to form an SCM.

Importantly, these methods discover SCMs among the variables $\rmZ$ only. Causal variables not included in the labels $\rmZ$ will be hidden confounders of the inferred SCMs.

\subsection{Independently Learned Structural Assignments} \label{crl:deepscm} \citet{pawlowski2020deep} introduce \method{DeepSCM}, a principled approach for estimating interventional and counterfactual distributions given the underlying causal DAG of the data-generating process. They propose to instantiate its SCM by learning the functional assignments $f_i$ for each variable $z_i := f_i(\epsilon_i ; \pa(z_i) )$ from its parents $\pa(z_i)$ and mutually independent noise terms $\epsilon_i$, using normalizing flows \cite{normalizing_flows_review} and variational inference \cite{original_vae}.

To estimate a given sample's counterfactuals, we need access to its exogenous noise values. Obtaining access to the noise values observed in a given sample is called the abduction step, as we covered in \Cref{def:cf}. To perform this step, \citet{pawlowski2020deep} propose to learn a mapping for each observed $z_i$ to its respective noise term, $\epsilon_i = f_i^{-1}(z_i; \pa(z_i))$ through normalizing flows in the low dimensional setting, and through variational inference in the high dimensional setting.
After learning this mapping, one can perform a counterfactual query by modifying the variables of choice in the causal graph and evaluating a prediction from the SCM with fixed noise values.

\Cref{fig:cf_brains} highlights some examples of the counterfactual samples produced. We note that this method has been applied with causal graphs of order 5. Scaling up this approach to larger causal graphs is an open research problem.


\subsection{Structural Assignments with a GNN}
\label{crl:vaca}
\citet{vaca} develop on the ideas of \method{DeepSCM} (\Cref{crl:deepscm}) and investigate how Graph Neural Networks (GNNs) can be used to solve the same task. Unlike DeepSCM, their approach simultaneously learns all (potentially nonlinear) structural assignments during training. They refer to their approach as \method{VACA}(VAriational Causal graph Autoencoder).

They express the causal graph as an adjacency matrix $\rmG$ embedded in each layer of the GNN. Both the encoder $p_{\params}(\vw \mid \vz, \rmG)$ and decoder $p_{\params}(\vz \mid \vw, \rmG)$ are GNNs that take $\rmG$ as input. They consider $\rmZ$ as representing the set of endogenous causal variables and identify each latent variable $W_i$ as capturing all the information about $Z_i$ that cannot be explained by $\pa(Z_i)$. Note that $\rmW$ does not necessarily correspond to the exogenous variables $\ex$, and $p(\vw) \neq p(\ex)$.

One limitation of this model is its expressivity w.r.t. the size of the underlying causal graph. \method{VACA} captures causal interventions if and only if the number of hidden layers in its decoder is greater than or equal to $\gamma-1$, with $\gamma$ being the length of the longest path between any two endogenous nodes in the true causal graph. As GNNs' performance tends to decrease sharply with depth due to over-smoothing \cite{oversmoothing, molecule_ssl}, \method{VACA} will struggle to perform well over large causal graphs.

\subsection{Group Structural Assignments}
\label{sec:dcevae_paragraph}
\begin{figure}[h]
    \centering
    \begin{subfigure}[t]{0.5\columnwidth}
    \includegraphics[width=\linewidth]{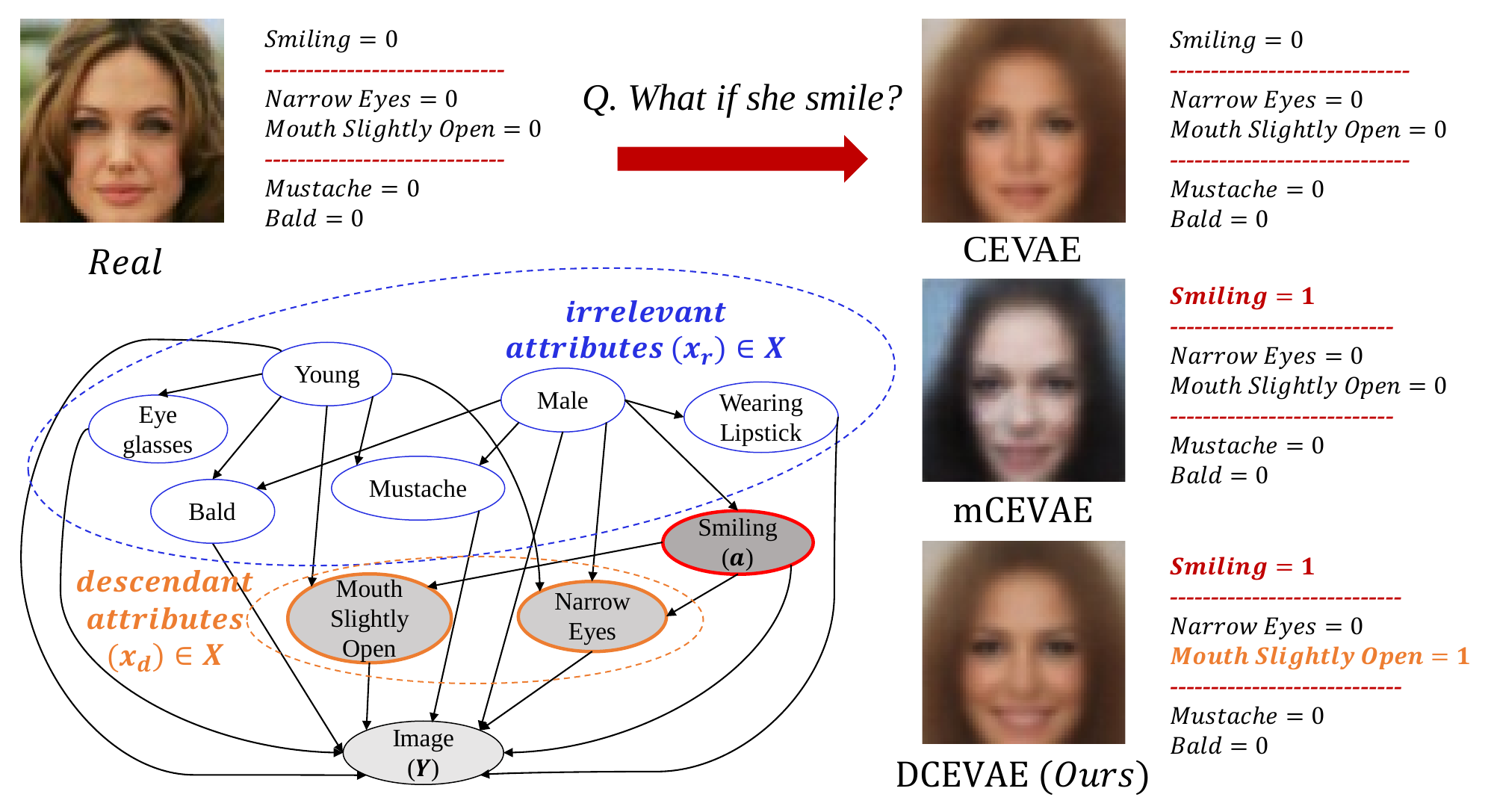}
  \caption{\textbf{Smiling Example}: A counterfactual image for $a_{\textit{Smiling}}=0$ should be labeled as $a_{\textit{Smiling}}=1$, and such change may cause downstream causal effects on descendant attributes of $a$, $x_d$ (i.e. \textit{Mouth Slightly Open}, \textit{Narrow Eyes}) by maintaining the other attributes intact, $x_r$. Prior works fail to maintain the irrelevant attributes of $a$.}
    \end{subfigure}    \begin{subfigure}[t]{0.48\columnwidth}
    \includegraphics[width = \columnwidth]{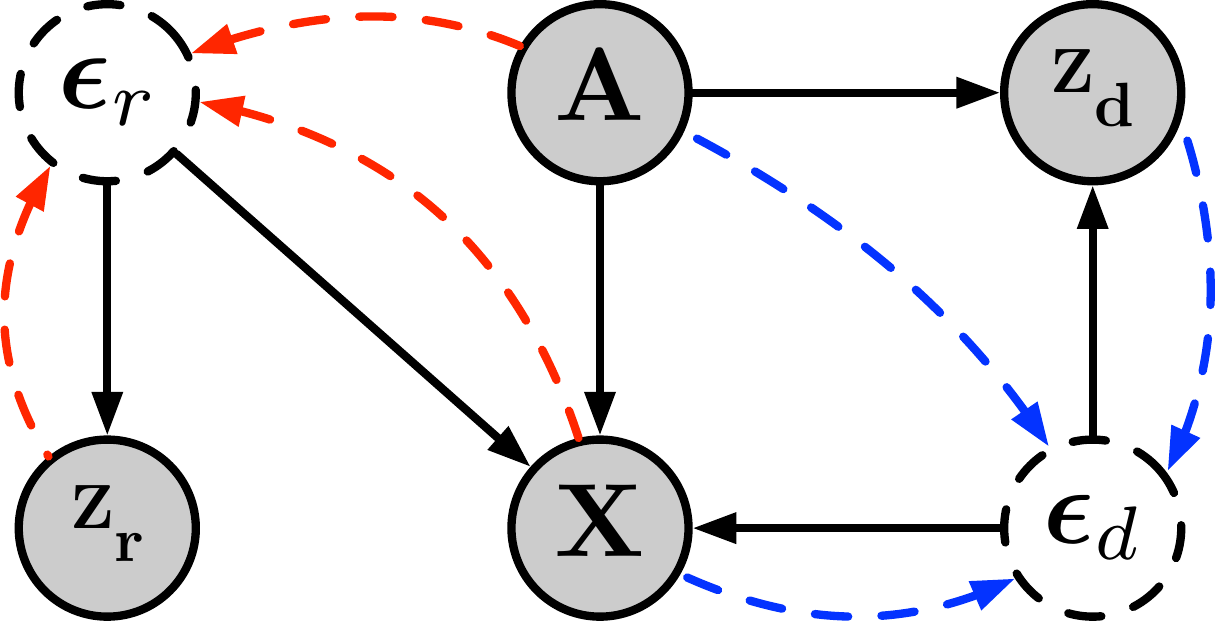}
    \caption{\textbf{Causal DAG}: $\rmA$ is one segment, its causal descendants another, and the last segment for all other variables. $\mX$ is generated from latent variables $\ex_r$, $\ex_d$ and attributes $\mA$. The colored arrows indicate latent variable modeling, where we learn $p(\ex_r \mid \vz_r, \va, \vx)$ and $p(\ex_d \mid \vz_d, \vz, \vx)$.}
    \end{subfigure}
    \caption{\textbf{Disentangled Causal Effect Variational Autoencoder (DCEVAE) \cite{disCEVAE}}, see \Cref{sec:dcevae_paragraph}.}
    \label{fig:dcevae}

\end{figure}

\citet{disCEVAE} develop a generative model, \method{DCEVAE}, that produces counterfactual samples by segmenting the causal DAG and learning three different segments (see \Cref{fig:dcevae}). These segments must depend upon the interventions we want to perform. They propose to cluster the causal graph based on which features undergo interventions, which may carry two benefits: i) it can reduce the problem of error propagation along Markov factorizations that \method{VACA} (\Cref{crl:vaca}) addressed, and ii) it allows for counterfactual sampling over higher-order causal graphs than the previous methods.

This method outputs an image and takes observed concept labels $\rmZ$, such as \emph{mustache} or \emph{gender} in CelebA data \cite{liu2015faceattributes}, as inputs. Given interventions on $\rmA \subset \rmZ$, the new interventional distribution $p(\vz^I \mid \doo(\va))$ is inferred, and $\vx^I$ is generated from $\vz^I$. The practitioner specifies which concept labels they want to intervene on, $\rmA$, and the causal descendants of the intervened variables $\rmZ_d$. The remaining attributes are $\rmZ_r = \rmZ \setminus \{\rmA \cup \rmZ_d \}$. The exogenous variables are split too, so $\ex_d, \ex_r$ are the exogenous causal parents of $\rmZ_d, \rmZ_r$, respectively. Given a sample $\vx$, the model produces a counterfactual sample $\vx_{\va}$ by identifying the generative variables $\vz$ and the sample exogenous factors $\ex_d$ and $\ex_r$.

The learning task is to estimate the encoder that maps the observed causal variables $\vz$ to the exogenous variables, $p_{\params}(\ex_r, \ex_d \mid \va, \vz_d, \vz_r, \vx)$ and the decoder $p_{\params}(\vz_d, \vz_r, \vx, \ex_d, \ex_r \mid \va)$ which generates the counterfactual sample. Due to their disentanglement of $\rmZ = \rmA \cup \rmZ_d \cup \rmZ_d$ and $\ex = \ex_d \cup \ex_r$, the encoders and decoders factorize neatly.

 \citet{disCEVAE} applied their method to the CelebA dataset \cite{liu2015faceattributes}and showed that it improves upon causal agnostic models in counterfactual sampling with downstream effects of interventions accounted for. However, the images they show are often blurry.

\subsection{Structural Assignments with a GAN}
\label{sec:dear}
Similarly to \method{DCEVAE}, \citet{shen2020disentangled} propose a latent variable model for counterfactual sampling called \method{DEAR} (Disentangled gEnerative cAusal Representation). In contrast to \method{DCEVAE}, they propose to learn all of the structural assignments among the latent variables. They employ a GAN \cite{goodfellow2014generative}, which removes the constraint of a Gaussian prior in VAE-based methods. The practitioner must have access to the causal ordering of the variables of interest, and labels for the causal variables in each sample, as found in the CelebA dataset \cite{liu2015faceattributes}. Given access to the causal ordering, a supergraph (i.e., a graph containing all subgraphs, akin to a superset) of the adjacency matrix $\rmG$ is initialized in the prior distribution $p_{\bbeta}(\vz)$, from which they identify $\rmG$ during training.

\begin{figure}
    \centering
    \includegraphics[width=\columnwidth]{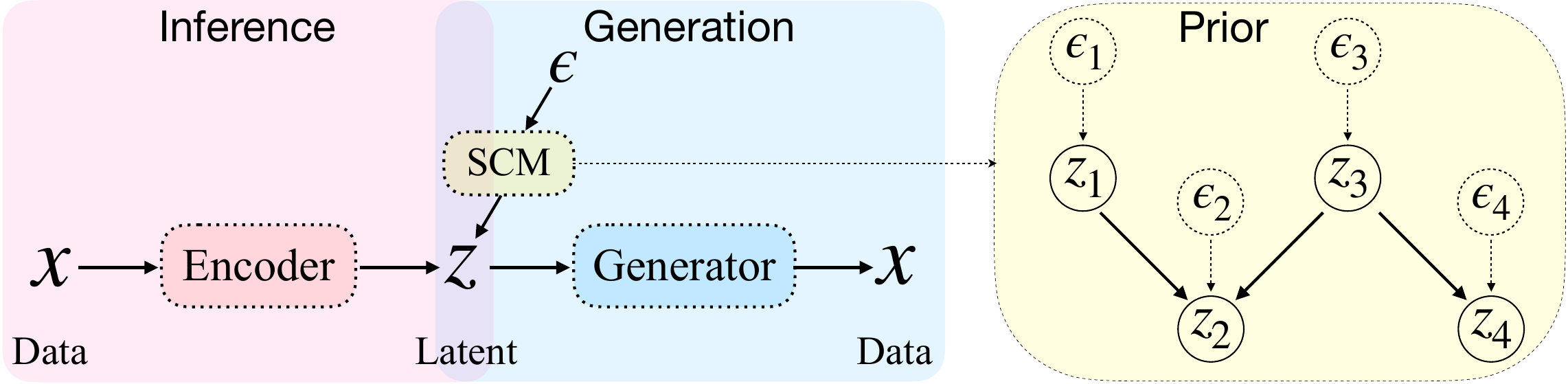}
    \caption{\textbf{\method{DEAR} \cite{shen2020disentangled}}: the prior $p_{\bbeta}(\vz)$ encodes the SCM among latents $\rmZ$.}
    \label{fig:dear}
\end{figure}

The authors propose to learn the generative model $p_{\params}(\vx \mid \vz)p_{\bbeta}(\vz)$ and the encoder $q_{\bphi}(\vz \mid \vx)$ with parameters $\params, \bbeta$ and $\bphi$, and a discriminator $D_{\bphi}$ through solving a bi-level optimization objective. For given exogenous factors $\ex$, $\vz$ is generated from the prior $p_{\bbeta}(\vz)$ by a function $F_{\bbeta}(\ex)$ of the form
\begin{align} \label{eq:dear_prior}
    \vz = F_{\bbeta}(\ex) := f\left( (\rmI - \rmG^\top)^{-1}h(\ex) \right),
\end{align}
where $\bbeta = \{f,h,\rmG \}$. They specify $f$ as invertible, enabling a modification of \Cref{eq:dear_prior} permitting simulation of interventions. Also, they enforce the encoder $q(\vz \mid \vx)$ to satisfy alignment between sample labels, which indicate causal variables of interest, and the latent variables encoded.

\section{Diffusion-Based Counterfactual Estimation}
\label{sec:diff_scm}
\begin{figure}[t!]
\centering
\begin{subfigure}[t]{0.25\columnwidth}
    \centering
    \includegraphics[width=\columnwidth]{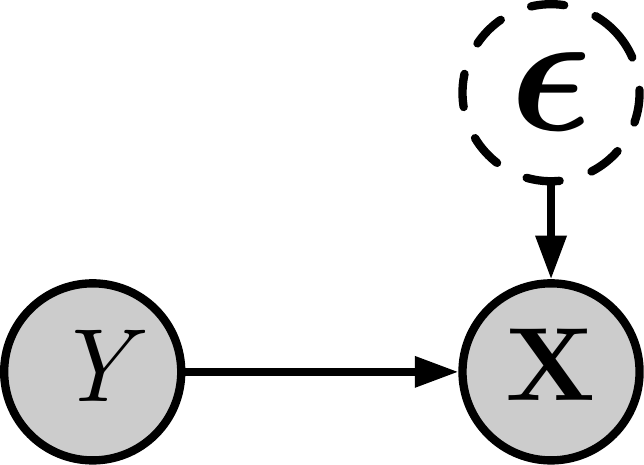}
    \caption{\textbf{Causal Graph}.}
    \label{fig:bivariate_causal_graph}
\end{subfigure}
\begin{subfigure}[t]{0.74\columnwidth}
    \centering
    \includegraphics[width=\columnwidth]{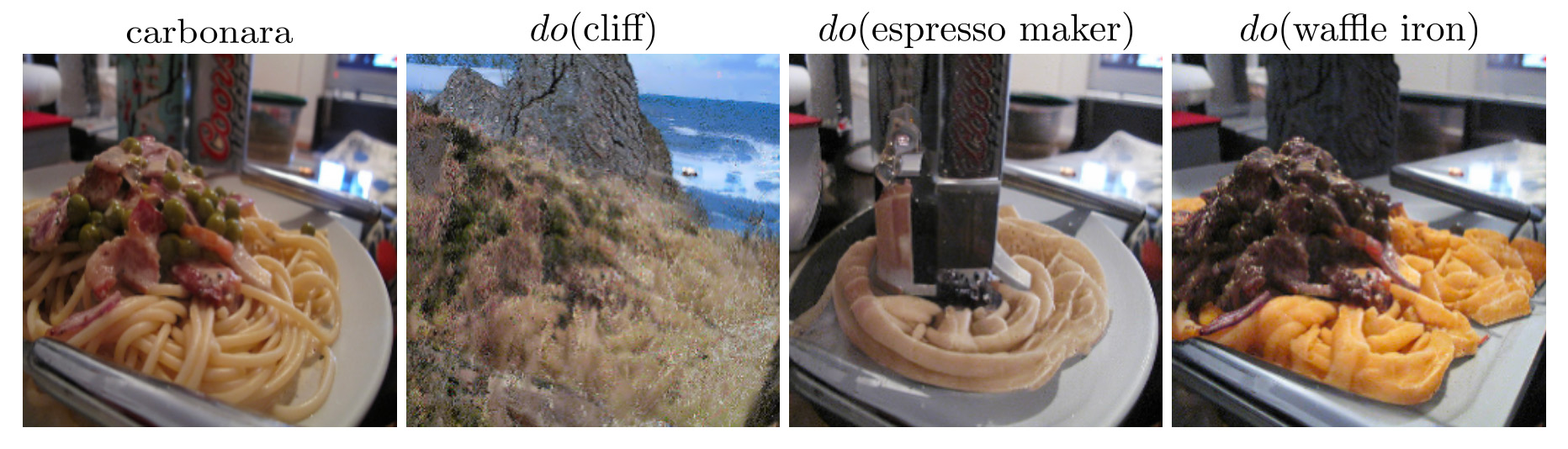}
    \caption{\textbf{Counterfactual samples.} The exogenous factors $\vu$ behind the 'carbonara' sample $\vx$ with interventions $y$.}
    \label{fig:diffusion_CF}
\end{subfigure}
\caption{\textbf{Diffusion-SCMs \cite{sanchez2022diffusion}}: It let us generate counterfactuals $\vx_y$ by intervening on $Y$ and inferring abducted noise $\vu$. The diffusion process encodes for the exogenous factors $\ex$, and the classifier guidance simulates interventions on $Y$. See \Cref{sec:diff_scm}.}
\end{figure}

Diffusion models have recently emerged as a highly effective generative modeling framework for images \cite{ho2020denoising}, and the classifier guidance framework proposed by \citet{dhariwal2021diffusion} permits conditional image generation.
\citet{sanchez2022diffusion} propose \method{Diff-SCM}, a framework for counterfactual sampling that employs Denoising Diffusion Models \cite{ho2020denoising, song2020score, dhariwal2021diffusion}. They implement a bi-variable causal model, where $Y \rightarrow \rmX$ as in \Cref{fig:bivariate_causal_graph}.

Following the score-based framework of \citet{song2020score} and the classifier guidance framework of \citet{dhariwal2021diffusion}, \method{Diff-SCM} uses an \emph{anti-causal predictor} as a means of classifier guidance. The anti-causal predictor steers the generation towards the counterfactual distribution, and the extent to which the guidance is balanced with the score-matching objective is managed by a parameter $s$. \citet{sanchez2022diffusion} view the forward diffusion as the encoding for the exogenous variables $\ex$ (abduction step) and view classifier guidance to simulate an intervention in the generation process.

In their work, they simulate interventions on the class label $Y$, thus producing counterfactual samples which share some qualitatively high-level similarities yet differ in classification label (see \Cref{fig:diffusion_CF}).

\section{Disentangled Mechanisms via Domain Knowledge} 
\label{sec:cgn}
\begin{figure}[t!]
    \centering
    \begin{subfigure}[t]{0.3\textwidth}
    \centering
    \includegraphics[width=\textwidth]{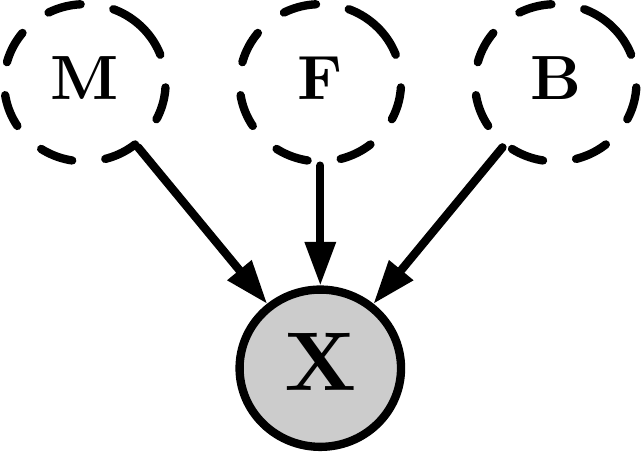}
    \caption{\textbf{Causal DAG} with mask $\rmM$, foreground $\rmF$, and background $\rmB$.}
    \label{fig:cgn_dag}
    \end{subfigure}
    \begin{subfigure}[t]{0.6\textwidth}
    \includegraphics[width=\columnwidth]{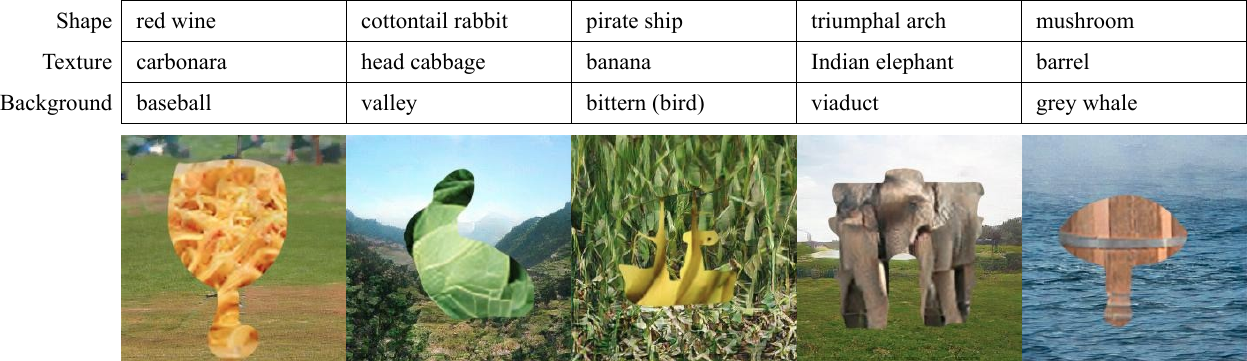}
    \caption{\textbf{ImageNet Counterfactuals.} CGN learns the disentangled FoVs and enables the generation of permutations thereof.}
    \end{subfigure}
    \caption{\textbf{Counterfactual Generative Networks \cite{sauer2021counterfactual}:} The factors of variation (FoVs) are object shape, object background and object texture (see \Cref{sec:cgn}).}
    \label{fig:cf_gen_nets}
\end{figure}

\citet{sauer2021counterfactual} propose \emph{Counterfactual Generative Networks} (\method{CGN}), a generative network motivated by invariant causal mechanisms (see \cref{sec:iml}). The authors employ three parallel \method{BigGAN} \cite{brock2018large} mechanisms that identify and modify different factors of variation (FoV): object shape, object texture, and background. The mechanisms are trained to specialize in one FoV and are composed through a composition module C, illustrated in \Cref{fig:cf_gen_nets}.

CGN relies upon the independence of the given FoVs, thus assuming a simple causal graph where each FoV causes $\rmX$, as shown in \Cref{fig:cgn_dag}. CGN takes noise vector $\ex$ and attributes $\rmA$ as input and outputs either an intervention sample (where the noise values $\ex$ vary between the mechanisms for given attributes $\va$) or a counterfactual sample (where the noise values and attributes are shared for each mechanism).

\section{Disentangled Mechanisms via Intervention Data}
\label{sec:weakly_sup_dis_wout_comp}
Similar to the goal of \method{CGN}, \citet{locatello2020weakly} propose \method{Ada-GVAE}, a generative model that identifies independent representations. This method demands two data conditions: (i) the data is a collection of paired samples before and after an unspecified intervention, and (ii) we know the number of intervened latent generative variables. Specifically, we have a collection of tuples $\{(\vx_i, \vx^I_i, P )\}_{i=1}^n$ where $P$ is the number of intervened generative variables, and $\vx^I$ denotes a sample after an intervention. In contrast to \method{CGN}, they do not require specification of the latent variables of interest, such as \emph{object shape}, but instead sample pairs and latent dimension specification.

For latent generative variables $\rmZ$, \citet{locatello2020weakly} assume that they are independent and admit a prior factorization $p(\vz)=\prod_{i=1}^{K} p\left(z_{i}\right)$ (thus a similar causal graph to that assumed in \Cref{fig:cgn_dag}). For $I \subset [K]$ the index set of intervened generative variables where $|I|=P$, and $\rmZ_{I} := \rmA$ the intervened generative variables, we have
\begin{align*}
    p(\vx, \vx^I, \vz, \vz_{I}, I) = p(\vx \mid \vz) p(\vx^I \mid \vz_{\bar{I}}, \vz_I) p(\vz) p(\vz_I) p(I).
\end{align*}
To identify the intervened variables $\rmZ_I$, the unintervened variables $\rmZ_{\bar{I}}$ are chosen to be the $K-P$ variables that induce the smallest values of $\KL(p_{\params}(z_i \mid \vx) \| p_{\params}(z_i \mid \vx^I))$. After identification, the posterior distributions for unintervened variables are set to equality between the sample pairs, while the posteriors for the intervened variables are left untouched. 

\section{Causal Disentanglement}
\label{sec:does_not_require_causal_dag}

\begin{table}[h]
    \centering
    \begin{minipage}{\textwidth}
        \centering
        \resizebox{\linewidth}{!}{%
    \begin{tabular}{|P{3.2cm}|P{7.8cm}|P{1.5cm}|}
        \toprule
        \bf Method & \bf Data Requirement & \bf Ref. \\
        \midrule
        \method{CausalVAE} \cite{causal_vae} & Causal variable labels & \Cref{sec:causal_vae} \\
        \hline
        \method{ILCM} \cite{brehmer2022weakly} &  Intervention sample pairs & \Cref{sec:weakly_supervised_caus_dis} \\
        \hline
        \method{CITRIS} \cite{lippe2022citris} & Sequential data with intervention target labels & \Cref{sec:citris} \\
        \bottomrule
    \end{tabular}
    }
    \caption{\textbf{Causal Disentanglement Methods.} Recent work on causal disentanglement and the data required by each.}
    \label{tab:CD_table}
    \end{minipage}
\end{table}

Next, we study methods that do not require the specification of any underlying causal graph. Instead, they identify both the underlying graph and the structural assignments between the variables, thus learning a set of causally disentangled representations \cite{statistical_to_causal_learning, scholkopf2021towards}.

\begin{mydef}{Causal Disentanglement \cite{scholkopf2021towards}}{causal_disentanglement}
We say a set of representations $\rmZ$, s.t. $\rmX = g(\rmZ)$ for some mapping $g$, are \emph{causally disentangled} if they permit the factorization
\begin{align}
     p(z_1,..,z_K) = \prod_{i=1}^K p(z_i \mid \pa(z_i)).
\end{align}
where $\pa(Z_i) \subset \{ Z_{j} \}_{j \neq i} \cup \epsilon_i$ and $\epsilon_i$ is the exogenous causal factor of $Z_i$.

\end{mydef}

These methods do not require access to the complete causal graph $\mathcal{G}$, instead requiring practitioner knowledge about the generative variables $\rmZ$ of interest. They learn how to reproduce observational distribution $p(\vx)$, interventional distribution $p(\vx \mid \doo(\va))$, and counterfactual distributions $p\left(\vx_{\rmA \rightarrow \va} \mid \vx \right)$.

\subsection{Exploiting Latent Labels}
\label{sec:causal_vae}

\citet{causal_vae} propose \method{CausalVAE}, which learns a causal model over latent variables from data and generates counterfactual samples. The dataset must contain labels on the latent causal variables of interest for each sample, following the identifiability framework outlined in \cite{khemakhem2020variational}. These labels represent the generative variables $\rmZ$.

Like \method{VACA} (\Cref{crl:vaca}), \method{CausalVAE} expresses the SCM as an adjacency matrix $\rmG$. However unlike \method{VACA}, \method{CausalVAE} learns $\rmG$ as well as linear structural assignments as in \Cref{eq:CausalVAE_SCM},
\begin{equation} \label{eq:CausalVAE_SCM}
    \vz=\rmG^{\top} \vz+\boldsymbol{\ex},
\end{equation} while \method{VACA} permits nonlinear structural assignments too.

We generate a counterfactual sample $\vx_{\va}$ according to attributes $\va$ as follows: for a given sample $\vx$, the encoder identifies the exogenous noise $\boldsymbol{\epsilon}$. The exogenous noise determines the causal latent variables $\vz$ via the SCM, \Cref{eq:CausalVAE_SCM}, to obtain $\vz = \left(\mathbf{I}-\mathbf{G}^{\top}\right)^{-1} \boldsymbol{\epsilon}$. Now that we have an encoding for $\vz$, we simulate an intervention on attributes $\va$. To obtain counterfactual $\vz_{\va}$, $\vz$ is input into \Cref{eq:CausalVAE_SCM} except with modifications on $\rmG$ that reflect the interventions $\va$. Then counterfactual $\vz_{\va}$ is input into the decoder to generate a counterfactual sample $\vx_{\va}$. The authors demonstrate the method's efficacy by applying it to the CelebA dataset \cite{liu2015faceattributes}. 

\subsection{Exploiting Intervention Samples}
\label{sec:weakly_supervised_caus_dis}
\begin{figure}
    \centering
    \includegraphics[width=\textwidth]{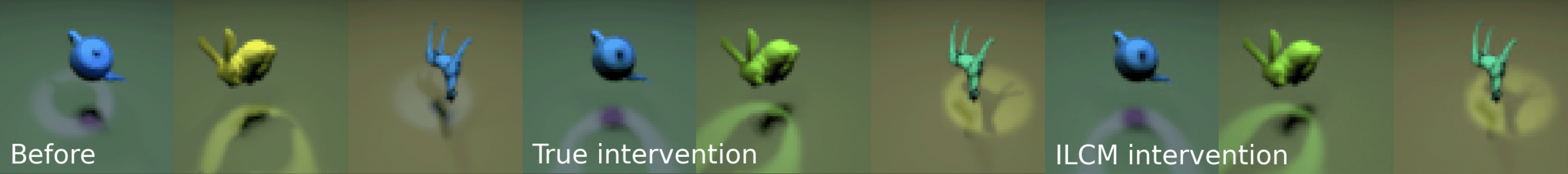}
    \caption{\textbf{Simulated interventions by \method{ILCM} \cite{brehmer2022weakly} compared to ground truth}: \method{ILCM} predicts the effects of a set of interventions $I$ on a set of causal variables $\{\vc_i \}_{i \in I}$, using the \emph{Causal3DIdent} dataset \cite{ssl_data_aug_scholkopf}. Given a sample $\rmX$, the model generates a simulation of the sample under intervention, as seen in the bottom row of the figure. See \Cref{sec:weakly_supervised_caus_dis} for explanation.}
    \label{fig:ILCM_images}
\end{figure}
\citet{brehmer2022weakly} improve upon \method{Ada-GVAE} by i) performing causal graph discovery and ii) removing the requirement that a practitioner has access to the number of intervened variables between sample pairs, proposing to learn a generative model over a collection of tuples $\{(\vx_i, \vx^I )\}_{i=1}^n$.

They propose two types of models: \emph{Explicit or Implicit Latent Causal Models}. For the former, \method{ELCM}, they introduce a prior over latent variables $p(\vz)$ that encodes the structure of a specified causal graph.

For the latter \method{ILCM}, they propose to learn a \emph{noise encoder} $p_{\params}(\ex, \ex^I \mid \vx, \vx^I)$ that maps the data $(\rmX, \rmX^I)$ to the exogenous noise values $(\ex, \ex^I)$ in the latent causal model, before and after interventions. No practitioner specifications of the underlying causal graph, the causal variables of interest, or the structural mechanisms are required. What \emph{is} required is the specification of the dimension of $\rmZ$, $K$, which determines the size of the causal graph. Further, \method{ILCM} relies upon the interventions $I$ only changing one noise value $\epsilon_i$, between $\ex$ and $\ex^I$. 

\method{ILCM} implicitly learns the generative variables $\rmZ$ since the exogenous noise values determine the variables in an SCM. To obtain the generative variables in explicit form, they learn \emph{solution functions} $s(.,.)$ s.t. $z_i = s(\epsilon_i, \ex_{-i})$. A causal graph is learned by applying intervention-based causal discovery algorithms over the generative variables $\rmZ$.


\subsubsection*{Exploit sequential data with intervention labels}
\label{sec:citris}
While \method{ILCM} exploits independence of exogenous noise values $\epsilon_i$ for causal graph identification, \citet{lippe2022citris} propose to exploit independence of the time indexed causal variables, conditioned on the previous time step and assuming no instantaneous effects. Their method, called \emph{Causal Identifiability from TempoRal Intervened Sequences} \method{CITRIS}, exploits data from a sequence $\{ (\vx^t, I^t) \}_{t=0}^T$, where each intervention target $I^t \in \{0,1\}^K$ labels the causal variable intervened upon to generate sample $\vx^t$. Given such data, \method{CITRIS} learns causal latent variables $\rmZ$ and a transition prior $p_{\bbeta}(\vz^{t+1} \mid \vz^t, I^{t+1} )$, which encodes the governing SCM and is trained using an ELBO objective.

\begin{figure}[h]
    \centering
    \begin{subfigure}[t]{0.3\textwidth}
    \centering
    \includegraphics[height=3.65cm]{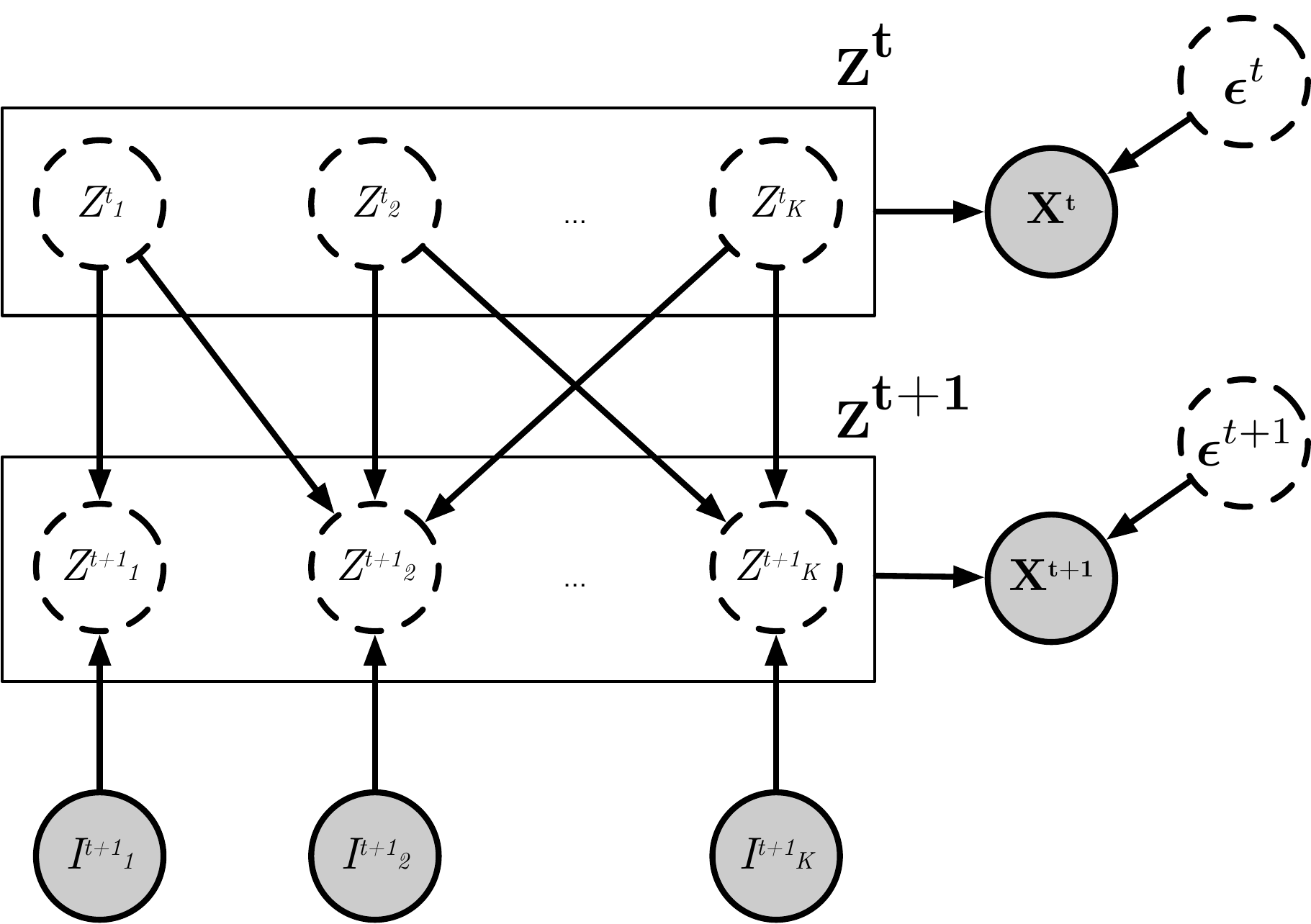}
    \caption{\textbf{Causal DAG} with latent variables $Z_i$, exogenous factors $\ex^t$, intervention targets $I^t_i$ and observations $\rmX^t$.}
    \label{fig:citris_dag}
    \end{subfigure}
    \begin{subfigure}[t]{0.6\textwidth}
    \centering
    \includegraphics[height=3.65cm]{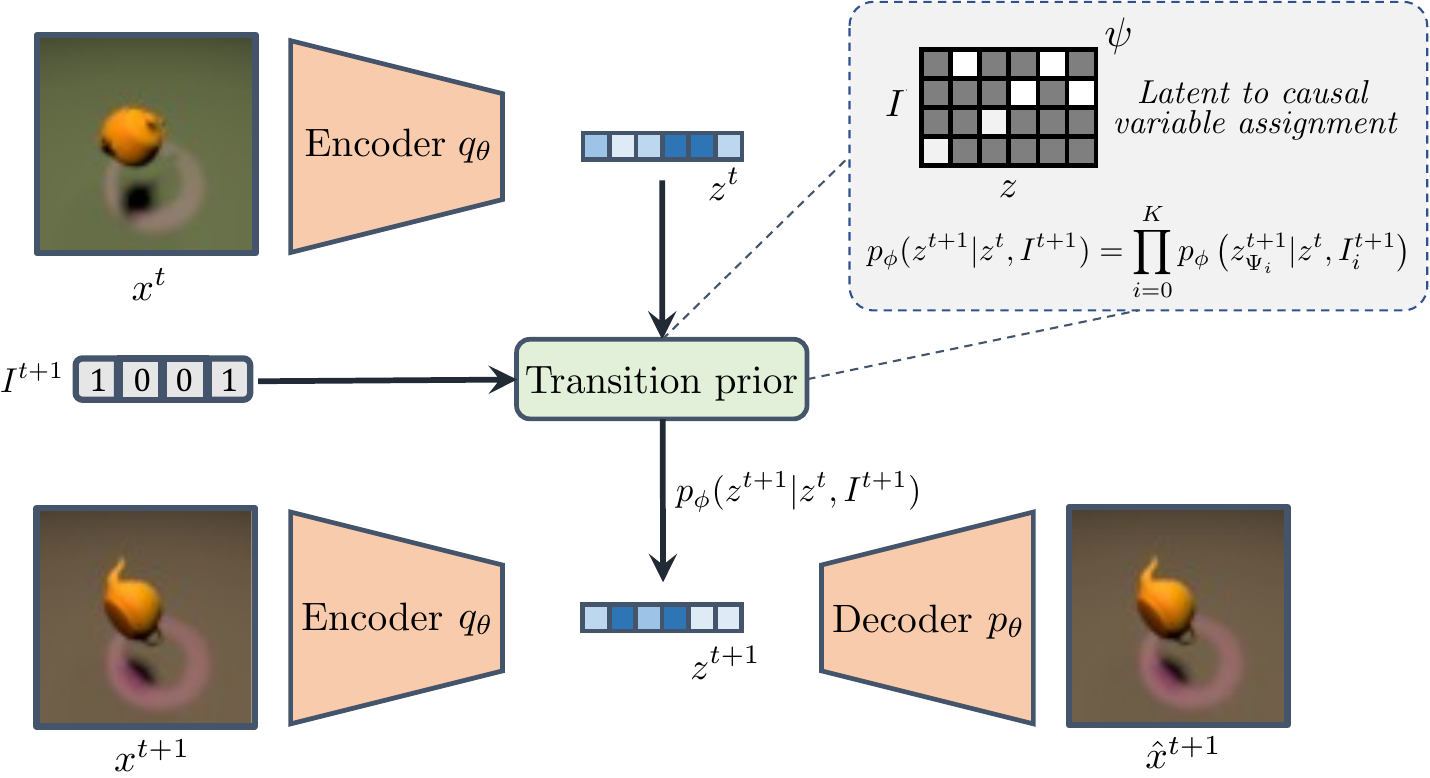}
    \caption{\textbf{Architecture}: the transition prior encodes the SCM between latents $\rmZ^t$ and $\rmZ^{t+1}$.}
    \end{subfigure}
    \caption{\textbf{CITRIS \cite{lippe2022citris}}: The transition prior $p(\vz^{t+1} \mid \vz^t, I^{t+1})$ captures the temporal causal relations from $\rmZ^t$ to $\rmZ^{t+1}$.}
    \label{fig:citris_fig}
\end{figure}

\method{CITRIS} also accommodates multi-dimensional causal variables, i.e., it groups latent variables of the same causal factor, such that the transition prior factorizes based on these multi-dimensional variables. For example, this property helps to learn one position variable to capture 3D coordinates instead of three independent axis variables because the ground truth axes are not always well-defined. 

\method{CITRIS} learns an assignment function $\psi: [\text{dim}(\rmZ)] \rightarrow [K]$ which factorizes the latent space into causal variables and exogenous noise, such that $\rmZ = \{Z_{\psi_1},..,Z_{\psi_K}, \ex \}$. We obtain the parents of any latent variable $Z_{\psi_{i}}$, and ultimately the causal graph, by identifying which of the latents $\{Z^t_{\psi_1},..,Z^t_{\psi_K} \}$ influences $Z^{t+1}_{\psi_{i}}$ in the transition prior.
Hence, the transition prior factorizes as
\begin{align}
    p_{\bbeta}\left(\vz^{t+1} \mid \vz^t, I^{t+1}\right) = p_{\bbeta}\left(\ex^{t+1} \mid \vz^{t}, I_{i}^{t+1}\right) \cdot \prod_{i=1}^{K} p_{\bbeta}\left(z_{\psi_{i}}^{t+1} \mid \vz^{t}, I_{i}^{t+1}\right).
\end{align}


Since \method{CITRIS} assumes no instantaneous effects in the data, it fails to discover the underlying causal graph when instantaneous effects are present. For instance, \cite{lippe2022icitris} highlight the example where both (i) flicking a switch, and (ii) its effect on a light bulb can occur within one time step of a sequence. In this case, both the variable \emph{light bulb} and \emph{switch} are not conditionally independent, given the previous time step information.

\citet{lippe2022icitris} propose to address this shortcoming with \method{iCITRIS} (\emph{instantaneous} CITRIS), employing a similar framework to \method{CITRIS}, and exploiting the same type of data. In contrast to \method{CITRIS}, which accommodated soft interventions, the authors assume that the interventions are perfect, rendering an intervened variable independent of its parents. The transition prior is augmented with instantaneous causal graph structure information, and the authors propose to learn the graph and generative model parameters by solving a bi-level optimization objective.

\section{Open Problems}

\subsection{Different Levels of Abstraction}
\label{sec:correct_abstraction}

Causal representation learning (CRL, \Cref{def:causal_rep_learning}) aims to find an SCM among latent generative factors $\rmZ$ of our data $\rmX$, where the generative factors are assumed to be \emph{causal variables}. This problem is ill-posed, however, unless we specify the correct level of abstraction in the SCM. For instance, we may choose to construct a causal model between each pixel in an image ($\text{dim}(\rmZ) = \text{dim}(\rmX)$) or between three types of possible \emph{objects} that may be observed in an image ($3 = \text{dim}(\rmZ) \ll \text{dim}(\rmX)$). Both models may deliver equal performance (given enough data and compute), yet the latter is more efficient and interpretable for a human.

It is not obvious how practitioners should determine a reasonable level of abstraction for causal representation learning, much less do so automatically \cite{chalupka2017causal, beckers2019abstracting, beckers2020approximate, kinney2020causal}. All of the approaches in \Cref{sec:does_not_require_causal_dag} required specifications that determined the abstraction of the SCM, for example. \method{CausalVAE} (\Cref{sec:causal_vae}) required variable labels in the data, while \method{ILCM} (\Cref{sec:weakly_supervised_caus_dis}) specified $\text{dim}(\rmZ)$ and required samples pairs before and after interventions. Can we identify other sensible specifications or even relax them? For instance, \method{ILCM} relaxed the specifications of \method{Ada-GVAE} (\Cref{sec:weakly_sup_dis_wout_comp}).

Important theoretical results determine the limits and the opportunities of how much we can automate causal representation learning. \citet{locatello2019challenging} demonstrate that unsupervised disentangled representation learning is ill-posed without the proper model inductive biases and some extra supervisory information. \citet{khemakhem2020variational} prove that it is possible to identify the underlying joint distribution between observed and latent variables, as long as a practitioner specifies a factorized prior of the latent variables and provides additional supervisory signals. \citet{gresele2020incomplete} derive identifiability results for identifying independent generative variables when considered from multiple environments, even under nonlinear mixing in the generative process. Finally, \citet{cohen2022towards} identifies that disentangled representations typically permit impossible combinations of causal variables and proposes a framework for interventions, incorporating relations between causal variables that govern what is physically possible.

Thus, what kinds of supervisory signals are optimal for CRL, what trade-offs exist, and which signals are reasonable to require, are all open problems.

\subsection{Scaling Structural Assignment Learning}
\label{sec:scale_SAL}

The structural assignment learning methods reviewed in \Cref{sec:requires_causal_dag} generate compelling counterfactual samples. However, they either operate on relatively low-order causal graphs or amortize the structural assignments and focus on a set of plausible interventions.

In applied sciences, we often want to model higher-order causal graphs and flexibly generate counterfactual samples from them to capture real-life dynamics, such as in cell biology \cite{sachs2005causal, guelzim2002topological}. Yet, the approaches reviewed in \Cref{sec:requires_causal_dag} do not demonstrate results over causal graphs of order larger than 5, except for \method{DCEVAE} (\Cref{sec:dcevae_paragraph}), which amortizes the graph assignments. \method{VACA} (\Cref{crl:vaca}) relies on GNNs, which struggle to scale to large-scale graphs due to over-smoothing \cite{oversmoothing}. Hence, how to tackle the scalability of the underlying causal graphs in structural assignment learning remains an open problem.

\subsection{Understanding Counterfactual Data Augmentation}
\label{sec:cf_vs_noncf_open_prob}

A growing trend has been counterfactual data augmentation (CFDA), as we saw in \Cref{sec:da_methods} and will see later in \Cref{rl:cda,cv:cda,nlp:cda,grl:cda}.
Here, we focus on CFDA using causal generative models instead of hand-crafted transformations. While many augmentations can be hand-crafted, such as color transformations and rotations over images \cite{shorten2019survey}, we can also use generative models to create data augmentations.
For instance, \method{GenInt} (\Cref{sec:da_methods}) use a GAN \cite{brock2018large} to create data augmentations. 

However, while \method{GenInt} uses a causal agnostic generative model, it is possible to use causal generative modeling to generate CFDAs. We explored the causal perspective on data augmentation in \Cref{sec:data_aug}, where we viewed an augmentation as an intervention on a spurious data feature. \citet{sauer2021counterfactual} \Cref{sec:cgn} compare the efficacy of training a classifier on datasets augmented with 1) the counterfactual samples and 2) non-counterfactual samples generated by a GAN and find that counterfactual samples improved the classifier's performance more.

Exploring when and by how much CFDA using causal generative models improves generalization remains an open problem. For hand-crafted data augmentations, some understanding of their efficacy already exists. For example, \citet{data_augmentation_regularization} conclude that it is more effective than explicit regularization, such as weight decay or dropout. \citet{vit_sam} show that conventional data augmentation flattens the loss surface and that it can achieve similar performance gains as flat-minima optimizer \cite{flat_minima_optimizer}. \citet{cubuk2018autoaugment,cubuk2020randaugment} offer automated augmentation generators. For CFDA, works and results like those mentioned above are missing.

Further, we may use CFDA as an evaluation proxy for comparing causal generative methods by their capability of improving an external predictor's generalization performance. For a fixed prediction model and training dataset, methods in this chapter can be compared to the prediction algorithm test loss when one trains the model using data augmentations generated by the counterfactual sampling method. Such a benchmarking method is under-explored.


\newcommand{\xf}{\vx^{\mathrm{F}}}
\newcommand{\xce}{\vx^{\mathrm{CE}}}
\newcommand{\xscf}{\vx^{\mathrm{SCF}}}

\chapter{Causal Explanations}
\label{chapter:explanations}

The goal of \emph{AI explainability} (or \emph{explainable AI}) is to output \emph{explanations} that make the decisions that a model constructs understandable to humans and provide an answer to \emph{why} the output was predicted \cite{miller2019explanation}. From the point of view of \emph{model interpretability} \cite{lipton2018mythos}, which refers to the degree to which an observer can understand the cause of a model prediction, an explanation falls under the category of \emph{post-hoc} interpretability: the model (prediction) is analyzed after training, while \emph{intrinsic} interpretability may refer to models that are restricted in their complexity and therefore sacrifice predictive performance \cite{molnar2022}. Many explanation methods have been proposed, and we refer the reader to \cite{burkart2021survey, dovsilovic2018explainable, ce_survey, survey_ac} for excellent surveys.

In this section, we focus on two classes of explanation techniques: \emph{feature attribution} and \emph{contrastive explanations}. We will explain how causality enters the picture in the corresponding subsections. All discussed methods provide \emph{local} explanations of individual predictions based on a single input, and are \emph{model-agnostic}, i.e., they can be used across different model types. In contrast, there also exist methods providing \emph{global} explanations, which describe the model's average reliance on each feature across the whole dataset, and model-specific methods, such as saliency maps for neural networks trained on images \cite{simonyan2013deep, adebayo2018sanity}. These latter two are not covered in this survey.

To motivate the usefulness of explanations, let us consider the case of someone who applied for a loan and was rejected by a financial institution's loan distribution model. In most cases, the individual wants to understand the model's reasoning in a bid to strengthen their next application. Fixing this scenario as a running example, we will introduce multiple techniques to produce explanations in the upcoming subsections.

\begin{table*}[h]
    \centering
    \begin{minipage}{\textwidth}
        \centering
        \resizebox{\linewidth}{!}{%
    \begin{tabular}{|P{2.3cm}|P{2.5cm}|P{7cm}|P{1.5cm}|}
        \toprule
        \bf Class & \bf Method & \bf Key Idea & \bf Ref. \\
        \midrule
        \multirow{7}{*}{\shortstack{Feature\\Attributions}}& \method{CXPlain} & Quantify causal influences of input features to model accuracy & \Cref{explanations:cxplain} \\ \cline{2-4}
        & \method{GCE} & Learn latent factors which cause change in the output & \Cref{explanations:gce} \\  \cline{2-4}
         & \method{ASVs} & Relaxes asymmetry axiom of Shapley values w.r.t. given causal knowledge & \Cref{explanations:asv} \\  \cline{2-4}
        & \method{CSVs} & Intervenes upon features instead of conditioning on them  & \Cref{explanations:csv} \\ \midrule
        \multirow{5}{*}{\shortstack{Contrastive\\Explanations}} & Counterfactual Explanations & $\xce$ satisfies minimal distance to $\xf$ & \Cref{explanations:counterfactual}  \\ \cline{2-4}
        & Algorithmic Recourse via Minimal Interventions & $\xscf$ satisfies minimal cost set of interventions on $\xf$ & \Cref{explanations:ac}  \\
        \bottomrule
    \end{tabular}
    }
    \caption{\textbf{Method Overview} of Causal Explanations.}
    \label{tab:problem_overview_explanations}
    \end{minipage}
\end{table*}

\notation
\begin{align}
&\begin{array}{ll}
\xf & \text{ Factual individual with undesired outcome } \\
\xce & \text{ Counterfactual explanation individual with desired outcome } \\
\xscf & \text{ Structural counterfactual individual with desired outcome (w.r.t. causal graph) } \\
\end{array} \nonumber
\end{align}

\section{Feature Attribution Explanations} \label{explanations:fa}
Attribution-based explanations assign a ranking to features, representing the marginal contribution of each feature to the output of the model. In our running example, such explanations may show that for the individual whose loan application was denied, the most important feature was their overall income, while their credit card debt did not contribute to the model's prediction. While this explanation does not provide an instruction of what the individual should do to get the loan applied, it informs the applicant about which features of their application may have caused the rejection to which degree.

Regarding our running example, we illustrate why a loan applicant would prefer \emph{causal} feature attribution explanations over associational ones, as follows. Imagine that the financial institution primarily cares about the applicant's yearly income: the higher the income, the higher the chances of providing a loan. Other applicant features may be spuriously correlated with the income: e.g., their education level, job, industry, whether they are self-employed, etc. 

A traditional, non-causal attribution method may assign uniformly high scores across all these correlated features. However, if the applicant knew that only their income matters, they likely would focus on improving it more directly: instead of even considering a career change, working directly on getting a promotion at their current job might be more effective. Hence, in this scenario, a causal attribution method would help the applicant to achieve their desired outcome more efficiently.

\subsection{CXPlain} \label{explanations:cxplain}
\citet{cxplain} transform the task of producing feature importance estimates for an existing predictive model into one of supervised learning. They train separate supervised \emph{causal explanation} (\method{CXPlain}) models to explain the predictive model. To train the explanation model, they use a causal influence (\Cref{sec:causal_influence}) function that quantifies both the attribution of each input feature and groups of input features to the accuracy of the prediction model.

The causal feature attribution $a_i \in \R$ measures to what extent the $i$th input feature causally contributed to the predictive model's output $\hat Y$, as the decrease in error, measured by a standard classification loss $\gL$. In other words, $a_i$ denotes the causal influence of adding that feature to the set of input features. Pre-computing the feature attributionsfor $N$ training samples $\{\vx_i\}_{i=1}^N$ takes $N(D+1)$ evaluations of the target predictive model at training time, where $D$ is the number of input features. 

They also provide confidence interval estimates of the level of uncertainty associated with each feature importance estimate $\hat{a}_{i}$ produced by a \method{CXPlain} model. By using bootstrap ensemble methods, $M$ explanation models are trained to estimate uncertainty. Each model is trained on $N$ training samples from $\{\vx_i\}_{i=1}^N$, sampled randomly with replacement from the original training set.

\subsection{Generative Causal Explanations} \label{explanations:gce}
\begin{figure}
    \centering
    \includegraphics[width=\columnwidth]{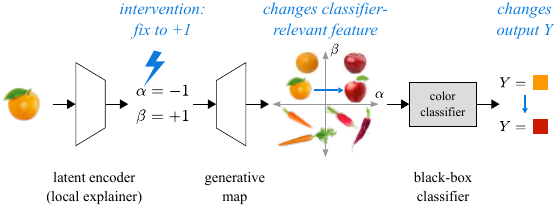}
    \caption{\textbf{Generative Causal Explanations \cite{causal_explanations}}: Changing learned latent factors produces a change in the classifier output statistics. To quantify the attribution of the latent factors, \citet{causal_explanations} use causal influence measures of variables in a SCM (\Cref{sec:causal_influence}).}
    \label{fig:gen_ce}
\end{figure}

\citet{causal_explanations} introduce the \emph{generative causal explanations} (\method{GCE}) framework for post-hoc explanations based on learned disentangled latent factors that produce a change in the classifier output distribution. For example, consider a black-box color classifier, for whose color classifications one wants to construct explanations. In this setup, one may learn a latent encoder (which they call the \emph{local explainer}) that learns a low-dimensional representation $(\alpha, \beta)$ describing the color and shape of inputs. Changing $\alpha$ (color) changes the output of the classifier, which detects the color of the data sample, while changing $\beta$ (shape) does not affect the classifier output. \Cref{fig:gen_ce} shows a pictorial depiction of this architecture.

To construct causal explanations, the authors propose two components: (i) a method for representing and moving within the distribution of the data and (ii) a metric that quantifies the causal influence (\Cref{sec:causal_influence}) of various aspects of the data on the classifier outputs. To ensure that the learned disentangled representations represent the data distribution, while encouraging a small subset of latent factors to exert a large causal impact on the classifier output, they formulate a corresponding optimization objective.

\subsection{Asymmetric Shapley Values} \label{explanations:asv}
\emph{Shapley values} provide a principled, model-agnostic way to explain model predictions. They are able to capture all the interactions between features that lead to a prediction by relying on coalitional game theory: the idea is to fairly distribute the ``payout'' among the features, which naturally quantifies which features contribute to a prediction \cite{lundberg2017unified, molnar2022}. They are principled because they uniquely satisfy four intuitive mathematical axioms.

However, \citet{asymmetric_shap_values} argue that Shapley values suffer from a significant limitation: they ignore all causal structure in the data. One of the axioms is \emph{Symmetry}: it places all features on equal footing in the model explanation, by requiring attributions to be equally distributed over features that are identically informative (i.e. redundant). \citet{asymmetric_shap_values} argue that when redundancies exist, we might instead seek a sparser explanation by relaxing this axiom. 

For example, two features might be bijectively related to one another if one is known to be the deterministic causal ancestor of the other. In this case, it makes sense to attribute all the importance to the ancestor and none to the descendant, in opposition to the Symmetry axiom. Otherwise, the explanations may obfuscate known causal relationships in the data. 

To relax the Symmetry axiom, the authors propose \emph{Asymmetric Shapley values} (ASVs), which uniquely satisfy the other axioms and reduce to Shapley values if the distribution over the ordering in which features are fed to the model is uniform. This relaxation allows the practitioner to place a non-uniform distribution over the ordering that incorporates causal understanding into the explanation. For example, one may place nonzero weight only on permutations in which an ancestor precedes its (known) descendants. In other words, only feature permutations that are consistent with the causal structure between the features have non-zero probability. This approach favors explanations in terms of distal (i.e. root) causes, rather than explanations towards proximate (i.e. immediate) causes. 

\citet{asymmetric_shap_values} discuss that ASVs span a continuum between maximally-data agnostic Shapley values, and causality-based explanation methods that often require the exact causal process underlying data. Hence, ASVs allow any knowledge about the data-generating process, however incomplete, to be incorporated into an explanation of its model, without the often-prohibitive requirement of full causal inference. For example, if causal knowledge is limited, even a single known causal ancestor can be ordered first, with permutations over remaining features uniformly weighted. 

\begin{mydef}{Distal distribution (ASVs) \cite{asymmetric_shap_values}}{asv}
Assume that there are $D$ input features, where $\Pi$ denotes the set of all permutations of them, and $\pi(j)<\pi(i)$ means that feature $j$ precedes feature $i$ under ordering $\pi$. Let $\Delta(\Pi)$ be the set of probability measures on $\Pi$, so that each $w \in \Delta(\Pi)$ is a map $w: \Pi \rightarrow[0,1]$ satisfying $\sum_{\pi \in \Pi} w(\pi)=1$. 

Asymmetric Shapley values replace the uniform distribution $w \in \Delta(\Pi)$ of Shapley values \cite{lundberg2017unified} with:
\begin{equation}
\begin{aligned}
    w_{\text {distal }}(\pi) \propto \begin{cases}1 & \text { if } \pi(i)<\pi(j) \text { for any known } \\ & \text { ancestor } i \text { of descendant } j \\ 0 & \text { otherwise }\end{cases}.
    \label{eq:asv}
\end{aligned}
\end{equation}
\end{mydef}

As a consequence of using \Cref{def:asv}, the ASVs of known causal ancestors indicate the effect these features have on predictions while their descendants remain unspecified. The ASVs of the descendants then represent their incremental effect upon specification. 

\subsection{Causal Shapley Values}
\label{explanations:csv}
\citet{janzing_fairness} argue that there is a misconception in previous work like \citet{lundberg2017unified} on Shapley values because these methods use observational conditional distributions rather than interventional ones. While these proposals are conceptually flawed, in practice, their software implementation still works because of its approximate nature.  

\citet{causal_shapley_values} remedy this conceptual flaw by proposing \emph{causal Shapley values} (CSVs) that explain the total effect of features on the prediction, taking into account their causal relationships. To incorporate causal knowledge, they replace the conventional conditioning by observation from conditional Shapley values with conditioning by intervention.

In contrast to ASVs, they point out that there is \emph{\say{no need to resort to asymmetric Shapley values to incorporate causal knowledge}}. Relaxing the Symmetry axiom is orthogonal to their approach. One can additionally get \emph{asymmetric causal Shapley values} that implement both ideas. As for ASVs, a practitioner needs to provide only a partial causal order and a way to interpret dependencies between features that are on an equal footing. 

One benefit of CSVs is that they permit a decomposition of the total effect that a feature has on a model's prediction into direct and indirect effects. The direct effect measures the expected change in prediction when the stochastic feature $X_{i}$ is fixed to $x_{i}$, without altering the other \say{out-of-coalition} features. The indirect effect measures the difference in expectation when the distribution of the other \say{out-of-coalition} features changes due to the additional intervention $\doo\left(X_{i}=x_{i}\right)$. 

\citet{true_to_model_or_data} argue that neither observational (\emph{\say{true to the data}}) nor interventional (\emph{\say{true to the model}}) conditional probabilities are preferable in general, but that the choice is application dependent. They present two real data examples from the domains of credit risk modeling and biological discovery to show how a different choice of value function performs better in each scenario, and how possible attributions are impacted by the choice of probability distribution.  

Similar to CSVs, \citet{pmlr-v162-jung22a} introduce $\doo$-Shapley values, which do not have the restriction that the model is \emph{accessible}, which means that it can be evaluated for arbitrary input features. Instead, their method is compatible with \emph{inaccessible} models, and uses semi-Markovian causal graphs (DAGs with bidirected edges).

\section{Contrastive Explanations}
\label{explanation:contrastive_explanations}
Sociological studies have shown that human explanations are typically \emph{contrastive}: they emphasize the causal factors that explain why an event occurred \emph{instead} of another event \cite{lipton1990contrastive}. By pruning the space of all causal factors, such explanations facilitate easier communication and reduce the cognitive load for both explainer and explainee \cite{contrastive_explanations_for_model_interpretability}. 

Contrastive explanations are typically \emph{counterfactual} in the sense that they estimate an altered version of an observation that would have changed the model's prediction, given our knowledge of the model's prediction from the original datapoint. The basic implementation of this idea, which we will henceforth refer to as \emph{counterfactual explanation}, is conceptually simple and does not require much causal machinery. Furthermore, we will look at \emph{algorithmic recourse} (AC): instead of providing an \emph{understanding} of the least \emph{distant} point in feature space that results in the desired prediction, AC aims at producing the minimal \emph{cost} set of actions for the individual, taking the dependencies among observed variables into account in the form of causal knowledge.

\subsection{Counterfactual Explanations}
\label{explanations:counterfactual}
\emph{Counterfactual explanations} (CE) explain a prediction by computing a (usually minimal) change of an individual's features that would cause the underlying model to classify it in a desired class \cite{counterfactual_explanations,verma2021counterfactual}. By showing feature-perturbed versions of the same person who would have received the loan, \emph{counterfactual explanations} can provide actionable information about what to do in the future to secure a better outcome. CE are \emph{counterfactual} in that they consider the alteration of an entity in the history of the event $P$, where $P$ is the undesired model output \cite{DBLP:conf/fat/KarimiSV21}.

For example, a feature instantiation that would have changed the prediction in the above example would be \say{you would have received the loan if your income was higher by \$$10$k}. In contrast to explanation methods that rely on approximating the classifier's decision boundary \cite{ribeiro2016should}, counterfactual explanations are always truthful w.r.t. the underlying model by using actual predictions of the algorithm.

\begin{figure}
    \centering
    \hspace*{\fill}
    \begin{subfigure}[t]{0.44\columnwidth}
        \includegraphics[width=\columnwidth]{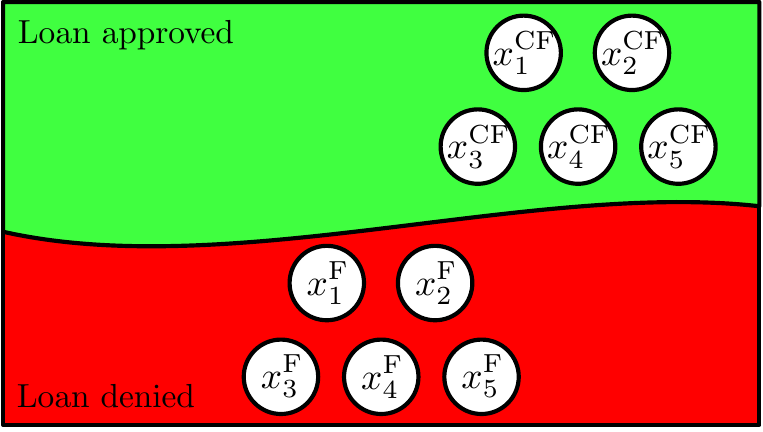}
        \caption{\textbf{Counterfactual Explanations \cite{counterfactual_explanations}:} The counterfactual instance corresponds to the \emph{nearest feature instantiation that lies on the other side of the decision boundary}.}
    \end{subfigure}
    \hfill
    \begin{subfigure}[t]{0.44\columnwidth}
    \includegraphics[width=\columnwidth]{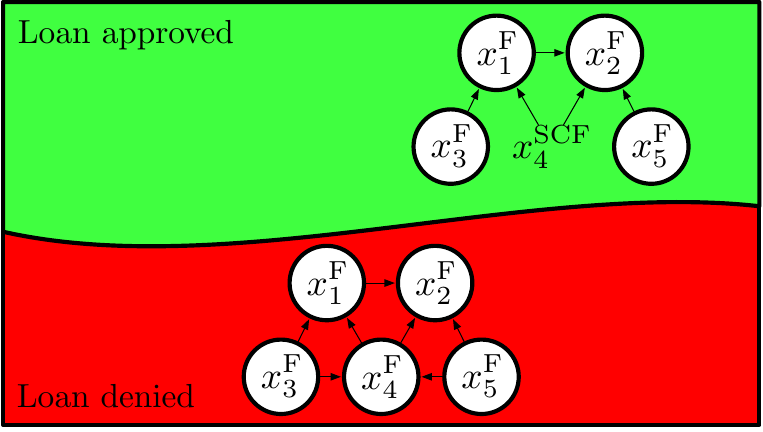}
    \caption{\textbf{Causal Algorithmic Recourse \cite{DBLP:conf/nips/KarimiKSV20}:} The counterfactual instance corresponds to the \emph{minimal cost set of interventions}; here, one intervention on $x_4$.}
    \end{subfigure}
    \hspace*{\fill}
    \caption{\textbf{Counterfactual explanations vs. Causal Algorithmic Recourse}, illustrated by an individual who got their loan application denied, represented by features $\vx^{\text{F}} = \{x_1^{\text{F}},\dots,x_5^{\text{F}}\}$. }
    \label{fig:cf_exp}
\end{figure}

\begin{mydef}{Counterfactual Explanation \cite{counterfactual_explanations}}{cfe}
A counterfactual explanation \(\mathbf{x}^{\mathrm{CE}}\) (or \emph{nearest contrastive explanation}) for an individual \(\mathbf{x}^{\mathrm{F}}\) is a solution of the following optimization problem:
\begin{equation}
\begin{aligned}
    \mathbf{x}^{\mathrm{CE}} \in \arg \min _{\mathbf{x} \in \mathcal{X}} \; \operatorname{dist}\left(\mathbf{x}, \mathbf{x}^{\mathrm{F}}\right) \; \text {s.t.} \; h(\vx) \neq h\left(\xf\right), \vx \in \mathcal{P},
    \label{eq:ce}
\end{aligned}
\end{equation}
where \(\operatorname{dist}(\cdot, \cdot)\) is a similarity metric on \(\mathcal{X}\), and $\mathcal{P}$ is an optional set of plausibility constraints to reflect feasibility, or diversity of the obtained counterfactual explanations \cite{DBLP:journals/corr/abs-1907-09615}.
\end{mydef}

Naively explaining predictions may interfere with constructive recourse, if they violate plausibility and actionability (feasibility) constraints for suggested feature changes. For example, actions such as asking the individual to reduce their age or change their race are not feasible. Therefore, common plausibility constraints include (i) domain-consistency, (ii) density-consistency, and (iii) prototypical-consistency, while actionability constraints distinguish between (i) actionable and mutable, (ii) mutable but non-actionable, and (iii) immutable (and non-actionable) \cite{survey_ac}. 

There are many flavors and variants of CE methods. \citet{interpretable_cf_exps} propose generating CEs in a white-box setting without an auxiliary model, by using the predictive uncertainty of the classifier. \citet{concept_counterfactuals} outlines an approach for producing counterfactual explanations in terms of human-understandable concepts. \citet{causal_ce} propose to preserve causal relationships among input features through feasibility constraints. For a more detailed review of CE methods, we refer the reader to \citet{ce_survey}.

\subsection{Causal Algorithmic Recourse via Minimal Interventions} 
\label{explanations:ac}
Previously, we looked at counterfactual explanations, which demonstrate \say{how the world would have (had) to be different for a desirable outcome to occur} \cite{counterfactual_explanations}. However, these explanations may not always be translated into optimal \emph{recourse actions}, a recommendable set of actions to help an individual to achieve a favorable outcome while respecting the costs of the actions \cite{actionable_recourse}. The field of \emph{Causal Algorithmic Recourse} (or \emph{consequential recommendations}) deals with generating such \say{what should be done in the future} recommendations while respecting both causal relations between features and as costs of actions \cite{DBLP:conf/nips/KarimiKSV20}.

From a causal perspective, actions correspond to interventions (\Cref{concept:intervention}). The modeling goal is to predict the effect of such interventions on one individual's situation to ascertain whether or not the desired outcome is achieved \cite{DBLP:conf/fat/KarimiSV21}.  
\begin{mydef}{\small Algorithmic Recourse via Minimal Interventions (ARMI) \cite{DBLP:conf/fat/KarimiSV21}}{ac}
Consider an SCM $\scm=\{\mathbb{F}, \mathbb{X}, \mathbb{U}\}$ consisting of observed variables $\mathbb{X} \in \gX$, exogenous variables $\mathbb{U} \in \mathcal{U}$, and structural equations $\mathbb{F}: \mathcal{U} \rightarrow \mathcal{X}$. ARMI seeks the minimal cost set of \textbf{interventions} that results in a counterfactual instance yielding the favorable output from $h(\cdot)$ :
\begin{align}
& \va^{*} \in \argmin_{\va \in \gA}  \operatorname{cost}\left(\va; \xf\right) \; \text { s.t. }  h\left(\vx^{\mathrm{SCF}}\right) \neq h\left(\xf\right) \\
& \vx^{\mathrm{SCF}}=\mathbb{F}_{\mathrm{A}}\left(\mathbb{F}^{-1}\left(\xf\right)\right), \quad  \xscf \in \mathcal{P}, \quad \va \in \gA \label{eq:ac}
\end{align}
where $\va^{*} \in \gA$ directly specifies the set of feasible actions to be performed for minimally costly recourse, with $\operatorname{cost}\left(\cdot ; \xf\right): \gA \times \mathcal{X} \rightarrow \mathbb{R}_{+}$, and $\vx^{* \operatorname{SCF}}=\mathbb{F}_{\mathrm{A}^{*}}\left(\mathbb{F}^{-1}\left(\xf\right)\right)$ denotes the resulting structural counterfactual.
\end{mydef}
Although $\xscf$ is a counterfactual instance, it does not need to correspond to the nearest counterfactual explanation $\xce$, resulting from \Cref{eq:ce}. Importantly, using the formulation in \Cref{eq:ac}, \citet{DBLP:conf/fat/KarimiSV21} prove formally that consequential recommendations generated by optimizing w.r.t the decision boundary may be suboptimal, as they do not benefit from the causal effect of actions towards changing the prediction.

Note that \Cref{def:ac} requires a causal model. In practice, we may learn this model from data under the assumption of no hidden confounding, but this assumption can be unrealistic for some scenarios.    \citet{algorithmic_recourse_unobserved_confounding} relax this assumption and propose a partial identification approach that is compatible with unobserved confounding and arbitrary structural equations. Their approach bounds the expected counterfactual effect of recourse actions. 

\section{Open Problems}
\subsection{Unifying Feature Attribution and Explanations}
Attribution and explanation methods can be complementary. For example, as \citet{ce_survey} point out that explanations need to emphasize what should remain the same \emph{in addition} to what should change to achieve the desired outcome. 

Imagine an ML model used for loan prediction that uses \say{income} and \say{years of employment} as inputs. The ML model rejected the loan request of an individual and recommended an increase in \say{income}. Therefore, the individual changed their job, and their feature \say{years of employment} was reset to zero. Despite the increase in \say{income} the model still rejected the loan request since it did not specify that the other feature should not change. 

Having access to causal attribution scores may prevent such situations: a low score for the feature ``years of employment'' would inform the individual that changing its value should not cause any interference with the desired outcome. First steps towards combining both approaches can be found in \cite{kommiya2021towards, galhotra2021feature, albini2021counterfactual,local_explanations_necessity_sufficiency,watson2022rational}.

\citet{recourse_cannot_be_robust} interpret a counterfactual explanation $\xce$ as attributions that describe a perturbation of $\xf$. Since the differences between $\xf$ and $\xce$ can be interpreted as the changes needed to flip the class, we can regard $\varphi_f = \xce - \xf$ as the attribution vector.

\subsection{Scalability and Throughput}

Computing explanations requires solving optimization problems that are often very expensive and difficult to solve. On the attribution side, the computation time often increases exponentially with the number of features, e.g., in the case of Shapley values, making their exact solution computationally intractable when we have more than a few features. On the contrastive explanations side, incorporating plausibility and actionability constraints makes the problem NP-hard or even NP-complete when solving for \cite{survey_ac}, e.g., integer-based variables \cite{artelt2019computation}, neural networks, or quadratic objectives and constraints. 

To make explanations deployable in large-scale systems with many users, approximate methods are highly needed. On the attribution side, one solution might be to compute contributions for only a few samples of the possible coalitions \cite{molnar2022}. \citet{lundberg2017unified} propose the approximate Kernel SHAP method, which assumes feature independence. As first steps towards that goal, \citet{mahajan2019preserving} learn a VAE which can generate multiple counterfactuals for any given input at once.

\subsection{Dynamics}
Most explanation methods assume a static black-box model that does not change over time. However, distribution shifts occur frequently in many real-life ML domains: user behavior may change over time, models get updated, and a decision-making system's utility function may be modified. Naturally, these issues arise for model explanations too, and there exists only limited work to address them \cite{ac_noisy, rawal2020algorithmic, ghorbani2020distributional}.

\subsection{Security and Privacy}
Consider a deployed ML system that provides explanations to its users, e.g., through access to an API. In such settings, an adversary may combine predictions and explanations to extract an approximate model as well as information about the data used to train it. 

\begin{figure}[t!]
\centering
\hspace*{\fill}
  \begin{tikzpicture}[xscale=0.4, yscale=0.4]

    \draw[step=1cm, gray, ultra thin, opacity=0.5] (0,0) grid (8,8);
    \draw[thick] (0,0) rectangle (8,8);
    \draw[black, line width = 0.06cm, dashed, -] (0,0) .. controls (0,8) and (8,0) .. (8,8);
    \draw[black, line width = 0.06cm, dashed, -] (4,8) .. controls (0,8) and (8,0) .. (8,8);

    \draw[olive, fill=olive, opacity=0.4, ultra thick] (1.9,1.75) circle (1.65cm);
    \draw[olive, fill=olive, opacity=0.4, ultra thick] (3,3) circle (0.95cm);
    \draw[olive, fill=olive, opacity=0.4, ultra thick] (5,2.5) circle (1.5cm);
    \draw[olive, fill=olive, opacity=0.4, ultra thick] (7.5,4.5) circle (0.47cm);
    \centerarc[olive, fill=olive, opacity=0.4, ultra thick](7,1)(72.5:197.5:3.3);
    \fill[olive, opacity=0.4] (8,4.155) -- (8,0) -- (3.845,0) -- cycle; 

    \draw[teal, fill=teal, opacity=0.4, ultra thick] (.5,3.5) circle (.47cm);
    \draw[teal, fill=teal, opacity=0.4, ultra thick] (2,5.3) circle (1.4cm);
    \draw[teal, fill=teal, opacity=0.4, ultra thick] (3.55,4.85) circle (0.7cm);
    \draw[teal, fill=teal, opacity=0.4, ultra thick] (5,4.35) circle (0.3cm);
    \centerarc[teal, fill=teal, opacity=0.4, ultra thick](1,7)(-122:32:1.88);
    \fill[teal, opacity=0.4] (0,5.4) -- (0,8) -- (2.6,8) -- cycle; 

    \draw[magenta, fill=magenta, opacity=0.4, ultra thick] (4,7) circle (1cm);
    \draw[magenta, fill=magenta, opacity=0.4, ultra thick] (6,6.1) circle (1.6cm);
    \draw[magenta, fill=magenta, opacity=0.4, ultra thick] (7.2,7.2) circle (0.745cm);

  \end{tikzpicture} \hfill
\begin{tikzpicture}[xscale=0.22, yscale=0.22, >=latex]

    \draw[very thick,->] (0.3,-3.5) -- +(0,7)   node[yshift=5pt] {};
    \draw[very thick,->] (0.3,-3.5) -- +(220:4) node[yshift=-5pt,xshift=-5pt] {};
    \draw[very thick,->] (0.3,-3.5) -- +(12,0)  node[xshift=6pt] {};

    \path[draw, ultra thick, name path=border3, dashed, -] (-2.5,-5) to[out=20,in=220] (5,4);
    \path[draw, ultra thick, name path=border6, dashed, -] (5,4)     to[out=10,in=160] (12,4);
    \path[draw, ultra thick, name path=border4, dashed, -] (12,4)    to[out=190,in=40] (4,-7);
    \path[draw, ultra thick, name path=border5, dashed, -] (4,-7)    to[out=120,in=0] (-2.5,-5);

    \shade[top color=gray!10, bottom color=gray!90, opacity=.30]
      (-2.5,-5)
      to[out=20,in=220] (5,4)
      to[out=10,in=160] (12,4)
      to[out=190,in=40] (4,-7)
      to[out=120,in=0]  (-2.5,-5);

    \filldraw[ball color=magenta, opacity=0.3] (2,2)    circle (2);
    \filldraw[ball color=magenta, opacity=0.3] (3,0)    circle (1);
    \filldraw[ball color=magenta, opacity=0.3] (-1,-1)  circle (3);
    \filldraw[ball color=magenta, opacity=0.3] (2,-2.5) circle (2);
    \filldraw[ball color=magenta, opacity=0.3] (5,1)    circle (1.5);
    \filldraw[ball color=magenta, opacity=0.3] (7,4)    circle (2);
    \filldraw[ball color=magenta, opacity=0.3] (10,5)   circle (1.5);

    \filldraw[ball color=olive, opacity=0.3] (10,2)      circle (1.5);
    \filldraw[ball color=olive, opacity=0.3] (11.5,3.5)  circle (.5);
    \filldraw[ball color=olive, opacity=0.3] (9,-1)      circle (3);k
    \filldraw[ball color=olive, opacity=0.3] (0,-5)      circle (1);
    \filldraw[ball color=olive, opacity=0.3] (1,-6)      circle (1.5);
    \filldraw[ball color=olive, opacity=0.3] (4,-6)      circle (2);
    \filldraw[ball color=olive, opacity=0.3] (5,-4)      circle (1.25);
    \filldraw[ball color=olive, opacity=0.3] (4.6,3)     circle (.5);
    \filldraw[ball color=olive, opacity=0.3] (5.5,-1.75) circle (1.25);

  \end{tikzpicture}
  \hspace*{\fill}
  \caption{\textbf{Model Stealing through Contrastive Explanations \cite{survey_ac}:} \citet{survey_ac} illustrate the model stealing process in 2D and 3D using hypothetical non-linear decision boundaries. ``How many optimal contrastive explanations are needed to extract the decision regions of a classifier?'' can be formulated as ``How many factual balls are needed to maximally pack all decision regions?''}
  \label{fig:ball_packing}
  \vspace{-3mm}
\end{figure}
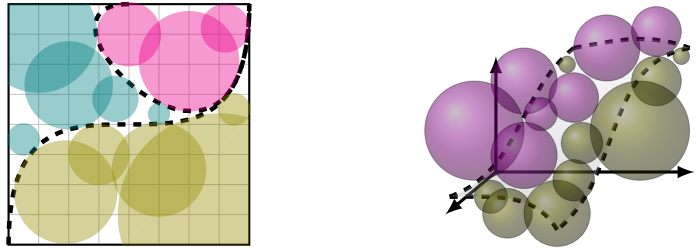

For example, in the case of contrastive explanations, three possible attack strategies are \emph{model extraction} \cite{aivodji2020model} by (i) augmenting the surrogate's model dataset with counterfactual instances/labels, and (ii) approximating its decision boundaries through optimal contrastive explanations, as visually illustrated in \Cref{fig:ball_packing} and formalized by \citet{survey_ac}, and (iii) \emph{membership inference attacks} which quantify the information explanations leak about the presence of a datapoint in the training set for a model \cite{shokri2021privacy}. 

In the case of models using causal knowledge about the DGP (\Cref{explanations:asv,explanations:csv,explanations:ac}), we hypothesize that it is important to guard this causal knowledge against causal-discovery-like model extraction attacks too. An attacker who is allowed to perform enough queries may use an active learning approach to actively intervene on input points and yield the causal graph of the explanation model \cite{he2008active}.

\citet{slack2021counterfactual} describes the vulnerabilities of counterfactual explanations and show that they may converge to drastically different counterfactuals under a small perturbation, indicating that they are not robust. To this end, they propose an objective to train seemingly fair models where counterfactual explanations find lower cost recourse under  perturbations. 

There has been very limited work on guarding explanation methods against such concerns. \citet{shokri2021privacy} propose differentially private (DP) model explanations. They design an adaptive DP gradient descent algorithm that finds the minimal privacy budget required to produce accurate explanations. \citet{naidu2021differential} investigate the interpretability of DP-trained medical imaging models.

\subsection{Robustness vs. Recourse Sensitivity}

Two desirable properties of explanation methods are \emph{robustness} and \emph{recourse sensitivity} \cite{recourse_cannot_be_robust}. Robustness refers to changes in the inputs: if we reason that similar users should get similar options for recourse, then small changes in the input $x$ should not cause large jumps in the explanation $\varphi_{f}(x)$, i.e. $\varphi_{f}$ should be continuous. A recourse sensitive attribution method permits that the user can always achieve sufficient utility by moving in the direction of the vector $\varphi_{f}(x)$. 

\citet{recourse_cannot_be_robust} point out that attribution methods and counterfactual explanations cannot be robust and recourse sensitive at the same time. Their main conclusion is that for any way of measuring utility, there exists a model $f$ for which no attribution method $\varphi_{f}$ can be both recourse sensitive and continuous. Future work should therefore consider workarounds to circumvent \citet{recourse_cannot_be_robust}'s impossibility result. 
\chapter{Causal Fairness}
\label{chapter:fairness}

Machine learning models increasingly assist in life-changing decisions like parole hearings, loan applications, and university admissions. Decisions in these areas may have ethical or legal implications, so model practitioners must consider the societal impact of their work. 

If the data used to train an algorithm contains demographic disparities against certain races, genders, or other groups, the algorithm will too. Instead of just maximizing measures that quantify desirable statistical properties of a predictor, \emph{algorithmic fairness} aims to provide criteria that can be used to assess a model's fairness and mitigate harmful disparities. 

Causality plays a significant role in investigating a model's fairness because it typically depends on the causal structure of the data: for some causal graphs, it can be fair to include certain input features, while for others, it is not. Unfortunately, statistical-based fairness measures are oblivious to distinguishing between different causal relationships among input variables and quickly fail to detect discrimination in the presence of statistical anomalies such as Simpson's paradox \cite{makhlouf2020survey}. 

For example, imagine that instead of considering the causal structure of the data, we would down-weight or discard sensitive attributes. Doing so may not result in a fair procedure as the sensitive attribute often correlates with other attributes. Especially in large feature spaces, sensitive attributes are often redundant given the other features. The classifier may learn a redundant encoding for the sensitive features in such settings.  

\citet{barocas-hardt-narayanan} illustrate this issue of learning classifiers using sensitive attributes without explicitly being asked to with the following example: Consider a fictitious start-up that sets out to predict your income from your genome. DNA can predict income better than random guessing because DNA encodes information about ancestry, which correlates with income in some countries, such as the United States. Hence, the learned classifier may use ancestry in an entirely implicit manner. Unfortunately, removing redundant encodings of ancestry from the genome is a difficult task that cannot be accomplished by removing a few individual genetic markers. 

In contrast, by embracing the sensitive attributes and their causal relationships with the other input variables, we can intervene on them and analyze how changing their values affects the model's predictions. A fair predictor would be invariant w.r.t. such interventions. In the DNA example, this means that -- all other things being equal -- a fair predictor remains invariant w.r.t. ancestry by ensuring that interventions on the genetic markers do not influence the predictions. 

In this chapter, we discuss two classes of causality-based fairness criteria: \emph{counterfactual} (CFF) and \emph{interventional} fairness (IF). CTF criteria assess the unfair influence of protected attributes on the outcome variable by analyzing counterfactuals. These criteria primarily differ based on how the counterfactuals are constructed. Further, IF approaches aim to relax some of the strong assumptions CFF requires to make them more practical in real-life settings.

\begin{table*}[h]
    \centering
    \begin{minipage}{\textwidth}
        \centering
        \resizebox{\linewidth}{!}{%
    \begin{tabular}{|P{0.9cm}|P{3cm}|P{6.8cm}|P{1.5cm}|}
        \toprule
        \bf Class & \bf Criterion & \bf Key Idea & \bf Ref. \\
        \midrule
         \multirow{7}{*}{Ctf.} & Counterfactual Fairness (CF)& A decision is fair towards an individual if it is the same in (a) the actual world and (b) a counterfactual world where the individual belonged to a different demographic group  & \Cref{sec:cf_fairness}  \\ \cline{2-4}
        & Path-Specific CF & Refinement of CF by distinguishing between fair and unfair pathways in causal DAG & \Cref{fairness:pscf} \\  \cline{2-4}
        & Causal Explanation Formula & Decomposition of CF into direct, indirect, and spurious discrimination & \Cref{fairness:ce} \\ \midrule
        \multirow{5}{*}{Int.} & Proxy Fairness & Intervention on proxy variables of protected attributes instead of the latter directly & \Cref{fairness:proxy}  \\ \cline{2-4}
        & Justifiable Fairness & Opens up the possibility of not requiring a full causal model: requires only separation of variables into admissible and inadmissible & \Cref{fairness:justifiable} \\
        \bottomrule
    \end{tabular}
    }
    \caption{\textbf{Criterion Overview} of Causal Fairness.}
    \label{tab:problem_overview_fairness}
    \end{minipage}
\end{table*}

\notation
\begin{nota}
    \rmA & \text{ Protected (or \say{sensitive}) attributes of an individual} \\
    \rmX & \text{ Remaining attributes of an individual } \\
    \rmU & \text{ Exogenous attributes of an individual } \\
    \rmW & \text{ Attributes between $\rmX$ and $Y$} \\
    Y & \text{ Outcome } \\
    \hat Y & \text{ Outcome predictor }\\
\end{nota}

\begin{figure}
\centering
\hspace*{\fill}
\begin{subfigure}[t]{0.2\textwidth}
        \includegraphics[width=\columnwidth]{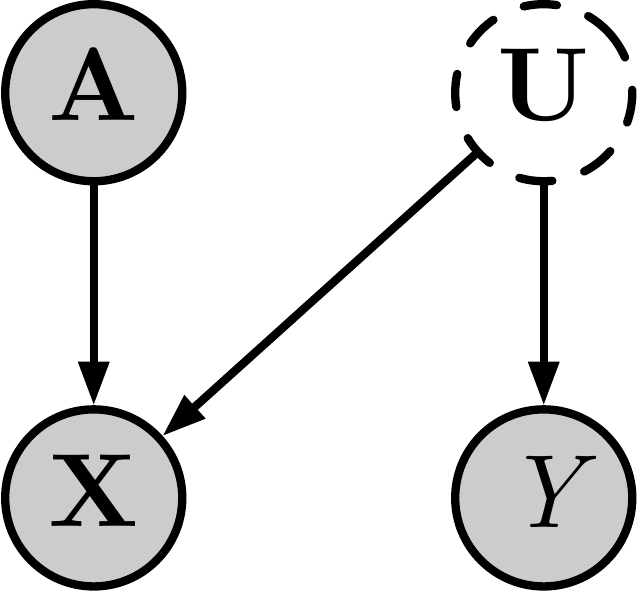}
\caption{}
\label{fig:cff_a}
\end{subfigure} \hfill \begin{subfigure}[t]{0.2\textwidth}
        \includegraphics[width=\columnwidth]{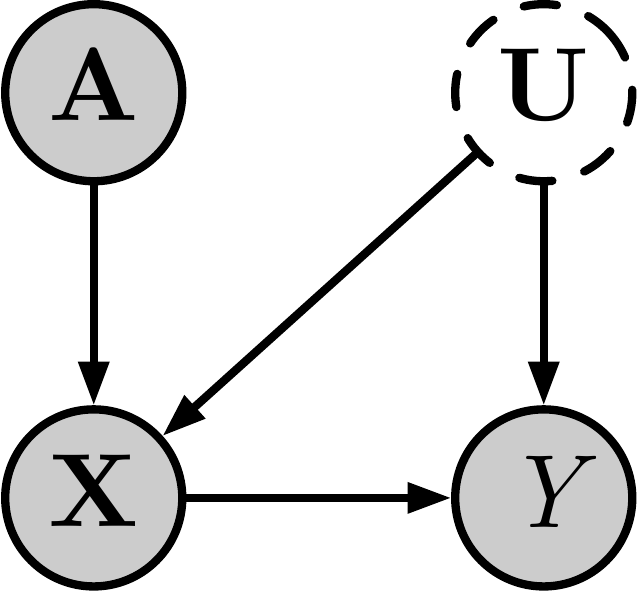}
\caption{}
\label{fig:cff_b}
\end{subfigure} \hspace*{\fill}
\caption{\textbf{Two possible causal graphs considered by Counterfactual Fairness \cite{counterfactual_fairness}}: (a) shows the setup where only the latent confounder $\rmU$ causes the outcome; in contrast, (b) considers both $\rmU$ and $\rmX$ influencing $Y$.}
 \label{fig:cf}
 \end{figure}

The aforementioned DNA example clarifies why one must account for the data's causal structure to ensure fair predictions. Now, we give two more detailed examples with concrete causal DAGs, as shown in \Cref{fig:cf}. These examples are taken from \citet{counterfactual_fairness}.

In both scenarios, using the seemingly fair attributes $\rmX$ to predict $Y$ is unfair. 
 
\Cref{fig:cff_a} illustrates a setting where only latent confounder $\rmU$ causes $Y$. Suppose, for example, a car insurance company wants to price insurance for car owners by predicting their accident rates $Y$, assuming there is an unobserved variable corresponding to aggressive driving $\rmU$, which (a) increases the likelihood that drivers will have an accident and (b) increases the likelihood of individuals preferring red cars, captured by $\rmX$. Furthermore, people with specific demographic characteristics $\rmA$ may prefer driving red cars. Using the red car feature $\rmX$ to predict the accident rate $Y$ seems unfair because these individuals are no more likely than anyone else to be aggressive or to get into accidents. 
 
\Cref{fig:cff_b} illustrates the situation where both $\rmU$ and $\rmX$ influence the outcome $Y$. Consider a crime prediction scenario in which a municipality wishes to estimate crime rates by neighborhood to allocate police resources. The data set contains residents' neighborhood $\rmX$, demographics $\rmA$, and binary labels $Y$ indicating a criminal arrest history. Due to historically segregated housing, the location $\rmX$ depends on $\rmA$, and locations $\rmX$ do not have equally many police resources (unobserved). The number of arrests $Y$ is higher in areas with more police resources $\rmX$. $\rmU$ represents the totality of socioeconomic factors and policing practices that influence where an individual may live and how likely they are to be arrested. While $Y$ depends on $\rmX$, the predictor $\hat Y$ could be influenced by values of $\rmA$ that are not explained by $\rmU$. 

\section{Counterfactual Fairness} 
\label{sec:cf_fairness}
We start with the \emph{counterfactual fairness} criterion by \citet{counterfactual_fairness}. It ensures that predictors of the outcome $Y$, which any individual with features $\{ \vx \cup \va \}$ receives in the real world, is the same as the one they would receive in a counterfactual world in which only the protected attribute of the individual $\rmA = \va^\prime$ has changed, everything else remaining equal. In other words, the predictor is invariant to both counterfactuals. We formalize this criterion by recalling the definition of a counterfactual in \Cref{concept:counterfactuals}.

\begin{mydef}{Counterfactual Fairness \cite{counterfactual_fairness}}{cff}
Given latent exogenous variables $\rmU$ that are not caused by any of $\rmX$ or $\rmA$, we say that the predictor $\hat{Y}$ is counterfactually fair if 
\begin{align}
   p\left(\hat{y}_{\va}(\vu) \mid \vx, \va\right) =p\left(\hat{y}_{\va^{\prime}}(\vu) \mid \vx, \va\right), \quad \forall \vx, \va, \va^\prime, y. \label{eq:counterfactual_fairness}
\end{align}
\end{mydef}

We interpret the final expression of \Cref{eq:counterfactual_fairness} as the probability that $\hat{Y}$ predicts $y$ for a given individual for which we observe features $\vx$ and protected attribute $\va$, had the protected attribute been $\va^{\prime}$ instead of $\va$. An analogous interpretation of the first expression asserts that it equals $p(y \mid \vx, \va)$.

\subsection{Path-Specific Counterfactual Fairness} \label{fairness:pscf}
The aforementioned counterfactual fairness criterion (Definition~\ref{def:cff}) considers the full effect of the sensitive attribute on the decision as problematic. However, this is not the case in specific scenarios, and we want to be more precise about their contribution to unfair predictions. \citet{nabi2018fair} identify that discrimination can be formalized as the presence of an effect of a covariate on the outcome along specific causal pathways, but not all of them.  

More concretely, \citet{chiappa2019path} substantiate this concern by using the famous Berkeley alleged gender bias case, commonly used as a textbook example of Simpson's Paradox. In this scenario, data on college admissions showed a bias for male applications overall. However, the university rejected female applicants more often than male applicants because they applied to more competitive departments with lower admission rates. Such an effect of gender through department choice is not unfair.

Motivated by this, \citet{chiappa2019path} propose \emph{path-specific counterfactual fairness}, a more fine-grained fairness criterion that deals with sensitive attributes affecting the decision along both fair and unfair pathways. It states that a decision is fair toward an individual if it coincides with the one the model would have taken in a counterfactual world where the sensitive attributes along the unfair pathways differed.

\begin{mydef}{Path-Specific Cf. Fairness \cite{nabi2018fair, chiappa2019path,loftus2018causal}}{pscf}
Define $\mathcal{P}_{\mathcal{G}_{A}}$ as the set of all directed paths from $\rmA$ to $Y$ in $\mathcal{G}$ which correspond to all unfair chains of events where $\rmA$ causes $Y$. Let $\rmX_{\mathcal{P}_{\mathcal{G}_{A}}^{c}} \subseteq \rmX$ be the subset of covariates not present in any path in $\mathcal{P}_{\mathcal{G}_{A}}$. Then, Predictor $\hat{Y}$ is path-specifically counterfactually fair w.r.t. path set $\mathcal{P}_{\mathcal{G}_{A}}$ if $\forall \vx, \va, \va^{\prime}, y:$
\begin{align}
p\left(\hat y_{\va, \vx_{\mathcal P^c_{\mathcal G_A}}}(\rmU) \mid \vx, \va\right) =
p\left(\hat y_{\va', \vx_{ \mathcal P^c_{\mathcal G_A}}}(\rmU) \mid \vx, \va\right).
\end{align}
\end{mydef}

Similar to \citet{chiappa2019path}, \citet{wu2019pc} define a similar version of path-specific counterfactual fairness. Their criterion enables us to reason about any population sub-group. Further, they address situations where identifiability may not hold by proposing a method that bounds the path-specific effects.

\subsection{Causal Explanation Formula} \label{fairness:ce}
\citet{zhang2018fairness} propose the \emph{causal explanation formula}, which allows practitioners to decompose the total observed disparity of decisions into three fine-grained measures.

First, the authors assume that there is a disadvantaged group $\va_1$ and an advantaged one, $\va_0$. Further, $\rmW$ denotes all observed intermediate variables between $\rmA$ and $Y$. Next, they define different effects; if the effect is non-zero, the predictor is unfair. They distinguish between direct and indirect spurious effects; the latter considers the back-door paths between $\rmA$ and $Y$, that is, paths with an arrow into $\rmA$.

\begin{mydef}{Counterfactual Effects \cite{zhang2018fairness}}{cf_effects}
We denote all observed mediator variables between $\rmA$ and $Y$ as $\rmW$. Then, the direct effect (DE), indirect effect (IE), and spurious effect (SE) are respectively defined as:
\begin{align}
\text{DE}_{\va_{0}, \va_{1}}(y \mid \va)&=p\left(y_{\va_{1}, \rmW_{\va_{0}}} \mid \va\right)-p\left(y_{\va_{0}} \mid \va\right), \\
\text{IE}_{\va_{0}, \va_{1}}(y \mid \va) &=p\left(y_{\va_{0}, \rmW_{\va_{1}}} \mid \va\right)-p\left(y_{\va_{0}} \mid \va\right), \\
\text{SE}_{\va_{0}, \va_{1}}(y)&=p\left(y_{\va_{0}} \mid \va_{1}\right)-p\left(y \mid \va_{0}\right).
\end{align}
\end{mydef}

Next, they aim to provide intuition on how the different effects relate to each other by decomposing the total variation (TV) into them. TV is the difference between the conditional distributions of $Y$ when passively observing $\rmA$ changing from $\va_0$ to $\va_1$. Formally, the TV of event $\rmA=\va_{1}$ on $Y=y$ (with baseline $\va_{0}$ is defined as:
\begin{align}
\text{TV}_{\va_{0}, \va_{1}}(y)=p\left(y \mid \va_{1}\right)-p\left(y \mid \va_{0}\right).
\end{align}

\begin{mydef}{Causal Explanation Formula \cite{zhang2018fairness}}{causal_explanation_formula}
The total variation (TV), spurious effect (SE), indirect effect (IE), and direct effect (DE) are related as
\begin{align}
\text{TV}_{\va_{0}, \va_{1}}(y) &=  \text{SE}_{\va_{0}, \va_{1}}(y)+\text{IE}_{\va_{0}, \va_{1}}\left(y \mid \va_{1}\right) - \text{DE}_{\va_{1}, \va_{0}}\left(y \mid \va_{1}\right), \\
\text{TV}_{\va_{0}, \va_{1}}(y)&= \text{DE}_{\va_{0}, \va_{1}}\left(y \mid \va_{0}\right) - \text{SE}_{\va_{1}, \va_{0}}(y)- \text{IE}_{\va_{1}, \va_{0}}\left(y \mid \va_{0}\right).
\end{align}
\end{mydef}

 For example, the first formula shows that the total disparity experienced by the individuals who have naturally attained $\va_{1}$ equals the disparity experienced via spurious discrimination, plus the advantage is lost due to indirect discrimination minus the advantage it would have gained without direct discrimination. 

\section{Interventional Fairness} 
Counterfactual fairness involves modeling counterfactuals on an individual level, which is problematic. For example, \cite{race, sex} point out that effects of race or gender are challenging to model \emph{even at the group level}. Thus, interventions on such (often ill-defined) protected attributes are typically hard to envision. \citet{Kilbertus20} illustrates this through the following thought experiment: imagine a pregnant woman's job application gets rejected. Could we imagine her life as a man in that world? Was she born male or perceived as male during the hiring process? Is she a pregnant man now? The authors argue that the notion of counterfactual fairness is meaningless if we compare (fictitious) individuals who are vastly different from each other.

\subsection{Proxy Fairness}\label{fairness:proxy}

To address these concerns, \citet{proxy_fairness} examines population-level interventional distributions, which they call \emph{proxy discrimination}. The idea is to separate the protected attribute $\rmA$ from its potential proxies, such as names, visual features, languages spoken at home, etc. Hence, an intervention based on proxy variables presents a more manageable problem. In practice, we are frequently limited to imperfect measurements of $\rmA$, so separating the root concept from a proxy is prudent \cite{Kilbertus20}.
A proxy $\rmP$ is a descendant of $\rmA$ in the assumed causal graph. 

\begin{mydef}{Proxy Fairness \cite{Kilbertus20}}{pf}
Predictor $\hat{Y}$ exhibits no proxy discrimination based on a proxy $\rmP$ if 
\begin{align}
    p\left(y \mid \doo\left(\vp\right)\right)=p\left(y \mid \doo\left(\vp^{\prime}\right)\right), \quad \forall \vp, \vp^{\prime}.
\end{align}
\end{mydef}

\subsection{Justifiable Fairness} \label{fairness:justifiable}
All previous approaches require knowledge of the causal graph. To deal with settings where a causal graph is missing but a partial knowledge of admissible variables is given, 
\citet{salimi2019interventional} formalize interventional fairness as a database repair problem. They present data pre-processing algorithms for providing fairness guarantees about classifiers trained on pre-processed training data.  

First, they assume that the causal graph is given and define the \emph{K-fair} criterion, which captures group-level fairness, similar to proxy fairness. 

\begin{mydef}{$\rmK$-Fair \cite{salimi2019interventional}}{kfair}
For a set of attributes $\rmK \subseteq \rmX$, we say that the predictor $\hat Y$ is $\rmK$-fair w.r.t. protected attributes $\rmA$ if $\forall \vk, \va, \hat y$:
\begin{align}
p\left(\hat y \mid \doo\left(\va, \vk\right)\right)=p\left(y \mid \doo\left(\va^{\prime}, \vk\right)\right).
\end{align}
\end{mydef}

A model is deemed \emph{interventionally fair} if it is $\rmK$-fair for every set $\rmK$. This notion differs from proxy fairness in ensuring that $\rmA$ does not affect $Y$ in any configuration obtained by fixing other variables to arbitrary values. As opposed to counterfactual fairness, it does not try to capture fairness at the individual level, so it employs level-2 interventions. 

Next, they define the \emph{justifiable fairness} criterion, which allows the user to distinguish only admissible and inadmissible variables. The former variables are a subset of the protected attributes $\rmA$ through which it is still permissible to influence the outcome. 

\begin{mydef}{Justifiable Fairness \cite{salimi2019interventional}}{justifiable_fairness}
A predictor $\hat Y$ is justifiably fair if it is $\rmK$-fair w.r.t. all supersets $\rmK \supseteq \rmA$. 
\end{mydef}

\section{Fairness under Distribution Shifts}

\citet{singh2021fairness} study the problem of learning fair prediction models under covariate shift, i.e., with test set covariates distributed differently from the training set. Given the ground-truth causal graph describing the data and anticipated changes, they propose an approach based on feature selection that exploits conditional independencies in the data to estimate accuracy and fairness metrics for the test set. 

\citet{schrouff2022maintaining} evaluates how realistic the assumptions made by \citet{singh2021fairness} as well as other works are \cite{fair_MAML, evaluating_model_robustness}. They categorize distribution shifts into four categories: \emph{demographic shift}, \emph{covariate shift}, \emph{label shift}, and \emph{compound shift}, independently of the causal graph considered. 
Their study examines two real-world applications in dermatology and electronic health records and shows that clinically plausible shifts directly affect aspects of the data distribution simultaneously. Therefore, and as empirically demonstrated, compound shifts impact the transferability of fairness properties in these applications.

\section{Open Problems}
\subsection{Alternatives To Equality} Most fairness definitions in the literature emphasize equality, ensuring that each individual or group receives the same resources, attention, or outcome. In contrast, \emph{equity} \cite{guy2012social, race_and_social_equity} has received little attention, which means that all individuals and groups have access to the resources they need to thrive. An exciting future direction is to operationalize this definition and examine how it enhances or contradicts existing definitions of fairness \cite{mehrabi2021survey}. 

Another alternative to equality might be considering a model's \emph{harm}: \citet{harm} propose a definition for a harm criterion that should be able to answer the following three questions precisely: \quo{\textbf{Q1:} Did the actions of an agent cause harm, and if so, how much? \textbf{Q2:} How much harm can we expect an action to cause before taking it? \textbf{Q3:} How can we identify actions that balance the expected harms and benefits?} They then propose a family of counterfactual objective functions that mitigate harm. 

\subsection{Fairness Beyond Predictions} 

In this section, we studied predictive fairness criteria that evolved around supervised models' predictive output. Often, however, we care about the non-discrimination of quanti-
ties we do not have full control over. For this reason, there is growing interest in assessing the fairness of ML techniques beyond prediction. 

\citet{equalizing_recourse} put forward that models should not limit the ability to achieve recourse to only those with access to expensive resources. Put another way, the model must fairly distribute the rights of recourse across (demographically defined) groups. \citet{von2020fairness} investigate the fairness of causal algorithmic recourse actions and conclude that much work remains. For example, whether it is appropriate to perform societal interventions on all individuals in a subgroup. \citet{huan2020fairness} deal with \emph{equality of efforts}: they seek to determine whether the efforts made to achieve the same outcome level are the same or different between those in the protected group and those in the unprotected group. \citet{kusner2019making} describe the \emph{Discriminatory Impact Problem}: How can one reduce discrimination arising from the real-world impact of decisions? To address this, they detail a method to both map the causal pathway of a single decision and also to model the effect of interference--how the impact on an individual
depends on decisions made about other people.

\subsection{Partial Identification}
Causal quantities, especially counterfactuals, are often unidentifiable (\Cref{sec:identifiability}). That means that we can not compute them from a statistical quantity. While previous causal fairness work has partly addressed these issues \cite{counterfactual_fairness, wu2019counterfactual,wu2019pc,kilbertus2020sensitivity}, we foresee more important work to be done in sensitivity analysis through realistic simulations to thoroughly test these methods and alter aspects of them to understand how violations of assumptions impact fairness estimation.

\subsection{Manipulability of Social Categories}
From a social science perspective, it is heavily debated whether social categories admit interventions. \citet{kohler2018eddie} criticize that counterfactuals require us to reduce race to only the signs of the category, e.g., the skin color or phenotype of a race. \citet{hu2022causation} argues that causal theories about social categories such as race involve \quo{ineliminable substantive moral and political considerations, a feature for which interventionism can not well account}. \citet{sex_fairness} challenge the validity of specifying a social group, such as gender, as a variable in a DAG while still assuming the model's modularity assumption. More broadly, \citet{kasirzadeh2021use} review various papers at this intersection between sociology and causal modeling, concluding that social categories often do not admit counterfactual manipulation.

\subsection{Trade-offs Between Fairness Criteria}
\citet{kleinberg2016inherent,corbett2017algorithmic,chouldechova2017fair} show that fairness criteria can be opposed to each other, and sometimes, one cannot satisfy multiple criteria simultaneously. Hence, the trade-off between multiple admissible criteria has to be balanced. \citet{beyond_impossibility} shed some first light on balancing certain measures using a multi-objective framework. Similarly, \citet{causal_fairness_consequences} highlight limitations and potential adverse consequences of causal fairness criteria. They demonstrate cases in which these criteria lead to policies that every stakeholder would disfavor.

\subsection{Unfair Models}
In debates about model fairness, a common assumption is that unfairness arises through biases in data. However, \citet{introduced_unfairness} point out that models can produce unfair predictions even when the training labels are fair. The authors refer to this phenomenon as \emph{introduced unfairness} and investigate the conditions under which it may arise. Taking a causal perspective, they show that the notion of introduced unfairness can be applied to causal definitions of fairness too. For example, they consider \emph{path-specific introduced effects} as the difference in some path-specific effects on labels and predictions. 

Minimal work exists investigating when fair training labels may not yield a fair model. \citet{introduced_unfairness} conclude that it is difficult to rule out unfair disparities, even when the criteria of causal fairness are met. Future research may stress-test the validity of this section, which introduced fairness criteria.

For another, even more exhaustive example, including calculations of statistical and causal fairness criteria using concrete data, we direct the reader to \citet{makhlouf2020survey}. 

\newcommand{\return}{\mathcal{R}}
\newcommand{\loss}{\ell}

\chapter{Causal Reinforcement Learning}
\label{chapter:rl}
Reinforcement Learning (RL) is a framework in which autonomous agents interact with their environments to learn optimal behaviors, improving over time through trial and error. Its central goal is learning how to map situations to actions while maximizing a numerical reward signal \cite{intro_to_rl}. RL researchers typically formalize their problem setup by using \emph{Markov decision processes} (MDPs), which include three ingredients: sensation (observation), action, and goal (reward). 

In this section, we highlight methods benefiting RL problems by exploiting the causal paradigm (and not the other way around, see, e.g., \cite{zhu2019causal}). We refer to this family of approaches as \emph{Causal Reinforcement Learning}. We summarize the benefits of these approaches in \Cref{tab:problem_overview_causal_rl}.

\begin{table*}[h]
    \centering
    \resizebox{\textwidth}{!}{%
        \begin{tabular}{|P{2.4cm}|P{5.8cm}|P{2.9cm}|P{1.3cm}|}
            \toprule
            \bf Problem & \bf Output & \bf Benefits over non-causal RL & \bf Ref. \\
            \midrule
            Causal Bandits & $\begin{aligned}\hat \pi = \argmin_{\pi \in \Pi} L_n(\pi)\end{aligned}$ & Optimal simple regret guarantees &  \Cref{rl:causal_bandits} \\ \midrule
            Model-Based RL & $\begin{aligned}\widehat \params = \argmin_{\params \in \Params} \loss \left(\params, \left(R_{t+1}, S_{t+1}\right) \right )\end{aligned}$ & Deconfounding & \Cref{rl:mbrl}  \\ \midrule
            Multi-Environment RL & $\begin{aligned}\hat \pi = \argmax_{\pi \in \Pi} \mathbb{E}_{c \sim p(c)}\left[\return\left(\pi,\mathcal{M}^{c}\right)\right]\end{aligned}$& Interpretable task embeddings, systematic generalization & \Cref{rl:multi_envs}  \\ \midrule
            Off-Policy Policy Evaluation & $\begin{aligned} \hat v_{\pi}(s) = \mathbb{E}_{\vx \sim d_{0}}\left[\sum_{t=0}^{T-1} \gamma^{t} r_{t} \mid \vx_{0}=\vx\right]\end{aligned}$ & Deconfounding & \Cref{rl:oppe}  \\ \midrule
            Imitation Learning & $\begin{aligned}\hat{\pi}=\underset{\pi \in \Pi}{\arg \min } \; \mathbb{E}_{\vx \sim d_{\pi^{*}}}\left[\ell\left(\vx, \pi, \pi^{*}\left(\vx\right)\right)\right]\end{aligned}$ & Deconfounding & \Cref{rl:il}  \\ \midrule
            Credit Assignment & $\begin{aligned}\scm_{a_t \rightarrow r_{t+k}} \text{ or } \scm_{a_t \rightarrow s_{t+1}} \text{ or } \scm_{a^i_t \rightarrow a^j_{t}} \end{aligned}$ & Intrinsic reward, Data-efficiency & \Cref{rl:ca}  \\ 
            \midrule
            Counterfactual Data Augmentation & $\begin{aligned} \tilde \tau = \{\tilde \vx_t, \tilde a_t, \tilde \vx_{t+1} \}_{t=1}^{T}  \end{aligned}$ & Data-efficiency & \Cref{rl:cda}  \\ \midrule 
            Agent Incentives & Incentive criteria and measures & Avoiding unintended harmful behavior& \Cref{rl:incentives} \\
            \bottomrule
        \end{tabular}%
    }
    \caption{\textbf{Problem Overview of Causal Reinforcement Learning.}}
    \label{tab:problem_overview_causal_rl}
\end{table*}

\notation
\begin{align}
&\begin{array}{ll}
t & \text { discrete time step } \\
T & \text { final time step of an episode } t \\
A_{t} & \text { action at time } t \\
\rmS_{t} & \text { state at time } t \\
L_{t} & \text{ regret/loss at time }t \\
R_{t} & \text { reward at time } t \\
\return & \text{ return } \\ 
\pi & \text { policy (decision-making rule) } \\
\pi(a \mid s) & \text { probability of taking action } a \text { in state } s  \\
\vs, \vs^{\prime} & \text { true states } \\
\vx, \vx^\prime & \text{ observed states} \\
v_{\pi}(s) & \text { value of state } s \text { under policy } \pi \text { (expected return) } \\
q_{\pi}(s, a) & \text { value of taking action } a \text { in state } s \text { under policy } \pi \\
\tau & \text{ trajectory, i.e., }\tau = \{\vx_t, a_t, \vx_{t+1}\}_{t=1}^T \\
\end{array} \nonumber
\end{align}

\section{Isn't RL already ``Causal''?}
The short answer is yes. For many years, researchers have argued that there exist links between certain flavors of RL and causal inference \cite{bottou2013counterfactual,bareinboim2015bandits, gershman2017reinforcement, causal_from_RL_perspective, EOCI, bannon2020causality, crl, learning_by_doing}. For example, \citet{bottou2013counterfactual} frame bandit problems and Markov decision processes (MDPs) as special cases of causal models. However, despite conceptual similarities, the two fields have mainly focused on different goals: on the one hand, the RL community has focused on building algorithms to \emph{maximize rewards}; on the other hand, the focus in the causality literature has been on the \emph{identifiability and} inferences of or based on given causal structure \cite{learning_by_doing}.  

We attribute one reason for different foci among both communities to the type of applications each tackles. The vast majority of literature on modern RL evaluates methods on synthetic data simulators, able to generate large amounts of data. For instance, the popular \method{AlphaZero} algorithm assumes access to a boardgame simulation that allows the agent to play many games without a constraint on the amount of data \cite{silver2017mastering}. One of its significant innovations is a \emph{tabula rasa} algorithm with less handcrafted knowledge and domain-specific data augmentations. Some may argue that \method{AlphaZero} proves Sutton's \emph{bitter lesson} \cite{bitter_lesson}. From a statistical point of view, it roughly states that given more compute and training data, general-purpose algorithms with low bias and high variance outperform methods with high bias and low variance. 

In contrast, in the causal inference literature, we are typically given a limited-size observational dataset from an unknown policy and unknown environment and cannot interact with the environment in an online fashion. The reason behind that problem convention is that much of causal inference methodology originated from domains like medicine, econometrics, online advertisements, and the social sciences, in which conducting experiments is infeasible due to ethical or cost/time-consuming reasons. Nonetheless, causal inference is typically applied in contexts where decisions directly influence human individuals. For example, \citet{bottou2013counterfactual} illustrates how causal inference can be used in the ad placement system associated with the Bing search engine.

In causal inference, instead of maximizing a reward function in expectation, a common target is to learn the heterogeneous treatment effect (HTE) (\Cref{rw:hte}): it quantifies the expected effect of changing the observed treatment $\vt$ to a different treatment $\vt^{\prime}$ for a certain subgroup characterized by covariates $\vx$, denoted as \begin{align}
    \tau\left(\vt^{\prime}, \vt, \vx\right) \triangleq \mathbb{E}\left[Y \mid \vx, \doo\left(\vt^{\prime}\right)\right]-\mathbb{E}\left[Y \mid \vx, \doo(\vt)\right].
\end{align} By assumption, we observe only one treatment and outcome pair per subgroup. By using an estimator $\widehat \tau$, decision-makers may reason about what treatments work well for what subgroups. Due to high risks in its applied domains, the HTE research communities prioritize strong theoretical guarantees like convergence rates as a function of the dataset size or analytical confidence intervals. This preference has led to advances like doubly-robust plug-in estimators capable of utilizing machine learning methods \cite{DML, kennedy2020optimal}. 

Somewhere in between the two is the more recent subfield of \emph{offline reinforcement learning} (ORL) \cite{levine2020offline}. Here, the goal is to learn optimal policies from a data set containing trajectories generated from an unobserved policy. \emph{Offline} refers to the fact that the algorithms have to exploit a batch dataset from an unknown environment without access to online exploration, which matches the common problem setup in causal inference. Despite these similarities, there are also differences between the two methodologies, which we list in \Cref{tab:orl_hte}. Nonetheless, we hope to see constructive cross-pollination between these two fields. 

\begin{table}[h]
    \centering
    \begin{tabular}{|P{2cm}|P{3cm}|P{2.2cm}|P{3.5cm}|}
    \toprule \bf Approach & \bf Input data & \bf Goal & \bf Typical Desiderata \\ \midrule 
        Offline RL & Multi-Step Trajectories $\{(\vx_t, \va_t, r_t)\}_{t=1}^{T}$ & Policy $\widehat \pi(\rmX)$ & Reward Maximization in Test Environments \\ \midrule
        HTE & One-Step Individuals $\{(\vx_i, \vt_i, y_i)\}_{i=1}^N$ & HTE  $\widehat \tau(\rmX, \rmT, \rmT^{\prime})$& Analytical Convergence Rates \\ \bottomrule
    \end{tabular}
    \caption{Differences between Offline RL and Heterogeneous Treatment Effect (HTE) estimation. Both methodologies share the motivation of extracting information for decision-making from observational data generated by unknown policies. The shown typical desiderata are simplistic, yet, we hope they clarify differences between the two schools of thought.}
    \label{tab:orl_hte}
\end{table}

\section{Causal Bandits} \label{rl:causal_bandits}
A bandit problem is a sequential game between a learner and an environment \cite{lattimore2020bandit}. The game is played over $n \in \mathbb{N}$ rounds, where $n$ is also called the \emph{horizon}. In each round $t \in[n]$, the learner first chooses an action $A_{t}$ from a given set $\mathcal{A}$ (also called \emph{arms}), and the environment then reveals a reward $R_{t} \in \mathbb{R}$.

When $|\gA| = k \in \sN$, we refer to the problem as $k$-armed bandits. When $k\geq2$, but $k$ itself is irrelevant, we simply call it \emph{multi-armed bandits}.

\citet{causal_bandits} formalize \emph{causal bandit problems}, a class of stochastic sequential decision problems in which rewards are given for repeated interventions on a fixed causal model. The motivation is to exploit the causal information for predicting outcomes of interventions without explicitly performing them. Thereby, utilizing non-interventional (i.e. observational) data may improve the rate at which the policy learns high-reward actions. This framework generalizes classical bandits and contextual stochastic bandit problems: in the former, we have no additional observations besides a reward, and in the latter, we observe the context \emph{before} an intervention is chosen. The causal bandit framework additionally allows us to use observations that occur \emph{after} an intervention. 

\citet{causal_bandits} describe the following scenario that motivates why using such extra observations can be helpful: Imagine a farmer who wants to maximize their crop yield. They know that crop yield is only affected by temperature, a particular soil nutrient, and moisture level, but their combination's precise effect is unknown. Each season, the farmer has enough time and money to intervene and control at least one of these variables: deploying shade or heat lamps will set the temperature to be low or high; the nutrient can be added or removed through a choice of fertilizer, and irrigation or rain-proof covers will keep the soil wet or dry. When not intervened upon, the temperature, soil, and moisture vary naturally from season to season due to weather conditions. These variations are all observed along with the final crop yield at the end of each season and can inform the farmer to conduct an experiment to identify the single, highest-yielding intervention in a limited number of seasons.

\begin{mydef}{Causal Bandit Problem \cite{causal_bandits}}{cbp}
Consider a causal model is given by a directed acyclic graph $\mathcal{G}$ over a set of multivariate random variables $\mathcal{X}=\left\{\rmX_{1}, \ldots, \rmX_{N}\right\}$ and a joint distribution $p(\mathcal{X})$ that factorizes over $\mathcal{G}$. 
For each state $\rmX \in \gX$, there is a reward variable $Y$ that takes on values in $\{0,1\}$. Further, given a set of allowed actions $\mathcal{A}$, we denote the expected reward for the action $a=\doo(\vx)$ by $\rmU_{a}:=\mathbb{E}\left[Y \mid \doo (\vx)\right]$ and the optimal expected reward by $\rmU^{*}:=\max _{a \in \mathcal{A}} \rmU_{a}$. 

The causal bandit game proceeds over $T$ rounds. In round $t$, the learner intervenes by choosing $a_{t}=\doo \left(\rmX_{t}=\vx_{t}\right) \in \mathcal{A}$ based on previous observations. In contrast to the conventional bandit problem, before the agent takes the next action, it observes further samples for all non-intervened variables $\rmX_{t}^{c}$ drawn from $p \left\{\rmX_{t}^{c} \mid d o\left(\rmX_{t}=\vx_{t}\right)\right\}$, including the reward $Y_{t} \in\{0,1\}$. 
\end{mydef}

Typically, the objective of the learner is to minimize the simple regret $R_{T}=\rmU^{*}-\mathbb{E}\left[\vu_{\hat{a}_{T}^{*}}\right]$. After $T$ observations, the learner outputs an estimate of the optimal action $\hat{a}_{T}^{*} \in \mathcal{A}$ based on its prior observations. This regret objective is sometimes referred to as a \emph{pure exploration} \cite{bubeck2009pure} or \say{best-arm identification} problem \cite{gabillon2012best} and is most appropriate when, as in drug and policy testing, the learner has a fixed experimental budget after which its policy will be fixed indefinitely.

We call this particular objective the \emph{parallel bandit} problem, in which we formalize each arm as a binary variable that is an independent cause of the reward variable. \citet{causal_bandits} propose and analyze an algorithm for achieving the optimal regret in this setting. The authors propose another, more general algorithm for general causal graphs but leave lower regret bounds for future work. Lastly, they empirically verify that both algorithms improve over the optimal successive elimination algorithm \cite{audibert2010best} through using causal knowledge of the parallel bandit problem under various conditions.

Similarly, \citet{structural_causal_bandits} show that whenever the underlying causal model of a decision-making process is not taken into account, the standard strategies of simultaneously intervening on multiple variables can lead to sub-optimal policies, regardless of the number of interventions performed by the agent in the environment. Therefore, the authors suggest reducing the search space for interventions that lead to optimal rewards and are not redundant. First, they formalize a \emph{minimal intervention set} (MIS) and then aim at finding \emph{possibly-optimal minimal intervention sets} (POMIS). 

\citet{causal_bandits_separating_sets,lu2021causal} study causal bandit problems with unknown causal graph structures. \citet{causal_bandits_separating_sets} formulate a causal bandit algorithm that uses outputs of causal discovery algorithms. They utilize a separating set, defined as a set $\rmS$ that renders a target variable $Y$ independent of a context variable $\rmI$ when conditioned upon, i.e., $\rmI \perp Y \mid \mathbf{S}$. The context variable encodes the intervention. Then, they separately model how interventions influence the separating set $\rmS$and the expected reward given $\rmS$. Their proposed estimator is unbiased and possesses lower variance than the sample mean. 

\citet{lu2021causal} propose an algorithm for a class of causal graph types and prove that it achieves stronger worst-case regret guarantees than non-causal algorithms. Formally, they demonstrate that their goal of exploiting meaningful causal relations among variables cannot be achieved for general causal graphs, and in the worst case, there is no chance to do better than standard algorithms. 

Similar to the causal bandit problem, \citet{cbo_ricardo} study the related problem of dose-response learning, i.e., how an outcome variable $Y$ (which can be a reward) varies under different levels of a control variable $\rmX$ (which can be an arm). The difference to causal bandits is that their goal is not to maximize $Y$ but rather learn the relationship $f(\vx) \equiv \mathbb{E}\left[Y \mid \doo(\vx)\right], \vx \in \mathcal{X}$, where $\mathcal{X} = \left\{\rmX_{1}, \ldots, \rmX_{N}\right\}$ is a pre-defined set of actions. Their method couples different Gaussian process priors that combine observational and interventional data. They also consider active learning schemes to choose arms informed by the GPs' uncertainties. 

Similarly motivated to both the causal bandit \cite{causal_bandits} and the dose-response learning problem, \citet{causal_bo} generalize Bayesian optimization to scenarios where causal information is available. Bayesian optimization is an efficient heuristic to optimize objective functions whose evaluations are costly and cannot be analytically described \cite{bo}, e.g., the final performance of a neural network after training with a specific set of hyper-parameters. \citet{causal_bo} argue that exploiting the causal graph noticeably increases reasoning capabilities about optimal decision-making strategies, decreasing the optimization cost and avoiding suboptimal solutions. By integrating real interventional data with estimated intervened effects computed using do-calculus, they propose an algorithm to balance two trade-offs: exploration vs. exploitation and observation vs. intervention.

\citet{bareinboim2015bandits} take a causal perspective on the conventional multi-armed bandit problem and show that the existence of unobserved confounders renders observational and interventional data distinct (recall our discussion in \Cref{example:interventions}). Further, they show that formalizing this distinction implies that previous bandit algorithms try to maximize rewards based on the observational distribution, which is not always the best strategy to pursue. Therefore, they propose an agent loss function that incorporates both observational and interventional distributions, improving over previous ones. 

\citet{ipl} develop a causal framework for characterizing the environmental shift problem in \emph{offline} contextual bandit problems. While the formerly discussed work has focused on exploiting causal knowledge for improving the finite sample performance or the regret bound in a single environment, their work focuses on modeling distributional shifts and the ability to generalize to new environments.

\citet{lu2022efficient} propose \emph{causal MDPs}, extending the idea behind causal bandits to MDPs. The motivation is similar: they use prior causal knowledge about the state transition and reward functions to obtain conditional independence relations among action, reward, and state variables and use them to develop efficient algorithms.

\section{Model-Based RL} \label{rl:mbrl}

Model-based reinforcement learning (MBRL) learns a model of the environment dynamics (state transition and reward function) in addition to the policy, effectively combining learning and planning methods (the latter have \emph{reversible} access to the MDP dynamics) \cite{moerland2020model}. Generative dynamics models are sometimes referred to as \emph{world models} \cite{ha2018world}. The enticing benefit of MBRL is the promise of improving sample efficiency compared to model-free methods by extracting valuable information from the observed trajectories and enabling the ability to sample simulated experiences from the model instead of the actual environment.

\subsection{Confounded Partial Models}
\label{sec:confounded_partial_models}
In prior works, dynamics models are often \emph{partial} in that they are neither conditioned on nor generate the full set of observed data. For example, the popular MuZero model \cite{muzero} is a partial model because it predicts the state observation $y_T$ at $T>1$ given actions $a_t$ and updated hidden states $h_t$ but using no observational data beyond the initial state $s_0$ at $t=0$. In other words, it generates $y_T$ directly without generating intermediate observations. 

\citet{rezende2020causally} demonstrate that the aforementioned partial models can be causally incorrect: they are confounded by the observations $y_{<T}$ that they do not model and can therefore lead to incorrect planning. The observations $y_{<T}$ are confounders because they are used by the policy to produce the actions $a_{<T}$, while the partial model is missing the $y_{<T}$ in its input. Therefore, the dynamics model is not robust against changes in the behavior policy.

\begin{figure}[h]
    \centering
    \begin{subfigure}[b]{0.3\textwidth}
        \centering
        \includegraphics[width=3cm]{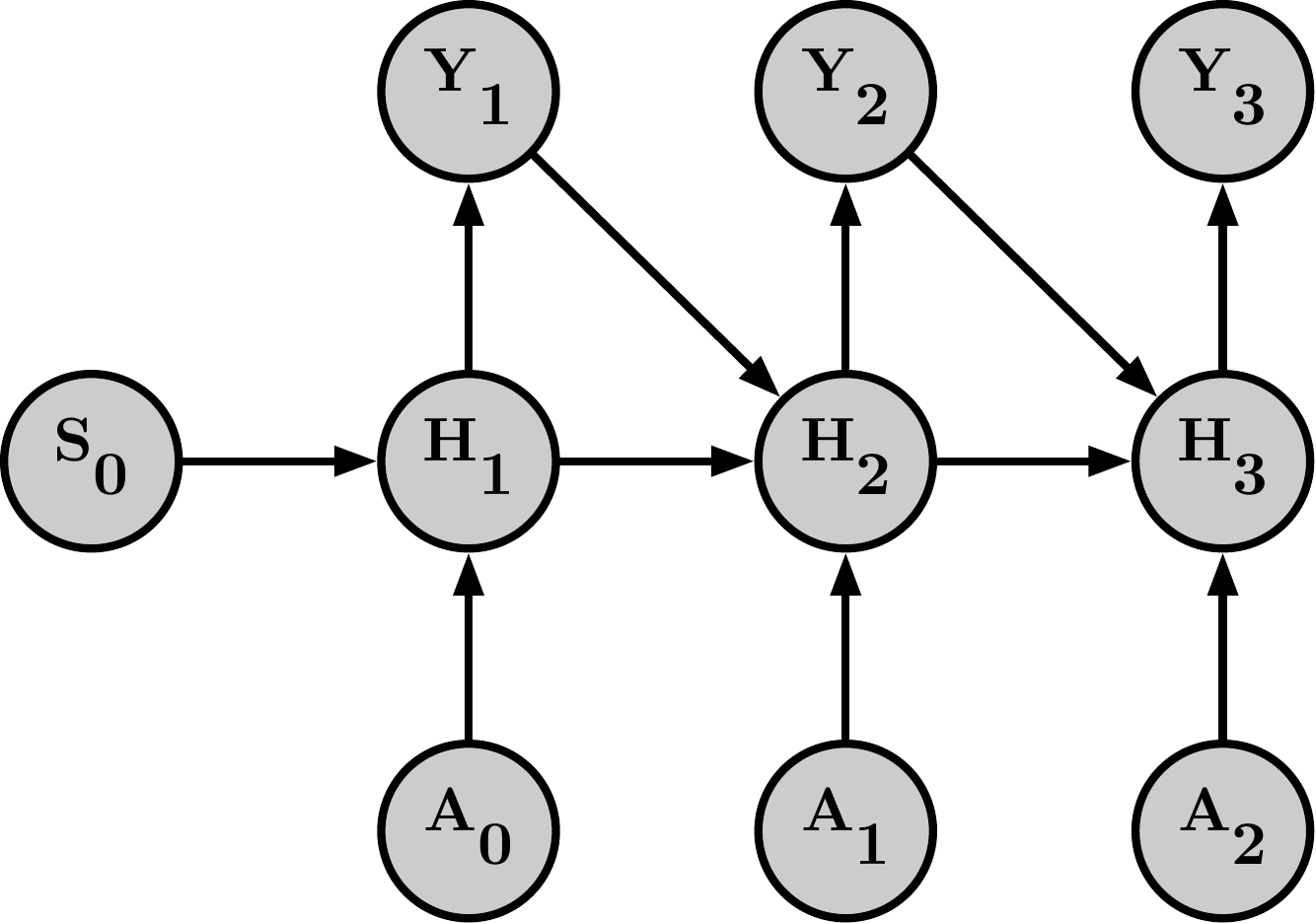}
        \caption{Standard autoregressive}
    \end{subfigure}
    \begin{subfigure}[b]{0.3\textwidth}
        \centering
        \includegraphics[width=3cm]{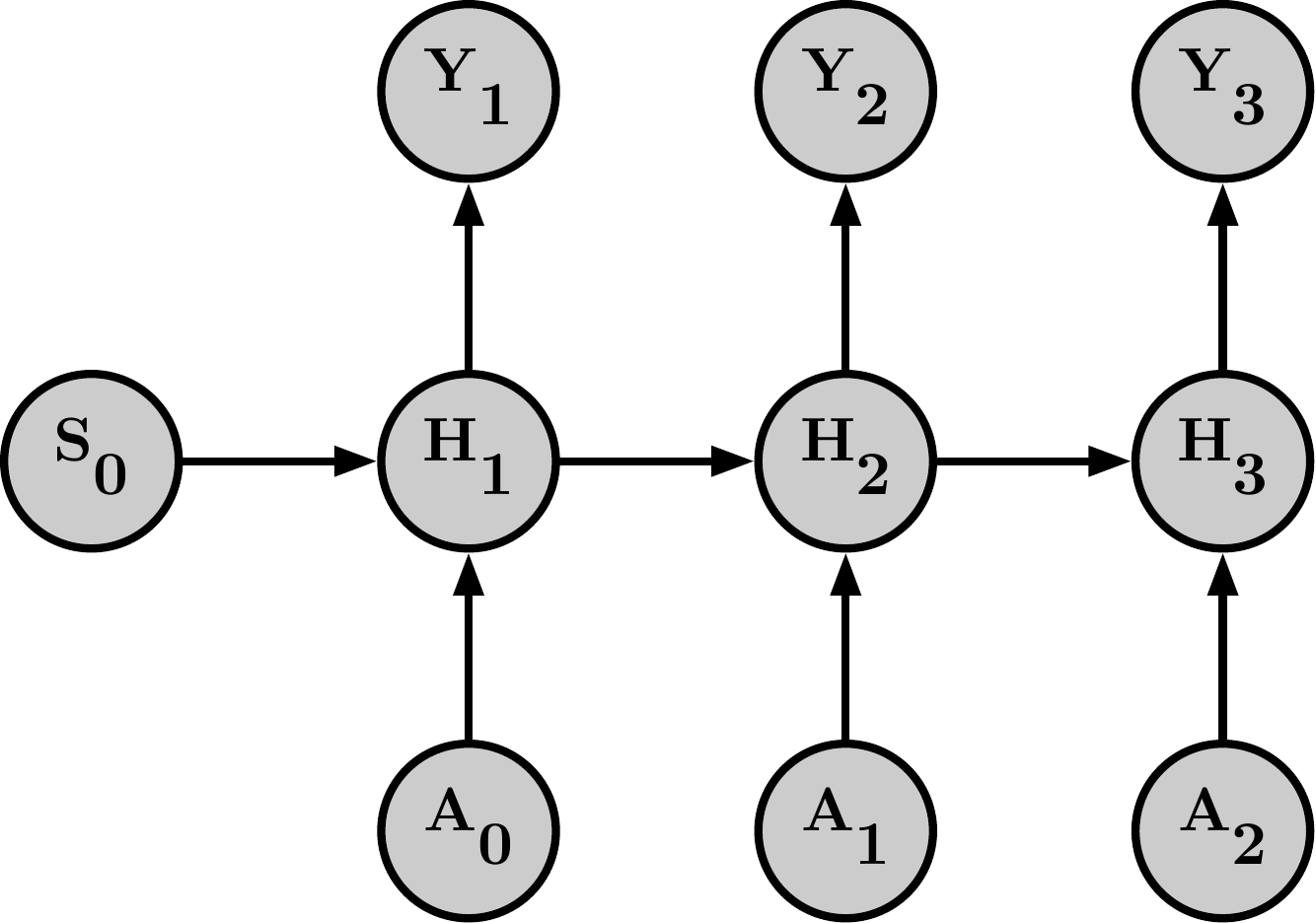}
        \caption{Non-Causal Partial Model}
    \end{subfigure}
    \begin{subfigure}[b]{0.3\textwidth}
        \centering
        \includegraphics[width=3cm]{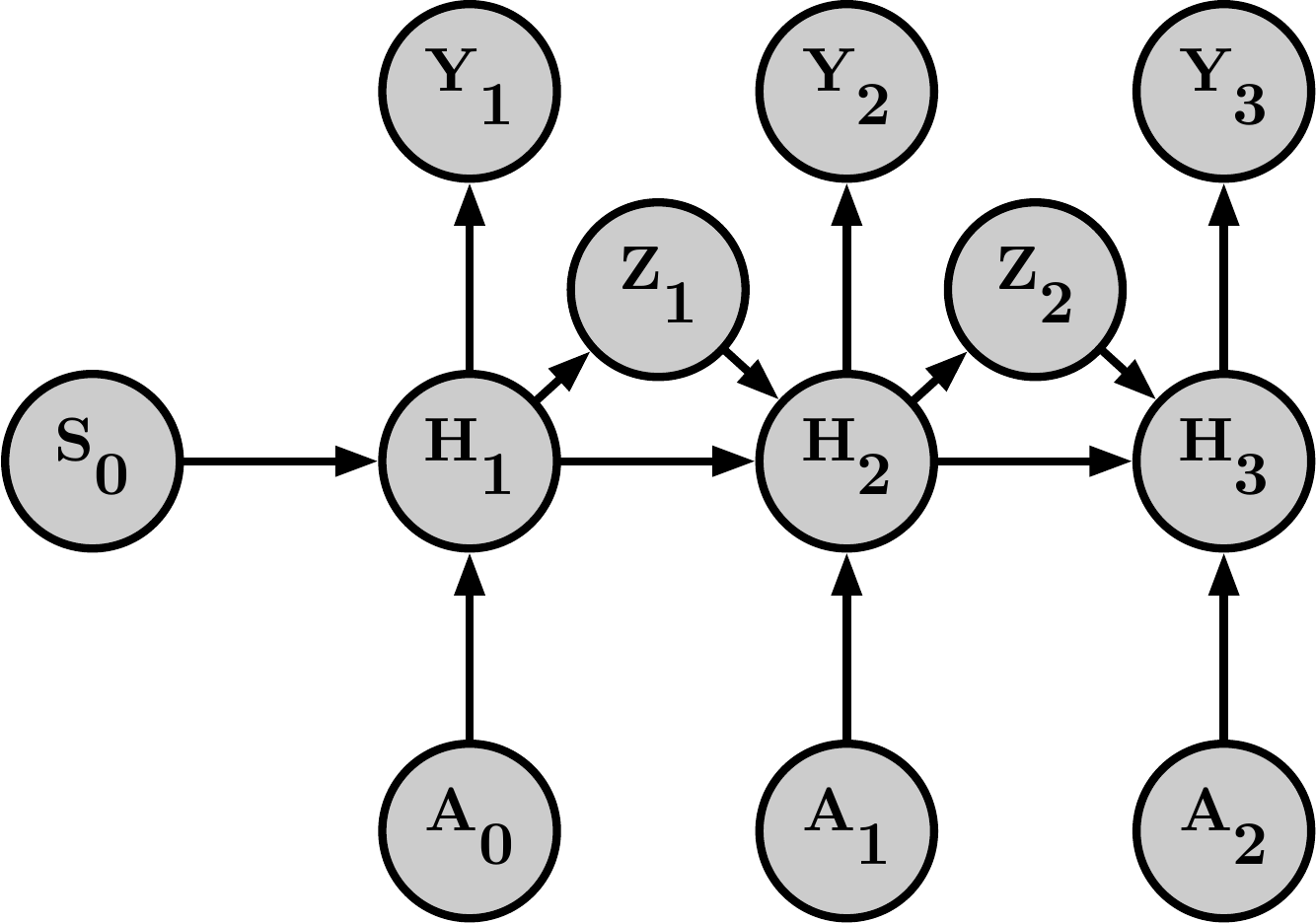}
        \caption{Proposed CPM}
    \end{subfigure}
    \caption{\textbf{Causal Partial Models (CPM) motivation \cite{rezende2020causally}}: Comparison of causal graphs behind common dynamics models in \Cref{sec:confounded_partial_models}}
    \label{fig:causal_partial_models}
\end{figure}

To remedy confounding, the authors propose to use backdoor adjustment (\Cref{sec:identifiability}) to make the action $a_t$ conditionally independent of the agent state $\vs_t$. They refer to models that are conditioned on the backdoor as \emph{Causal Partial Models} (CPM) and models that are not as \emph{Non-Causal Partial Models} (NCPM), see \Cref{fig:causal_partial_models}. They list multiple possible choices for the backdoor and discuss their trade-offs. For future work, they suggest focusing on dynamics model robustness against other types of interventions in the environment besides policy changes.

\subsection{Causal World Models}
Learned models of the environment dynamics are sometimes referred to as \emph{world models}, particularly when the dynamics are observed from a sequence of high-dimensional raw pixel frames. Fundamentally, they are still used to estimate the observational conditional of the state transition function $p\left(\mathbf{s}^{t+1} \mid \mathbf{s}^{t}, a^{t}\right)$ (analogously, the reward function).

\begin{figure}[h]
    \centering
    \begin{subfigure}[b]{0.3\columnwidth}
    \includegraphics[width=\columnwidth]{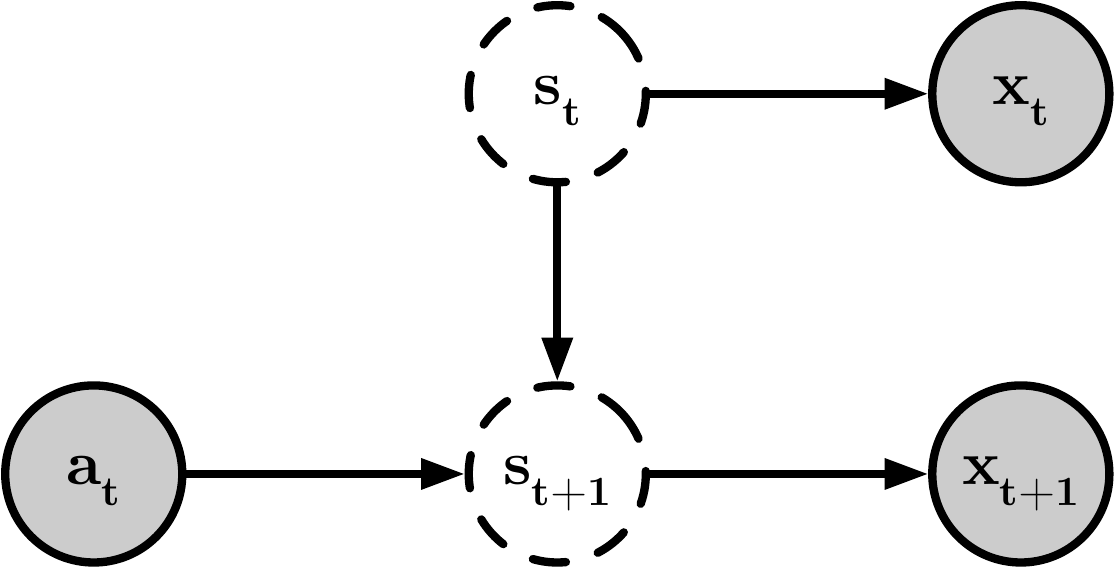}
    \caption{}
    \label{subfig:graph_noconf}
    \end{subfigure} \begin{subfigure}[b]{.3\columnwidth}
    \includegraphics[width=\columnwidth]{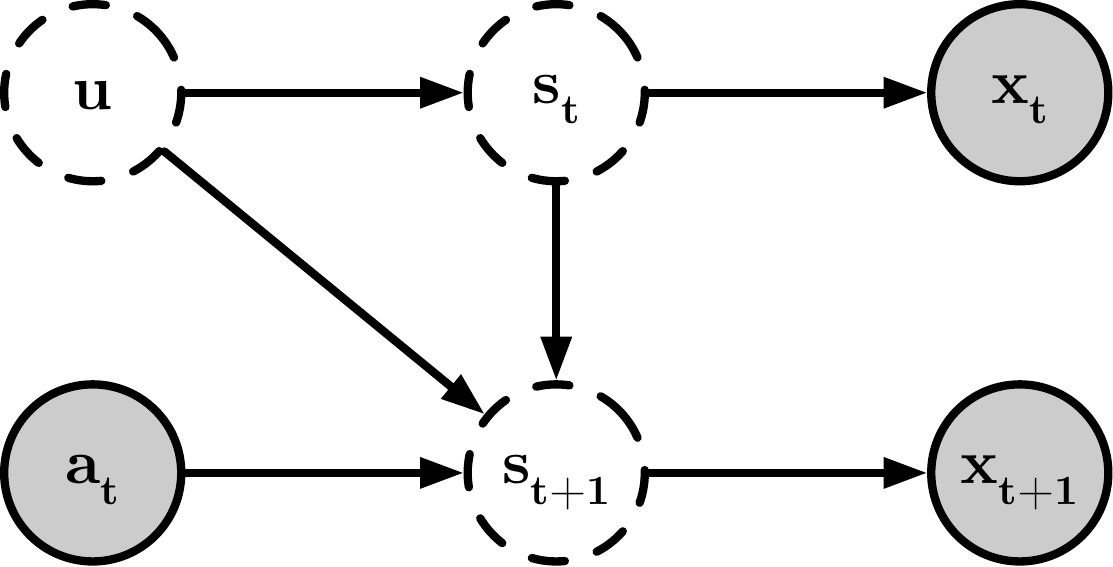} \caption{}     \label{subfig:graph_conf}
    \end{subfigure} \begin{subfigure}[b]{0.3\columnwidth}
    \includegraphics[width=\columnwidth]{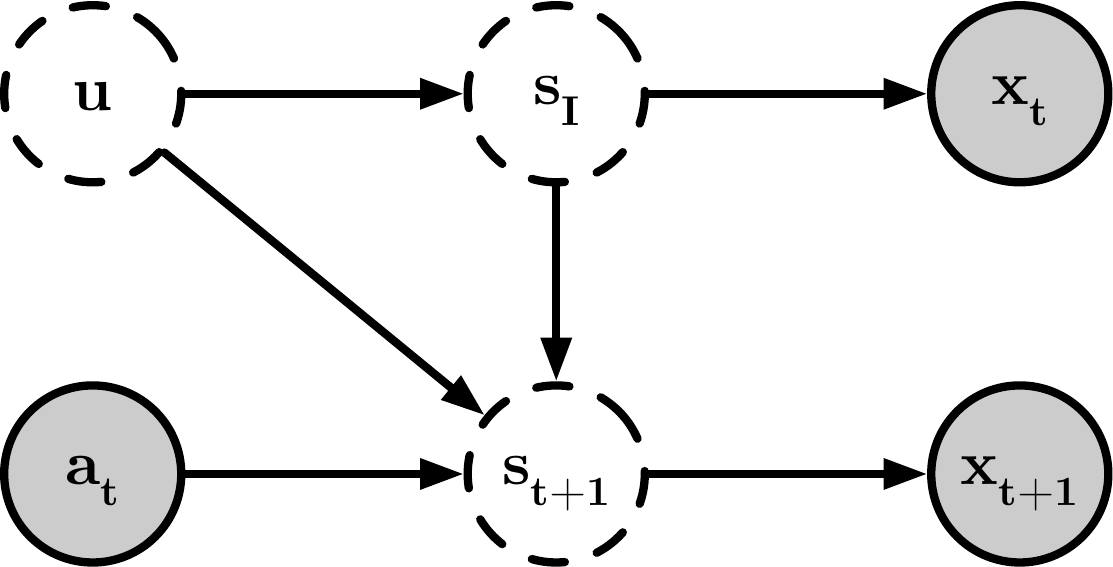}
    \caption{} \label{subfig:graph_do_s}
    \end{subfigure} \caption{The graphical model of (a) POMDPs used by conventional world models \cite{li2020causal}, (b) causal POMDPs used by causal world models (\method{CWMs}), and (c) \method{CWMs} after intervention (\textbf{do}-operation).}
    \label{fig:cwm}
\end{figure}

\citet{li2020causal} argue that such observational conditional (world) models can be biased in the real world where confounding factors exist. To make this issue apparent, they first contrast partially-observable MDPs (POMDPs) (\Cref{subfig:graph_noconf}) used by conventional models with the causal POMDPs (\Cref{subfig:graph_conf}) that their \emph{causal word model} (\method{CWM}) aims to learn from.

Given the causal POMDP, the authors define their quantity of interest as the interventional conditional $p\left(\mathbf{s}^{t+1} \mid \doo\left(\mathbf{s}^{t}=\vs_I\right), a^{t}\right)$. Then, they argue that in many real-world cases, the observational conditional and the interventional conditional are different because of the existence of confounding factors $\mathbf{u}$. The observational conditional can be written as
\begin{align}
p\left(\mathbf{s}^{t+1} \mid \mathbf{s}^{t}, a^{t}\right) &=\int_{\mathbf{u}} p\left(\mathbf{s}^{t+1} \mid \mathbf{u}, \mathbf{s}^{t}, a^{t}\right) p\left(\mathbf{u} \mid \mathbf{s}^{t}, a^{t}\right) d \mathbf{u} \nonumber \\
&=\int_{\mathbf{u}} p\left(\mathbf{s}^{t+1} \mid \mathbf{u}, \mathbf{s}^{t}, a^{t}\right) \textcolor{red}{\frac{p\left(\mathbf{s}^{t}, a^{t} \mid \mathbf{u}\right)}{p\left(\mathbf{s}^{t}, a^{t}\right)}} p(\mathbf{u}) d \mathbf{u},
\end{align}
and the interventional conditional as
\begin{align}
    p\left(\mathbf{s}^{t+1} \mid \doo \left(\mathbf{s}^{t}\right), a^{t}\right)=\int_{\mathbf{u}} p\left(\mathbf{s}^{t+1} \mid \mathbf{u}, \mathbf{s}^{t}, a^{t}\right) p(\mathbf{u}) d \mathbf{u}.
\end{align}
The difference is highlighted in \textcolor{red}{red}.

Consequently, CWMs infer the interventional query \say{Given that we have observed $\vx^{t: T}=\vx^{t: T}$ in the real world, what is the probability that $\vx^{t+1: T}$ would have been $\vx_I^{t+1: T^{\prime}}$ if $\vx^{t}$ were $\vx_I^{t}$ in the dream world?} Technically, they want to intervene upon the abstract state variable as shown in \Cref{subfig:graph_do_s}, where $\boldsymbol{s}_I^{t} \in \mathcal{S}$ is the counterfactual value in the dream environment. This intervention is then rendered as an observable change applied to $\vx^{t}$ (such as, for instance, object displacement or removal) by the conditional observation distribution $\mathcal{U}\left(\vx^{t}=\vx_I^{t} \mid \mathbf{d o}\left(\mathbf{s}^{t}=\right.\right.$ $\left.\left.\boldsymbol{s}_I^{t}\right)\right)$, where $\vx_I^{t} \in \mathcal{O}$ represents the value of the counterfactual observation.

\subsection{Task-Independent State Abstractions}
Standard, non-causal MBRL dynamics models are \emph{dense} in the sense that they predict the next step value of each variable based on the action and all variables in the current state. As such, they are sensitive to spurious associations. 

\begin{figure}
  \centering
  \begin{subfigure}[t]{0.54\columnwidth}
  \includegraphics[width=\columnwidth]{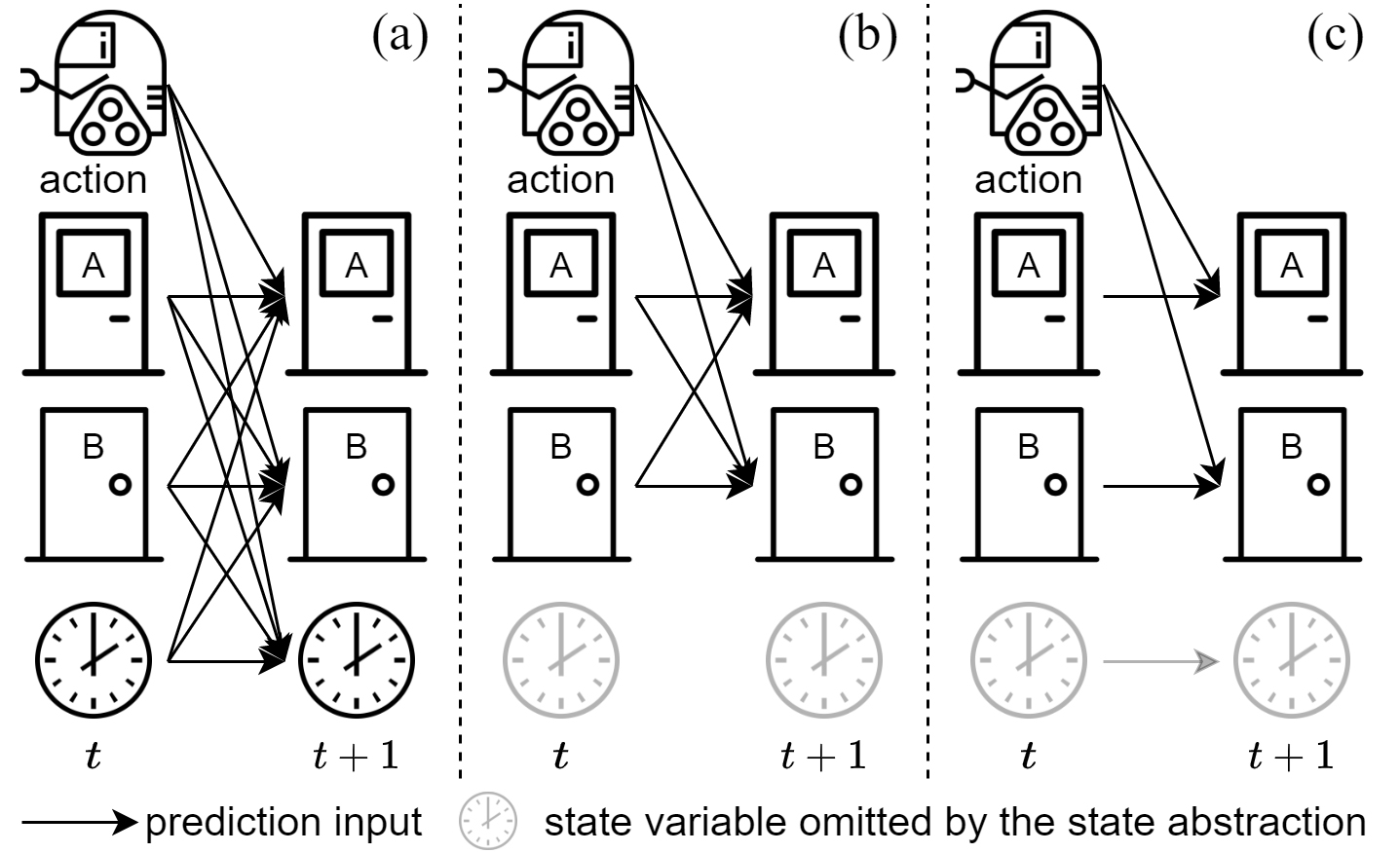}
  \caption{\textbf{Abstraction example.} Imagine an environment with two doors that the robot can open and go through and a clock on the wall. \emph{(a)}: Standard MBRL dynamics model predicts dynamics by unnecessarily using all variables. \emph{(b)}: State abstractions learn to omit the clock based on a pre-defined reward but still use a dense model for the remaining variables. \emph{(c)}: \method{CDL} only keeps necessary dependencies (doors A and B depend on the action individually) and derives a state abstraction independent from any reward function.}
  \label{fig:cdl_example}
  \end{subfigure}
  \begin{subfigure}[t]{0.45\columnwidth}
  \centering
  \includegraphics[width=\columnwidth]{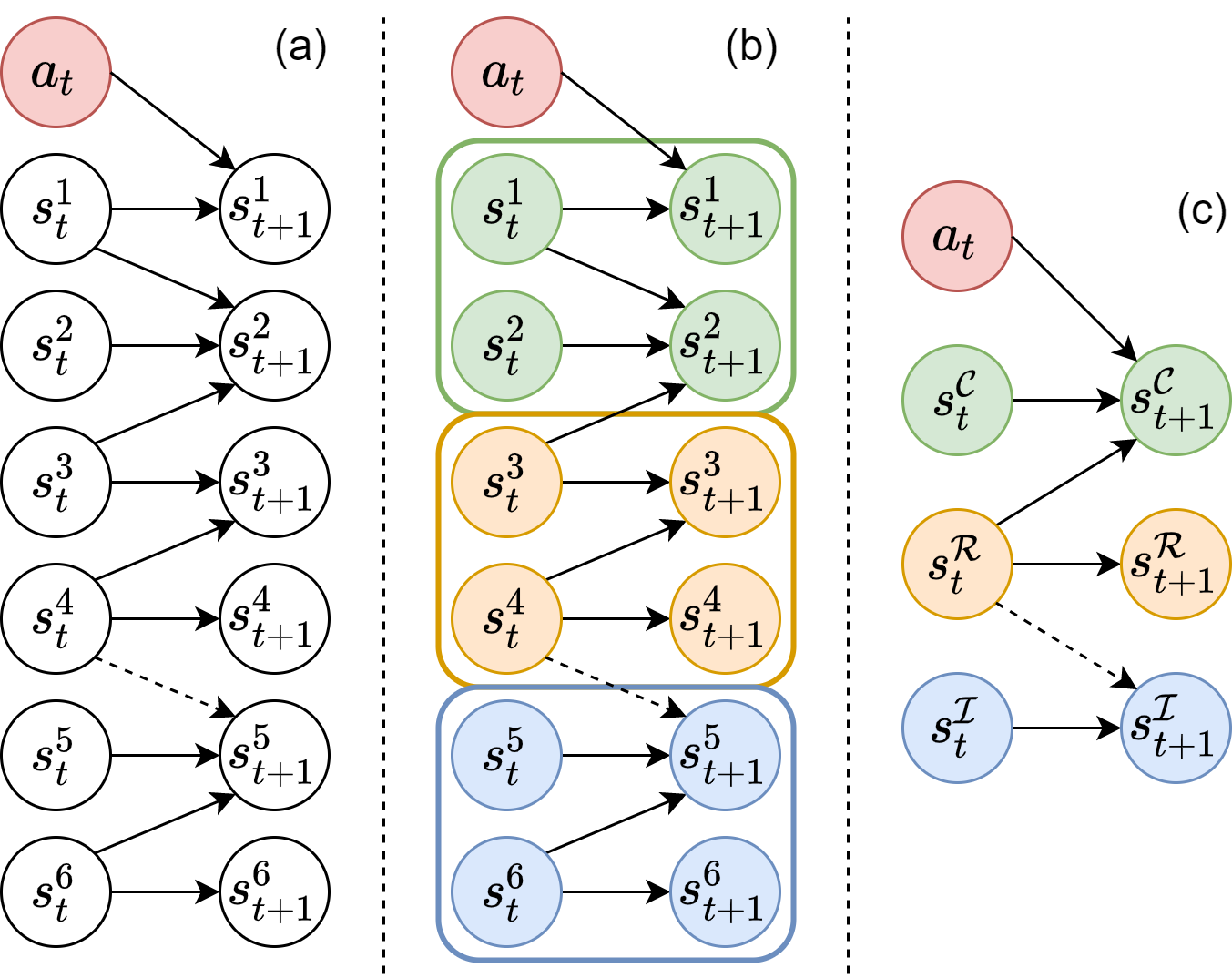}
  \caption{\textbf{State variables can be split into three types.} \emph{(a)} Complete causal dynamics model. \emph{(b)} Split: controllable (green), action-relevant (orange), and action-irrelevant (blue). The dashed arrow represents whether it exists does not affect $s^5$ to be action-irrelevant. \emph{(c)} Causal graph can be split into three subgraphs, one for each type of state variable.}
  \label{fig:cdl_illustration}
\end{subfigure}
\caption{\textbf{Causal Dynamics Learning (\method{CDL}) for Task-Independent State Abstraction \cite{cdl}.}}
\label{fig:cdl}
\end{figure}

For example, consider the example in \Cref{fig:cdl_example}, where a robot faces two doors it can open and observes a wall clock. Subfigure (a) shows a dense model. When door B is at angles unseen during training, or the clock is at unseen times, this dense model's prediction of door A can be inaccurate due to unnecessary dependence on the other variables. This issue has motivated state abstraction techniques \cite{chapman1991input,mccallum1996reinforcement}, which group many states into an abstract state by omitting some state variables, as illustrated in Subfigure (b). However, \cite{cdl} argue that the generalization issues persist, as dense models are still used for the remaining variables, leaving unnecessary dependencies in the abstract states. 

To this end, \citet{cdl} introduce \emph{Causal Dynamics Learning (CDL) for Task-Independent State Abstraction}, as illustrated in Subfigure (c). This approach learns a causal model that explains which actions and state variables affect which variables from data. In the above example, its predictions about door A do not rely on door B and the clock and are, therefore, more robust to spurious associations than dense models.

More generally speaking, if there exist certain state variables that no other variables depend on (e.g., the clock), they can be omitted for planning. This motivates a novel form of state abstraction: \citet{cdl} suggest partitioning the state variables into three groups: (i) those that the model can change with its actions (\emph{controllable variables}, e.g., doors A and B), (ii) those that it cannot change but that still influence actions' results on those that it can (\emph{action-relevant variables}, e.g., an obstacle that blocks door A's motion), and (iii) the remainder which can be omitted entirely (\emph{action-irrelevant variables}, e.g., the clock). 

As shown in \Cref{fig:cdl_illustration}, their approach represents the transition dynamics through a causal graphical model, which is then split into subgraphs corresponding to the three previously described partitions. To learn the causal dynamics model, a key challenge is to determine whether a causal edge exists between two state variables, i.e., whether $\vs_{t}^{i} \rightarrow \vs_{t+1}^{j}$ holds. For that, they leverage \citet{mastakouri2021necessary}'s conditional independence test, which relies on approximating conditional mutual information.  

To expose the causal relationships thoroughly, one has to collect trajectories with wide coverage of the state space. For that exploratory phase, they use a reward function that is the prediction difference between the dense predictor and the causal predictor learned so far: 
\begin{align}
r_{t}=\tanh \left(\tau \cdot \sum_{j=1}^{d_{\mathcal{S}}} \log \frac{\hat{p}\left(s_{t+1}^{j} \mid \vs_{t}, a_{t}\right)}{\hat{p}\left(s_{t+1}^{j} \mid \pa_{s^{j}}\right)}\right),
\end{align}
where $\tau$ is a scaling factor and $\tanh$ bounds the reward. This reward encourages the exploratory agent to take transitions where the dense predictor is better than the causal predictor, which usually suggests the learned causal graph is inaccurate.

Lastly, to solve downstream tasks, \method{CDL} simultaneously learns a transition dynamics model (including reward function) and uses a planning algorithm depending on that model for action selection, as most MBRL algorithms do. In experiments, they verify that this improves generalization and sample efficiency on the learned dynamics models and the policies for downstream tasks. 

\section{Multi-Task RL} \label{rl:multi_envs}
Multi-Task RL learning refers to settings where we expect an agent to solve multiple environments\footnote{We use task and environment interchangeably.}. Further, some works assume that the environments have never been encountered during training, but there exist some invariances across all environments (e.g., the same dynamics but observation functions with different noise sources), or there may exist an adaptation phase with limited exposure to the new task (also referred to as \emph{meta-reinforcement-learning} \cite{weng2019metaRL}).

\textbf{Causal Curiosity.} \label{sec:cc}
\citet{causal_curiosity} consider the setting in which the environment dynamics depend on hidden parameters \cite{hi_params} that differ across environments or over time. For example, if a body in an environment loses contact with the ground, the coefficient of friction between the body and the ground no longer affects the outcome of any action. Likewise, the outcome of an upward force applied by the agent to a body on the ground is unaffected by the friction coefficient. 

However, in contrast to previous approaches that learn latent task representations, e.g., \cite{PEARL, PAML}, they aim to recover \emph{disentangled} representations for the factors, i.e., independent causal mechanisms that affect the outcomes of behaviors in each environment. The motivation is that disentangled embeddings of the causal factors make the changing behaviors interpretable.

\textbf{Block MDPs.} \label{sec:block_mdps}
\citet{zhang2020invariant} consider the problem of learning state abstractions that generalize in block MDPs, families of environments in which the observations may change, but the latent states, dynamics, and reward function are the same. By leveraging ideas from \method{IRM} (\Cref{concept:irm}), they propose to learn invariant state abstractions from stochastic observations across different interventions on variables in the observation space (e.g., the background color of 3D-rendered Physics simulation). 

The block structure assumption states that each observation $\rmX$ can uniquely determine its generating state $\rmS$. Two additional causal assumptions the authors make about the causal structure of the environment are 1) the environment state at time $t$ can only affect the values of the state at time $t+1$ and the reward at time $t$, and 2) each environment corresponds to an intervention on a single variable in the observation space. For example, in one of their experiments, they intervene on the background color of the environment and set it to a value sampled at random. 

\textbf{Model-Invariant State Abstractions.}
\citet{MISA} introduce \emph{model-invariant state abstractions} for systematic generalization to unseen states in a single-task setting. These abstractions are built on two ideas: (1) \emph{causal sparsity} in the transition dynamics over state variables and (2) \emph{causal invariance} in the learned representations. (1) means that given a set of state variables, each variable only depends on a small subset of those variables in the previous timestep. (2) dictates that given a set of features, the learned representations comprise only those features that are consistently necessary for predicting the target variable of interest across different interventions. Therefore, it likely contains the true causal features and will generalize well to possible shifts in the data distribution.

\textbf{Schema Networks.}
\citet{kansky2017schema} propose \emph{Schema Networks}, a generative model for model-based reinforcement learning. Their approach relies on \emph{Schemas}, which are local cause-effect relationships involving one or more object entities. The model's knowledge about the class of environments is represented with schemas such that in a new environment, these cause-effect relationships are traversed to guide action selection. Thereby, experience from one scenario can be transfered to other similar scenarios that exhibit repeatable structure and sub-structure. For example, they demonstrate that Schema Networks are able to generalize to variations of the Atari Breakout \cite{mnih2015human} game with perturbed positions of objects, while model-free baselines fail to do so.

\textbf{Systematic Generalization.} \label{sec:sg}
Systematic generalization aims at learning universal (causal) relations of the environment dynamics from interactions with a few environments, such that we can deal with unseen states in a single-task setting, or approximately solve unseen other environment without further interactions in a multi-task setup.

\citet{mutti2022provably} define the following systematic generalization problem. First, they define a \emph{universe}: a large, potentially infinite, set $\mathbb{U}$ of environments modeled as discrete MDPs without rewards,
\begin{align}
\mathbb{U}:=\left\{\mathcal{M}_{i}=\left(\left(\mathcal{S}, d_{S}, n\right),\left(\mathcal{A}, d_{A}, n\right), P_{i}, \mu\right)\right\}_{i=1}^{\infty}.
\end{align} The agent's goal is to acquire sufficient knowledge to approximately solve any task that can be specified over the universe $\mathbb{U}$ by drawing a finite amount of interactions. 

A task is defined as any pairing of an MDP $\mathcal{M} \in \mathbb{U}$ and a reward function $r$. Solving it refers to providing a slightly sub-optimal policy via planning, i.e., without taking additional interactions. 

\begin{mydef}{Systematic Generalization \cite{mutti2022provably}}{sec:sg}
For any latent MDP $\mathcal{M} \in \mathbb{U}$ and any given reward function $r: \mathcal{S} \times \mathcal{A} \rightarrow$ $\left[0,1\right]$, the systematic generalization problem requires the agent to provide a policy $\pi$, such that $V_{\mathcal{M}, r}^{*}-V_{\mathcal{M}, r}^{\pi} \leq \epsilon$ up to any desired sub-optimality $\epsilon>0$.
\end{mydef}

Since the set $\mathbb{U}$ is infinite, the authors posit the presence of common causal structure underlying the transition dynamics of the universe. This assumption makes the problem feasible and allows them to yield a provably efficient algorithm that achieves systematic generalization with polynomial sample complexity. They verify their algorithm's effectiveness on a synthetic universe, where each environment is a person, and the MDP represents how a series of actions the person can take influences their weight and academic performance. 

\textbf{Variational Causal Dynamics.}
\label{rl:VCD} \citet{variational_causal_dynamics} learn a causally factorized latent state-space dynamics model. They sparsity regularization from causal discovery to capture sparse causal structures describing the environment dynamics. The motivation behind their model is that it can adapt to novel environments by intervening on learned latent independent mechanisms without affecting the others. 

\begin{figure}
\centering
\begin{subfigure}[t]{.32\textwidth}
  \centering
  \includegraphics[width=\columnwidth]{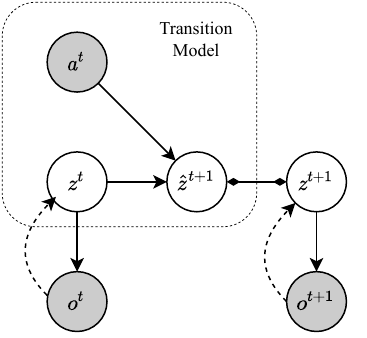}
  \caption{\textbf{Latent state space model}.}
  \label{fig:vcd_world_model_ELBO}
\end{subfigure}
\begin{subfigure}[t]{.32\textwidth}
  \centering
  \includegraphics[width=\columnwidth]{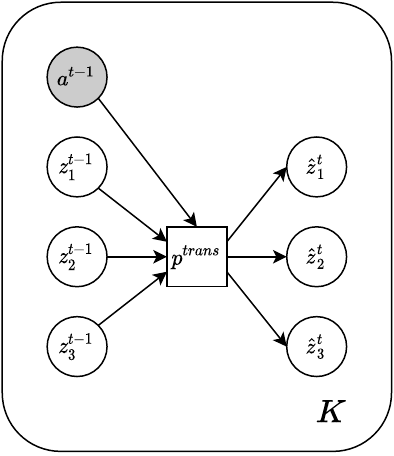}
  \caption{\textbf{Previous transition models.}}
  \label{fig:vcd_unstructured_transition}
\end{subfigure}
\begin{subfigure}[t]{.32\textwidth}
  \centering
  \includegraphics[width=\columnwidth]{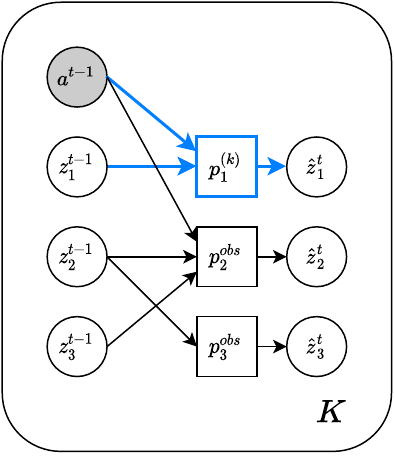}
  \caption{\textbf{VCD transition model.}}
  \label{fig:vcd_transition}
\end{subfigure}
\caption{\textbf{Variational Causal Dynamics (VCD) \cite{variational_causal_dynamics}}. The shown DAGs illustrate one timestep and $K$ environments. The VCD model combines causal discovery with learning transition dynamics, aiming to generalize across $K$ environments by identifying sparse changes in the underlying causal graph. Each latent state variable $z_i$ is represented by a separate conditional distribution $p_i$ that is conditioned on a subset of the previous states and the action at each timestep. In \textcolor{blue}{blue}, we highlight the intervened mechanism specific to environment $k$. In black, we denote mechanisms shared across environments.}
\label{fig:VCD_architecture}
\end{figure}

In contrast to the other approaches discussed in this section, they parameterize the belief over the causal adjacency matrix $\rmG$ as a random binary matrix. Each entry of that matrix follows a Bernoulli distribution with learned scalar variables. To jointly learn $\rmG$ as well as the latent intervention masks for each environment, they utilize variational inference and derive an ELBO objective. \Cref{fig:VCD_architecture} summarizes their approach. 

\section{Off-Policy Policy Evaluation} \label{rl:oppe}
The purpose of policy evaluation is to measure the expected return $\mathbb{E}_{\mathfrak{p}^{\pi}}[G]$ of a target policy $\pi$, where $G$ denote some return, e.g., the discounted return $\sum_{k=0}^{\infty} \gamma^{k} R_{t+k+1}$ at time step $t$ with discount rate $\gamma \in [0,1]$ and reward $R$. This evaluation uses sample trajectories $\gD=\left\{\hat{\vh}_{T}^{i}\right\}_{i=1, \ldots, N}$ consisting of logged episodes $\hat{\vh}_{T}^{i}=\left(\hat{\vx}_{1}^{i}, \hat{a}_{1}^{i}, \ldots \hat{a}_{T-1}^{i}, \hat{\vx}_{T}^{i}\right)$ generated either by the same policy $\pi$ (on-policy) or by another policy $\rmU$ (off-policy). When using the latter, we refer to the evaluation as \emph{off-policy policy evaluation} (OPPE) and the policy used to generated the behavior is called the \emph{behavior policy}. 

In principle, OPPE is an attractive problem with many potential use cases in which online learning is not feasible, e.g., due to cost or ethical constraints on experimentation. From a technical perspective, it can guide policy learning methods to reuse off-policy experiences and find good policies more data-efficiently. However, it is generally difficult because there is a distributional mismatch between the trajectories generated by the target policy and the behavior policy. This mismatch often causes policy evaluation methods to have high variance and slow convergence \cite{intro_to_rl}. 

Interestingly, \citet{bannon2020causality} emphasize that the causal inference task \emph{counterfactual inference} (CFI) and OPPE are two different approaches to similar (under certain conditions identical) problems. 

\begin{cp}{Off-Policy Policy Evaluation \cite{bannon2020causality}}{oppe}
\emph{Observations}, \emph{interventions} and the \emph{query variable} in CFI directly correspond to \emph{off-policy episodes}, the \emph{target policy}, and the \emph{expected return}, respectively. Similarly, \citet{generalizing_ope} argue that formulating the task of OPPE from a generalized and causal perspective opens the possibility for \emph{counterfactual} or \emph{retrospective} off-policy evaluation at the level of individual units (e.g. patient-level) of the population. 
\end{cp}

Observing this correspondence begs the question if the techniques developed in both fields are complementary. In the following, we discuss methods that answer in the affirmative.

\subsection{Counterfactual Policy Evaluation} \label{sec:cf_gps}
\citet{cf_gps} present \emph{Counterfactually-Guided Policy Search} (\method{CF-GPS}), which uses counterfactual inference (\Cref{concept:counterfactuals}) for off-policy evaluations (CF-PE) in the setting of \emph{model-based (off-policy) policy evaluation} (MB-PE). In this setting, we want to evaluate the policy on synthetic data sampled from a model $\mathcal{M}$, i.e., we can estimate the expected return $\mathbb{E}_{p^{\pi}}\left[G\right]$ by sampling the \emph{scenarios} $\rmU \sim p_{\vu}$ (all aspects of the environment that cannot be influenced by the agent, e.g., the initial state distribution), and then simulating a trajectory $\tau$ from the functions $f_{i}$ and computing its return. 

Assuming that we use an SCM $\mathcal{M}_{\text{SCM}}$ instead of a statistical model $\mathcal{M}_{\text{ST}}$, \citet{cf_gps} show that CF-PE should be less biased than MB-PE. First, they show that one can represent any given partially-observable Markov decision process (POMDP) under a policy $\pi$ as SCM $\scm$ over trajectories $\gT$ (for details on POMDPs, we refer the reader to \cite{intro_to_rl}). 

\begin{figure}
    \centering
    \includegraphics[width=10cm]{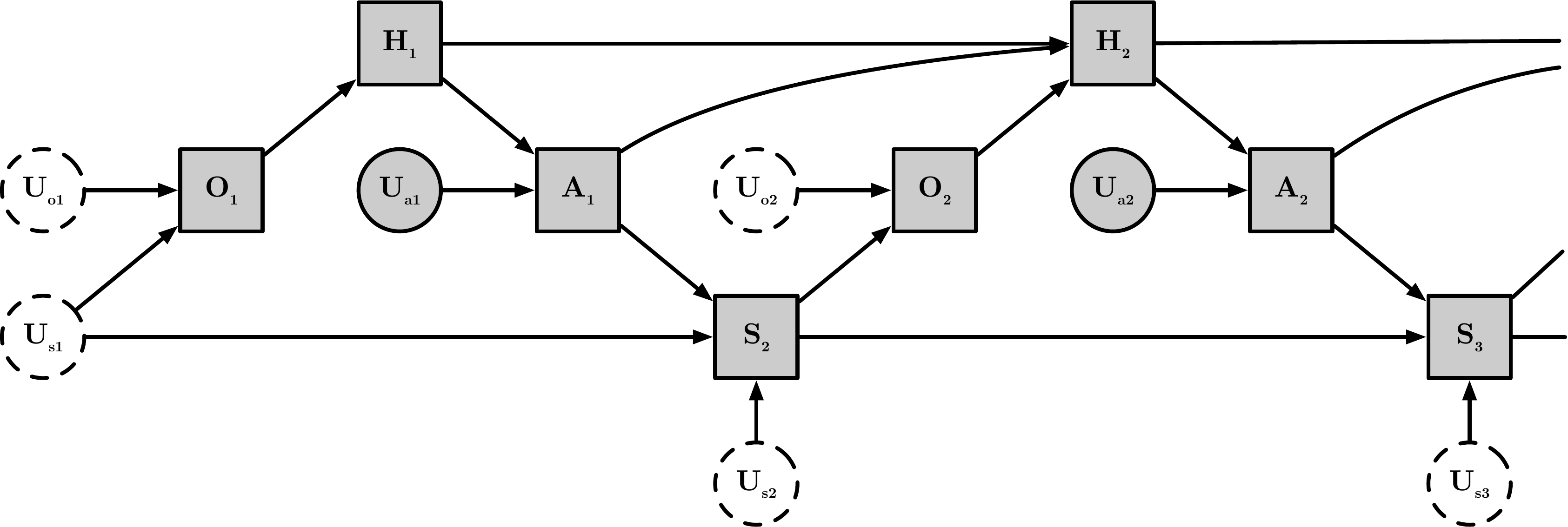}
    \caption{
    \textbf{Representing a POMDP as an SCM \cite{cf_gps}:} We denote initial state $\rmU_{\vs_1}=\rmS_1$, states $\rmS_t$ and histories $\rmH_t$. The mechanism that generates the actions $A_t$ is the policy $\pi$. Scenarios $\rmU$ summarize immutable aspects, some of which are observed (grey), some not (white).} 
    \label{fig:POMDP_AS_SCM}
\end{figure}

\begin{cp}{POMDPs \cite{cf_gps}}{pomdps}
We can represent any given Partially-Observable MDP (POMDP) by an SCM $\scm$ over trajectories $\mathcal{T}$ in the following way. By exploiting ideas related to the reparameterization trick commonly used in variational inference (e.g., VAEs) \cite{original_vae}, we can express all conditional distributions, e.g. the state transitions  $p\left(\rmS_{t+1} \mid \rmS_{t}, A_{t}\right)$, as deterministic functions with independent noise variables $\rmU$, such as $\rmS_{t+1}=f_{st}\left(\rmS_{t}, A_{t}, \rmU_{st}\right)$. The DAG of $\scm$ is shown in \Cref{fig:POMDP_AS_SCM}.
\end{cp}

Based on this representation, they then prove that counterfactual inference (\Cref{concept:counterfactuals}) yields an unbiased estimator of $p^{\doo(I)}(\vx)$, the POMDP's distribution over trajectories after intervention $I$ took place. In contrast, any bias in a statistical model $\mathcal{M}_{\text{ST}}$ propagates from $p^{\pi}$ to the estimate $\mathbb{E}_{p^{\pi}}[G]$. 

\Cref{alg:everything_cf} summarizes the \method{CF-GPS} procedure. Given data $\gD \sim \mathfrak{p}^{\vu}$ and assuming no model mismatch, i.e. $\mathfrak{p}^{\vu}=p^{\vu}$, we can regard the task of off-policy evaluation of $\pi$ as a counterfactual query with data $\hat{h}_{T}^{i}$, intervention $I(\rmU \rightarrow \pi)$ and query variable $G$ (reward). Then, the difference to MB-PE is that instead of sampling from the prior $\vu^{i} \sim p(\vu)$, we sample scenarios from the posterior $\vu^{i} \sim p^{\vu}\left(\vu \mid \hat{h}_{T}^{i}\right)$ where we inferred scenarios $U$ in hindsight from given off-policy data $\hat x_o$. Then, we evaluate the agent on these specific scenarios. 

Sampling from the posteriors $N^{-1} \sum_{i=1}^{N} p^{\vu}\left(\vu \mid \hat{h}_{T}^{i}\right)$ has access to strictly more information than the prior $p(\vu)$ by taking into account additional data $\hat{h}_{T}^{i}$ (\Cref{concept:counterfactuals}). This semi-nonparametric distribution can help to de-bias the model by effectively winnowing out parts of the domain of $\rmU$ which do not correspond to any real data.

\begin{algorithm}[t] \small
    \caption{Counterfactual Policy Evaluation \cite{cf_gps}}
    \label{alg:everything_cf}
    \begin{algorithmic}[1] 
        \Statex // Counterfactual inference (CFI)
        \Procedure{CFI}{data $\hat \vx_o$, SCM $\scm$, intervention $I$, query $\rmX_q$}
            \State $\hat \vu\sim p(\vu\vert\hat \vx_o)$ \Comment{Sample noise variables from posterior}
            \State $p(\vu)\leftarrow \delta (\vu-\hat \vu)$ \Comment{Replace noise distribution in $p$ with $\hat \vu$}
            \State $f_i\leftarrow f_i^I$ \Comment{Perform intervention $I$}
            \State \textbf{return} $\vx_q\sim p^{\doo(I)}(\vx_q\vert \hat \vu)$ \Comment{Simulate from the resulting model $\scm^I_{\hat \vx_o}$}
        \EndProcedure    
        \newline
        \Statex // Counterfactual Policy Evaluation (CF-PE)
        \Procedure{CF-PE}{SCM $\scm$, policy $\pi$, replay buffer $\gD$, number of samples $N$}
            \For{$i\in \lbrace 1,\ldots N\rbrace$} 
                \State $\hat \vh^i_T\sim D$  \Comment{Sample from the replay buffer}
                \State $g_i=\mathrm{CFI}(\hat \vh^i_T, \scm, I(\rmU\rightarrow\pi), G)$ \Comment{Counterfactual evaluation of return}
            \EndFor
            \State \textbf{return} $\frac{1}{N}\sum_{i=1}^N g_i$ \Comment{Return averaged counterfactual return}
        \EndProcedure
    \end{algorithmic}
\end{algorithm}

After demonstrating that CF-PE outperforms MB-PE in partially observed grid-world settings, they conclude that one may expect CF-PE to outperform MB-PE when the transition and reward kernels $f_{s t}$ are accurate models of the environment dynamics, but the marginal distribution over the noise sources $P_{U}$ is difficult to model. 

Following a similar methodology, \cite{counterfactual_oppe_gm} use counterfactually-generated trajectories to highlight episodes where the target and behavior policy returns differ substantially. They interpret this as a helpful procedure for off-policy \say{debugging} in high-risk settings, such as healthcare. In contrast to \cite{cf_gps}, who use counterfactuals to approximate draws from the interventional distribution, they treat the counterfactual distribution as the primary object of interest and demonstrate their method's utility in a sepsis management simulation. 

\subsection{Unobserved Confounding} \label{chapter:rl_confounding}
In OPPE, the caveat of not using the current policy for evaluation is that we almost inevitably face unobserved confounders (\Cref{sec:confounding})
that causally affect the behavior policy. \citet{confounding_robust} illustrate the problem with the following clinical example: Suppose we want to compare the efficacy of different drugs. During normal clinical practice, we observe the outcomes (rewards) of those prescribed the drug. A drug may appear less clinically effective if those prescribed it were more unhealthy to begin with and, therefore, would have had less successful outcomes regardless. Conversely, if the drug was given only to patients who would benefit most, it could be mistakenly regarded as beneficial to everyone. While these issues can potentially be addressed by controlling for more factors that may have affected treatment decisions, they can never be entirely eliminated. Healthcare databases are often incomplete about medical histories, patient severity notes, etc., so they are especially susceptible to unobserved confounding.  

To this end, \citet{confounding_robust} study the setup of \emph{confounding-robust policy improvement} (including OPPE) to account for possible unobserved confounding (UC). They develop a method for minimizing the worst-case estimated regret of a candidate policy against a baseline policy over a set of propensity weights that control the extent of UC. As a result of their theoretical analysis, they obtain generalization guarantees that ensure that their policy will be safe when put into practice. Further, it will yield the best possible uniform control on the range of all possible population regrets consistent with UC's extent.

\textbf{Infinite-Horizon.}
\citet{crpe_inf} and \citet{bennett2021off} consider the above setup extended to the infinite-horizon setting, as e.g., commonly considered in continuous control of physical systems \cite{lillicrap2015continuous} or quantitative trading \cite{charpentier2021reinforcement}. Similarly, \citet{namkoong2020off} study the case when UC occurs only at a single time step, as when an expert makes an initial decision based on unrecorded information and then follows a set of protocols based on well-recorded observations.

\textbf{POMDPs.}
\citet{OPPE_POMDP} consider OPPE in partially-observable MDP (POMDP) environments, where the unobserved variables may have confounding effects, motivating them to propose the \emph{decoupled POMDP} model, which is a class of POMDPs for which observed and unobserved variables are distinctly partitioned. \citet{proximal_RL} utilize the framework of proximal causal inference to reveal POMDP settings where identification of the target policy value is possible. Further, they construct semi-parametrically efficient estimators for these settings.

\textbf{Combining Offline and Online Data.}
\citet{gasse2021causal, wang2021provably} focus on using offline data to warm start online RL. \citet{gasse2021causal} suggests learning a latent-based transition model that explains both the interventional and observational regimes and then inferring the standard POMDP transition model. \citet{wang2021provably} propose \emph{confounded MDPs}, which naturally captures both the offline and online setting as well as their mismatch due to confounding. They then construct deconfounding algorithms in the episodic setting with linear function approximation.  

\begin{figure}
    \centering
    \hspace*{\fill}
    \begin{subfigure}[t]{0.32\columnwidth}
    \includegraphics[width=0.94\columnwidth]{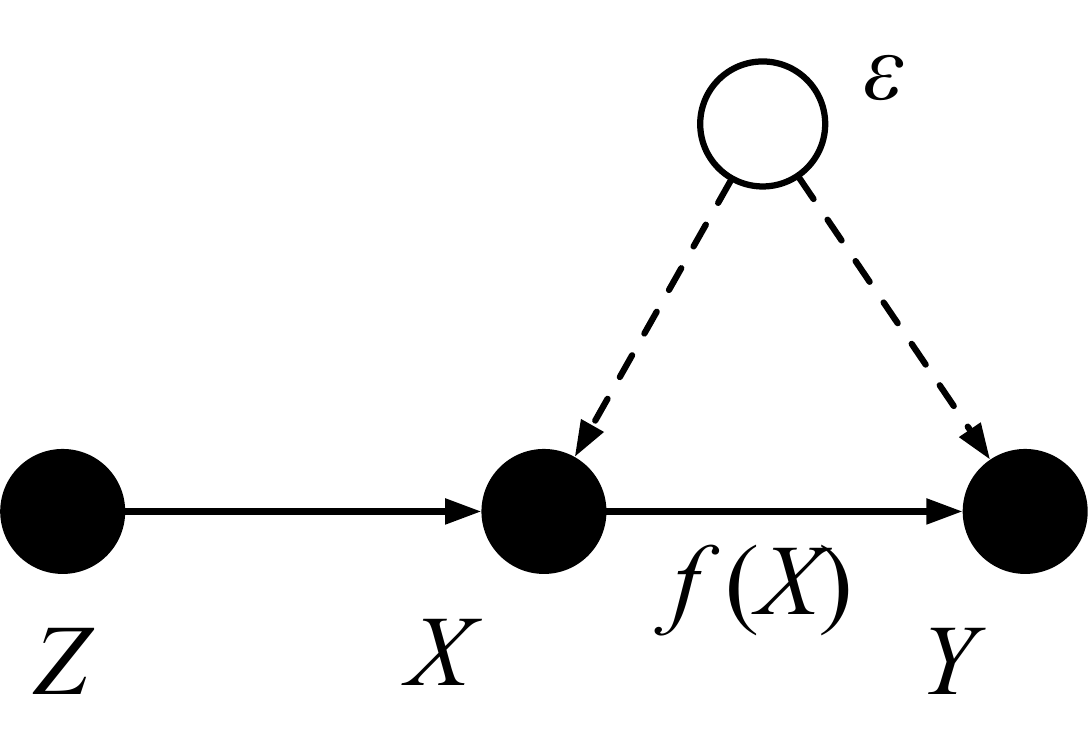}
    \caption{IVR.}
    \end{subfigure}
    \hfill
    \begin{subfigure}[t]{0.32\columnwidth}
    \includegraphics[width=\columnwidth]{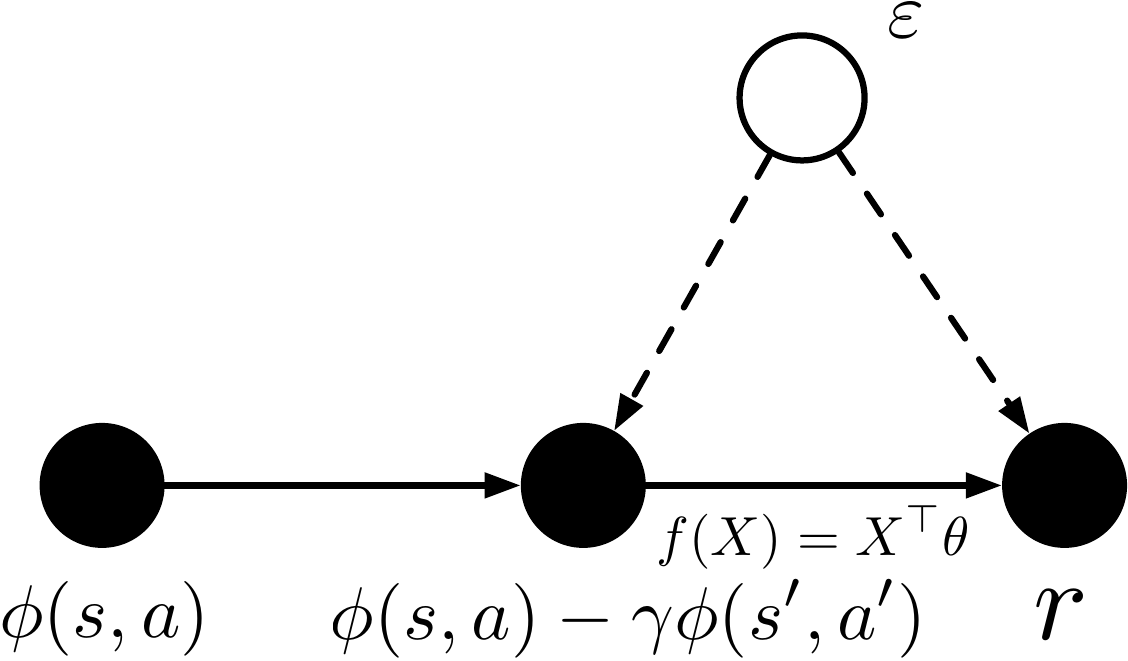}
    \caption{LSTD.}
    \end{subfigure}
    \hfill
    \begin{subfigure}[t]{0.32\columnwidth}
    \includegraphics[width=\columnwidth]{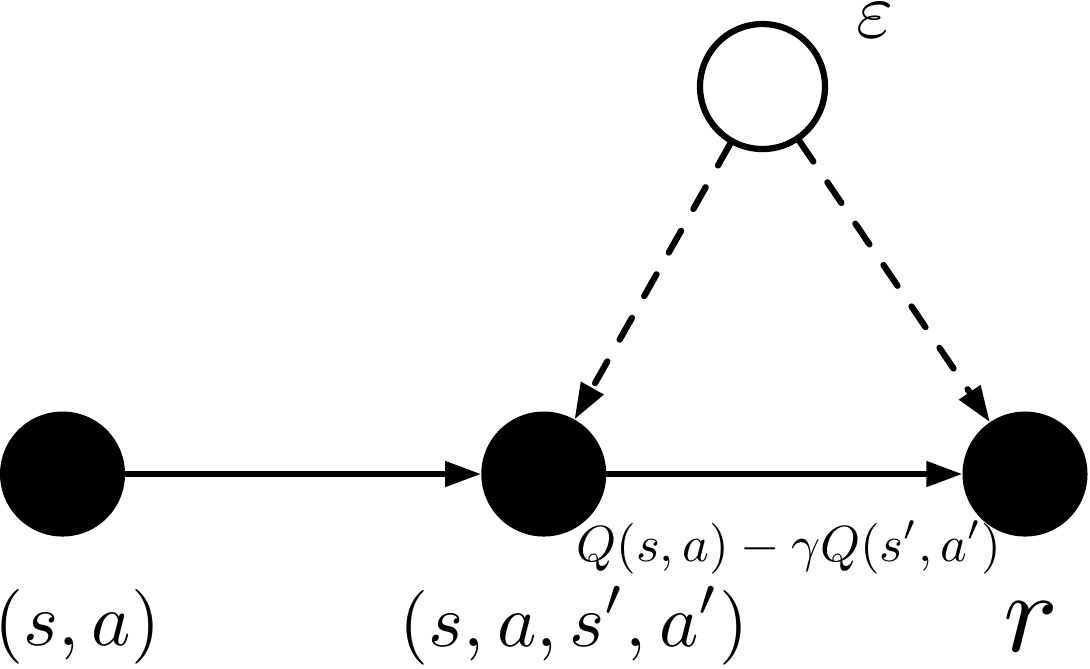}
    \caption{Non-Linear Q-Function.}
    \end{subfigure}
    \hspace*{\fill}
    \caption{\textbf{Causal DAGs revealing the relationship between Instrumental Variable Regression (IVR), Least-Squared Temporal Difference (LSTD) and non-linear Q-functions \cite{ivr_ope}.}}
    \label{fig:ivr_ope}
\end{figure}

\textbf{Relationship to Instrumental Variable Regression.}
Many OPPE rely on estimates of the state-action value (Q-) function by minimizing the mean squared Bellman error. \citet{ivr_ope} reveal that this strategy leads to a regression problem with an unobserved confounder between the inputs and the output noise: \Cref{fig:ivr_ope} displays that the causal relationships between instrumental variable regression, the least-squared temporal difference (linear Q-function) and non-linear Q-functions are equivalent.

\section{Imitation Learning}\label{rl:il}
Imitation learning (IL) aims to learn control policies directly from examples of demonstrations provided by human experts \cite{hussein2017imitation}. The motivation behind it is to remove the need for extensive interaction with the environment during policy learning and/or designing reward functions specific to the task. Cameras and sensors of today collect and transmit vast amounts of data quickly, and processors with significant computing power are getting quicker in mapping the sensory inputs to actions. Hence, developing real-time perception and reaction models assisted through IL opens up many potential applications, such as humanoid robots, self-driving vehicles, or human-computer interaction systems.

One challenge in IL is confounding. In the following, we consider confounded settings in which it is difficult for the imitator to recover the expert's performance. At their core, these issues stem from similar roots as the ones we have seen in the subsection on OPPE (\Cref{rl:oppe}), where one aims at learning policies based on trajectories that were not controlled by the agent itself but an external entity. However, as the goal of IL is different from OPPE, different remedies have been developed. 

By and large, \emph{causal} imitation learning (CIL) aims at solving confounding issues. The following works aim at deconfounding observed expert trajectories such that imitator policies achieve similar performances or define necessary and testable conditions for such. 

\subsection{Causal Confusion} \label{sec:causal_misidentification}
\citet{causal_confusion} study the setup of \emph{causal confusion}: in which the inputs to the expert policy are fully observed; however, the mechanisms of the expert policy are latent, i.e., it is not known which of the observed input variables are actual causes of the expert's actions and which are not (i.e., \say{nuisance variables}). Here, the expert and imitator observe the same contexts, but the causal diagram is not available to the imitator.

The authors show that in this setting, the phenomenon of \emph{causal misidentification} (CM) can occur: if there is a shift between inputs causing the expert's actions and the imitator's actions (e.g., when the latter relies on nuisance variables), then access to more data can yield worse performance. In other words, CM happens when cloned policies fail by misidentifying the true causes of expert actions, and training them on more observations can exacerbate the performance. 

\begin{figure*}
    \centering
    \includegraphics[width=\textwidth]{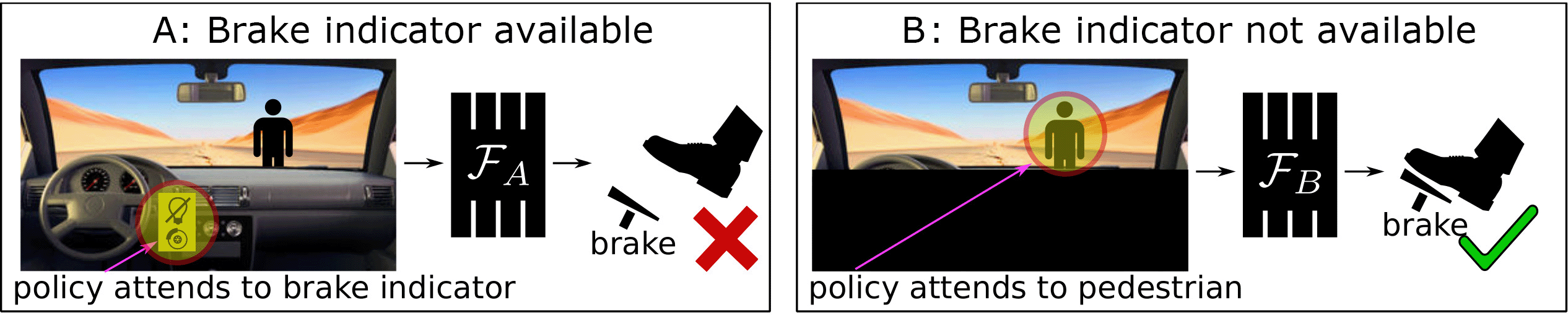}
    \caption{\textbf{Causal misidentification: more data can yield worse imitation learning performance  \cite{causal_confusion}.} There exists a spurious association between braking and the car's brake indicator. A naive behavior-cloned model ($\gF_A$) relying on the braking indicator will apply the brake only when the brake light is on. This model works well in the left Scenario A; however, if the braking indicator is unavailable (Scenario B, right), the imitator will fail. A better model ($\gF_B$) decides whether to brake by attending to the pedestrian.}
    \label{fig:causal_confusion}
\end{figure*}

Consider \Cref{fig:causal_confusion} as an example: we aim to train a neural network to drive a car. The model input for scenario A is an image of the dashboard and windshield. The input to the model (with identical architecture) is the same image, but the dashboard is masked out. Although both cloned policies achieve low training loss, model B performs well when tested on the road, while model A does not. The dashboard has an indicator light that comes on immediately when the brake is applied, and model A wrongly learns to apply the brake only when the light is on. Even though the brake light is an effect of braking, model A could achieve low training error by misidentifying it as a cause.

To remedy these issues, the authors suggest finding the true causal model of the expert's actions. A two-stage pipeline can automate this: in stage 1, we jointly learn policies corresponding to various causal graphs; in stage 2, we perform targeted interventions to search the hypothesis set for the correct causal model efficiently.

\citet{tien2022study} provide a systematic study of causal confusion in the context of learning reward functions from human inputs that take the form of pairwise preferences or rankings. Specifically, they demonstrate across three different robot learning tasks that causal confusion over the true reward function occurs, even with large numbers of pairwise preferences over trajectories. The authors investigate different factors affecting causal confusion: type of training data, reward model capacity, and the preference training data generation mechanism. One of their conclusions is that while human demonstrations seeking to be pedagogic lead to better sample efficiency, all the preference data collection methods were prone to causal confusion.

\subsection{Self-Delusion}\label{il:self_delusion}
\citet{ortega2021shaking} demonstrate that the common perception that popular sequence models \say{do not understand the cause and effect of their actions} is the result of conditioning on actions rather than treating them as causal interventions. The reason is that the model update differs depending on whether the collected data originated from within the model (i.e., from actions) or outside of it (i.e., from a third person whose policy we do not know), and mixing them up results in incorrect inferences.

\begin{figure}
    \centering
    \includegraphics[width=9cm]{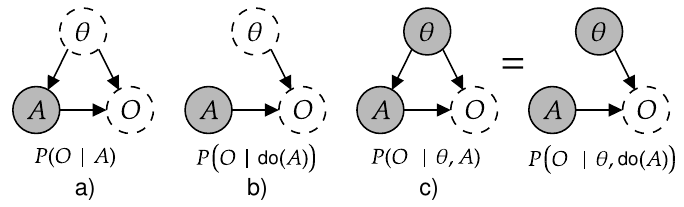}
    \caption{\textbf{The Self-Delusion problem illustrated through a prize-or-frog game \cite{ortega2021shaking}:} Panel~(a) illustrates the self-delusion problem: conditioning on the self-generated action leads to wrong inferences about~$O$, since~$A$ and~$O$ are confounded by~$\theta$. Panel~(b) shows that treating the self-generated action as a causal intervention circumvents the self-delusion because no information can flow backward from~$A$ into~$\theta$. Panel~(c) corresponds to the fully observable case. When~$\theta$ is observed, conditioning or intervening on the self-generated action~$A$ leads to the same prediction over~$O$.}
    \label{fig:prize_or_frogs}
\end{figure}

To give intuition, the authors give a minimal example, as illustrated in \Cref{fig:prize_or_frogs}. Consider a prize-or-frog problem, where there are two boxes $(1 \& 2)$, one containing a prize and the other a frog, respectively. The objective is to open the box containing the prize. For simplicity, let us assume that the two configurations are equiprobable. The true DGP has a joint distribution $p(\theta, A, O)$ over the box configuration $\theta$, the chosen box $A(1$ or 2$)$, and the observed content $O$ ($\pm 1$ reward). 

Suppose we learn a probabilistic model from data generated by an expert who opens the correct box when told where the prize is. Crucially, unlike the expert, we would not observe $\theta$. The source of the delusion can be seen by comparing the two posterior beliefs \(p(\theta \mid a, o)\) and \(p(\theta \mid \doo(a), o)\) over the task parameter, that is, where the past action is treated as a condition and an intervention, respectively (differences are highlighted):
\begin{align}
p(\theta \mid a, o&)=\frac{p(\theta) \textcolor{red}{p(a \mid \theta)} p(o \mid \theta, a)}{\sum_{\theta^{\prime}} p\left(\theta^{\prime}\right) \textcolor{red}{p\left(a \mid \theta^{\prime}\right)} p\left(o \mid \theta^{\prime}, a\right)} \\ 
p(\theta \mid \doo(a), o)&=\frac{p(\theta) p(o \mid \theta, a)}{\sum_{\theta^{\prime}} p\left(\theta^{\prime}\right) p\left(o \mid \theta^{\prime}, a\right)}.
\end{align}
Here, we can see that the intervention causes the dismissal of the evidence $\textcolor{red}{p(a \mid \theta)}$ produced by the self-generated action $A=a$.

\subsection{Unobserved Confounding} \citet{causal_imitation_learning} tackle the setting in which a causal graph is provided, but \emph{unobserved} confounders (UC) affect both actions and outcomes of the expert demonstrations. This means that the expert's input observation may differ from the ones available to the imitator. For instance, self-driving cars rely solely on cameras or lidar, ignoring the auditory dimension. However, most human demonstrators can exploit this data, especially in dangerous situations (car horns, screeching tires) \cite{kumor2021sequential}.

To formalize and study this problem in single-stage settings, they introduce \emph{partially observable SCMs}, a particular type of SCMs that explicitly allows one to model the unobserved nature of some endogenous variables. Then, they show that the problem of imitability is orthogonal to identifiability and introduce a data-dependent, graphical criterion for determining whether imitating an expert's performance is feasible. Based on this criterion, they provide an algorithm that checks whether there are instruments in the data, permitting it to satisfy the graphical imitability criterion. Finally, for the case of the criterion being met, they propose a procedure for estimating an imitating policy.

\subsection{Sequential Data} \label{cil_seq} \citet{kumor2021sequential} extend \citet{causal_imitation_learning}'s results to sequential settings, where the imitator must make multiple decisions per episode. To this end, they introduce a generalized backdoor criterion that allows one to learn imitating policies across a sequence of states and actions. They prove that their criterion is \emph{necessary} for imitability.  

\subsection{Multiple Environments} 
\begin{figure}
    \centering
    \includegraphics[width=\columnwidth]{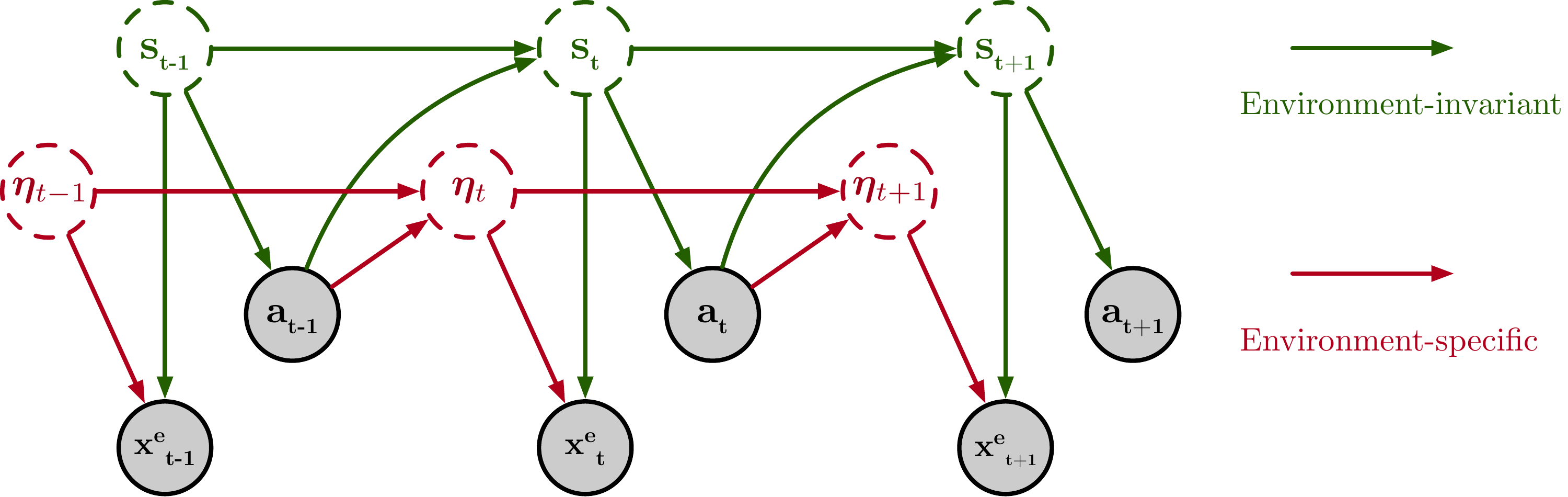}
    \caption{\textbf{Problem setup of Invariant Causal Imitation Learning \cite{invariant_cil}}. The observations are decomposable into environment-invariant features $\vs_t$ and environment-specific $\bm{\eta}_t$ encapsulating spurious associations with the actions. The goal is to recover $\vs_t$ such that the learned policy $\pi(\cdot \mid \vs_t)$ generalizes well to new environments.}
    \label{fig:icil}
\end{figure}

\citet{invariant_cil} consider the imitation learning setting with multiple environments. Their goal is to learn a policy that matches the expert behavior in all possible environments from a class of environments that share a certain structure for the observations and the transition dynamics, similarly, as we have discussed in \Cref{rl:multi_envs}. \Cref{fig:icil} visualizes the causal graph of their setup.  

To incentivize the learned imitation policy to stay within the distribution of the expert's observations, the authors propose to minimize the energy of the next observation obtained by following $\pi$ given the current observation. This way, they assign a low loss to the imitation policy for staying within high-density areas of the expert's occupancy measure and a high loss for straying from it. 

\subsection{Temporally Correlated Noise}
\citet{cil_under_temporally_correlated_noise} hypothesize that in many imitation learning settings, recordings of the expert may be corrupted by an unobserved confounder that is temporally correlated, i.e., \emph{temporally correlated noise} that is not independently distributed across timesteps. \Cref{tab:cil_tcn} illustrates settings where such noise can occur. The first row, for example, considers a quadcopter pilot who might have been flying under the persistent wind. An imitation learner will likely reproduce these deviations and perform poorly in test-time environments with different weather conditions. This setup is related to \Cref{cil_seq} but makes stronger assumptions on the unobserved confounder (additive noise).

To this end, \citet{cil_under_temporally_correlated_noise} leverage instrumental variable regression (IVR) to deconfound the data (\Cref{rw:ivr}). IVR leverages an \emph{instrument}, a source of random variation independent of the confounder, to deconfound inputs to a model via conditioning on the instrument. 

The key idea behind their approach is to frame past states as instruments to break the spurious association between states and actions caused by the unobserved confounder. The motivation behind this is that historical transitions are unaffected by future confounding \cite{hefny2015supervised}. 

They formalize the TCN setup by an SCM and the target estimand as the interventional effect of the expert policy $\mathbb{E}\left[\pi_{E}(\vs) \mid \vs\right]=\mathbb{E}[a \mid \doo(\vs)] \neq \mathbb{E}[a \mid \vs]$ based on IVR. To estimate that estimand, they propose two approaches inspired by generative modeling and game-theoretic IVR approaches. The former utilizes access to a simulator, while the latter can be run entirely offline. Further, they derive performance bounds, which assume that the same distribution of TCN affects the learner at test time.   

\begin{table}[t]
\vskip 0.15in
\begin{center}
\scalebox{0.8}{
\begin{tabular}{|c|c|c|c|c|}
\toprule
 \bf{Setting} & \bf{State} & \bf{Expert Action} & \bf{Observed Action} & \bf{Confounder}\\
 \midrule
 Quadcopter Flying \cite{ng2003autonomous} & Position & Intended Heading & Actual Heading & Persistent Wind \\
 \midrule
Product Pricing \cite{wright1928tariff} & Demand & Profit Margin & Price & Raw Materials Cost \\
 \midrule
ICU Treatment \cite{desautels2017prediction} & Symptoms & Intent to Treat & Patient Treated & Comorbidity \\  \midrule
Shared Autonomy \cite{cil_under_temporally_correlated_noise} & User State & Intended Action & Executed Action & Assistance \\
\bottomrule
\end{tabular}
}
\end{center}
\caption{\textbf{Examples of Temporally Correlated Noise in Causal Imitation Learning \cite{cil_under_temporally_correlated_noise}:} In these settings, spurious associations (\Cref{sec:confounding}) between states and actions can lead to inconsistent estimates of the expert's policy. \label{intertiatable}}
\label{tab:cil_tcn}
\end{table}

\begin{figure}[t]
\centering
  \includegraphics[width=\columnwidth]{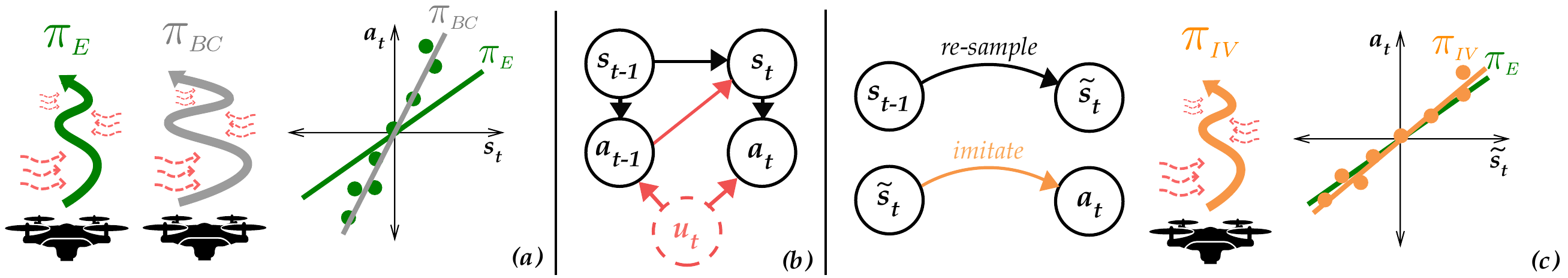}
  \caption{\textbf{Causal Imitation Learning under Temporally Correlated Noise (TCN) \cite{cil_under_temporally_correlated_noise}:}
  \textit{(a)} Unobserved confounders (e.g., wind) induce spurious associations between states and expert actions. Non-causal imitation learning approaches like behavioral cloning can learn policies that rely on this noise, leading to poor test-time performance. \textit{(b)} Causal DAG illustrating how the TCN confounder $\vu_t$ affects both the input ($\vs_t$) and output ($a_t$). \textit{(c)} Using instrumental variables \Cref{rw:ivr}, We can re-simulate state transitions from a past state, producing fresh samples ($\widetilde{\vs_t}$). We can then regress from these sampled states to observed expert actions to recover the expert's policy as the noise on inputs, and outputs is no longer correlated.}
\label{fig:TCN}
\end{figure}

\section{Credit Assignment}\label{rl:ca}
One key issue in RL is \emph{credit assignment} (CA), the task of understanding the causal relationship between actions and rewards and determining to what extent an outcome is due to external, uncontrollable factors \cite{minsky1961steps, cca_mfrl}. That means that a correct value function allows us to disentangle the relative aspects of \say{skill} and \say{luck} in an agent's performance. However, unfortunately, because of partial observability, scale, long time horizons, or increasing number of actions, each action taken by an agent may have only a vanishing effect on the outcome, making it increasingly difficult to learn proper value functions.

In the following, we look at methods that assign credits by measuring the causal influence of an action on the observed reward (\Cref{sec:causal_influence}).

\subsection{Causal Influence Detection} \label{sec:ci_rl}
\citet{seitzer2021causal} postulate that an agent can only influence its environment in certain situations. If, for instance, a robotic arm is placed in front of an object of interest on a table, the object can only be moved once contact is made between the robot and the object. Some situations have an immediate causal effect, while others do not. 

The authors argue that \emph{causal action influence} (CAI), which quantifies the causal influence of states, can help guide the learning algorithm to seek out states with stronger (predicted) influence, even enabling it to discover helpful behavior in the absence of task-specific rewards. To derive CAI, they introduce a causal model, which allows them to quantify whether the agent is in control given its state by conditional mutual information (CMI), which they learn through neural network models. Based on their experimental results, they conclude that CAI improves sample efficiency and performance by (i) better state exploration through an exploration bonus, (ii) causal action exploration, and (iii) prioritizing experiences with causal influence during training. 

\subsection{Counterfactual Credit Assignment}
\citet{cca_mfrl} propose \emph{counterfactual credit assignment} (CCA), a framework that uses the notion of counterfactuals to deal with the credit assignment issue. They propose to implicitly perform counterfactual policy evaluation (\Cref{sec:cf_gps}) by conditioning value functions on future event embeddings, which learn to extract relevant information from a trajectory. They then use the estimated counterfactual returns to form unbiased and lower variance estimates of the policy gradient by building future-conditional critics. 

While conventional state-action functions estimate the return for all actions, they do so by averaging all possible futures. In contrast, CCA estimates the return for different actions while keeping many external factors constant between the return and the counterfactual estimate. This makes CCA potentially finer-grained and improves data efficiency in environments with complex credit assignment structures.

The critical ingredient for the FC-PG algorithm is to learn the hindsight statistics $\Phi$, which embed a trajectory while not precluding any action $a$, which was possible for $\pi$ from having potentially produced $\Phi$. For example, $\Phi_t = A_t$ does \emph{not} satisfy this condition because it precludes any action $a \ne A_t$ from having produced $\Phi_t$. The authors discuss several options to learn $\Phi$ and mainly focus on using a hindsight network $\varphi$ which extracts $\Phi$ directly from the observed state-action-reward triplets, i.e., $\Phi_{t}=\varphi\left(\rmX, A, R\right)$. 

By establishing a connection to counterfactually-guided policy search \cite{cf_gps}, introduced in \Cref{sec:cf_gps}, the authors explain that the CCA estimator can be understood as sidestepping modeling the environment by discarding information from the whole trajectory $\tau$ which would require a model, and left with still useful information $\Phi_{t}$ with which counterfactuals can be computed in a model-free way.

\subsection{Multi-Agent Reinforcement Learning}
Multi-agent systems consist of autonomous, distributed agents that interact with each other in a shared environment \cite{weiss1999multiagent,marl_survey}. Each agent strives to accomplish an assigned goal, and the interplay between agents can vary depending on the task, such that agents cooperate or act competitively to beat the competition. 

Here, the credit assignment problem arises in cooperative settings when the collection of the agents' actions generates only a global, shared reward, making it difficult for each agent to deduce its contribution to the team's success \cite{chang2003all,coma}. 

Additionally to the shared rewards given to all agents, \cite{social_influence} proposes that agents be rewarded for influencing the actions of other agents. Agents receive this intrinsic reward in addition to their immediate reward, reflecting their effect on another. They refer to it as \say{causal influence}, similarly as \citet{seitzer2021causal} in the single-agent setup discussed in \Cref{sec:ci_rl}. Using counterfactual reasoning, an agent simulates possible actions and determines the effect they would have had on another agent's behavior at every time step. Actions that significantly affect the other agent's behavior are considered highly influential and rewarded. Generally, this added reward helps agents learn interpretable communication protocols and obtain higher collective rewards than baselines, which sometimes fail to learn completely. 

\section{Counterfactual Data Augmentation} \label{rl:cda}
The following techniques apply the idea of counterfactual data augmentation (CFDA) (\Cref{chapter:cgm}) to RL trajectories. 

\subsection{Local Causal Models \cite{local_causal_models}}
\begin{figure}[!t]
	\centering
	\includegraphics[width=0.98\textwidth]{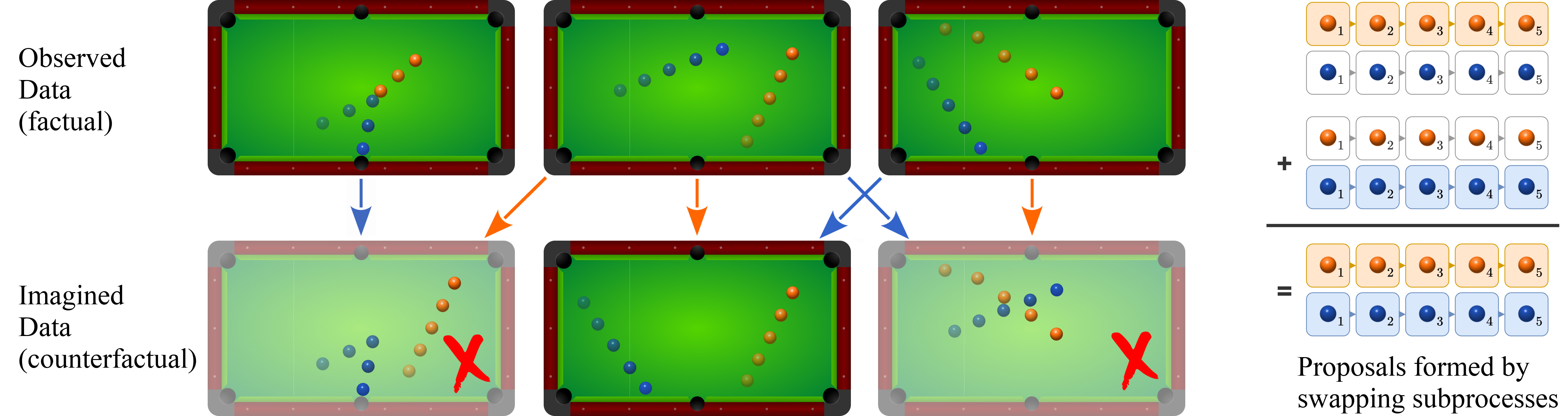}
	\caption{\textbf{Pool example motivating Counterfactual Data Augmentation (\method{CoDA}) \cite{local_causal_models}}. Knowing the local causal structure allows us to mix and match factored subprocesses to form counterfactual samples based on three factual samples. We reject the first proposal because one of its factual sources (the blue ball) is not localized. Due to the swapped proposal not being locally factored in, the third proposal is rejected. Accepted proposal two may be used as additional training data for reinforcement learning agents.
	}
	\label{fig:coda_intro_diagram}
\end{figure}
Consider a game of billiards, as illustrated in \Cref{fig:coda_intro_diagram}: each ball can be seen as a physical process. Before the opening break, due to their initial placement, each ball has a non-zero chance of colliding with the others. Hence, to predict the expected outcome of the opening break, we require a transition dynamics model that considers all balls. However, besides the initial timestep, most interactions between balls remain sparse. In other words, only a small subset of all balls are involved at most timesteps. 

\citet{local_causal_models} take advantage of the fact that during the time between their interactions (\emph{locally}), the subprocesses are causally independent: they propose a counterfactual data augmentation technique that is compatible with any agent architecture and does not require a forward dynamics model. By inferring whether or not local interactions occur, they swap factorized subspaces of observed trajectory pairs whenever two trajectories have the same local factorization. 

\begin{figure}[t]
	\centering
	\includegraphics[width=\textwidth]{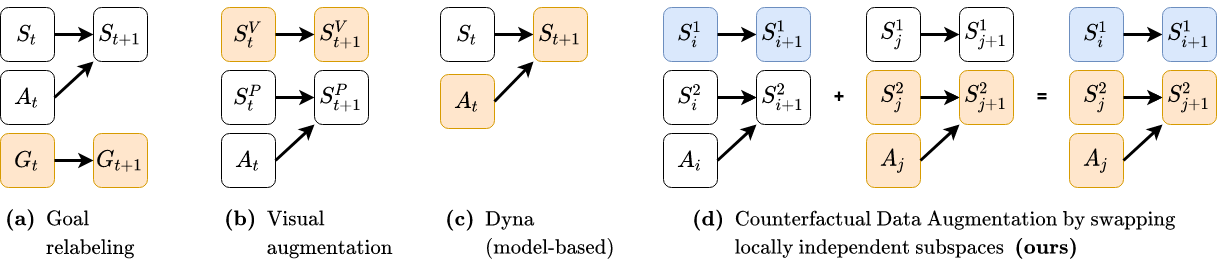}
	\vspace{-\baselineskip}
	\caption{\textbf{Existing RL data augmentation techniques can be interpreted as specific instances of \method{CoDA} \cite{local_causal_models}:} Orange nodes are relabeled, exogenous noise variables omitted for clarity. \textbf{(a)} Goal relabeling \cite{kaelbling1993learning}, including \method{HER} \cite{andrychowicz2017hindsight}, augments transitions with counterfactual goals. \textbf{(b)} Visual feature augmentation \cite{andrychowicz2020matters,laskin2020reinforcement} changes visual features $S_t^V$ (e.g., lighting, camera positions, etc.) that do not impact the physical state $S^P_{t+1}$. \textbf{(c)} Dyna \cite{sutton1991dyna}, including \method{MBPO} \cite{janner2019trust}, augments observed states with new actions and re-samples the next state using a learned dynamics model. \textbf{(d)} Given two transitions that share local causal structures, \method{CoDA} swaps connected components to form new transitions.}
	\label{fig:coda_examples}
\end{figure}

The key idea behind \method{CoDA} is to leverage the principle of independent mechanisms (\Cref{def:pim}). Assume that we can decompose state and action space into multiple subspaces, e.g., represented through subgraphs $\mathcal{G}_{i}, \mathcal{G}_{j} \subset \mathcal{G}$, where $\gG$ is the causal DAG of the global transition dynamics. Then, the causal mechanisms represented by $\mathcal{G}_{i}, \mathcal{G}_{j}$ are independent when $\mathcal{G}_{i}$ and $\mathcal{G}_{j}$ are disconnected in $\mathcal{G}$. In other words, when the global dynamics governed by $\gG$ are divisible into two (or more) connected components, we can reason about each subgraph as an independent causal mechanism. Existing RL data augmentation techniques rely on similar global independence relationships, as illustrated in \Cref{fig:coda_examples}.

\subsection{Personalized Policies \cite{Luetal20}}
\citet{Luetal20} propose a CFDA method to tackle environment heterogeneity by \emph{personalized policies}, e.g., for healthcare settings, where patients may exhibit different responses to identical treatments. Similar to \method{CF-GPS} (\Cref{sec:cf_gps}), they formalize the transition dynamics process as an SCM. Then, similar to methods in \Cref{rl:mbrl}, they learn that SCM, including its structural functions and exogenous noise variables (\Cref{sec:scm}), they use \emph{Bidirectional Conditional GANs} \cite{jaiswal2018bidirectional}. 

The authors propose two algorithms, differing in whether we assume latent environment heterogeneity in the data or not. If so, they explicitly include an environment variable in the causal system to consider variability across observations.

\section{Agent Incentives} \label{rl:incentives}
The development of artificial general intelligence (AGI), systems that equal or exceed human intelligence in a wide variety of cognitive tasks, raises serious safety concerns \cite{bostrom_superintelligence,everitt2018agi}. The key concern underlying AGI safety research is that these systems may produce autonomous agents much more intelligent than humans, which consequently pursue goals that conflict with our own. One way to take precautions against potential catastrophes caused by AGIs is to \emph{align} their \emph{incentives} with ours, also referred to as \emph{AI alignment} \cite{yudkowsky2016ai,christian2021alignment}.

Many agent incentive concepts rely on causal relationships among variables influencing the agent's utility. For example, \citet{incentives_are_causal} postulate that classifiers designed to incentivize improvement \emph{necessarily} require to perform causal inference. In other words, the authors show that any agent that strategically adapts performs causal modeling \say{in disguise}. 

In the following, we follow \citet{everitt_carey_hammond_fox_langlois_legg_2021} and summarize recent progress on formalizing agent incentives causally. 

\subsection{Causal Influence Diagrams}
\begin{figure}
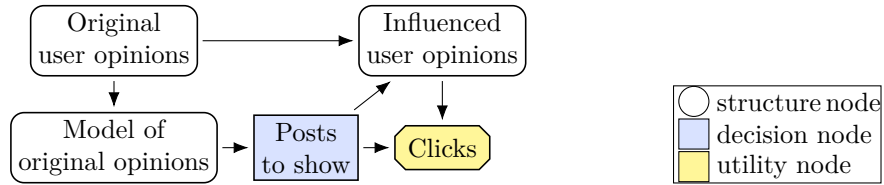

\centering \hspace*{\fill}
  \begin{subfigure}[t]{0.64\textwidth}
      \centering
      \begin{influence-diagram}
  \setrectangularnodes
  \setcompactsize
  \tikzset{node distance=4mm and 3.5mm}
  \node (D) [decision, anchor=west] {Posts\\ to show};
  \node (M) [left = of D] {Model of\\ original opinions};
  \node (P1) [above =3.5mm of M] {Original\\ user opinions};
  \node (U) [right = of D, utility] {Clicks};
  \node (P2) at (U|-P1) {Influenced\\ user opinions};
\draw[->]
    (P1) edge (M)
    (M) edge[information] (D)
    (P1) edge (P2)
    (D) edge (P2)
    (D) edge (U)
    (P2) edge (U)
  ;
\end{influence-diagram}
  \end{subfigure}\hfill
  \begin{subfigure}[t]{0.16\textwidth}
    \begin{influence-diagram}
    \cidlegend[]{
      \legendrow{}{structure\! node} \\
      \legendrow{decision}{decision node}\\
      \legendrow{utilityc, chamfered rectangle xsep=1.5pt, chamfered rectangle ysep=1.5pt}{utility node}\\
}
\end{influence-diagram}
  \end{subfigure} \hspace*{\fill}
  \caption{\textbf{Example of a Causal Influence Diagram (CID) in the context of a content recommendation system \cite{causal_influence_diagram}.} The goal of a content recommendation system is to choose posts that will maximize the user's click rate. However, the system's designers prefer the system not to manipulate the user's opinions to obtain more clicks.~\looseness=-1
      }
\label{fig:cid_example}
\end{figure}
One formalism for studying agent incentives is a \emph{causal influence diagram} (CID) \cite{jern2011capturing,kleiman2015inference,causal_influence_diagram}, a causal DAG with decision  (\begin{tikzpicture}
  \node [draw,decision, minimum size=\ucht, inner sep=0mm] {};
\end{tikzpicture}) and utility (\begin{tikzpicture}
  \node [draw,utilityd, minimum size=\ucht, inner sep=0mm] {};
\end{tikzpicture}) variables. After defining a CID, one can analyze incentives through graphical criteria, e.g., the \emph{value of information} (VoI) \cite{howard1966information}, which measures which variables a decision-maker would benefit from knowing before making a decision.  

\Cref{fig:cid_example} shows an example: Imagine a content recommendation engine recommending a series of posts (decision node) to a user. For the algorithm designers to maximize the number of clicks (utility node), the content should be tailored to the user's interest (structure node). They do not, however, want the algorithm to use provocative content to manipulate the user's opinion (structure node). \citet{user_tampering_recommender_systems} refers to this problem as \emph{user tampering}. Using a simulation study, the researcher demonstrates that a Q-learning algorithm learns how to exploit its opportunities to polarize users with its early recommendations to have more consistent success with later recommendations that cater to that polarization. 

More generally, we define a CID as follows.

\begin{mydef}{Causal Influence Diagram \cite{jern2011capturing,kleiman2015inference,causal_influence_diagram}}{cid}
A \emph{causal influence diagram} (CID) is a directed acyclic graph $\gG$ with vertices $\gV$ partitioned into $\gV = \{\gX, \gD, \gU\}$ with \emph{structure nodes} $\gX$, \emph{decision nodes} $\gD$ and \emph{utility nodes} $\gU$. Utility nodes have no children. Edges into decisions indicate what information is available at the time of the decision, so we call them \emph{information links}.
\end{mydef}

Next, to define incentive concepts, \citet{causal_influence_diagram} use a \emph{structural causal influence model} (SCIM) \cite{dawid2002influence}, which is similar to an SCM (\Cref{def:scm}), but not equivalent. There are three key differences: (i) its structural functions are only defined for non-decision endogenous variables, (ii) the vertices are partitioned as described in \Cref{def:cid}, and (iii) it holds that the utility variable domains are a subset of the real numbers, i.e., $\gU \subset \R$.

\begin{figure}
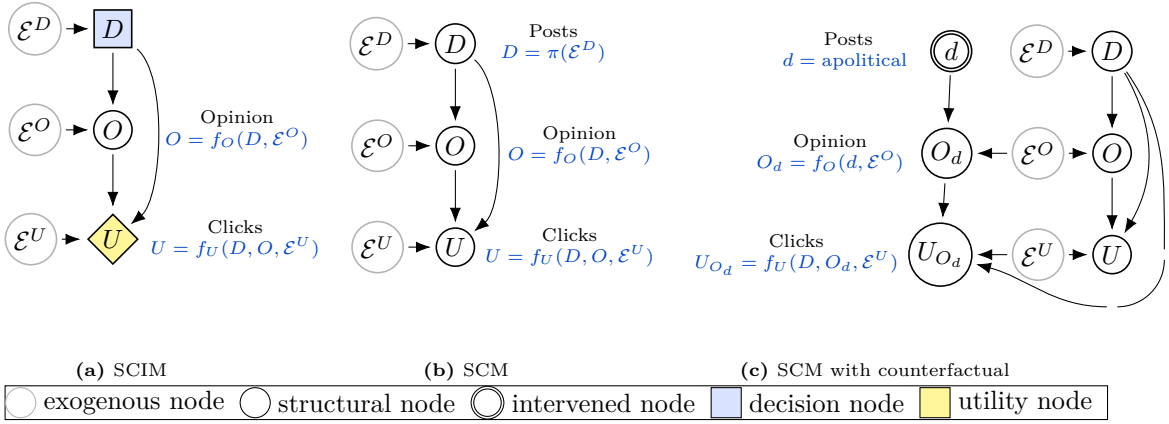

  \centering
  \begin{subfigure}[t]{0.295\textwidth}
    \centering
    \begin{minipage}{\textwidth}
        \centering
        \resizebox{\linewidth}{!}{
    \begin{influence-diagram}
      \setcompactsize
      \setinnersep{0.5mm}
      \node (D) [decision] {$D$};
      \node (X) [below =7mm of D] {$O$};
      \node (Y) [below =7mm of X,utility] {$U$};
        \node (ed) [left = 3mm of D, exogenous] {$\exovarv{D}$};
        \node (ex) [left = 3mm of X, exogenous] {$\exovarv{O}$};
        \node (ey) [left = 2.5mm of Y, exogenous] {$\exovarv{U}$};
      \path
      (ed) edge[->] (D)
      (ex) edge[->] (X)
      (ey) edge[->] (Y)
      (D) edge[->] (X)
      (X) edge[->] (Y)
      ;
      \draw[->] (D) to[in=40,out=-40,looseness=.6] (Y);
      \begin{scope}[
        every node/.style={draw=none,rectangle,align=center,inner sep=0mm}
        ]
        \scriptsize
\node[right = 3mm of X] {Opinion \\ \textcolor{dmblue500}{$O = f_O(D, \exovarv{O})$}};
        \node[right = 0mm of Y] {Clicks\\ \textcolor{dmblue500}{$U = f_U(D, O,\exovarv{U})$}};
      \end{scope}
      \node (h2) [below = 4mm of Y, inner sep=0mm, minimum size=0mm] {};
    \end{influence-diagram}
    }
    \caption{SCIM}\label{fig:scim-example}
    \end{minipage}
\end{subfigure}
  \begin{subfigure}[t]{0.295\textwidth}
    \centering
    \begin{minipage}{\textwidth}
        \centering
        \resizebox{\linewidth}{!}{
    \begin{influence-diagram}
      \setcompactsize
      \setinnersep{0.5mm}
      \node (D) [] {$D$};
      \node (X) [below =7mm of D] {$O$};
      \node (Y) [below =7mm of X] {$U$};
        \node (ed) [left = 3mm of D, exogenous] {$\exovarv{D}$};
        \node (ex) [left = 3mm of X, exogenous] {$\exovarv{O}$};
        \node (ey) [left = 3mm of Y, exogenous] {$\exovarv{U}$};
      \path
      (ed) edge[->] (D)
      (ex) edge[->] (X)
      (ey) edge[->] (Y)
      (D) edge[->] (X)
      (X) edge[->] (Y)
      ;
      \draw[->] (D) to[in=40,out=-40,looseness=.6] (Y);
      \begin{scope}[
        every node/.style={draw=none,rectangle,align=center,inner sep=0mm}
        ]
        \scriptsize
        \node[right = 2mm of D] {Posts \\ \textcolor{dmblue500}{$D = \pi(\exovarv{D})$}};
        \node[right = 3mm of X] {Opinion \\ \textcolor{dmblue500}{$O = f_O(D, \exovarv{O})$}};
        \node[right = 0mm of Y] {Clicks\\ \textcolor{dmblue500}{$U = f_U(D, O,\exovarv{U})$}};
      \end{scope}
      \node (h2) [below = 4mm of Y, inner sep=0mm, minimum size=0mm] {};
    \end{influence-diagram}
    }
    \caption{SCM}\label{fig:counterfactual1}
    \end{minipage}
\end{subfigure}
  \begin{subfigure}[t]{0.39\textwidth}
    \centering
    \begin{minipage}{\textwidth}
        \centering
        \resizebox{\linewidth}{!}{
    \begin{influence-diagram}
      \setcompactsize
      \setinnersep{0.5mm}
      \node (D) [] {$D$};
      \node (X) [below =7mm of D] {$O$};
      \node (Y) [below =7mm of X] {$U$};
      \node (ed) [left =2.5mm of D, exogenous] {$\exovarv{D}$};
      \node (ex) [left =2.5mm of X, exogenous] {$\exovarv{O}$};
      \node (ey) [left =2.5mm of Y, exogenous] {$\exovarv{U}$};
      \node (d) [double,left = of ed] {$d$};
      \node (Xd) [left = of ex] {$O_d$};
      \node (YXd) [left = of ey] {$U_{O_d}$};
      \path
      (ed) edge[->] (D)
      (ex) edge[->] (X)
      (ex) edge[->] (Xd)
      (ey) edge[->] (Y)
      (ey) edge[->] (YXd)
      (D) edge[->] (X)
      (d) edge[->] (Xd)
      (X) edge[->] (Y)
      (Xd) edge[->] (YXd)
      (D) edge[->,bend left] (Y)
      ;
      \node (h1) [right = 3mm of Y, inner sep=0mm, minimum size=0mm] {};
      \node (h2) [below = 3mm of Y, inner sep=0mm, minimum size=0mm] {};
      \path (D) edge[out=-55,in=90] (h1)
      (h1) edge[out =-90, in=0] (h2)
      (h2) edge[->, out =180, in=-15] (YXd)
      ;
      \begin{scope}[node distance=0.1cm and 0.1cm,
        every node/.style={ draw=none, rectangle,align=center,
          inner sep=0mm }
        ]
        \scriptsize
        \node [left = of d] {Posts \\ \textcolor{dmblue500}{$d=\text{apolitical}$ }};
        \node [left = of Xd] { Opinion\\
          \textcolor{dmblue500}{$O_d = f_O(d, \exovarv{O})$}};
          \node [left =0mm of YXd] {Clicks\\
          \textcolor{dmblue500}{$U_{O_d} = f_U(D,O_d,\exovarv{U})$}};
        \end{scope}
    \end{influence-diagram}
    }
\caption{SCM with counterfactual}\label{fig:counterfactual3}
\end{minipage}
\end{subfigure}
\begin{minipage}{\textwidth}
        \centering
        \resizebox{\linewidth}{!}{
\begin{influence-diagram}
    \cidlegend[]{
      \legendcol{exogenous}{exogenous node} 
      \legendcol{}{structural node} 
      \legendcol{double}{intervened node} 
      \legendcol{decision}{decision node}
      \legendcol{utilityc, chamfered rectangle xsep=1.5pt, chamfered rectangle ysep=1.5pt}{utility node} \\
}
\end{influence-diagram}
}
\end{minipage}
  \caption{
    \textbf{An example of a Structural Causal Influence Model (SCIM) in a content recommendation scenario \cite{causal_influence_diagram}.} The system aims to select posts that maximize click-through rates without manipulating user opinions. 
    \textbf{(a)} Displaying political or apolitical posts $D$ will affect the user's opinion $O$. $D$ and $O$ influence the user's clicks $U$.
    \textbf{(b)} Given a policy, the SCIM becomes an SCM. \textbf{(c)}
    We can use this SCM to estimate counterfactuals. For example, the counterfactual $U_{O_d}$ represents
    the number of clicks if the user has the opinions they would arrive at after viewing apolitical content.}
  \label{fig:scim}
\end{figure}

Returning to the content recommendation example, \Cref{fig:scim} illustrates its SCIM. By specifying a policy $\pi$, which comprises a structural function for decision nodes $\gD$, we  turn a SCIM into an SCM, allowing us to use the standard definitions of causal interventions and counterfactuals.  

Being equipped with CIDs and SCIMs, we can now analyze agent incentives and other safety-related concerns. 

\subsection{Incentive Concepts}
We summarize five incentive concepts informally and direct readers interested in more details to \citet{causal_influence_diagram}. 

\textbf{Materiality \cite{fagiuoli1998note}}: Which observations are \emph{material} for optimal performance? Nodes can be identified as immaterial based on the graphical structure, e.g., if they are \emph{d-separated} (i.e., conditionally independent) from the utility nodes. 

\textbf{Value of Information \cite{howard1966information}} Generalization of materiality to unobserved nodes to assess which variables a decision-maker would benefit from knowing \emph{before} making a decision.

\textbf{Response Incentive \cite{causal_influence_diagram}} An alternative generalization of generalizing a material observation from the perspective of it being one that influences optimal decisions. Under this interpretation, variables that have a response incentive are the ones that influence the optimal agent's decisions. 

\textbf{Value of Control \cite{shachter1986evaluating}} Can the agent benefit from setting a variable's value? More specifically, can the attainable utility be increased by letting the agent decide the structural function for the variable?

\textbf{Instrumental Control Incentive \cite{causal_influence_diagram}} What is the agent interested in and able to control? Would that choice matter if the agent chooses decision $D \in \gD$ to influence $X \in \gX$ independently of how $D$ influences other aspects of the environment? \citet{causal_influence_diagram} call it an \emph{instrumental} control incentive because the control of $X$ is a means to achieve utility. 

\subsection{User Interventions}
Sometimes it is desirable to constrain or modify the policies of agents when they are deployed in practice. For example, action constraints have been used to prevent damage when training robot policies, and transformations to the policy can ensure that an agent stays within some safe region of state space \cite{human_intervention,berkenkamp2017safe}. Whenever it is difficult to specify unsafe states formally, a human overseer might interrupt the agent \cite{everitt_carey_hammond_fox_langlois_legg_2021}. 

\citet{langlois2021rl} address this issue by introducing \emph{modified-action MDPs} (MAMDPs), where the agent's policy outputs the agent's decision rather than the actions. The actions are assumed to be not fully under the agent's control but can also be influenced by the action modification. Since the agent might learn the interruption scheme from interruptions during training, the corresponding CID includes an information link from the action modification to the policy.

The MAMDP framework, including its CIDs, enables the authors to analyze the causal assumptions that prototypical RL algorithms make about the environment and to understand how they adapt to interruption via a graphical criterion for path-specific response incentives. 

They find that black-box optimization algorithms, such as evolutionary strategies, take all causal relationships of the environment into account and try to both obscure its policy and disable its off-switch. In contrast, with the action chosen by the agent, Q-learning and \method{SARSA} assume that the next action will be taken optimally, with no action modification. In other words, these two algorithms ignore causal effects from the action modification to the actual action taken. Another version of \method{SARSA}, based on the modified action, ignores the effect of action modification on the current action but considers the effect on subsequent actions and tries to disable the off-switch.

\subsection{Reward Tampering}
By default, an agent aims to maximize its observed reward, given by a (programmed) reward function. However, sometimes, it may be incentivized to tamper with this function, which we call \emph{reward tampering} \cite{reward_tampering}. \citet{everitt_carey_hammond_fox_langlois_legg_2021} give three examples: 
\begin{enumerate}[leftmargin=*]
    \item \textbf{Wireheading}: Rewriting the source code of its implemented reward function.
    \item \textbf{Feedback tampering}: Influencing users that train a learned reward model.
    \item \textbf{Input tampering}: Manipulating the reward function's inputs.
\end{enumerate}

\newcommand{\ThetaR}{\Theta^{\mathrm{R}}}
\newcommand{\thetaR}{\theta^{\mathrm{R}}}
\newcommand{\hThetaR}{\hat\Theta^{\mathrm{R}}}
\newcommand{\hthetaR}{\hat\theta^{\mathrm{R}}}
\newcommand{\tThetaR}{\tilde\Theta^{\mathrm{R}}}
\newcommand{\tthetaR}{\tilde\theta^{\mathrm{R}}}
\newcommand{\ThetaT}{\Theta^{\mathrm{T}}}
\newcommand{\thetaT}{\theta^{\mathrm{T}}}
\newcommand{\ThetaO}{\Theta^{\mathrm{O}}}
\newcommand{\Thetadiamond}{\ThetaR_{\textrm{diamond}}}
\newcommand{\Thetarock}{\ThetaR_{\textrm{rock}}}
\newcommand{\Thetatdiamond}{\ThetaR_{\textrm{diamond}, t}}
\newcommand{\Thetatrock}{\ThetaR_{\textrm{rock}, t}}
\newcommand{\Thetasdiamond}{\ThetaR_{*\textrm{diamond}}}
\newcommand{\Thetasrock}{\ThetaR_{*\textrm{rock}}}
\newcommand{\hThetadiamond}{\hat\Theta^{\mathrm{R}}_{\textrm{diamond}}}
\newcommand{\hThetarock}{\hat\Theta^{\mathrm{R}}_{\textrm{rock}}}

\newcommand{\ThetaOdiamond}{\ThetaO_{\textrm{diamond}}}
\newcommand{\ThetaOrock}{\ThetaO_{\textrm{rock}}}
\newcommand{\wall}{wall \cellcolor{dmgray300}}
\newcommand{\goal}{\cellcolor{dmteal100}}
\newcommand{\tamper}{\cellcolor{dmpurple100}}
\newcommand{\good}{\cellcolor{dmyellow300!50}}

\begin{figure}
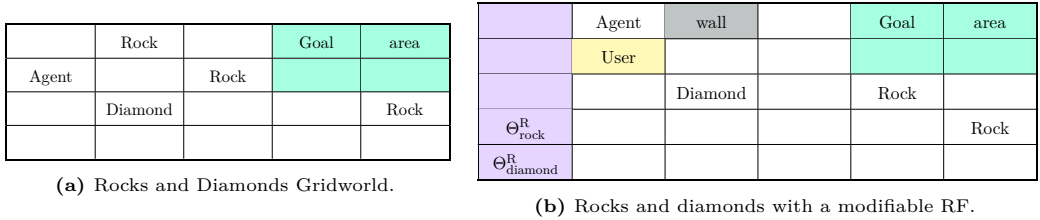

  \centering
  \begin{subfigure}[t]{0.44\textwidth}
    \resizebox{\textwidth}{!}{
  \renewcommand{\arraystretch}{1.5}
  \begin{tabular}{V{3}x{\cw}|x{\cw}|x{\cw}|x{\cw}|x{\cw}V{3}}
    \hlineB{3}
    & Rock & & Goal \goal & area \goal \tabularnewline\hline
    Agent & & Rock &\goal & \goal \tabularnewline\hline
    & Diamond & & & Rock \tabularnewline \hline
    & & & & \tabularnewline\hlineB{3}
  \end{tabular}
}
  \caption{Rocks and Diamonds Gridworld.}
  \label{fig:rocks-and-diamonds}
\end{subfigure}  
  \begin{subfigure}[t]{0.55\textwidth}
    \resizebox{\textwidth}{!}{
      \renewcommand{\arraystretch}{1.5}
      \begin{tabular}{V{3}x{\cw}|x{\cw}|x{\cw}|x{\cw}|x{\cw}|x{\cw}V{3}}
        \hlineB{3}
        \tamper & Agent & \wall & &Goal \goal & area \goal \tabularnewline\hline
        \tamper & User\good & & & \goal & \goal \tabularnewline \hline
        \tamper & & Diamond & &Rock & \tabularnewline \hline
        $\Thetarock \tamper$& &  & &  & Rock \tabularnewline\hline
        $\Thetadiamond \tamper$ &  & &  &  & \tabularnewline\hlineB{3}
      \end{tabular}
    }
    \caption{Rocks and diamonds with a modifiable RF.
}
    \label{fig:rd-irf}
  \end{subfigure}
  \caption{\textbf{Gridworld Example of Reward Tampering \cite{reward_tampering}}. The agent can avoid getting the implemented reward function updated and even change it.}
  \label{fig:mdp-mrf}
\end{figure}
\Cref{fig:mdp-mrf} showcases an example of an MDP with a modifiable implemented reward function. Imagine a gridworld where the agent can push rocks and diamonds by walking towards them from an adjacent cell, as illustrated in \Cref{fig:rocks-and-diamonds}.
At time $t$, the reward function rewards diamonds and punishes rocks for each item pushed to the goal area: 
\begin{equation}R_{t}=\# \text { diamonds in goal area }-\# \text { rocks in goal area. }
\end{equation}

Now, let us distinguish between \emph{intended rewards} that encourage completion of the intended task and \emph{observed rewards}, which are the rewards received by the agent. Generally speaking, we parameterize the returning reward function as $R_t = R(S_t;\ThetaR)$, where $\ThetaR_t$ will denote the parameter for an implemented reward function at time $t$, and $\ThetaR_*$ the parameter of an intended reward function. 

To model the possibility of the agent influencing the intended function, we include a user and two reward parameters $\Thetarock$ and $\Thetadiamond$. The new, modifiable reward function is \begin{equation}
    \label{eq:observed-reward}
    R_t = \Thetatdiamond\cdot (\#\text{diamonds in goal area}) + \Thetatrock\cdot
    (\#\text{rocks in goal area}),
  \end{equation} where the reward parameters toggle between $-1$ and $+1$ if the agent stands on top of them. Further, the reward parameters are set to their intended value of diamond-gathering
  ($\Thetarock=-1$ and $\Thetadiamond=1$) when the agent visits the user tile.

The intended agent's task is to gather diamonds.
However, in its initial implementation, $\ThetaR_1$ rewards rocks instead of diamonds, which is corrected when the agent passes the user. According to the intended reward function, the agent should not walk around the user or visit the reward parameter tile. Unfortunately, by breaking either of these conditions, the agent can observe more rewards. 

\citet{reward_tampering} claim that a standard RL agent may have an instrumental goal to tamper with its implemented reward function. By modeling the above problems with CIDs and analyzing their  previously proposed solutions, the authors find that the latter avoid undesirable incentives by removing causal links in a way that avoids instrumental control incentives. However, there are still differences between the proposed methods, e.g., whether they use previously observed or latent variables to formulate an objective avoiding reward tampering.

\subsection{Delicate States} A \emph{delicate} state is as part of the environment which is hard to define a reward for (\emph{subtle}) and vulnerable to deliberate manipulation towards bad outcomes (\emph{manipulable}). \citet{path_specific_objectives} try to remove any incentive for the agent to control the delicate part of the state-space (as having a control incentive over it is dangerous). 

To remove it, they declare a \emph{stable} state condition, which is robust against unmotivated action and renders side-effects unlikely to be bad. They formalize these conditions through CIDs' graphical criteria for the presence of incentives to control certain parts of the state.

\section{Open Problems}

\textbf{Lack of unified evaluation environments}
For example, in causal model-based RL, all methods share the same goal of learning the environment transition dynamics (e.g., to improve sample efficiency for policy search). However, there is no overlap in their evaluation environments. \citet{rezende2020causally} evaluate on \method{AvoidFuzzyBear} (proposed in the paper), \method{MiniPacman} \cite{minipacman}, and visual 3D \cite{gregor2019shaping} environments. \citet{cdl} use a Chemical and manipulation environment. \citet{li2020causal} conduct their experiments on the \method{CoPhy} benchmark \cite{Baradel2020CoPhy} (\Cref{benchmark:cophy}). Similar inconsistent evaluation procedures occur in causal multi-task RL.

\subsection{Unification of Formalisms}
Due to the parallels between causality and reinforcement learning and the lack of maturity at the intersection of both, we fear redundancy among papers, i.e., an analogy of \emph{reinventing the wheel} \cite{riw} taking place. An increasing number of formalisms in the literature may lead to decision paralysis for practitioners. Similarly, as we have argued in the previous paragraph, it slows down progress by increasing barriers to comparing approaches. We suggest that future work compares the applicability of different formalisms more rigorously and possibly unifies them.

For example, consider a set of MDPs with shared dynamics (e.g., the set of all physically feasible pendulums), where each single MDP is specified by additional dynamics parameters (e.g., the length and mass of a pendulum, see \cite{PAML} for more concreteness). Let us assume that these parameters are unobserved.

In the literature, we notice a myriad of MDP formalisms proposed for that type of setup: MDP with Unobserved Confounders \cite{zhang2016markov}, Confounded MDPs \cite{wang2021causal}, Causal POMDPs 1 \cite{li2020causal}, Causal POMDPs 2 \cite{causal_curiosity} (they are not the same formulation), Causal MDPs \cite{lu2022efficient}, and discrete MDPs (a \emph{universe}, \Cref{sec:sg}) \cite{mutti2022provably}. Even in the non-causal RL literature, we find multiple formalisms with similar motivations, e.g., Contextual MDPs \cite{cmdps}, Hidden Parameter MDPs \cite{hi_params}, and Bayes Adaptive MDPs \cite{bayes_adapt}. We understand that these formalisms may not be technically identical but hypothesize that unification is possible and can facilitate progress.

\subsection{Deconfounding Offline RL}
Whenever we aim to learn policies from an offline dataset without access to observed policy or the environment, the dataset likely includes confounding biases. As we have seen throughout this section, a significant motivation for using causality for RL is to deconfound observed data, i.e., transform it so that confounding biases are removed. We believe that deconfounding observational data is under-explored in pure offline RL settings, and only a few works have addressed this setup \cite{crpe_inf}.    

\subsection{Counterfactual Decision-Making}
\citet{crl} expose that inference of counterfactuals is neglected in the RL literature, possibly due to the difficulty of learning correct SCMs. However, counterfactual reasoning in agents may bring additional benefits to agents only reasoning based on observational distributions, such as less bias during policy search (\Cref{sec:cf_gps}), assigning credit to single agents in a shared-reward cooperative multi-agent system \cite{coma} or considering the intended actions of humans in a human-in-the-loop system \cite{zhang2022can}.

\citet{bottou2013counterfactual} illustrate how counterfactual reasoning can assist large-scale real-world machine learning systems, such as the ad placement system associated with the Bing search engine. They propose a causal model for ad placement and apply it to optimizing ad auction tuning parameters. This work is very detailed and discusses multiple fundamental causal inference techniques, connections to RL, and different design choices. We hope to see more carefully-designed counterfactual agents for real-world systems that may take inspiration from \citet{bottou2013counterfactual}.

\newcommand{\orw}[1]{\textcolor{green}{\textbf{#1}}}
\newcommand{\mow}[1]{\textcolor{red}{\textbf{#1}}}
\chapter{Modality-specific Applications}\label{sec:applications}
In the previous chapters, we familiarized ourselves with different methodologies that applied two causal primitives, interventions and counterfactuals, throughout 5 different problem domains; ranging from how to learn invariances in data up to dealing with confounding in sequential decision making settings.

In this chapter, we review methods designed for particular data modalities, namely: image (computer vision), text (natural language processing), and graph (graph representation learning) data. Some of these works are off-the-shelf applications of the core methodology introduced in the previous chapters; some are not but are too modality-specific to be presented in the earlier chapters. 

We observe a few common themes across all three areas, which we use to divide the methods within each modality. For convenience, we recap and briefly summarize them here:
\begin{itemize}[leftmargin=*]
\item \textbf{Causal Supervised Learning} (\Cref{chapter:cil}): Methods disposing of spurious associations by extracting (invariant) feature maps containing only the causal parents of the predictor variable $Y$.
\item \textbf{Counterfactual Data Augmentation} (\Cref{chapter:cgm}): Methods augmenting training data with counterfactual modifications to a subset of the causal factors, leaving the rest of the factors untouched. 
\item \textbf{Counterfactual Explanations} (\Cref{explanations:counterfactual}): Methods explaining model predictions by computing a (minimally) altered features instantiation of an individual that would cause the underlying model to classify it in a different class.
\item \textbf{Causal Fairness} (\Cref{chapter:fairness}): Methods ensuring that model predictions are fair w.r.t. causal relationships involving protected attributes (e.g., gender, race, etc.).
\item \textbf{Miscellaneous}: Everything else that does not fit into the above categories.
\end{itemize}

\section{Causal Computer Vision}\label{app:cv}
\subsection{Causal Supervised Learning}

\textbf{Long-Tail Classification}
Stochastic gradient descent (SGD) methods are central to neural network optimization \citep{bottou2018optimization}. Modern SGD variants (e.g., Adam, SGD-M, etc.) commonly used for image classification often involve some form of \emph{momentum} (or \emph{acceleration}), which accumulates historical gradients to speed up convergence, akin to a heavy ball rolling down the loss function landscape.

\citet{momentum_confounder} interpret momentum $\rmM$ as a confounder in long-tailed classification settings. If a dataset is balanced, each class contributes equally to the momentum. Long-tailed datasets, however, will be dominated by \emph{head} samples (i.e., samples belonging to the majority class). Long-tailed datasets are very common in practice, e.g., in NLP due to Zipf's law \cite{zipf} or in image segmentation, where increasing the images of tail class instances like \say{remote controller} requires more head instances like \say{sofa} or \say{TV} simultaneously. 

Let us consider input features $\rmX$, momentum $\rmM$, classification logits $Y$, and $\rmD$, which denotes $\rmX$'s projection on the head feature direction that eventually deviates $\rmX$. In a long-tailed dataset, few head classes possess most training samples, which have less variance than the data-poor but the class-rich tail. Therefore, the moving averaged momentum will point to a stable head direction. Specifically, the authors demonstrate that one can decompose any random feature vector $\rmX$ into $\rmX=\ddot{\rmX}+\rmD$, where $\rmD$ is a function of the exponential moving average features and the number of training iterations (for brevity, not shown here). 

\begin{figure}[t]
   \includegraphics[width=\linewidth]{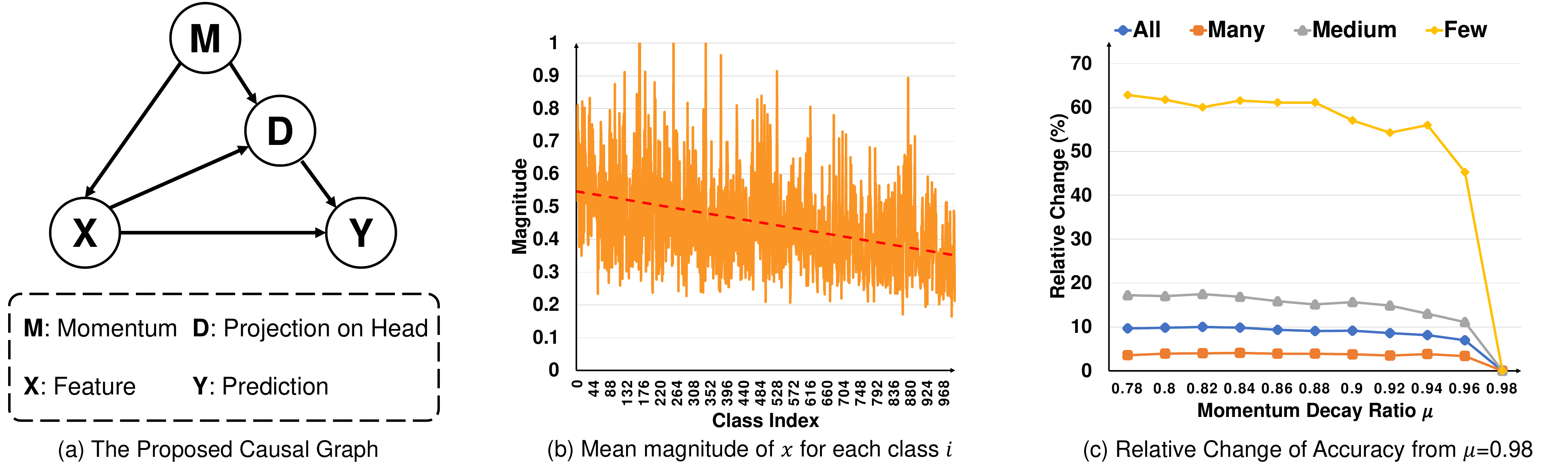}
   \caption{\textbf{SGD Momentum as Confounder \cite{momentum_confounder}.} (a) Causal DAG explaining the causal effect of momentum. (b) Sorted mean magnitudes of feature vectors for each class \(i\) after training with momentum $\beta=0.9$. (c) The relative accuracy change from  $beta=0.98$ as a function of $\beta$ indicates that the few-shot tail is more sensitive to it.}
   \label{fig:momentum_confounder} 
   \vspace{-5mm}
\end{figure}

The authors show that the following causal relationships emerge: $\rmM$ confounds the input features $\rmX$ ($\rmM \rightarrow \rmX$) and the classification logits $Y$ (via $\rmM \rightarrow \rmD \rightarrow Y$), as illustrated in \Cref{fig:momentum_confounder}(a). The backdoor path $\rmX \leftarrow \rmM \rightarrow \rmD \rightarrow Y$ causes a spurious association between $\rmX$ and $Y$. The authors call this association a \say{bad} bias. \Cref{fig:momentum_confounder}(b,c) demonstrate that momentum contributes to this bias empiricially. Further, the mediation path $\rmX \rightarrow \rmD \rightarrow Y$ is considered a \say{good} bias, because it respects the inter-relationships of the semantic concepts in classification. 

To remedy the \say{bad} momentum bias, \citet{momentum_confounder} apply backdoor adjustment to account for the backdoor confounding path, while keeping the \say{good} mediation path. Further, for the final prediction logits, they compute the intervention $p(y \mid \vx)$ to yield the direct causal effect of $\rmX \rightarrow Y$. Their method improves over baselines in long-tailed image classification and instance segmentation benchmarks.

\textbf{Few-Shot Learning} \label{sec:int_few_learn} 
Few-shot learning (FSL) methods assume that deployed models in production will likely encounter novel tasks for which only a few samples with labels are available; yet, these data-poor tasks may have some structural similarity to other data-rich tasks \cite{wang2020generalizing,pml1Book}. \citet{pml1Book} give the following example: consider the task of classifying endangered bird species, which are by definition rare. However, birds bear many structural similarities across species (wings, feathers, etc.). Hence, training a model on a large dataset of non-endangered species and then transferring that knowledge to the small datasets of endangered birds may result in better performance than training on the small datasets alone.

A common strategy to deal with such settings is \emph{transfer learning}. A simple transfer learning strategy is \emph{fine-tuning}, which consists of two phases: First, we perform a \emph{pre-training} phase, in which we train a model on a large source dataset. Second, we freeze some pre-trained parameters and continue training the rest on the few-shot target (training) dataset of interest.

Another, more advanced transfer learning strategy is  \emph{meta-learning}, which aims to learn a meta-model that quickly adapts to different few-shot datasets. We will not discuss how these methods work but direct interested readers to \cite{wang2020generalizing}.

\begin{figure}
    \centering
     \begin{subfigure}[b]{0.69\columnwidth}
         \centering
    \includegraphics[width=\columnwidth]{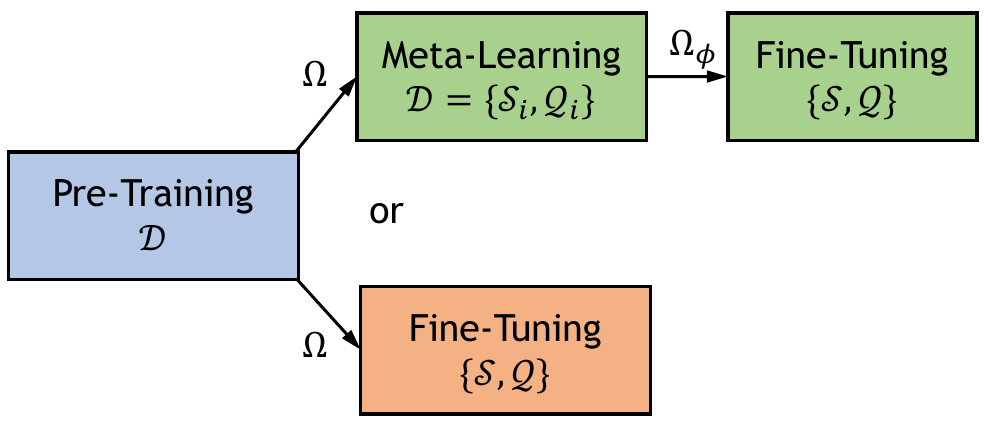}
         \caption{relationships among different FSL paradigms}
         \label{fig:ifsl_relationship}
     \end{subfigure}
      \begin{subfigure}[b]{0.29\columnwidth}
         \centering
    \includegraphics[width=\columnwidth]{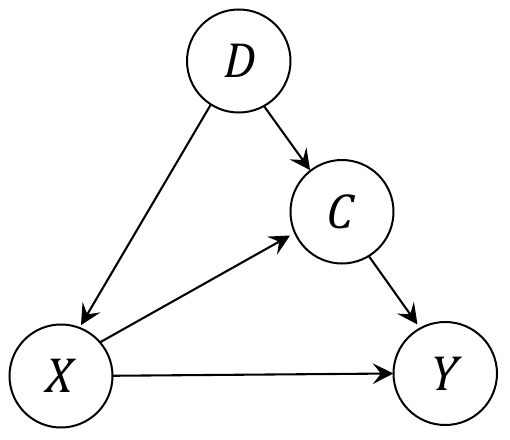}
         \caption{\textbf{Causal DAG}. Pre-trained knowledge $\rmD$ confounds $\rmX$ and $\rmC$.}
         \label{fig:ifsl_graph}
     \end{subfigure}
    \caption{\textbf{Confounding (\Cref{sec:confounding}) in Few-Shot-Learning \cite{interventional_fsl}:} Pre-trained weights, obtained through conventional pre-training or meta-learning, can be interpreted as a confounder that may introduce transfer deficiencies (\Cref{sec:int_few_learn}).}
    \label{fig:ifsl}
\end{figure}

\citet{interventional_fsl} study both FSL strategies from a causal perspective and find that pre-trained knowledge is a confounder that can limit their performance. By adjusting for this confounder, the authors develop three algorithms that achieve new state-of-the-art results on several FSL benchmarks. 

For illustration, recall the exemplary problem in \Cref{fig:spurious_associations}, where we are interested in predicting the cow label based on cow-specific features and not the background features (e.g., the mountains). Using pre-trained knowledge $\rmD$ (e.g., a large dataset $\gD$ or a third-party pre-trained model with parameters $\params$), $p(y \mid \vx)$ may fail to generalize well, because it induces spurious associations: the pre-trained weights generate features ($\rmD \rightarrow \rmX$), and semantics ($\rmD \rightarrow \rmC \rightarrow Y$) that may overly rely on mountain specifics.

The authors argue that to make FSL more robust, we need to pursue the true causality between $\rmX$ and $Y$, i.e.,  the causal intervention $p(y \mid \doo\left(\vx \right))$. To illustrate why this is, they contrast FSL with many-shot learning (MSL), which refers to fine-tuning with a larger target dataset. Naturally, we would expect MSL to work better simply because we use more data; yet, the authors argue that this would not answer why MSL converges to the true causal effects as the number of samples approach infinity. \Cref{cp:fsl} explains this in more detail. 

\begin{figure}
\centering
\captionsetup{font=footnotesize,labelfont=footnotesize}
\begin{subfigure}[t]{.3\textwidth}
  \centering
  \captionsetup{width=\linewidth}
  \includegraphics[width=\linewidth]{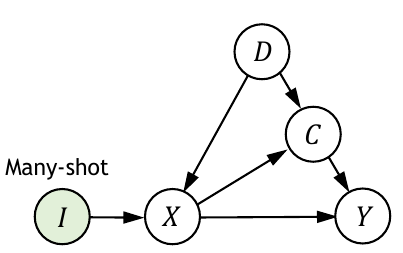}
  \caption{MSL, where $p(y\mid \vx) \approx p(y\mid\doo(\vx))$}
  \label{fig:msl}
\end{subfigure}%
\begin{subfigure}[t]{.3\textwidth}
  \centering
  \captionsetup{width=\linewidth}
  \includegraphics[width=\linewidth]{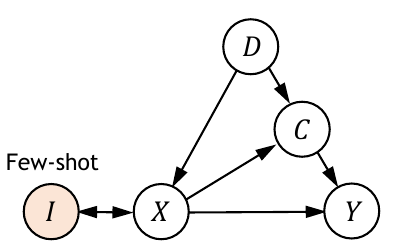}
  \caption{FSL, where $p(y\mid\vx)\not\approx p(y \mid \doo(\vx)$}
  \label{fig:fsl}
\end{subfigure}
\begin{subfigure}[t]{.2\textwidth}
  \centering
  \captionsetup{width=\linewidth}
  \includegraphics[width=\linewidth]{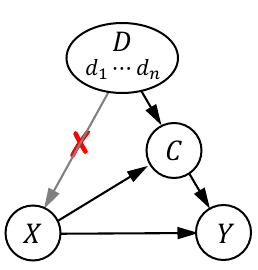}
  \caption{IFSL, where $p(y \mid \doo(\vx))$ through backdoor adjustment.}
  \label{fig:ifsl_backdoor}
\end{subfigure}
\caption{\textbf{Many-shot (MSL) vs. few-shot (FSL) vs. interventional few-shot learning (IFSL) \cite{interventional_fsl}.} MSL and IFSL are robust against confounding of $\rmD$, while FSL is not.} 
\label{fig:msl_fsl_ifsl}
\end{figure}

\begin{cp}{Many-Shot vs. Few-Shot Learning \cite{interventional_fsl}}{fsl}
\citet{interventional_fsl} postulate that $p(y \mid \doo\left(\vx\right)) \approx p(y \mid \vx)$ in MSL while $p(y \mid \doo(\vx)) \not \approx p(y \mid \vx)$ in FSL. To show why this holds, let us introduce the sample ID $I$ and assume that $\vx \sim p(\vx \mid i)$. Naturally, we may assume that we can use $p(y \mid i)$ to estimate $p(y \mid \vx)$, so we can incorporate $I$ into the estimation of $p(y \mid \vx)$, by writing \begin{align}
    p\left(y \mid \vx_{i}\right):= \mathbb{E}_{\vx \sim p(\vx \mid i)} \left[Y \mid \vx, i \right]=p(y \mid i).
\end{align}
Now, we compare how $I$ enters the causal graphs for MSL and FSL, illustrated in \Cref{fig:msl,fig:fsl}. For MSL, we find that $I \rightarrow \rmX$, and not $\rmX \rightarrow I$, because tracing $\rmX$'s ID out of many samples is like \say{finding a needle in a haystack}. This makes $I$ an instrumental variable (\Cref{rw:ivr}), effectively meaning that $I$ and $\rmD$ are independent and $p(y \mid \vx)=p(y \mid i) \approx p\left(y \mid \doo\left(\vx\right)\right)$ (details in \cite{interventional_fsl}'s Appendix). However, $\rmX \rightarrow I$ persists in FSL, because it is easier to guess the correspondance, e.g., in the 1-shot extreme case that has a trivial 1:1 mapping for $\rmX \leftrightarrow I$. 
\end{cp}

Next, the authors propose \emph{Interventional FSL}, where the idea is to use backdoor adjustment to estimate $p(y \mid \doo(\vx))$ without the need for many-shot samples, as illustrated in \Cref{fig:ifsl_backdoor}. This adjustment requires observing and stratifying the confounding variable, which is non-trivial when $\rmD$ is a third-party delivered pre-trained network. The authors suggest three implementations for this: (i) feature-wise adjustment, (ii) class-wise adjustment, and (iii) combined adjustment. They show that these implementations improve the baselines across all query hardnesses.

\textbf{Attention Models}
\label{sec:causal_attention}
\begin{figure}[t]
\begin{center}
\includegraphics[width=0.8\textwidth]{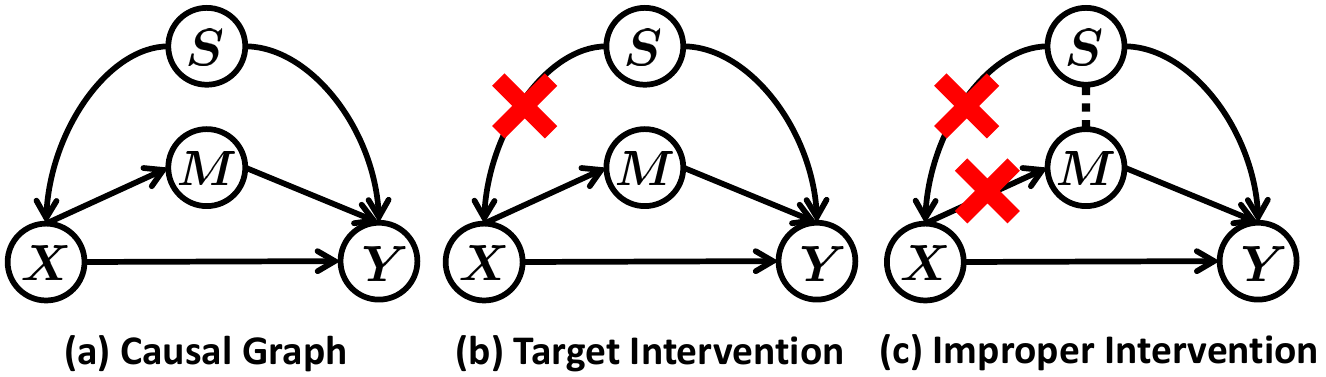}
\end{center}
  \caption{\textbf{Causal DAG for Causal Attention Module \cite{wang2021causal}}, consisting of image $\rmX$, label $Y$, mediator $\rmM$ and unstable context confounder $\rmS$. The goal is to learn the causal association $\rmX \rightarrow Y$, while disposing assocations along the pathways of $\rmX \leftarrow \rmS \rightarrow Y$.}
\label{fig:caam}
\end{figure}
To make visual attention-based models less susceptible to capture spurious correlations and more robust in OOD settings, \citet{wang2021causal} propose a \emph{causal attention module} that annotates confounders in an unsupervised way. \Cref{fig:caam} shows their causal graph: $\rmX \rightarrow Y$ is the desired causal effect from image $\rmX$ to label $Y$. Further, we assume an invariant mediator variable $\rmM$ that contains discriminative object parts, e.g., a bird's wing. $\rmS$ is the style confounder, e.g., the non-discriminative background of an image (sky when the bird flies, ground when its wings are down). The target intervention is to disentangle the background $\rmS$ and the mediator $\rmM$, as shown in \Cref{fig:caam}b). However, since a perfect intervention is typically not easy to obtain when working with visual datasets, the authors propose to perform an improper intervention, illustrated in \Cref{fig:caam}c).

To identify the content and style variables, they train two separate attention mechanisms $f_a(\cdot), f_{\bar a}(\cdot)$ in an adversarial fashion, respectively. Similar to training Generative Adversarial Networks \cite{goodfellow2014generative}, each iteration of the training pipeline corresponds to solving a bilevel optimization problem consisting of a minimization and maximization step. The minimization step optimizes the invariant feature extractor $f_a(\vx)=\vc$, while the maximization step updates the style confounder extractor $f_{\bar a}(\vx)=\vs$.

\begin{figure}
    \centering
    \begin{subfigure}[t]{0.45\textwidth}
    \includegraphics[width=0.8\textwidth]{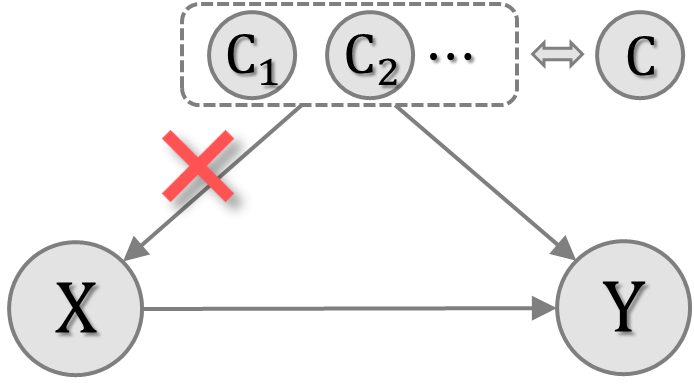}
    \caption{When the confounders are observed, we can apply the back-door criterion to adjust for them. }
    \label{fig:cifc_backdoor}
    \end{subfigure}
    \begin{subfigure}[t]{0.45\textwidth}
    \includegraphics[width=0.8\textwidth]{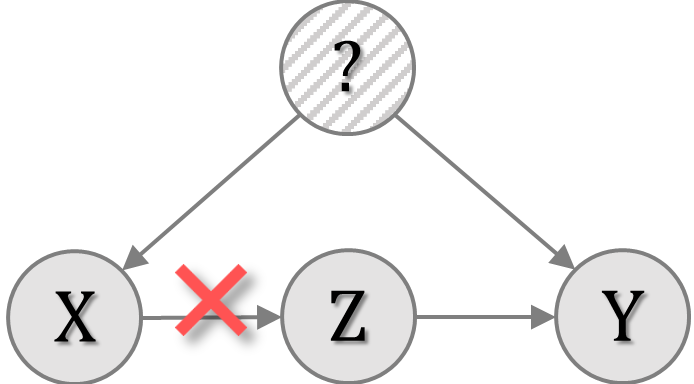}
    \caption{If the confounders are unobserved but the causal association between $\rmX$ and $Y$ is mediated through $\rmZ$, we can use the front-door criterion.}
    \label{fig:cifc_frontdoor}
    \end{subfigure}
    \caption{\textbf{Confounder Identification-free Causal Feature Learning \cite{causal_visual_feature_learning}.}}
    \label{fig:causal_visual_feature_learning}
\end{figure}

\textbf{Unobserved Confounders}
Most previous approaches adopt the backdoor criterion to mitigate the effect of confounders. However, the backdoor criterion requires the explicit identification of confounders. Since it can be challenging to identify confounders in many real-world scenarios, \citet{causal_visual_feature_learning} explore a confounder identification-free method using the front-door criterion. \Cref{fig:causal_visual_feature_learning} illustrates the difference between using the backdoor and front-door criterion. 

The front-door criterion does not require identifying the confounders if we have access to an intermediate variable $\rmZ$, such that $\rmX \rightarrow \rmZ \rightarrow Y$. The authors propose a strategy to simulate interventions on $\rmZ$, relying on meta-gradients. They also connect this strategy to gradient-based meta-learning methods and explain why methods like MAML \cite{maml} work from a causal perspective. Overall, their method improves the cross-domain performance of vision models.

\begin{figure}[h]
    \centering
    \begin{subfigure}[t]{0.3\textwidth}
    \centering
    \includegraphics[width = \textwidth]{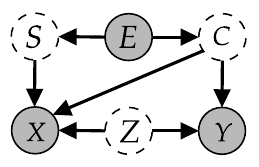}
    \caption{\textbf{Causal Graph} with domain invariant $\rmZ$, domain specific $\rmC$, and style $\rmS$. Video index $E$ confounds $\rmS$ and $\rmC$.}
    \label{fig:causal_motion_forecasting_dag}
    \end{subfigure}
    \begin{subfigure}[t]{0.6\textwidth}
    \includegraphics[width = \textwidth]{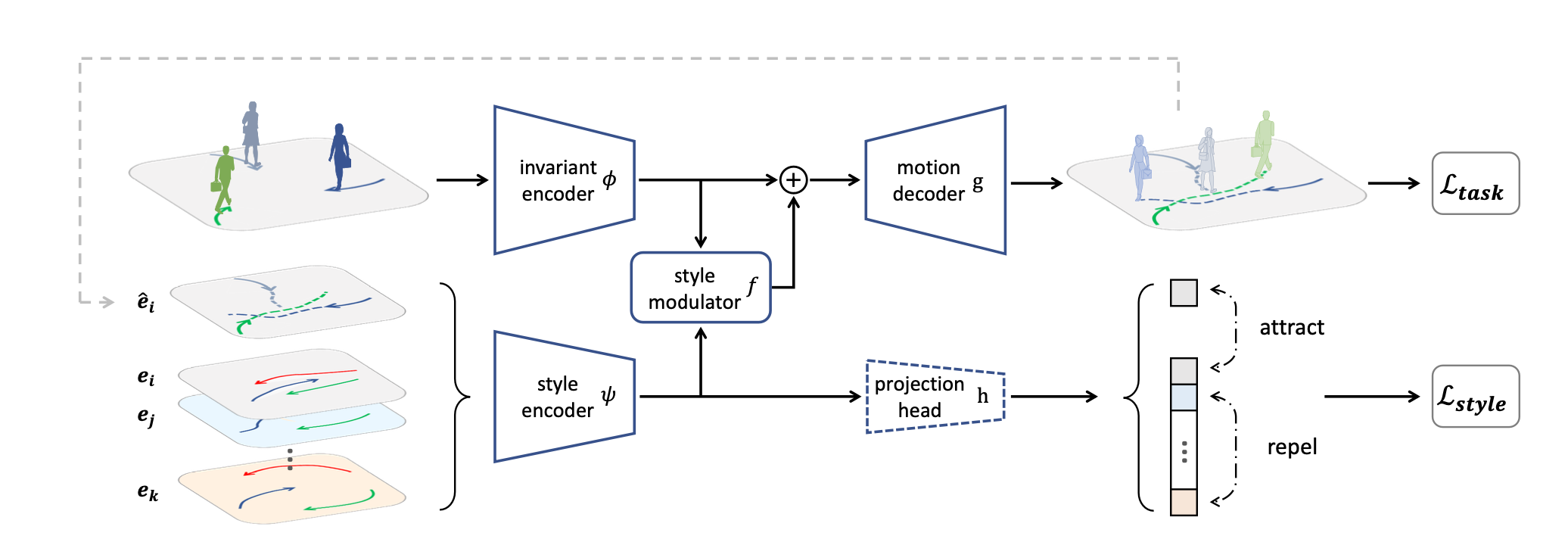}
    \caption{\textbf{Model Architecture}: The invariant encoder estimates domain invariant variables $\rmZ$, while the style encoder separates between the intra-domain predictor variables $\rmC$ and the spurious variables $\rmS$. $\rmC$ and $\rmZ$ are concatenated to make a prediction.}
        \label{fig:causal_motion_forecasting_architecture}
    \end{subfigure}
    \caption{\textbf{Causal Motion Forecasting \cite{causalmotionforecasting}:} An approach for forecasting the location of an object over time that generalizes across environments.}
\end{figure}

\textbf{Motion Forecasting}
\label{sec:causal_motion_learning}
\citet{causalmotionforecasting} study motion forecasting, the task of forecasting the location of a tracked object from a video. \Cref{fig:causal_motion_forecasting_dag} illustrates their causal DAG: they treat each video as a domain $e \in \mathcal{E}$, and include style $\rmS$, content $\rmC$, as well as \emph{domain invariant variables} $\rmZ$, in contrast to $\rmS$ and $\rmC$ which have dependence on the domain $e \in \mathcal{E}$. For example, $\rmZ$ may capture laws of physics.

\Cref{fig:causal_motion_forecasting_architecture} illustrates their model architecture: an invariant encoder $\bphi(\cdot)$ models domain invariant relationships, the style encoder models $\bpsi(\cdot)$ models domain-specific relationships, and the style modulator $f(\cdot)$ extracts useful domain-specific information while ignoring spurious information. A contrastive loss $\mathcal{L}_{\text{style}}$ is employed to encourage learning useful representations of domain specific information.

\textbf{Weakly-Supervised Semantic Segmentation}
\label{sec:context_adjustment}
\citet{wsss} present a framework called \emph{Context Adjustment} (\method{CONTA}) to deconfound pixel-level pseudo-masks in weakly-supervised semantic segmentation (WSSS). Specifically, the goal is to obtain non-spurious pixel-level pseudo-masks $\rmM$ of image segments $Y$. The authors identify that three types of style features $\rmS$\footnote{The authors refer to this variable as \say{context} in their work, but it is semantically equivalent to our concept of \say{style} (\Cref{def:scd}) that we use consistently across methods.} lead to inaccurate pseudo-masks: 
\begin{enumerate}[leftmargin=*]
    \item \textbf{Object ambiguity}: Objects often co-occur with each other under certain contexts. For example, if most horse images contain people riding horses, there exists a spurious association between these two distinct objects. A model relying on it will learn that most horses are with people, outputting pseudo-masks that do not draw boundaries between a person and a horse.
    \item \textbf{Incomplete background}: Backgrounds are often composed of multiple (unlabeled) semantic objects. Hence, there is a co-occurrence of foreground, and background objects, e.g., some parts of the background \say{floor} can be misclassified as the foreground \say{sofa}. 
    \item \textbf{Incomplete foreground}: Semantic parts of the foreground object may co-vary with different contexts, e.g., the window of a car may include reflections of its surroundings. Hence, a model may learn to segment less discriminative parts to represent the foreground, e.g., the \say{wheel} of a \say{car} (instead of the whole car).
\end{enumerate}

\begin{figure}[h]
    \centering
    \begin{subfigure}[t]{0.5\columnwidth}
            \includegraphics[width=0.8\columnwidth]{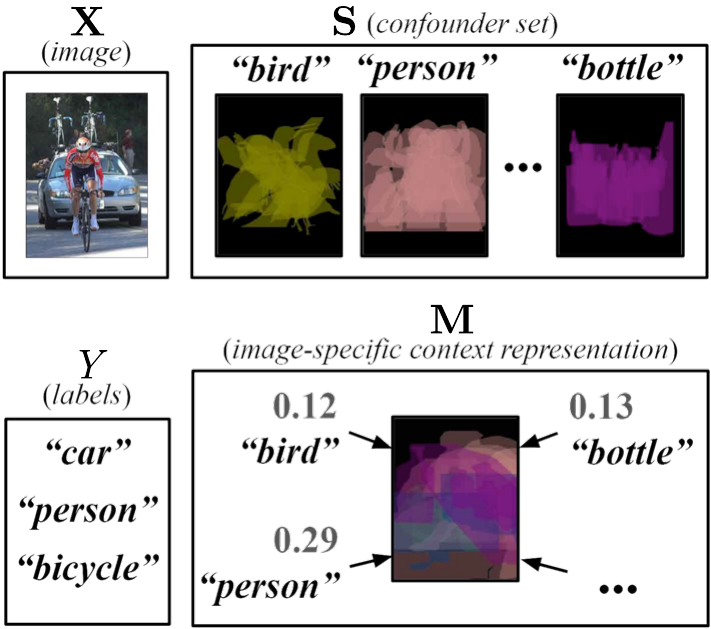}
    \caption{\textbf{Pictorial description of each component.} }
    \end{subfigure}
    \begin{subfigure}[t]{0.45\columnwidth}
        \includegraphics[width=0.8\columnwidth]{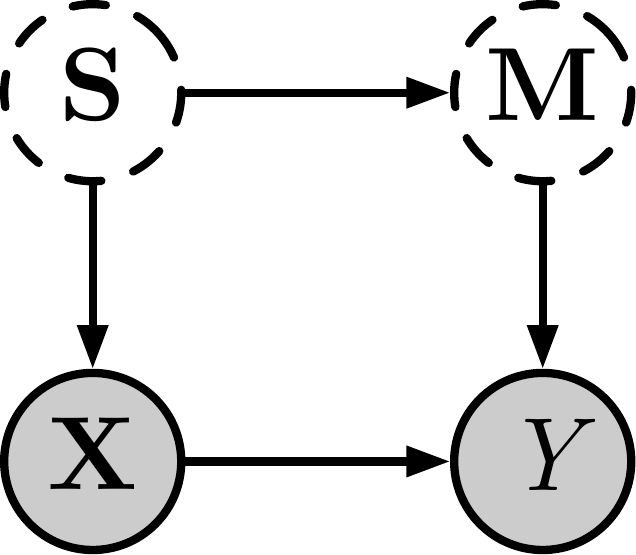}
    \caption{\textbf{Causal DAG} with image $\rmX$, label $Y$, style features $\rmS$, and pseudo-masks $\rmM$.}
    \end{subfigure}
    \caption{\textbf{Context Adjustment (\method{CONTA}) for Weakly-Supervised Semantic Segmentation \cite{wsss}}. The goal is to deconfound pixel-level pseudo-masks.}
    \label{fig:wsss}
\end{figure}

Therefore, they propose an SCM capturing the causal relationships between images, styles, and class labels, as illustrated in \Cref{fig:wsss}. Their deconfounding method, named \emph{context adjustment}, removes the confounding bias in semantic segmentation by simulating an intervention $p(y \mid \doo\left(\vx)\right)$ using an approximation of the unobserved $\rmS$.

\textbf{Visual Grounding}
Visual grounding is the task of mapping a free-form natural language query (phrase or sentence) onto its corresponding region of the image, e.g., \say{the dog next to the person who is driving a car} \cite{deconfounded_visual_grounding}. By investigating failure cases of existing grounding methods, \citet{deconfounded_visual_grounding} reveal spurious associations between certain subjects and their locations in common grounding datasets. For example, pictures corresponding to queries including \say{sheep} are often located in the central area; \say{corner of} tend to be of smaller size; \say{standing} tend to display standing people in the center because they are commonly the focus of the photographer; however, if encounter images where most people stand aside, this spurious association does not hold anymore.

\begin{figure}[t]
\centering
\hspace*{\fill}
\begin{subfigure}[t]{0.4\textwidth}
\includegraphics[width=\columnwidth]{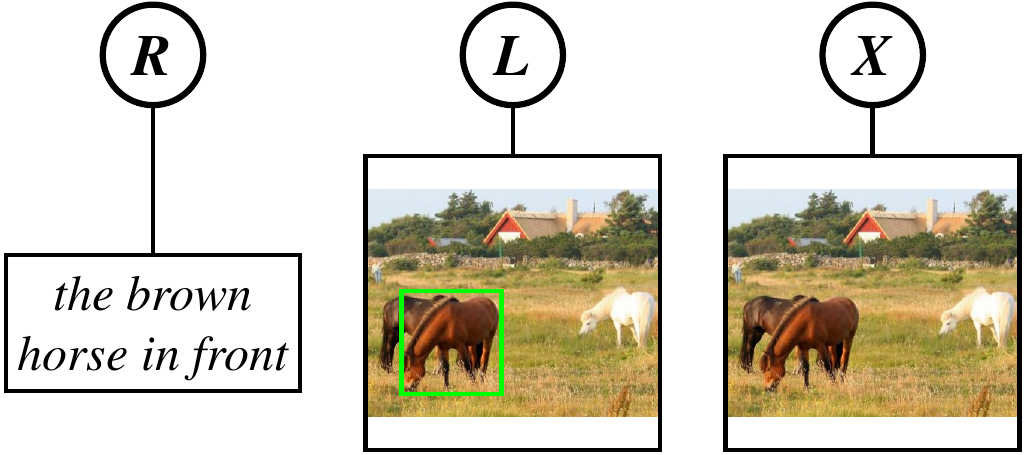}
\end{subfigure}\hfill
\begin{subfigure}[t]{0.3\textwidth}
\centering
\includegraphics[width=\columnwidth]{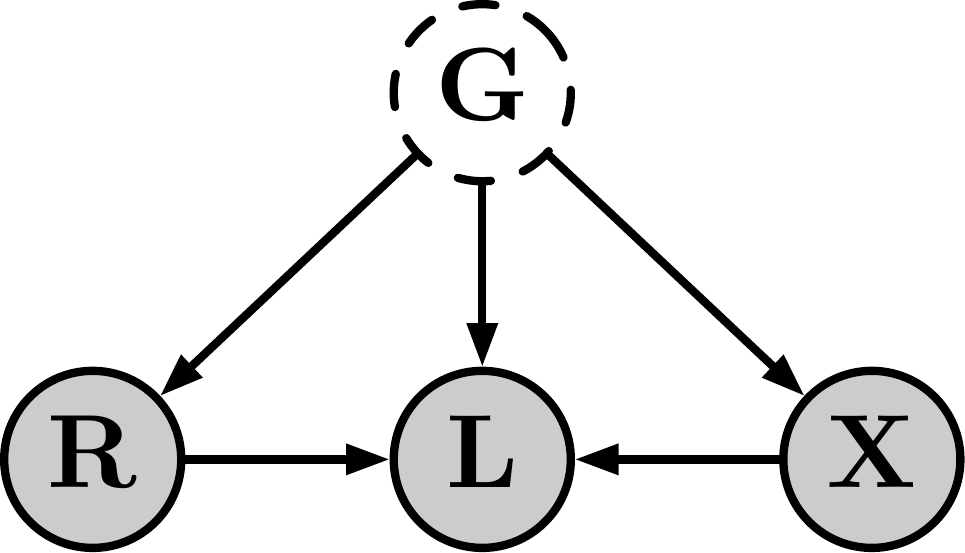}
\end{subfigure}\hspace*{\fill}
\caption{\textbf{Deconfounded Visual Grounding \cite{deconfounded_visual_grounding}:} $\rmG$: unobserved confounder, $\rmX$: pixel-level image, $\rmR$: language query, $\rmL$: the location for the query.}
\label{fig:dvg}
\end{figure}

\Cref{fig:dvg} shows the causal graph \citet{deconfounded_visual_grounding} propose. The causal estimand of interest is the interventional distribution $p(l \mid \doo(\vr), \vx) \neq p(l \mid \vr, \vx)$, where $\rmX$ denote the image, $R$ the language query, and $\rmL$ the object location. The inequality is due to the unobserved confounder $\rmG$. To deal with this unobserved confounder, the authors exploit the previously proposed \emph{Deconfounder} algorithm by \citet{deconfounder}. This algorithm allows one to learn a generative model of the substitute confounder $\hat \rmG$, which \citet{deconfounded_visual_grounding} use to perform the backdoor adjustment. We note that this algorithm has caused some controversy, and \citet{ogburn2019comment,d2019multi} questioned its legitimacy.   

\textbf{Video moment retrieval}
The task of identifying the start and end of a segment in a video that corresponds to a textual query is known as \emph{video moment retrieval} (VMR). \citet{vmr_challenges} argue that VMR models often exploit spurious temporal location biases in datasets rather than learning the cross-modal matching. In other words, the temporal location of moments is a hidden confounder that spuriously correlates user queries and moment locations, making the model ignore the actual video content. 

\citet{nan2021interventional} propose to remove such spurious correlations from the video grounding model by considering interventions on the text and video input. They call this method \emph{Interventional Video Grounding (IVG)}, as it uses backdoor adjustment to model $p\left(y \mid \doo\left(\vx\right)\right)$. They then propose a method to approximate the unobserved confounder as one cannot sample it directly from the dataset. They learn representations of the video and text inputs and enforce alignment between these representations of similar concepts with a contrastive approach. 

Similarly and concurrently, \citet{yang2021deconfounded} \citet{yang2021deconfounded} construct an SCM consisting of four variables: $\rmQ$ (query), $\rmV$ (video moment), $Y$ (prediction), and $\rmL$ (moment location). Then, they replace the non-causal query $p(y \mid \vq, \vv)$ with $p(y \mid \doo(\vq, \vv))$. Thereby, the query is forced to fairly interact with all possible locations of the target based on the intervention. 

\subsection{Counterfactual Explanations}
\citet{cf_ve} produce counterfactual explanations \Cref{explanations:counterfactual}) to scrutinize the prediction of image classification models. Given a \emph{query} image $\rmX$ with class label $y$, the goal of the counterfactual visual explanation is to identify how to change the image such that an image classification model would predict a different class $y^\prime$. 

Their methods work as follows: First, one has to select a \emph{distractor} image $\rmX^\prime$ that the model predicts as the class $y^\prime$. Then, they identify spatial regions in the images $\rmX$ and $\rmX^\prime$, replacing the spatial region in $\rmX$ with the area in $\rmX^\prime$ would lead to predicting the class $y^\prime$. Beyond providing explanations for model prediction, the authors find that counterfactual visual explanations can help human users distinguish different classes better. 

\begin{figure}
    \centering
    \includegraphics[width=0.8\columnwidth]{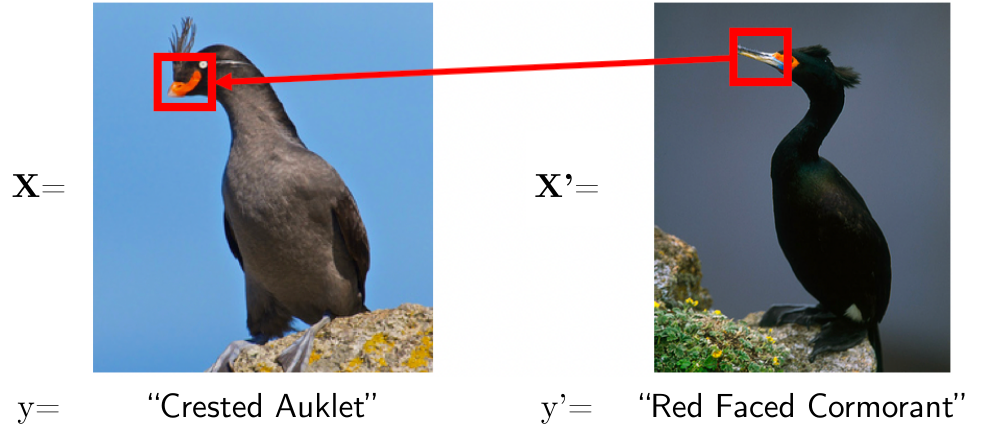}
    \caption{\textbf{Counterfactual visual explanations (CVE) \cite{cf_ve}:} Consider two images of two similar looking yet different birds: $\rmX, \rmX^\prime$ is the query and distractor image with label $y, y^\prime$, respectively. CVE identify regions in both images such that if the highlighted region in $\rmX$ looked like the highlighted region in $\rmX^\prime$, the resulting image $\rmX^{*}$ would be classified more confidently as $y^{\prime}$.}
    \label{fig:cf_visual_exp}
\end{figure}

\citet{cf_exp_nl} propose to generate counterfactual explanations by inspecting which evidence an input data point misses but might contribute to a different classification decision if present in the image. The motivation is to generate post-hoc natural language explanations about what attributes might change classification decisions if present in an image, e.g., ``This is not a Scarlet Tanager because it does \emph{not} have black wings.''

\subsection{Causal Generative Modeling}

\textbf{Counterfactual Data Augmentation}
\label{cv:cda}

\textbf{Zero-Shot Learning.}
\citet{yue_counterfactual_2021} suggest to generate counterfactual samples for out-of-distribution generalization of classifiers, by means of a generative model which disentangles class attributes $Y$ and sample-specific attributes $\rmZ$ of $\rmX$. Then, they generate counterfactual samples from a counterfactual distribution where $\rmZ$ has been intervened upon while keeping $Y$ constant, $\tilde{\vx} \sim p\left(\vx \mid y, \doo\left(\tilde{\vz}\right)\right)$.

To ensure that $\tilde{\vx}$ lies in the true distribution of the seen or unseen samples, they enforce counterfactual faithfulness by applying the contrapositive of the Consistency Rule \cite{consistency_rule}: if $\vx$ is dissimilar to $\tilde{\vx}$, the ground-truth attribute of $\vx$ cannot be $y$. We can implement this criterion by training a binary classifier that distinguishes between seen and unseen data $\rmX$.

\textbf{Cross-Domain Pose Estimator.} 
\citet{zhang2021learning} propose using causal representation learning to improve cross-domain 3D pose estimation tasks. Specifically, they train a counterfactual feature generator that takes domains and contents as input. They change domains to simulate interventions and steer the model to produce counterfactual features. This strategy facilitates the model to learn transferable features across domains.

\textbf{Counterfactual Trajectory Generation}
Our goal is to generate a counterfactual sequence $\vx^{\text{CF}}_{1:T}$ under a new initial condition $\vx^{\text{CF}}_0$, given an observed initial condition $\vx_0$ and a sequence of $T$ frames $\vx_{1:T}$.

\citet{causal_discovery_from_videos} propose the \emph{Visual Causal Discovery Network} (V-CDN), which is akin to tackling all three levels of causal representation learning (\Cref{def:causal_rep_learning}). It consists of three modules: (i) a perception network employing an unsupervised keypoint detection algorithm \cite{kulkarni2019unsupervised} to extract useful features from a video; (ii) a structural inference module using a GNN to learn a causal graph; and (iii) a dynamics prediction model conditioned on the causal graph and the current state features to make predictions into the future. The model tackles a task that purely associational object learners fail on: modeling interactions between objects beyond collisions only.

\begin{figure}
    \begin{center}
        \includegraphics[width=\linewidth]{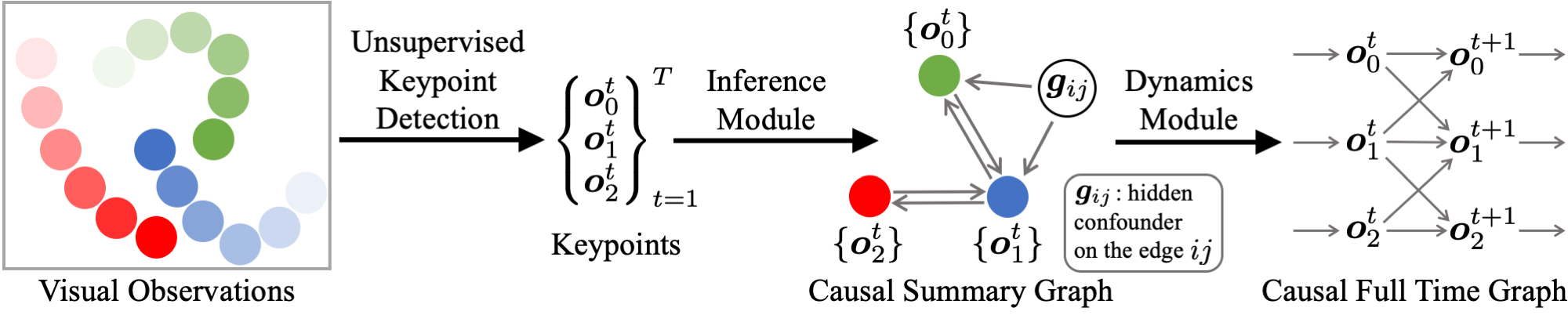}
    \end{center}
    \caption{{\bf Visual Causal Discovery Network \cite{causal_discovery_from_videos}:} it extracts unsupervised keypoints as the state representation, infers a causal graph, and then learns a dynamics module conditioned on both.
    }
    \label{fig:vcdn}
\end{figure}

Following up on \citet{causal_discovery_from_videos}, \citet{janny2022filtered} complement learning latent representations based on the unsupervised discovery of keypoints with additional information. Specifically, this addresses two problems: First, one must encode the shape, geometry, and relationships between multiple moving objects through the relative positions of points because each object is only discriminated through its 2D positions. Second, a 2D keypoint space may not be the optimal representation for modeling the dynamics of a physical system because of the additional imaging process that may confound the data.  

The authors deploy their proposed architecture on video data from a 3D simulator of stacked blocks. An encoder learns a representation of the blocks in the form of 2D keypoints and information coefficients which encode for shape and appearance. Latent confounders are extracted from this representation, and a dynamic model forecasts a trajectory from a representation of an observation. A decoder subsequently maps the trajectory prediction into video data. The authors also introduce a novel benchmark for counterfactual trajectory prediction, which is discussed in \Cref{sec:cv_benchmarks}.

\section{Causal Natural Language Processing}
CausalML has been used in a variety of natural language processing (NLP) tasks such as natural language explanations \cite{DBLP:journals/corr/abs-2005-04118,DBLP:journals/corr/abs-2107-07150}, fairness \cite{DBLP:journals/corr/abs-2011-06485}, text classification \cite{DBLP:journals/corr/abs-1809-10610}, and machine translation \cite{DBLP:conf/naacl/LiuKB21}. Next, we start introducing some representative work.

\subsection{Causal Supervised Learning}

\textbf{Visual Question Answering}
Imagine working at a social network company, and your team's task is to build models detecting hate speech. Colleagues of yours have already developed great models for text data only; however, a remaining challenge is detecting it in memes, where the model needs to understand the combined meaning of the words and pictures. 

\begin{figure}[h]
    \centering
    \includegraphics[width=\columnwidth]{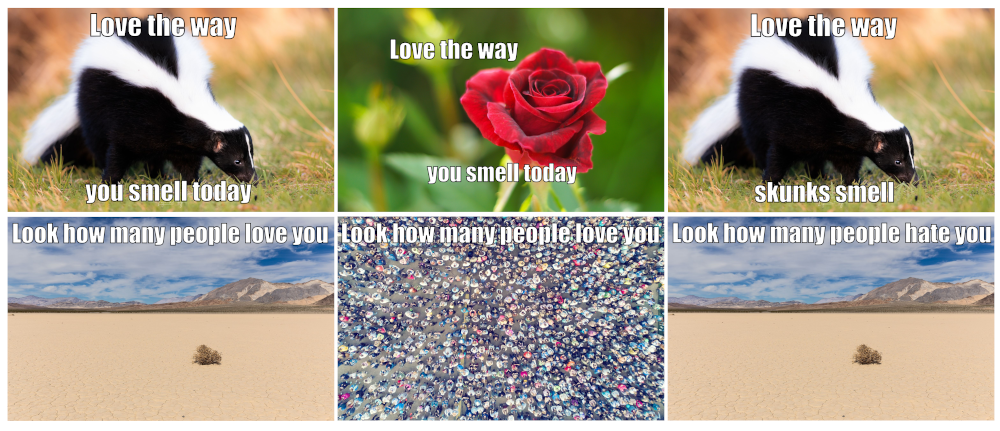}
    \caption{\textbf{Multimodal ``mean'' memes \cite{hateful_memes}}. \emph{Left:} Mean memes, \emph{Middle:} Benign Image Confounders, \emph{Right:} Benign Text Confounders.}
    \label{fig:hateful_memes}
\end{figure}

This problem falls into the category of \emph{Visual Question Answering} (VQA), i.e., the task of answering previously unseen questions framed in natural language about a previously unseen image (e.g,. \say{Does this image include hate speech?}). The classical approach for VQA is to learn a model from a training set made up of image $\vv$, question $\vq$ and answer $a$ triplets $\mathcal{D}=$ $\left\{\left\langle\boldsymbol{q}_{i}, \boldsymbol{v}_{i}, a_{i}\right\rangle\right\}_{i=1}^{n}.$ The model infers an embedding of the questions $\ve^{q}=f_{q}(\boldsymbol{q})$, an embedding of the image $\ve^{v}=f_{v}(\boldsymbol{v})$ and a fusion function of the two $\vz=h\left(\ve^{q}, \ve^{v}\right)$ into what is known as the \emph{joint space}. 

One challenge for such models is to deal with spurious associations between the image and text modalities. For instance, \Cref{fig:hateful_memes}(\emph{Left}) illustrates common mean memes; memes with text like \say{love the way you smell today} might be spuriously associated with images of unpleasant smells. These spurious associations make it harder for models to truly capture multimodal understanding, as shown by \citet{hateful_memes}. The authors develop a benchmark with \emph{benign confounders}, which are minimum replacement images/texts that flip the labels for a given multimodal meme from hateful to non-hateful. Benign Image and Text confounders are shown in the \emph{Middle} and \emph{Right} of \Cref{fig:hateful_memes}, respectively. They find that when evaluated on this benchmark, state-of-the-art methods perform poorly compared to humans.

In the following, we look at two causal VQA methods that propose to break spurious associations between text and image. 

\textbf{Counterfactual Vision and Language Learning.}
\citet{abbasnejad_counterfactual_2020} suggest to train VQA models on both observational and generated counterfactual samples to improve generalization. The motivation behind this procedure is to force the model to use both input modalities instead of relying on correlations from one modality only. Typical VQA models possess feature extractors $\ve^q = f_q(\vq)$ and $\ve^v = f_v(\vv)$ for question and visual input $\vq$ and $\vv$, respectively. In contrast, \citet{abbasnejad_counterfactual_2020} construct an SCM in which the feature extractors depend on exogenous variables: $f_{v}$ is replaced by $\tilde{f}_{v}\left(\boldsymbol{v}, \vu^{v}\right)$ and $f_{q}$ by $\tilde{f}_{q}\left(\vq, \vu^{q}\right)$ where $\vu^{v}$ and $\vu^{q}$ are exogenous variables for image (vision module) and question (language module), respectively. During training, they intervene on either $\vq$ or $\vv$, denoted by $\tilde{\boldsymbol{q}}$ and $\tilde{\boldsymbol{v}}$, respectively, and yield the corresponding embeddings $\tilde{\ve}^{q}$ and $\tilde{\ve}^{v}$. They find their approach effective on both unimodal vision and language tasks as well as multi-modal vision-and-language tasks.

\textbf{Counterfactual VQA.}
\begin{figure}
    \centering
    \begin{subfigure}[t]{\columnwidth}
    \includegraphics[width=\linewidth]{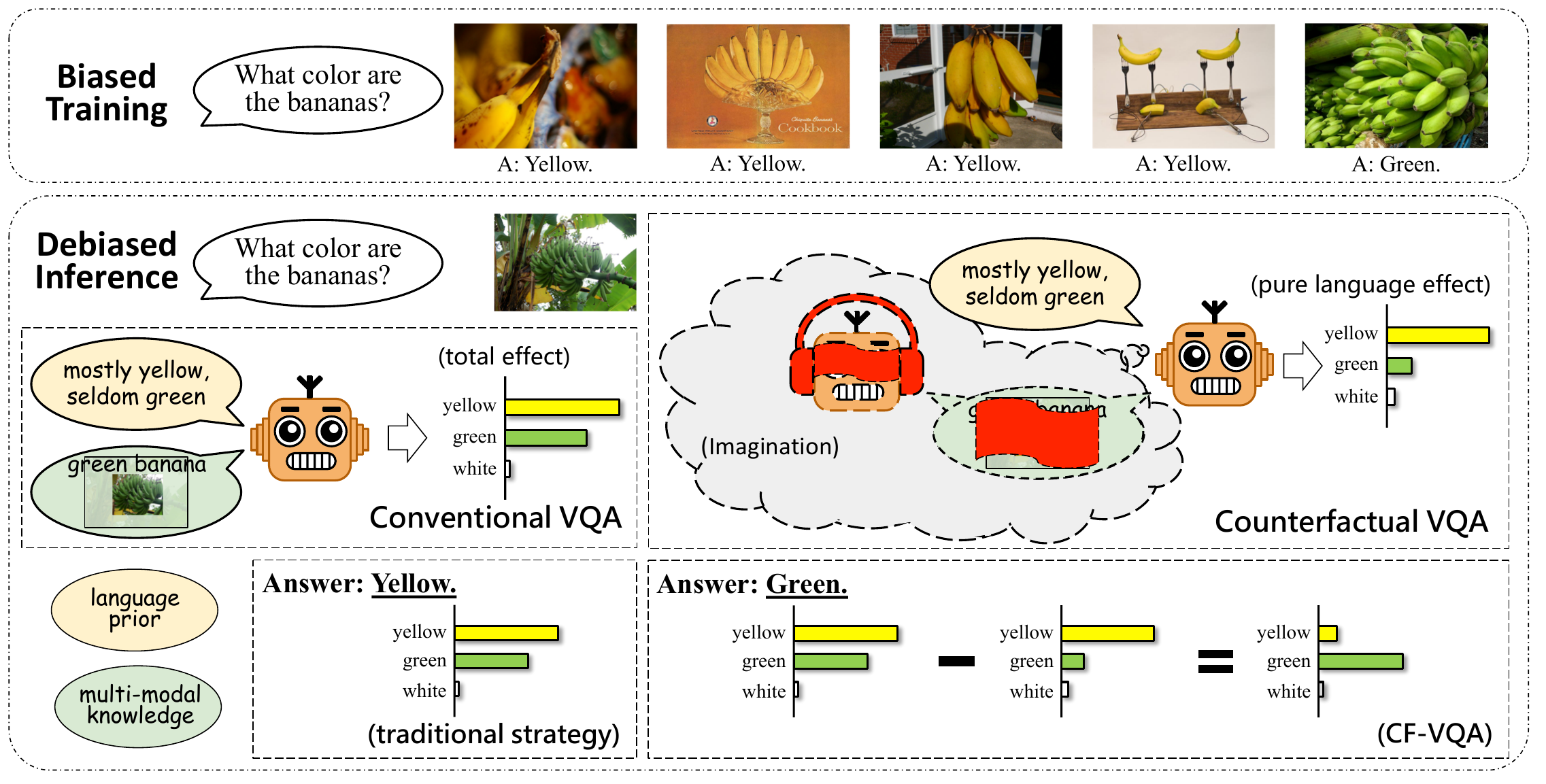}
    \caption{\textbf{Example of spurious associations between questions and answers}. The query \say{What color are the bananas?} is most often associated with the answer \say{yellow}.}
    \label{fig:cfvqa_teaser}
    \end{subfigure}
    \begin{subfigure}[t]{\columnwidth}
    \begin{subfigure}[t]{0.32\columnwidth}
        \includegraphics[width=\linewidth]{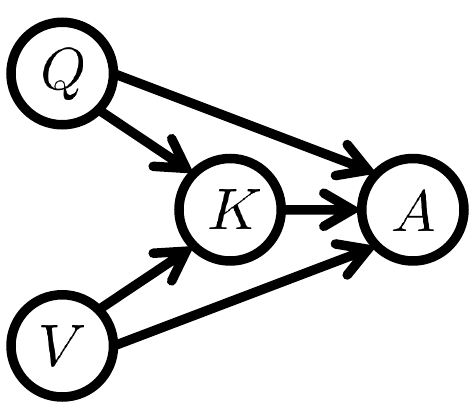}
        \caption{\textbf{Causal DAG for VQA.} $\rmQ$: question. $\rmV$: image. $\rmK$: multi-modal knowledge. $A$: answer.}
    \end{subfigure}
    \begin{subfigure}[t]{0.64\columnwidth}
        \includegraphics[width=\linewidth]{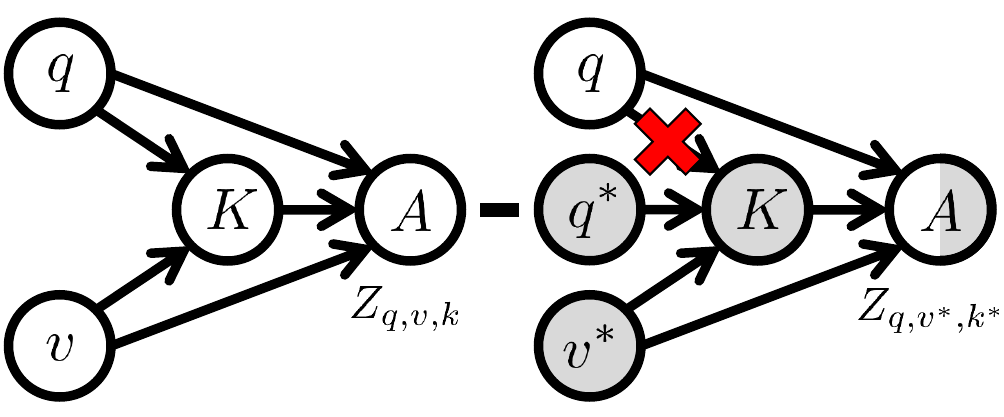}
        \caption{\textbf{Comparison between conventional VQA (left) and counterfactual VQA (right).} Unshaded nodes possess the value $\rmV\!=\!\vv$ and $\rmQ\!=\!\vq$. Shaded nodes are at the value $\rmV\!=\!\vv^*$ and $\rmQ\!=\!\vq^*$, where $\vv^*, \vq^*$ denote the non-treatment condition where $\vv$ and $\vq$ are not given.}
        \label{fig:cfvqa_dag}
    \end{subfigure}
    \end{subfigure}
    \caption{\textbf{Counterfactual VQA \cite{CF_VQA}}. Conventional VQA relies on spurious associations. \method{CFVQA} isolates the language effect by imaging the scenario where the model reads the question but the multi-modal knowledge (including image) is removed. By subtracting the pure language effect from the total effect, \method{CFVQA} deconfounds the answer} 
    \label{fig:cfvqa}
\end{figure}
\citet{CF_VQA} formulate language confounding as the direct causal effect of questions on answers. To deconfound such spurious associations, they propose to subtract the direct language effect from the total causal effect. They refer to their approach as \emph{Counterfactual VQA} (\method{CFVQA}).

\Cref{fig:cfvqa_teaser} illustrates how conventional and counterfactual VQA differ in their workings.
Conventional VQA asks \say{What will answer $A$ be, if machine hears question $Q$, sees image $V$, and extracts the multimodal knowledge K?}. Doing so cannot disentangle the single-modal linguistic effect and the multimodal reasoning effect. 

To isolate the linguistic effect, \citet{CF_VQA} consider the following counterfactual question: \say{What would have happened if the model had not performed multimodal reasoning?}. This corresponds to the counterfactual query in which the model the machine considers $Q$, but the multimodal knowledge $K$ is intervened under the no-treatment condition, i.e., as if $V$ and $Q$ had not been accessible. Since the response of $K$ to $Q$ is now blocked, the model can only rely on the single-modal effect, effectively isolating the language bias. \Cref{fig:cfvqa_dag} depicts the corresponding causal DAGs. 

By inferring the counterfactual query \say{What would $A$ be, if the model reads $Q$, but had not extracted $K$ or seen $V$?}, \method{CFVQA} extracts the language bias. Then, to reduce it in the final query, it is subtracted from the total effect of $\rmV = \vv$ and $\rmQ = \vq$ on $A=a$, also referred to as the \emph{total direct effect} (TDE). This TDE estimand is different than conventional VQA's posterior distribution $p(a \mid \vv, \vq)$.

\textbf{Certified Robustness Against Natural Language Attacks} \label{nlp:caus_int_nlp}

\citet{alzantot-etal-2018-generating} expose that sentiment analysis models can be fooled by synonym substitution attacks, as illustrated by their adversarial examples in \Cref{tab:text_adversarial_examples}. This has motivated a myriad of works making NLP models more robust against such attacks \cite{scpn,hotflip,safer,adversarial_text}.

\begin{table}[!t]
\centering
\begin{tabular}{|P{\columnwidth}|}
\hline
Original Text Prediction = \textbf{Negative}. (Confidence = 78.0\%) \\
\hline 
\textit{This movie had \orw{terrible} acting, \orw{terrible} plot, and \orw{terrible} choice of actors. (Leslie Nielsen ...come on!!!) the one part I \orw{considered} slightly funny was the battling FBI/CIA agents, but because the audience was mainly \orw{kids} they didn't understand that theme.} \\
\hline
\hline
Adversarial Text Prediction = \textbf{Positive}. (Confidence = 59.8\%)  \\
\hline
\textit{This movie had \mow{horrific} acting, \mow{horrific} plot, and \mow{horrifying} choice of actors. (Leslie Nielsen ...come on!!!) the one part I \mow{regarded} slightly funny was the battling FBI/CIA agents, but because the audience was mainly \mow{youngsters} they didn't understand that theme.} \\
\hline
\end{tabular}
\caption{\textbf{Natural Language Adversarial Examples for the sentiment analysis task \cite{alzantot-etal-2018-generating}.} We highlight modified words in green and red for the original and adversarial texts, respectively.}
\label{tab:text_adversarial_examples}
\end{table}

\begin{figure}
    \centering
    \begin{subfigure}[t]{0.48\columnwidth}
        \includegraphics[width=\linewidth]{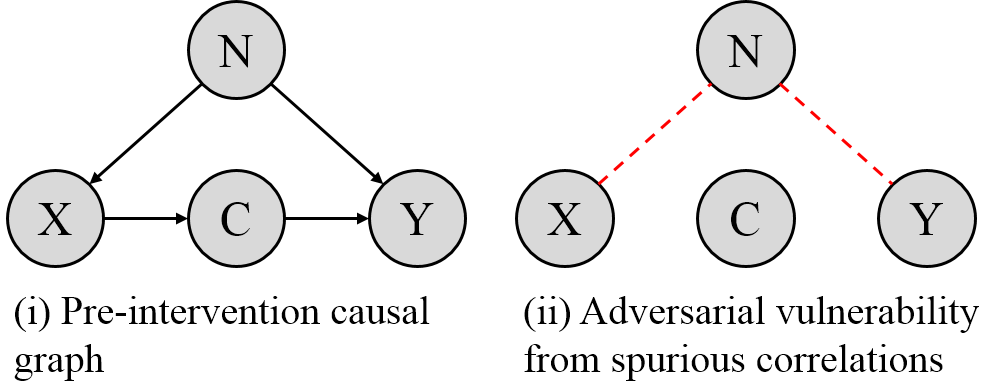}
        \caption{\textbf{Causal DAG for CISS.} We denote the text by $\rmX$, label by $Y$, content variable by $\rmC$, and manipulatable confounder by $\rmN$.}
        \label{fig:ciss_dag}
    \end{subfigure}
    \begin{subfigure}[t]{0.48\columnwidth}
        \includegraphics[width=\linewidth]{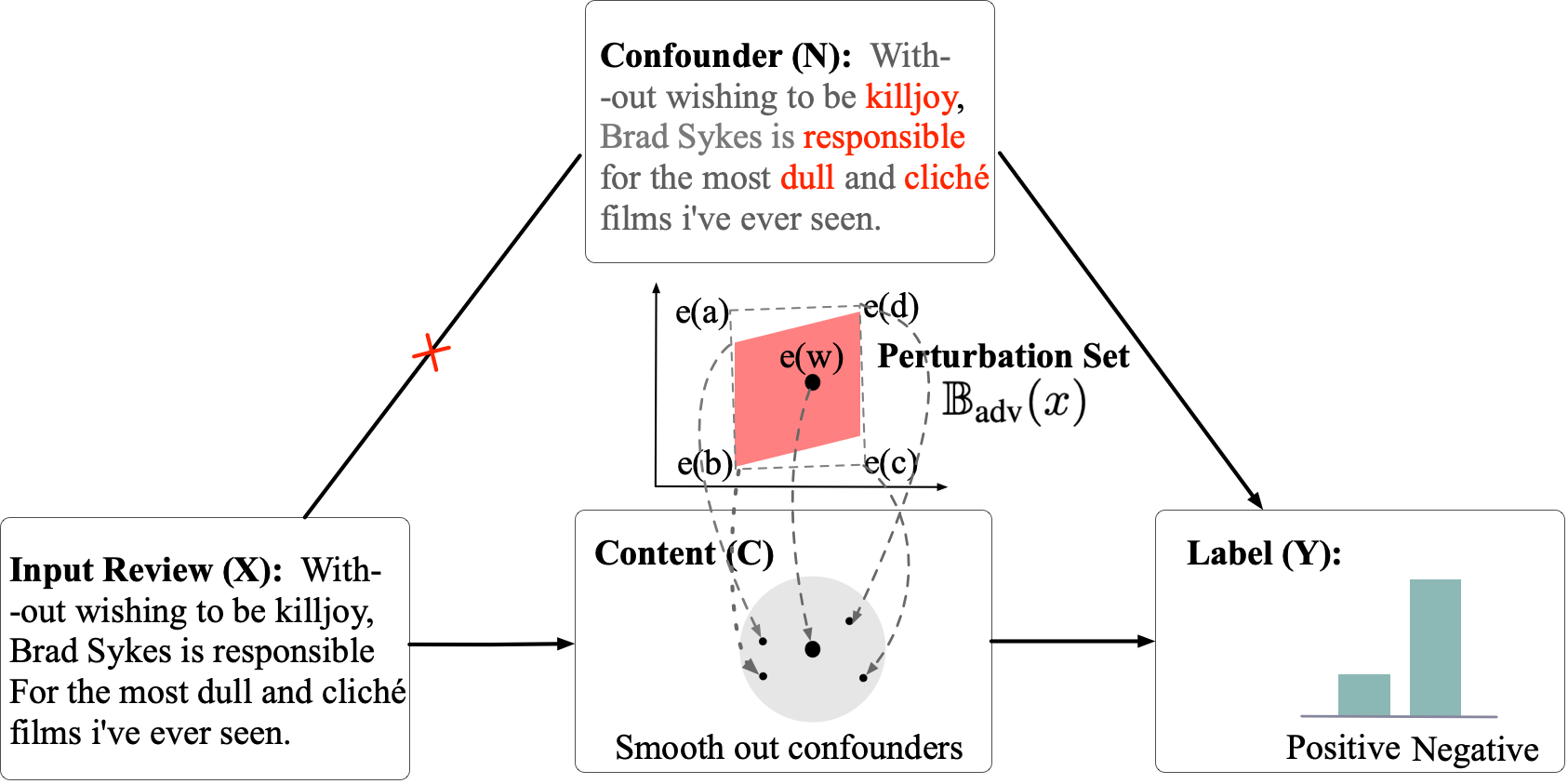}
        \caption{\textbf{Exemplary illustration of CISS pipeline.} Noise added to the latent representations smooths out the confounder effect that otherwise is exploitable by potential attackers.}
        \label{fig:ciss_pipeline}
    \end{subfigure} 
    \caption{\textbf{Causal Intervention by Semantic Smoothing (CISS) \cite{zhao2022certified}:} A framework towards robustness against natural language attacks by learning $p(y \mid \doo(\vx))$.}
    \label{fig:ciss}
\end{figure}

\citet{zhao2022certified} take a causal perspective on the natural language attack problem and frame the source of adversarial vulnerability as the spurious association induced by confounders. \Cref{fig:ciss_dag} illustrates their Causal DAG. For instance, when considering a movie review ($\rmX$) from the IMBD dataset \cite{imdb}, a professional reviewer likely uses jargon ($\rmN$), has high standards, and therefore, is likelier to assign a lower score on average. This spurious association can be exploited, e.g., by adding more jargon words to a positive movie review. 

To defend models against such attacks, \citet{zhao2022certified} 
show that a Gaussian-based randomized classifier models the interventional distribution $p(y \mid \doo(\vx))$ and is therefore robust against $l_{2}$-bounded attacks. However, textual input spaces are not continuous, and text substitutions do not follow Gaussian distributions. To circumvent these issues, they propose smoothening out the latent semantic space of the learned content variable $\rmC$ instead, as illustrated in \Cref{fig:ciss_pipeline}. They refer to this framework as \emph{Causal Intervention by Semantic Smoothing} (CISS).

\subsection{Counterfactual Explanations}

\textbf{Causal Model Explanation Through Counterfactual Language Models} \label{app:causalm}
Understanding the effect of a concept in the input on a model is crucial for explanation and model dissemination. However, this usually requires generating counterfactual sequences by dropping/replacing the concept of interest, which is challenging for existing text generation models. \citet{causaLM} propose a framework named CausalLM to generate counterfactual representations instead of counterfactual sequences. To this end, \citet{causaLM} fine-tune deep contextualized embedding models with auxiliary adversarial tasks to encourage the model to ``forget'' the concept of interest. The representation of the input sequence and the counterfactual representation forgetting the concept of interest are fed into a classifier to measure the effect of the concept on the classifier's prediction.

\textbf{Investigating Gender Bias Using Causal Mediation Analysis}
\label{nlp:gender_bias}
Many text corpora contain spurious associations due to gender stereotypes propagated or amplified by NLP systems. For example, the sentence \say{he is an engineer} is more likely to appear in a corpus than \say{she is an engineer} due to the current gender disparity in engineering \cite{bio_bias}. 

\begin{figure}[t]
    \centering
    \includegraphics[width=\linewidth]{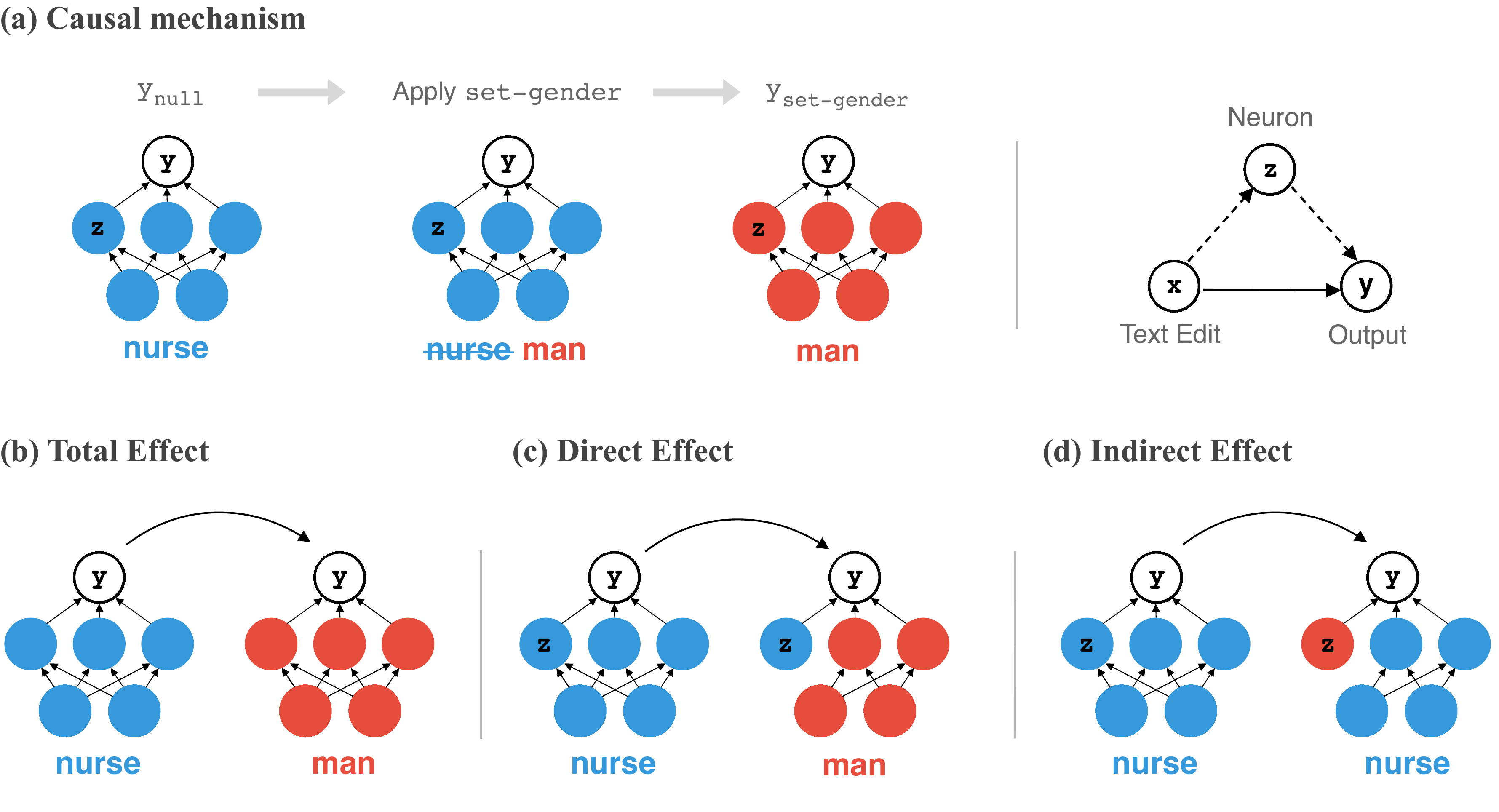}
    \caption{\textbf{Investigating Gender Bias using Mediation Analysis \cite{gender_bias}.} Given a prompt $\vu$ such as ``The nurse said that'', we ask a language model to generate a continuation. A biased model may assign a higher likelihood to \textit{she} than to \textit{he}. To understand the role of model components on this biased prediction, we perform the $\doo$-operation $\vx=\texttt{set-gender}$, which changes $\vu$ from \textit{nurse} to \textit{man} in this example. By inferring direct and indirect effects, we can analyze the causal role of specific mediators (neurons) between $\vx$ and $\vy$.} 
    \label{fig:mediation}
\end{figure}
\citet{gender_bias} propose a method to interpret which parts of a model are biased. Their method uses causal mediation analysis to locate which parts of a neural model are causally implicated, as illustrated in \Cref{fig:mediation}. This approach is arguably better than other analysis tools like probing, as probing can only measure whether the information is encoded in hidden representations rather than whether the information is used by the model.

\textbf{Counterfactual Fairness in Text Classification through Robustness} Text classifiers are sensitive to specific contents of the input. For example, a toxicity model predicts that the toxicity of "Some people are gay" is 98\%, while the toxicity of "Some people are straight" is 2\%. \citet{cf_fairness_text_classification} study counterfactual fairness in text classification by asking the counterfactual question ``How would the output of a classifier change if some sensitive tokens were different?''. They define a metric named \emph{counterfactual token fairness} to measure the differences between the outputs before and after substituting tokens associated with identity groups. They further evaluate three methods to promote fairness: replacing all sensitive tokens
with a special token, counterfactual data augmentation, and a regularization called \emph{counterfactual logit pairing}. 

\subsection{Counterfactual Data Augmentation}
\label{nlp:cda}

\textbf{Counterfactual Generator}
\citet{counterfactual_generator} propose a new data augmentation algorithm for entity recognition with counterfactual inference. Each input sequence is divided into two parts, including entity and context. An entity in the input sequence is replaced with another entity of the same entity type. The augmented example is kept if a discriminator correctly recognizes the replaced entity. \citet{counterfactual_generator} find that this data augmentation method can improve the generalization ability in low-resource settings. Besides, training on augmented examples can partially eliminate spurious correlations between contexts and output labels.

\textbf{Counterfactual Data Augmentation for Neural Machine Translation} 
\citet{DBLP:conf/naacl/LiuKB21} design a counterfactual data augmentation method for neural machine translation. The method interprets language models and phrasal alignment causally. It creates augmented parallel translation corpora by answering the question ``Given that $Y_j$ is aligned to $X_i$, what would $Y_j$ have looked like, had $X_i=\hat{x}_i$ and all other phrases $\mathcal{X}_{-i}, \mathcal{Y}_{-j}$ had been held constant?''. They use a masked language model and a translation language model to replace the source and target phrases, respectively. Compared to previous work, this method takes context and alignment
into account for data augmentation.

\textbf{Mitigating Gender Stereotypes in Languages with Rich Morphology}

In \Cref{nlp:gender_bias}, we learned that many text corpora include gender biases. Typically, such biases exist in many different languages around the world.
Most NLP research has focused on mitigating gender stereotypes in English \cite{NIPS2016_a486cd07,zhao-etal-2017-men}. However, these approaches often create ungrammatical sentences in morphologically rich languages like Spanish. To this end, \citet{zmigrod-etal-2019-counterfactual} present a counterfactual data augmentation approach for mitigating gender stereotypes associated with nouns representing people in such languages. 

Their unsupervised approach uses dependency trees, lemmata, part-of-speech tags, and morpho-syntactic tags from Universal Dependencies corpora \cite{universal_dependencies}. It consists of four steps: (1) Analysis of the sentence (including parsing, etc.), (2) Intervention on a gendered word, (3) Inference of the new morpho-syntactic tags, and (4) Reinflection of the lemmata to their new forms. Using four different languages, they demonstrate that their approach reduces gender stereotyping by an average of 2.5 without sacrificing grammar.

\subsection{Causal Traces}
The advent of transformer-based Large Language Models (LLMs) \citep{DBLP:journals/corr/abs-2303-08774,DBLP:journals/corr/abs-2312-11805,claude2,DBLP:journals/corr/abs-2305-10403,DBLP:journals/corr/abs-2307-09288, anthropic2024claude, DBLP:journals/corr/abs-2307-10169, llama3modelcard} marked a significant advancement in NLP. The downside of scaling model parameters and pre-training dataset sizes is that the models get harder to interpret \cite{bricken2023monosemanticity}.

\textbf{Causal Traces}
The advent of transformer-based Large Language Models (LLMs) \citep{DBLP:journals/corr/abs-2303-08774,DBLP:journals/corr/abs-2312-11805,claude2,DBLP:journals/corr/abs-2305-10403,DBLP:journals/corr/abs-2307-09288,anthropic2024claude,DBLP:journals/corr/abs-2307-10169, llama3modelcard} marked a significant advancement in NLP. The downside of scaling model parameters and pre-training dataset sizes is that the models get harder to interpret \cite{bricken2023monosemanticity}.

\begin{figure}
    \centering
    \includegraphics[width=\linewidth]{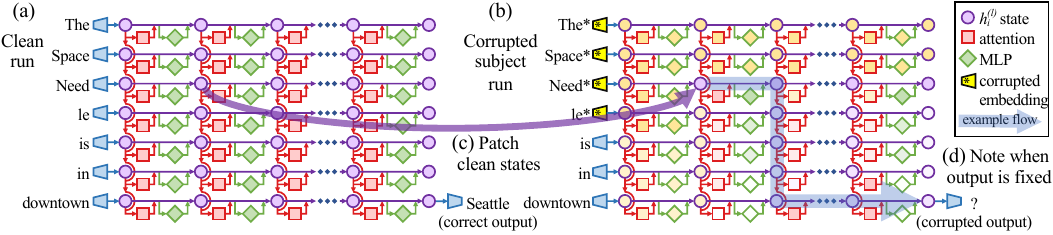}
    \caption{\textbf{Causal Tracing} \cite{meng2022locating} assesses the causal impact of neuron activations on a model's factual predictions by the following steps: (a) run the model on the original input and record its activations and output; (b) corrupt the subject token in the input (e.g., by adding noise) and run the model again, resulting in an altered prediction; (c) starting from the corrupted input, selectively restore certain internal activations to their original "clean" values from step (a); (d) if the output returns to the original correct prediction, it indicates that the restored activations are causally responsible for that prediction. The light blue path highlights an example of the information flow through these key activations.}
    \label{fig:causal_tracing}
\end{figure}

\citet{meng2022locating} analyze how factual knowledge is stored within such LLMs like GPT2 \cite{gpt2}, representing facts as knowledge tuples $t = (s, r, o)$, where $s$ is the subject, $r$ is the relation, and $o$ is the object. They introduce a method called \emph{causal tracing} (illustrated in \Cref{fig:causal_tracing}), based on \citet{gender_bias}'s method using causal mediation analysis that we discussed in \Cref{nlp:gender_bias}. To trace the flow of factual information through the model's hidden states, causal tracing measures the causal effect of individual hidden states on the model's output by intervening on these states during the forward pass and observing changes in the prediction of $o$. Their findings indicate that specific hidden states, particularly in mid-layer MLP modules at the last token of the subject, have a strong causal effect on predicting factual associations. 

Building on this insight, they propose \emph{Rank-One Model Editing (ROME)}, a method to modify the model's weights to insert or update specific factual associations by applying a rank-one update to the weights of targeted MLP modules. 

\subsection{Miscellaneous}

\textbf{Causal Direction of Data Collection} \label{nlp:data_collection}
The supervised learning problem aims to predict a label $Y$ based on features $\rmX$. From a causal perspective, we can further distinguish this problem into two cases: \emph{causal} and \emph{anti-causal} learning, determined by the data collection process being $\rmX \rightarrow Y$ or $Y \rightarrow \rmX$, respectively. In words, if, during the data collection process, $\rmX$ is generated first, and then $Y$ is collected based on $\rmX$ (e.g., through annotation), we say that $\rmX$ causes $Y$ (causal learning). Vice versa, if $Y$ is generated first, and then $\rmX$ is collected based on $Y$, we say that $Y$ causes $\rmX$ (anti-causal learning).

Previous works show that this simple distinction has important implications for scenarios such as covariate shift, transfer learning, semi-supervised learning \cite{anticausal}, and adversarial examples \cite{KilParSch18}.

\begin{table}[h]
    \centering
    \begin{minipage}{\textwidth}
        \centering
        \resizebox{\linewidth}{!}{%
    \begin{tabular}{|P{3cm}|P{10cm}|}
    \toprule
    \textbf{Category} & \textbf{NLP Tasks} \\ \midrule
    \textbf{Causal} & Summarization, parsing, tagging, data-to-text generation, information extraction \\ \hline
    \textbf{Anticausal} & Author attribute / review sentiment classification \\ \hline
    \textbf{Mixed} & Machine translation, question answering, question generation, text style transfer, intent classification \\
    \bottomrule
    \end{tabular}
    }
    \caption{\textbf{Classification of typical NLP tasks into causal, anticausal, and mixed learning tasks \cite{Jinetal21}}. Causal learning means that the model takes the cause as input and predicts the effect; anticausal refers to settings where the model takes the effect as input and predicts the cause. Some tasks do not have a clear causal interpretation of the data collection process, or a mixture of both types of data is typically used.}
    \label{tab:nlp_task_class}
    \end{minipage}
\end{table}

\citet{Jinetal21} study the causal direction of the data collection process for common NLP datasets, as summarized in \Cref{tab:nlp_task_class}. For example, they observe that language sentence pairs are mixed up in machine translation regardless of their original source-to-target direction (e.g., whether a sentence originated in English and was translated into Spanish or vice versa). Splitting the data into subsets reveals that they exhibit different properties, such as performance differences in self-supervised or domain adaptation settings. In light of these findings, the authors make various recommendations for future research, e.g., annotating the causal direction when collecting new NLP data and incorporating it into the model.

\section{Causal Graph Representation Learning}

In numerous applied fields, graphs represent systems of relations and interactions. Examples include social networks \cite{fan2019graph}, molecular graphs \cite{molecule_ssl}, protein associations \cite{jing2021fast}, and programming code syntax trees \cite{hu2020open}. Graph Neural Networks (GNNs) form an effective framework for learning representations over graph-structured data.

\subsection{Causal Supervised Learning}

\textbf{Discovering Invariant Rationales for GNNs} Intrinsic interpretability of graph neural networks aims to discover a small subset of an input graph that contributes most to the model prediction. However, most work on intrinsic interpretability is prone to discover data biases and spurious correlation, failing to capture causal patterns. Besides, these methods suffer from performance degradation on out-of-distribution data as they are sensitive to spurious correlation. \citet{wu2022discovering} propose a new method by discovering invariant rationale. A rationale generator is first applied to split each input graph into causal and non-causal subgraphs. Next, they encode causal and non-causal subgraphs into hidden representations. Then causal interventions are conducted on the non-causal representations to create perturbations. Finally, they utilize two classifiers for the final prediction, trained through an invariant risk loss function.

\textbf{Invariance Principle Meets Out-of-Distribution Generalization on Graphs}
\citet{chen2022invariance} propose the Graph Out-Of-Distribution (GOOD) framework, akin to the Style- and Content Decomposition (\Cref{def:scd}), to learn invariant graph features. To this end, a featurizer is designed to distinguish invariant subgraphs from the other parts of the graph that domain shifts can easily perturb. The authors demonstrate that this approach improves out-of-domain generalization.

\begin{figure}
    \centering
    \begin{subfigure}[t]{0.4\columnwidth}
    \includegraphics[width=0.9\columnwidth]{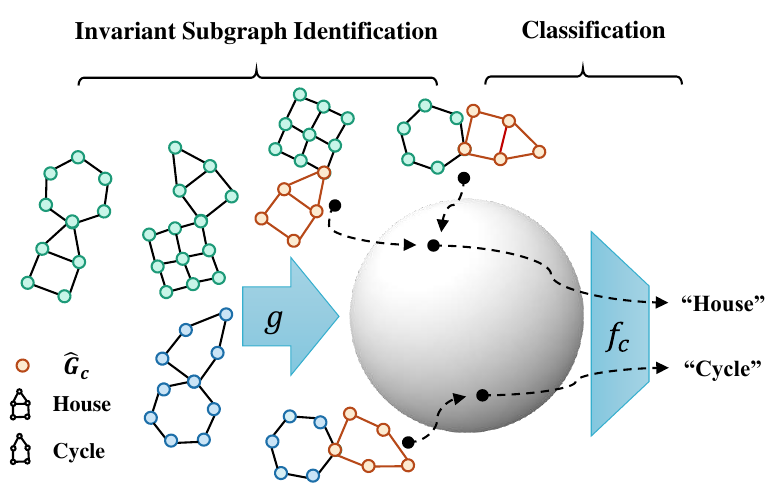}
    \caption{\textbf{Invariant Subgraph Identification:} The encoder $g(\cdot)$ extracts invariant subgraphs causing the labels by maximally preserving invariant intra-class information.} \label{fig:good_subgraph}
    \end{subfigure}    \begin{subfigure}[t]{0.5\textwidth}
    \begin{subfigure}[t]{0.32\textwidth}
		\resizebox{!}{\columnwidth}{\tikz{
				\node[latent] (S) {$\rmS$};%
				\node[latent,left=of S,xshift=0.5cm] (C) {$\rmC$};%
				\node[latent, below=of C,xshift=0.5cm,yshift=0.75cm] (EG) {$E_G$};%
				\node[latent,below=of C,xshift=-0.5cm,yshift=0.5cm] (GC) {$\rmG_c$}; %
				\node[latent,below=of S,xshift=0.5cm,yshift=0.5cm] (GS) {$\rmG_s$}; %
				\node[obs,below=of GC,xshift=1cm,yshift=0.5cm] (G) {$\rmG$}; %
				\edge[dashed,-] {C} {S}
				\edge {C} {GC}
				\edge {S,EG} {GS}
				\edge {GC,EG,GS} {G}
			}}
			\caption{$\gG$-SCM}\label{fig:graph_gen}
	\end{subfigure}
	\begin{subfigure}[t]{0.32\textwidth}
		\resizebox{!}{\columnwidth}{\tikz{
				\node[latent] (E) {$E$};%
				\node[latent,below=of E,yshift=0.5cm] (S) {$\rmS$}; %
				\node[obs,below=of E,xshift=-1.2cm,yshift=0.5cm] (Y) {$Y$}; %
				\node[obs,below=of E,xshift=1.2cm,yshift=0.5cm] (G) {$\rmG$}; %
				\node[latent,below=of Y,xshift=1.2cm,yshift=0.5cm] (C) {$\rmC$}; %
				\edge {E} {S}
				\edge {C} {Y,G}
				\edge {S} {G}
				\edge {C} {S}
			}}
	\caption{FIIF SCM}
	\label{fig:scm_fiif}
	\end{subfigure}	\begin{subfigure}[t]{0.32\textwidth}
		\resizebox{!}{\columnwidth}{\tikz{
				\node[latent] (E) {$E$};%
				\node[latent,below=of E,yshift=0.5cm] (S) {$\rmS$}; %
				\node[obs,below=of E,xshift=-1.2cm,yshift=0.5cm] (Y) {$Y$}; %
				\node[obs,below=of E,xshift=1.2cm,yshift=0.5cm] (G) {$\rmG$}; %
				\node[latent,below=of Y,xshift=1.2cm,yshift=0.5cm] (C) {$\rmC$}; %
				\edge {E} {S}
				\edge {C} {Y,G}
				\edge {S} {G}
				\edge {Y} {S}
			}}
	\caption{PIIF SCM}
	\label{fig:scm_piif}
    \end{subfigure}	
\end{subfigure}
    \caption{\textbf{Graph Out-Of-Distribution (GOOD) Generalization Framework \cite{chen2022invariance}}: A Style- and Content Decomposition (\Cref{def:scd}) perspective on graph distribution shifts.}
    \label{fig:good}
\end{figure}
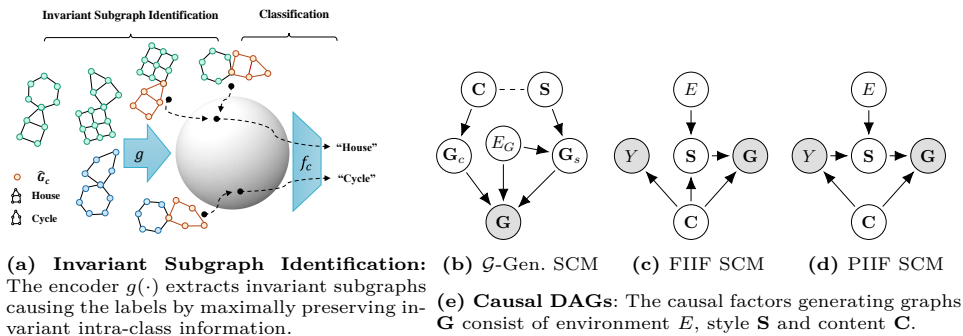

Assuming the true underlying causal DAG in \Cref{fig:graph_gen}, \citet{chen2022invariance} categorize the interactions between content $\rmC$ and style $\rmS$ into three possible DAGs, depending on whether $\rmC$ is fully informative about $Y$, i.e., $(\rmS, E) \Perp Y \mid \rmC$: Fully Informative Invariant Features (FIIF, \Cref{fig:scm_fiif}), Partially Informative Invariant Features (PIIF, \Cref{fig:scm_piif}). The main difference is that $\rmS$ is directly controlled by $\rmC$ in FIIF, and indirectly controlled by $\rmC$ through $Y$ in PIIF, which can exhibit different behaviors in the observed distribution shifts. 

To extract the invariant subgraph containing the latent content features $\rmC$ from the complete observed graph $\rmG$, the authors explicitly align the two causal mechanisms during the graph generation, i.e., $C \rightarrow G$ and $\left(\rmG_{s}, E_{G}, \rmG_{c}\right) \rightarrow G$. \Cref{fig:good_subgraph} illustrates the alignment procedure: they decompose a GNN into two sub-components: i) a featurizer GNN $g: \mathcal{G} \rightarrow \mathcal{G}_{c}$ aiming to identify the desired $\rmG_{c} ;$ and ii) a classifier GNN $f_{c}: \mathcal{G}_{c} \rightarrow \mathcal{Y}$ that predicts $Y$ based on the estimated $\rmG_{c}$, where $\mathcal{G}_{c}$ refers to the space of subgraphs of $\rmG$.

\textbf{Deconfounded Training for Graph Neural Networks} 
\citet{Sui2021DeconfoundedTF} aims to discriminate each input graph into critical and trivial parts. Since the trivial part is a confounder between the critical part and the label, this opens a backdoor and leads to spurious correlations. Therefore, \citet{Sui2021DeconfoundedTF} propose deconfounded training to mitigate the confounding effect. They use attention modules to split each graph into a critical and trivial subgraphs. Then they eliminate the confounding effect of trivial subgraphs by performing the backdoor adjustment. 

\subsection{Counterfactual Data Augmentation}
\label{grl:cda}

\textbf{Learning from Counterfactual Links for Link Prediction}
A common GNN task is to predict edges (or \emph{links}) between node pairs. For example, links can refer to citations between papers \cite{sen2008collective}, relations in knowledge graphs \cite{hu2020open}, or molecule interactions \cite{molecule_ssl}.  

\citet{counterfactual_graph_learning} point out that the causal relationship between the graph structure and link existence has been largely ignored. The authors give the following example of a social network. Imagine Alice and Adam live in the same neighborhood, and they are close friends. Associating neighborhood belonging with friendship could be too strong to identify the essential components of friendship, like mutual interests or family ties. Such spurious associations may also explain why they live in the same neighborhood. 

Consequently, \citet{counterfactual_graph_learning} propose to examine the counterfactual query \say{would Alice and Adam still be close friends if they were not living in the same neighborhood?}. Generally speaking, any counterfactual link prediction queries follow the structure of \say{would the link exist or not if the graph structure was different from that observed?}. Hence, if we train link prediction models on such corresponding counterfactuals, they rely less on spurious associations, as described above. 

\begin{figure}
    \centering
    \begin{subfigure}[t]{.2\linewidth}
        \centering
        \includegraphics[width=\columnwidth]{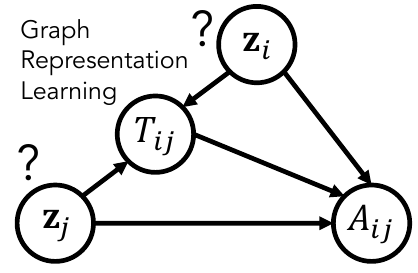}
        \caption{\textbf{Causal DAG}: $\vz_{i}$ and $\vz_{j}$ are representations of nodes $v_{i}$ and $v_{j}$, $T_{i,j}$ the neighborhoods, and the outcome $A_{i, j}$ is the link existence between $v_{i}$ and $v_{j}$.}
        \label{fig:cflp_dag}
    \end{subfigure}    \begin{subfigure}[t]{0.79\linewidth}
        \begin{subfigure}[t]{.42\linewidth}
            \centering
            \includegraphics[width=\columnwidth]{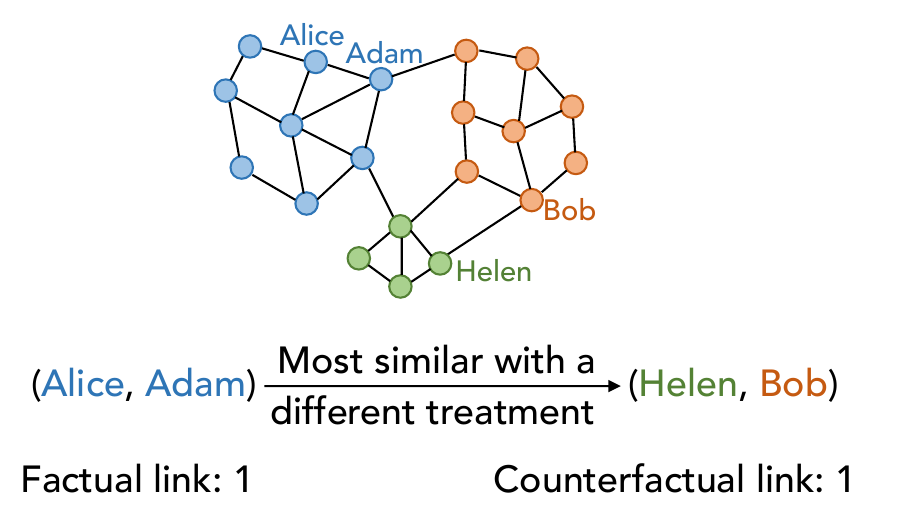}
            \caption{\textbf{Step 1:} Find counterfactual link based on most similar node pair with a different intervention (neighborhood).}
            \label{fig:cflp_framework1}
        \end{subfigure}
        \begin{subfigure}[t]{.57\linewidth}
            \centering
            \includegraphics[width=\columnwidth]{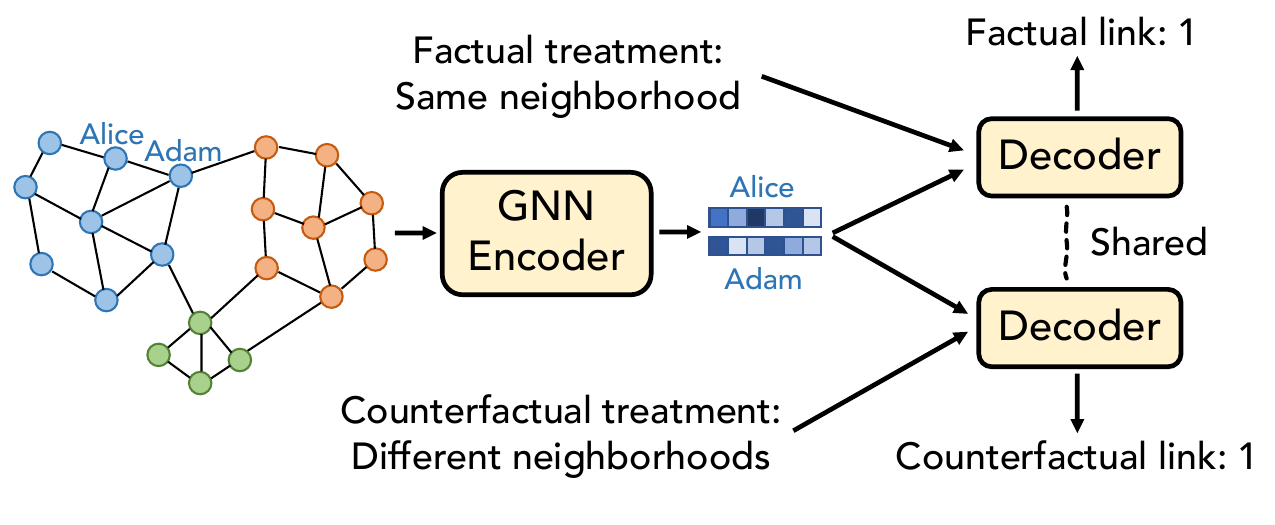}
            \caption{\textbf{Step 2:} Train a GNN-based link predictor which predicts factual and counterfactual links given the node pairs and neighborhoods.}
            \label{fig:cflp_framework2}
        \end{subfigure}
        \caption{\textbf{Two-step procedure}.}
        \end{subfigure}
    \caption{\textbf{Counterfactual Link Prediction Framework \cite{counterfactual_graph_learning}.}}
        \label{fig:cflp}
\end{figure}

\citet{counterfactual_graph_learning} generate counterfactual links to augment the training data for link prediction tasks. Their approach estimates the causal relationship between the observed graph structure (considered as the intervention) and link existence (the outcome). \Cref{fig:cflp} summarizes their approach. 

\subsection{Miscellaenous}

\textbf{Tackling Over-Smoothing by Causal Effect Uncertainty}
Real-world graphs usually exhibit locally varying structures, i.e., uneven distributions of properties such as homophily and degree. However, a commonly-faced issue faced by GNNs is \emph{over-smoothing}, which means that the representations of the graph nodes of different classes become indistinguishable, often causing local structure discrepancies \cite{over-smoothing}.

\begin{figure}[t]
	\centering
	\includegraphics[width=0.7\textwidth]{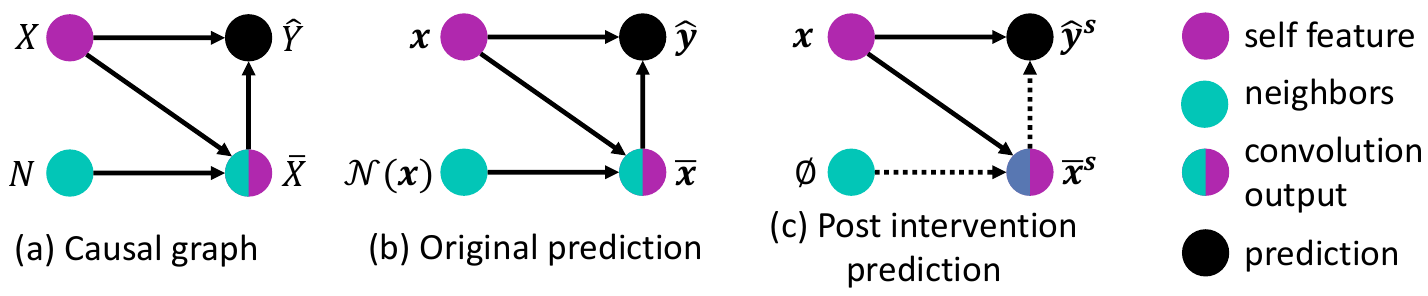}
	\caption{\textbf{Causal Graph Convolutional Network Inference (\method{CGI}) \cite{feng2021should}.} (a) Causal graph of \method{CGI}; (b) Original prediction; (c) causal intervention $\doo(N = \emptyset)$ where the dashed arrow indicates that the effect from the predecessor is blocked.
	}
	\label{fig:app_cgi}
\end{figure}

\citet{feng2021should} aim to resolve such node-specific structure discrepancies \emph{at inference time} by taking into account the uncertainty of the causal effect of the local structure on the prediction. The causal effect uncertainty indicates whether a test node should be trusted when predicting its label based on its local structure. For example, if the local structure exhibits different properties from those observed, the causal effect uncertainty will be high, suggesting that the predicted class may be incorrect.  

\Cref{fig:app_cgi} illustrates this approach: Using the causal graph (a), we can compare the original (b) and post intervention prediction, which estimates what the prediction would be if the target node has no neighbors (c). By computing the difference between the two, we assess the causal effect of the neighborhood on the label prediction. Depending on this causal effect and a decision threshold, we include the node or discard it from the training data. 

\textbf{Relating Graph Neural Networks to Structural Causal Models}
Graph neural networks are a powerful tool for learning with relational data. \citet{scm_gnn} provide a theoretical analysis to establish several connections between graph neural networks and SCMs. They design a new neural-causal model class by formalizing interventions for graph neural networks. The authors also provide theoretical results and proofs on the feasibility, expressivity, and identifiability of this new neural-causal model class for causal inference. 

\chapter{Causal Benchmarks} \label{sec:benchmarks}
This section gives an overview of benchmarks designed for \causalml tasks and includes interventional or counterfactual ground-truth data. We discuss the current limitations of benchmarks more critically in \Cref{bad:benchmarks}.

We believe it is worth mentioning that for some tasks, one can (and should) evaluate \causalml methods on \say{conventional} benchmarks developed without causality-specific design choices. For example, in \Cref{chapter:cil}, we discussed methods that can be used for out-of-distribution (OOD) tasks. One may evaluate these approaches on standard OOD benchmarks, such as some causal invariance learning methods (\Cref{prediction:causal_transportability}).

Similarly, conventional RL benchmarks rely on a simulation engine to generate trajectory data. Famous examples within the RL community are \emph{MuJoCo} \cite{todorov2012mujoco}, a physics engine offering a suitable playground for continuous control tasks, or the Arcade Learning Environment, allowing agents to play Atari 2600 games \cite{bellemare13arcade}. One can use these existing benchmarks for evaluating CausalRL methods simply by interpreting (PO)MDPs as SCMs (\Cref{cp:pomdps}).

Nevertheless, not all conventional ML benchmarks are suitable for evaluating the techniques discussed in this work. For example, to probe visual reasoning models on causal comprehension (e.g., \say{What effect will pushing this object have?}), we require question and answer pairs that encompass causal relationships.

\section{Reinforcement Learning} \label{benchmarks:rl}

Several authors designed RL simulators with high-level causal variables to facilitate the symbiosis of Causality and RL. While conventional benchmarks, such as MuJoCo, can be tweaked to allow an agent to intervene on environment variables (such as masses or lengths of the physical system \cite{PAML}), CausalRL benchmarks such as \emph{CausalWorld} \cite{causal_world} offer a well-defined API and underlying causal graph to simplify and extend environment interventions. 

\begin{figure}
    \centering
    \begin{subfigure}[t]{0.44\columnwidth}{
    \includegraphics[width=\columnwidth]{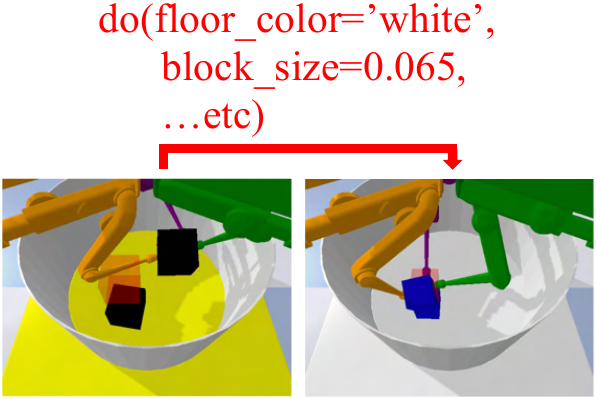}
    \caption{\textbf{CausalWorld \cite{causal_world}}: A robotic manipulation benchmark for causal structure and transfer learning. Tasks consist of constructing 3D shapes from a given set of blacks, inspired by how children learn to build complex structures.}
    }
    \end{subfigure} \begin{subfigure}[t]{0.48\columnwidth}{
    \includegraphics[width=\columnwidth]{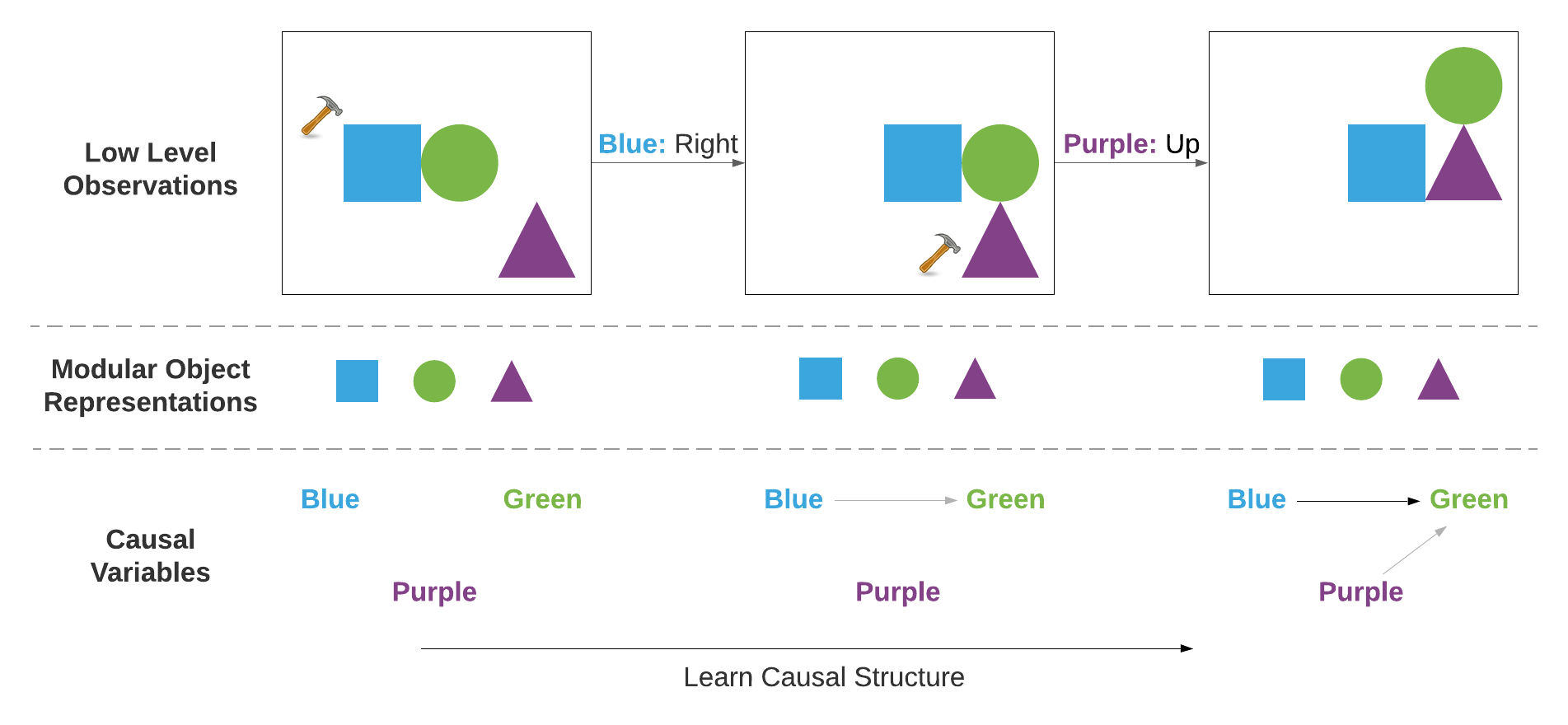}
    \caption{\textbf{Visual Causal Discovery \cite{visual_causal_discovery_rl_benchmark}}: A benchmark for learning representations of high-level causal variables from raw pixel data. Environments have objects that interact according to the underlying causal graph.}
    }
    \end{subfigure} \caption{\textbf{Causal RL benchmarks.}}
\end{figure}

\textbf{CausalWorld.}
\emph{CausalWorld} \cite{causal_world} is a robotic manipulation simulator that provides a combinatorial family of tasks with common causal structures and underlying factors (e.g., robot and object masses, colors, and sizes). The user or the agent can intervene on a subset of causal variables that determine the environment dynamics, allowing them to control how similar tasks are. 

\textbf{Visual Causal Discovery.}
\citet{visual_causal_discovery_rl_benchmark} note that RL agents often only observe low-level variables like pixels in images, and have to induce high-level causal variables (\Cref{sec:correct_abstraction}). To evaluate the ability of model-based RL methods to identify these causal variables and structures, they design a suite of physics and chemistry environments whose underlying causal graphs are parameterizable. 

\textbf{Alchemy.}
\begin{figure}
    \centering
    \includegraphics[width=\columnwidth]{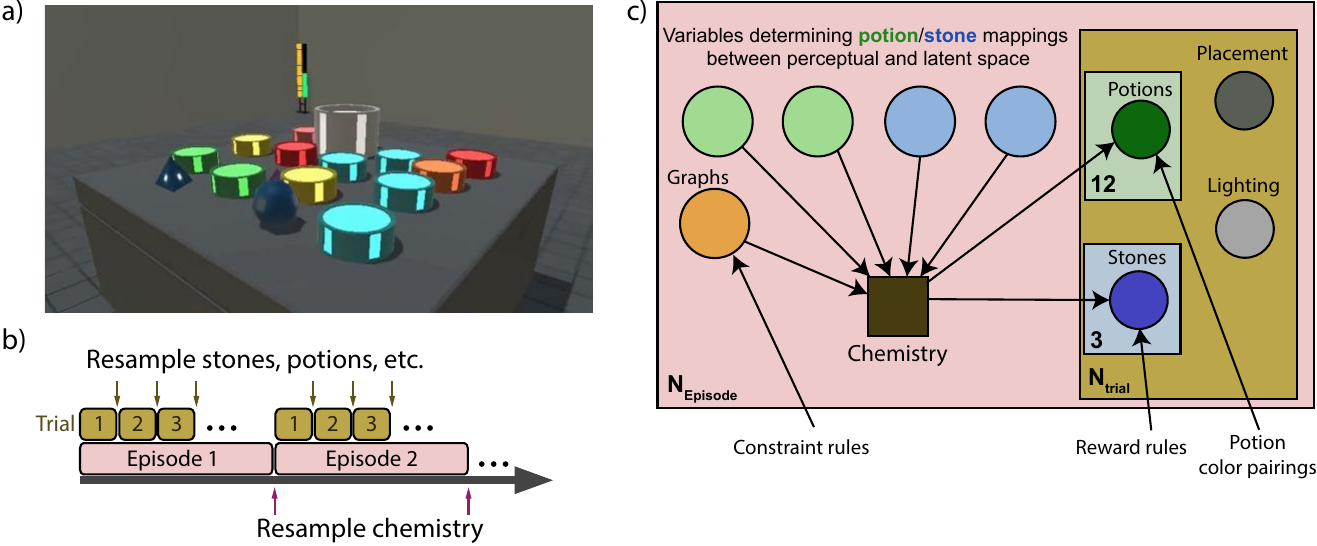}
    \caption{\textbf{Alchemy \cite{wang2021alchemy}}: A 3d video game implemented with the Unity game engine, revealing an accessible ground-truth task parameterization. It involves a causal structure (\emph{c}) that is periodically resampled (\emph{b}), offering a testbed for RL, structure learning, online inference, and hypothesis testing. It provides two observation spaces; one is symbolic and the other is image-based.}
    \label{fig:benchmark_alchemy}
\end{figure}
\citet{wang2021alchemy} propose \emph{Alchemy}, a 3D video game with a latent causal structure that is re-sampled procedurally in each episode. This environment provides a task distribution whose parameterization is accessible to the researcher yet, yields challenging tasks to solve, as demonstrated through experiments in which non-trivial deep RL methods fail. Based on probing experiments and analyses, they conclude that agents have to identify relevant parts of the latent structure to solve the tasks.

\textbf{CausalCity.}
\begin{figure}
    \centering
    \includegraphics[width=\columnwidth]{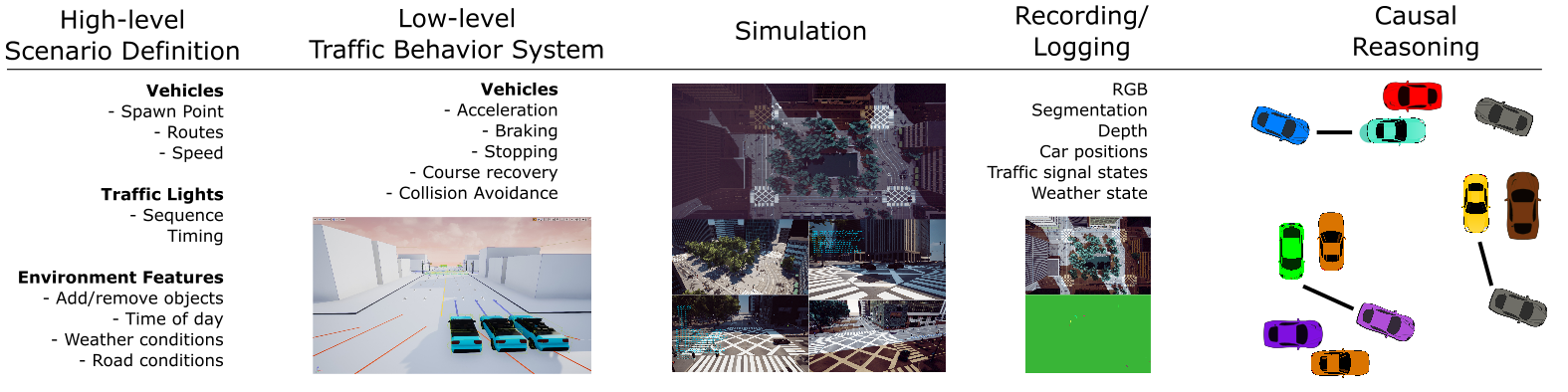}
    \caption{\textbf{\method{CausalCity}} \cite{mcduff2022causalcity}: A multi-agent driving simulator for complex temporal causal events, such as driving and vehicle navigation. It introduces \emph{agency} abstractions, such that one can simulate complex scenarios by only defining high-level variables.}
    \label{fig:benchmark_causalcity}
\end{figure}\citet{mcduff2022causalcity} develop \method{CausalCity}, a high-fidelity simulator for causal reasoning in the safety-critical context of driving. The goal is to navigate vehicles that have \say{agency}, high-level configurations controlling their sequence of actions (e.g., turn left at the next intersection), deciding their low-level behaviors (e.g., their speed). The environment is designed to simulate scenarios with complex causal relationships, including different types of confounders (e.g., weather conditions)

\section{Computer Vision}
\label{sec:cv_benchmarks}

\begin{figure}[t]
\centering
\includegraphics[width=\textwidth]{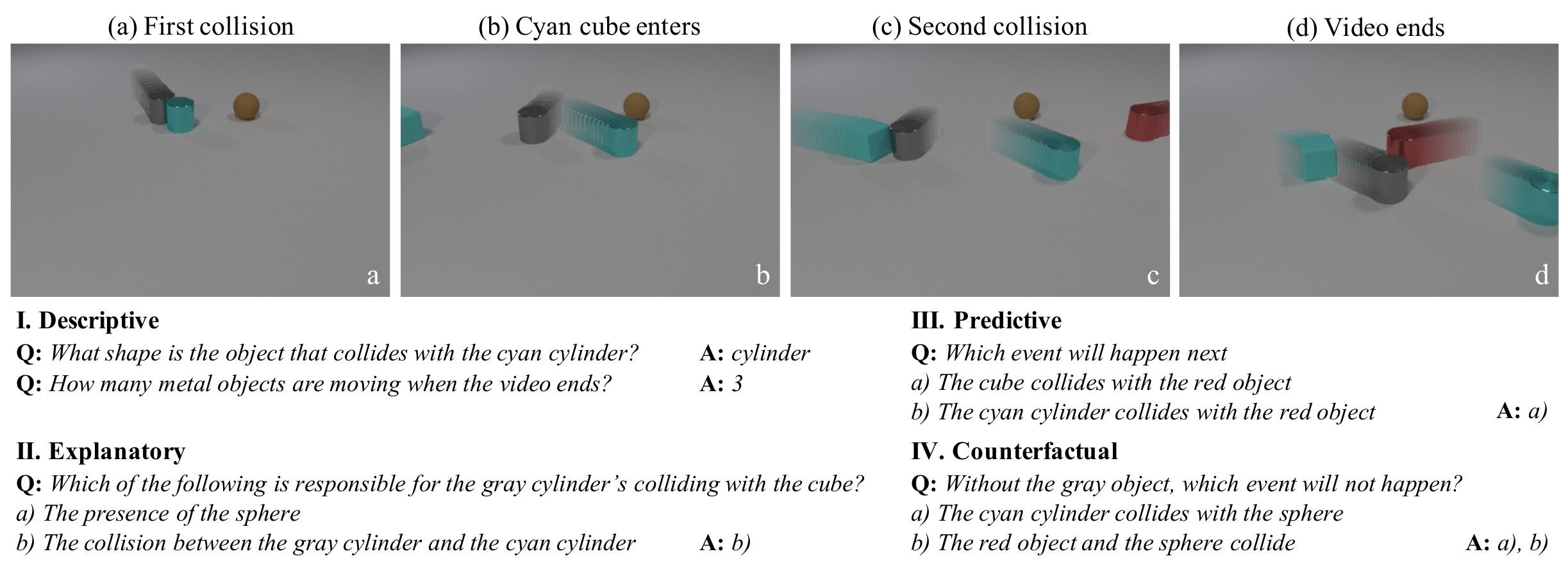}
\caption{\textbf{Sample video, questions, and answers from the CLEVRER dataset \cite{CLEVRER}}, designed to evaluate whether visual reasoning models understand the following classes of questions: (i) \textbf{descriptive}, (ii) \textbf{explanatory}, (iii) \textbf{predictive}, and (iv) \textbf{counterfactual}. All tasks, except (i), are considered causal. We include captions for the readers to understand the frames better, but they are not part of the dataset.} 
\label{fig:CLEVRER}
\end{figure}

\textbf{Clevrer.} \citet{CLEVRER} introduce the \emph{CoLlision Events for Video REpresentation and Reasoning} (CLEVRER) dataset. 
This video dataset allows us to evaluate models on four reasoning tasks: (i) descriptive, (ii) explanatory, (iii) predictive, and (iv) counterfactual. The authors interpret (ii)-(iv) as causal tasks and show that various state-of-the-art models for visual reasoning perform poorly on these. They conclude that future methods must learn the underlying causal relations between the objects seen in the images.

\begin{figure}
    \centering
    \includegraphics[width=\columnwidth]{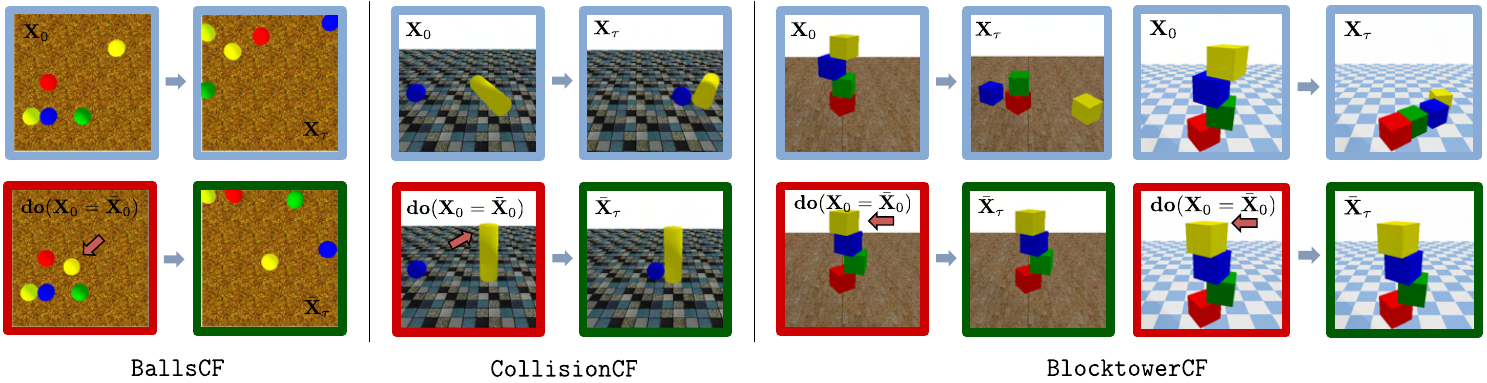}
    \caption{\textbf{Counterfactual Physics benchmark (\method{CoPhy}) \cite{Baradel2020CoPhy}}. Given an observed frame $\rmX_0$ and a sequence of future frames $\rmX_{1:\tau}$, it generates ground-truth counterfactual trajectories corresponding to if we had intervened upon $\rmX_0$ and set it to $\rmX_0$ (e.g., changing the initial positions of objects in the scene).}
    \label{fig:cophy_benchmark}
\end{figure}

\textbf{CoPhy.} \label{benchmark:cophy} \citet{Baradel2020CoPhy} develop the \method{CoPhy} benchmark, a synthetic 3D environment for causal physical reasoning. It consists of a number of physical dynamics scenarios, such as \emph{tower of blocks falling}, \emph{balls bouncing against walls} or \emph{objects colliding}, as illustrated in \Cref{fig:cophy_benchmark}. It allows us to intervene on the initial conditions of a trajectory and simulate its ground-truth counterfactual.

\textbf{Filtered-CoPhy.} \citet{janny2022filtered} propose a benchmark to test for counterfactual trajectory prediction, where given an observed initial condition $\vx_0$ and a sequence of $T$ frames $\vx_{1:T}$, one aims to predict a counterfactual sequence $\vx^{\text{CF}}_{1:T}$ under a new initial condition $\vx^{\text{CF}}_0$. There are three types of video data available to test: 
\emph{BlocktowerCF}, \emph{BallsCF} and \emph{CollisionCF}, and confounders that affect the prediction are designed to be obtainable with a sufficiently powerful representation.

\begin{figure}[t!]
    \centering
    \hfill
    \begin{subfigure}[t]{0.28\textwidth}
        \centering
        \includegraphics[width=\textwidth]{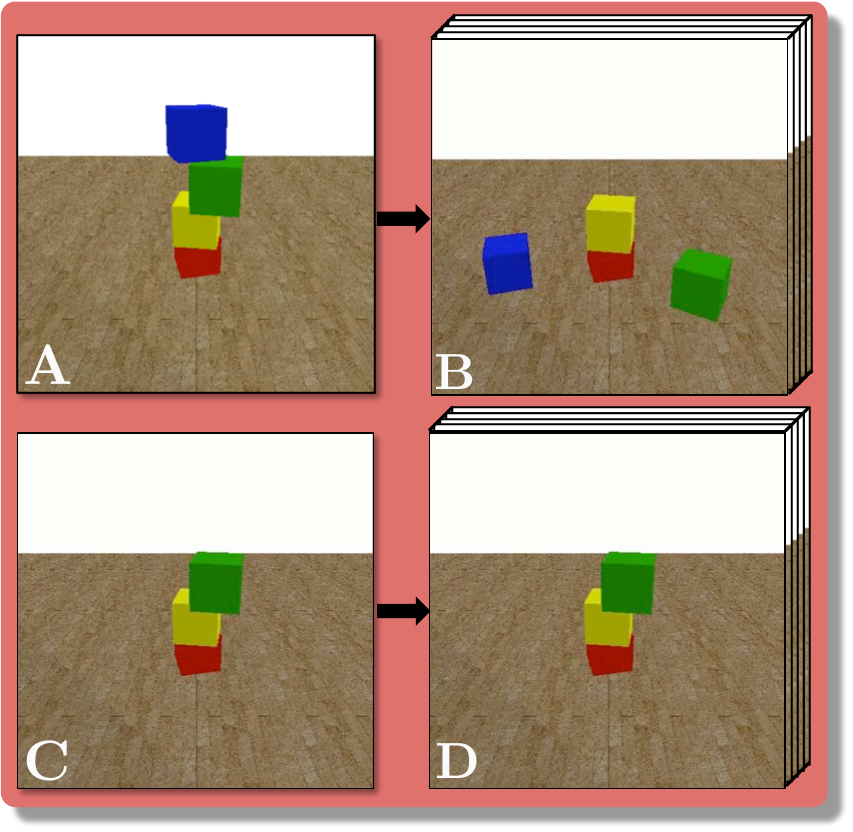}
        \caption{\texttt{BlocktowerCF} (BT-CF)}
    \end{subfigure}
    \hfill
    \begin{subfigure}[t]{0.28\textwidth}
        \centering
        \includegraphics[width=\textwidth]{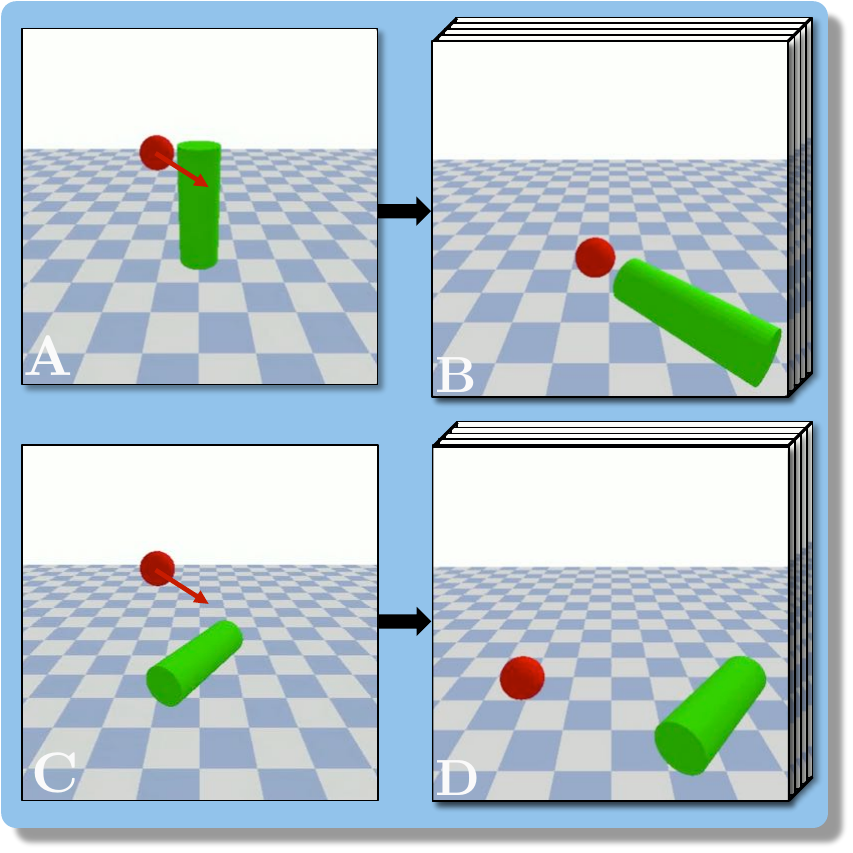}
        \caption{\texttt{CollisionCF} (C-CF)}
    \end{subfigure}
    \hfill
    \begin{subfigure}[t]{0.28\textwidth}
        \centering
        \includegraphics[width=\textwidth]{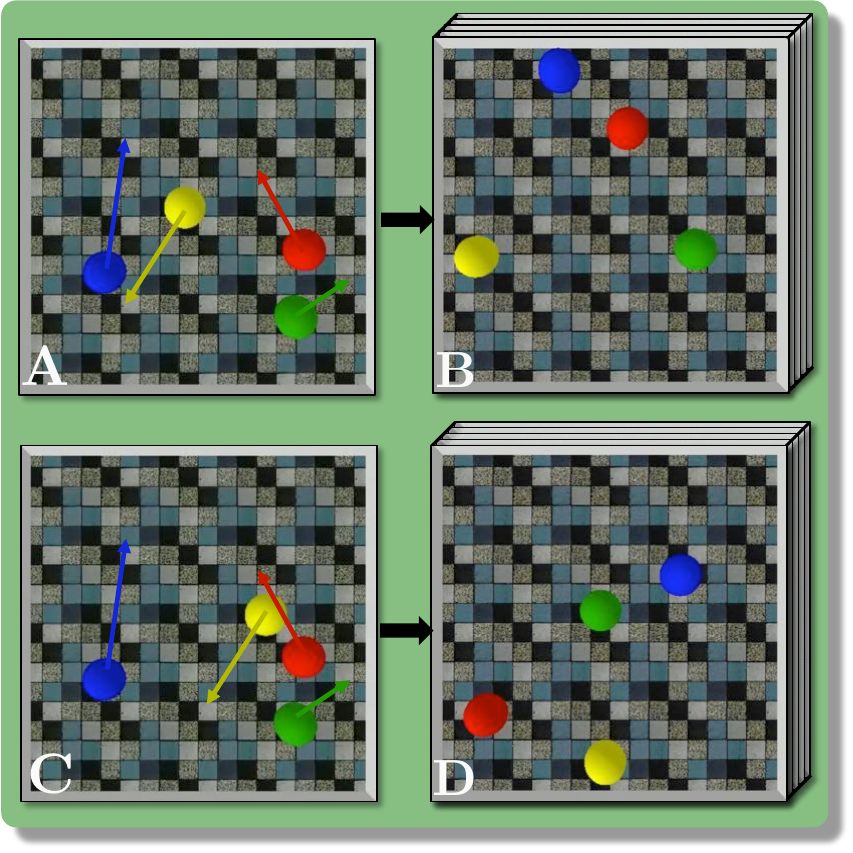}
        \caption{\texttt{BallsCF} (B-CF)}
    \end{subfigure}
    \hfill
    \caption{\textbf{Filtered-CoPhy benchmark suite \cite{janny2022filtered}}. It contains three challenging scenarios involving 2D or 3D rigid body dynamics with complex interactions, including collision and resting contact. Similar to \method{CoPhy}, we can intervene on the initial conditions $\rmX_0$ and set them to $\hat \rmX$. Arrows indicate the initial motion.}
    \label{fig:cophy}
\end{figure}

\textbf{Causal3DIdent.} \citet{ssl_data_aug_scholkopf} introduce an image dataset of 3D objects with causal dependencies, that can be used to study the effectiveness of data augmentations for style and content decomposition techniques (\Cref{sec:ifl}), which aim at isolating invariant content and discarding varying style.

\textbf{Spawrious} \citet{lynch2023spawrious} create an image classification benchmark suite
that includes spurious correlations between classes and backgrounds. They show that current robustness methods are unable to achieve over 70\%
accuracy on the hardest split using a ResNet50 pretrained on ImageNet.

\section{Natural Language Processing}

Constructing causal benchmarks involving text is challenging as the vocabulary size of a text corpus is usually large (e.g., the vocabulary of BERT \cite{DBLP:conf/naacl/DevlinCLT19} has around 30K tokens) and ground-truth data for multiple interventions is rare in natural language processing \cite{DBLP:journals/corr/abs-2109-00725}.

\textbf{Counterfactually-Augmented Data.} 
\citet{Kaushik2020Learning} provide two counterfactual NLP datasets for sentiment analysis and natural language inference, respectively. Given initial documents and labels, the authors recruited humans to revise the documents to conform to a counterfactual target label while ensuring internal consistency and avoiding gratuitous changes to facts unrelated to the target label.

\textbf{CausaLM.}
\citet{causaLM} propose four NLP datasets designed for causal explanations, three of which include ground-truth counterfactual examples for a given concept. These datasets allow researchers to evaluate counterfactual representations that dropped a given concept of interest, as further explained in \Cref{app:causalm}.

\textbf{CRASS.}
\citet{frohberg2021crass} introduce \method{CRASS}, a counterfactual reasoning assessment benchmark for language models that is part of the BIG-bench suite \cite{big_bench}. For example, they showcase the query \quo{A woman sees a fire. What would have happened if the woman had fed the fire?} with three possible answers (a) \quo{The fire would have become larger}, (b) \quo{The fire would have become smaller}, and (c) \quo{That is not possible}.
\quo{A women sees a fire} is the base premise and \quo{What would have happened if the woman had fed the fire?} is a questionized counterfactual conditional (QCC). The possible answers, i.e., a correct consequence and a set of potential effects as distractors, define a so-called \emph{premise-counterfactual tuple} (PCT). More formally, A QCC has the form \quo{What would have happened if $\rmA^{\text{CF}}$?}, where $\rmA^{\text{CF}}$ is some altered version of the base premise $\rmA^{\text{O}}$, effectively generating the counterfactual. 

\textbf{Fine-grained Causal Reasoning.} 
\citet{yang2022towards} propose a fine-grained causal reasoning dataset, including the three tasks of \emph{causality detection}, \emph{fine-grained causality extraction} and \emph{Causal QA} tasks. To motivate the need for finer granularity, they give the following example: 
\say{the spread of COVID-19 has \emph{led} to the boom in online shopping, [cause], but it also has \emph{deterred} [prevent] people from going shopping centres}. The authors argue that previous datasets only considered the above-annotated cause relation, but not more fine-grained causal events like \emph{enable} or \emph{prevent}. 

Now consider a different passage \say{COVID-19 has accelerated change in online shopping, and given Amazon's ... it will result in economic returns for years to come and offering more competitive prices compared to an offline business that brings pressures for the offline business recruitment.}. Previous datasets allow models to extract facts such as \say{COVID-19 causes an increase in online shopping}, yet, they cannot detect the subsequence for Amazon to \say{offer more competitive prices}, and the negative influence on offline business recruitment. Both of these can be useful, e.g., if we ask the what-if question \say{What if COVID stops?}, whose correct answer in the above context should include \say{there will be more offline business recruitments.} Their experiments reveal a significant gap between model and human ceiling performance (74.1\% vs. 90.53\% accuracy), providing evidence that statistical models still struggle to solve causal reasoning problems.  
\chapter{The Good, the Bad and the Ugly} \label{sec:gbu}
In this section, we want to provide our perspective on what benefits \causalml may buy us (the \emph{good}), what challenges the field is currently struggling with (the \emph{bad}), and what price one has to pay for \causalml (the \emph{ugly}). In other words, this section provides an informative discussion of the advantages and disadvantages of \emph{current} \causalml methods. 

\section{The Good}

We discussed many methods making explicit use of the various causal formalisms, such as SCMs (\Cref{sec:scm}), interventions (\Cref{concept:intervention}), or counterfactuals (\Cref{concept:counterfactuals}). 
Causality presents a framework (SCMs) to formally express assumptions about the data-generating process that we would like to encode into our models and a mathematical tool (the do-operator) that can enforce properties thereof during the model training.

In the following, we briefly summarize some of the benefits causal formalisms can bring us across the discussed problem fields. Many of these benefits cannot be recovered with purely statistical reasoning.

With Causal Supervised Learning (\Cref{chapter:cil}), we can improve predictive models: Invariant feature learning approaches (\Cref{sec:ifl}) attempt to identify a set of content variables $\rmC$ that represented the causal parents of $Y$, and learn a predictor $p(y \mid \vc)$ that is invariant to interventions on the style variables $\rmS$. 
In contrast, invariant mechanism learning approaches (\Cref{sec:iml}) model the change of distribution caused by interventions on a set of independent unobserved confounders $\rmU$, and they learn separate mechanisms to model for each. To summarize, these methods build the foundation to learn domain-robust, reusable features or mechanisms \cite{scholkopf2021towards}.

Causal Generative Modeling (\Cref{chapter:cgm}) offers a principled framework for the task of controllable generation (i.e., generation with control over certain attributes). The structural assignment learning (\Cref{sec:requires_causal_dag}) approaches allow practitioners to add domain knowledge of an underlying causal graph for the DGP. These generate counterfactual samples that consider causal dependencies between the attributes of interest and the other generative variables. The causal disentanglement (\Cref{sec:does_not_require_causal_dag}) approaches go one step further and perform causal graph discovery as well as structural assignment learning, which allows for controllable generation by exploiting weaker forms of available causal domain knowledge.

Through Causal Explanations (\Cref{chapter:explanations}), we gain interpretability of model predictions. Without access to the underlying causal dependencies, we can use feature attribution methods (\Cref{explanations:fa}) to identify which variables under intervention are most relevant to changes in the model output. With access to the underlying causal graph, we can generate contrastive explanations (\Cref{explanation:contrastive_explanations}), which suggest counterfactual model outputs dependent on actionable alternative inputs. 

By using Causal Fairness (\Cref{chapter:fairness}) criteria, we obtain counterfactual, or interventional, quantities that allow us to evaluate the fairness of prediction models dependent on sensitive attributes of interest. With access to underlying causal dependencies in the DGP, we can enforce case-specific fairness criteria that deconfound potential sources of selection bias.

In \emph{Causal Reinforcement Learning} (\Cref{chapter:rl}), we reviewed publications that use interventions for formalizing actions (\Cref{rl:causal_bandits}), changes in the environment (\Cref{rl:multi_envs}), deconfounding of observed trajectory data (\Cref{rl:il,rl:oppe}), the effects of actions on the reward (\Cref{rl:ca}), changing structure in the state space, exposing opportunities to generate counterfactual data augmentations, allowing us to recycle already observed data. 

\section{The Bad}

\subsection{Lack of Open-Source Ecosystem}
\label{bad:software}
Most open-source machine learning software packages or model hubs focus on observational models. Automatic differentiation frameworks like PyTorch \cite{paszke2017automatic}, \method{Tensorflow} \cite{abadi2016tensorflow} and \method{JAX} \cite{jax2018github} (including \method{Flax} \cite{flax2020github} and \method{Haiku} \cite{haiku2020github}) paired with model libraries like \method{Transformers} \cite{wolf-etal-2020-transformers}, \method{timm} \cite{rw2019timm}, or \method{PyG} \cite{pyg} facilitate rapid prototyping of model pipelines. Importing datasets, state-of-the-art (pre-trained) models, and launching a training loop can be done in a few lines of code. Model hubs like \href{https://www.tensorflow.org/hub}{Tensorflow Hub}, \href{https://huggingface.co/models}{Huggingface Models} or \href{https://pytorch.org/hub/}{Pytorch Hub} provide thousands of pre-trained models across multiple modalities ready for fine-tuning and deployment.

Unfortunately, there is a much smaller open-source ecosystem for learning or performing inference with SCMs. At the time of this writing, we are unaware of software libraries providing APIs for convenient identification tests, manipulating SCMs, or importing causal benchmarks (\Cref{sec:benchmarks}) as well as pre-trained SCMs to facilitate progress on novel research ideas. 


We believe that the ingredients to either (a) build novel \causalml-focused platforms or (b) incorporate \causalml techniques and models into existing ones are ready. For example, causality researchers often prove causal estimand identification formally by hand, although, e.g., \citet{xia2021causal} develop an algorithm that verifies identifiability automatically for differentiable SCMs. Providing convenient APIs for such algorithms would be immensely helpful.

For estimating causal estimands using machine learning techniques, there exist some libraries. For example, \citet{scutari2009learning} and \citet{kalisch2012causal} propose R packages for causal discovery; \citet{sharma2020dowhy,chen2020causalml,bach2021doubleml} propose Python packages for treatment effect estimation (see more details on these in \Cref{sec:ml_for_causality}). \citet{james_fox-proc-scipy-2021} introduce a Python library for (multi-agent) causal influence diagrams, which are commonly used for decision-making problems under uncertainty.

\subsection{Lack of Comparisons to Non-Causal Methods}
Several \causalml papers lack experimental comparisons to non-causal approaches that solve similar, if not identical, problems. While the methodology may differ, e.g., depending on whether causal estimands are involved, some of these methods claim to improve performance on non-causal metrics, such as accuracy in prediction problems or sample efficiency in RL setups. This trend of not comparing against non-causal methods evaluated on the same metrics harms the measure of progress and practitioners who have to choose between a growing number of methods. 

One area in which we have identified indications of this issue is invariance learning (\Cref{sec:ifl}). Some of these methods are motivated by improving a model's generalization to out-of-distribution OOD data; however, they do not compare their method against typical domain generalization methods, e.g., as discussed in \citet{lost_DG}. For example, \citet{mouli2022asymmetry} tackle OOD tasks by learning counterfactually-invariant representations with asymmetry learning (discussed in \Cref{sec:asymmetry_learning}); yet, in their experiments, they do not compare against any non-causal OOD method (see, e.g., \citet{lost_DG,wang2021generalizing}).

Another area is the causal model-based RL literature (\Cref{rl:mbrl}). \citet{causal_curiosity} (\Cref{sec:cc})  learn disentangled latent task embeddings, arguing these are more interpretable than previous (non-causal) latent task embedding methods, which define their setup using the formalisms of Hi-Param MDPs \cite{hi_params} and BAMDPs \cite{bayes_adapt, zintgraf2019varibad}. In the experiments, they compare their agent in terms of sample efficiency against baselines not having access to any task embeddings, excluding Hi-Param MDP and BAMPD approaches (e.g., \cite{PEARL,zintgraf2019varibad}). \citet{zhang2020invariant}, presented in \Cref{sec:block_mdps}, argue that their method makes stronger assumptions on how tasks relate to each other compared to common multi-task RL methods. However, their experiments do not compare against any multi-task RL method (e.g., \cite{PEARL, maml, gupta2018meta}). \citet{mutti2022provably} aim at learning an agent that can systematically generalize across a universe, i.e., an infinite set of possible MDP tasks (\Cref{sec:sg}). Their experiments evaluate how well their method approximates the optimal value function but do not compare against non-causal models.

\section{The Ugly}

\subsection{Difficulties of Obtaining Ground-Truth Evaluation Data}
\label{bad:benchmarks}
One of the biggest open problems in \causalml is the lack of public benchmark resources to train and evaluate causal models. \citet{cheng2022evaluation} find that the reason for this lack of benchmarks is the difficulty of observing interventions in the real world because the necessary experimental conditions in the form of randomized control trials (RCTs) are often expensive, unethical, or time-consuming. In other words, collecting interventional data involves actively interacting with an environment (i.e., \emph{actions}), which, outside of simulators, is much harder \footnote{Real-world experiments in forms of randomized control trials are often expensive, time-consuming, or even impossible due to ethical concerns.} than, e.g., crawling text from the internet and creating passively-observed datasets (i.e., \emph{perception}). Evaluating estimated counterfactuals is even worse: by definition, we cannot observe them, rendering the availability of ground-truth real-world counterfactuals impossible \cite{rubin1974estimating}. 

The \emph{pessimistic} view is that yielding \say{enough} ground-truth data for \causalml to get deployed in real-world industrial practice is unlikely soon. Specifying how much data is \say{enough} is task-dependent; however, in other fields that require active interactions with real-world environments, too (e.g., RL), progress has been much slower than in fields thriving on passively-collected data, such as NLP. For example, in robotics, some of the best-funded ML research labs shut down their robotics initiatives due to \say{not enough training data} \cite{openai_shutting_down_robotics}, focusing more on generative image and language models trained on crawled internet data. 

Moreover, a curse of the need for simulated data is its manipulability and the consequent lack of consistency across papers. Authors can easily create novel or modify existing simulations tailor-made for an empirical verification for their particular setup/method, which may not generalize to other setups. Such inconsistencies hinder progress because it is harder to compare methods. Real-world fixed datasets such as \method{ImageNet} \cite{imagenet} are less prone to such problems since papers altering them raise more suspicion.

The \emph{optimistic} view is that the lack of benchmarks is simply due to the field being in its infancy, and both more available RCT datasets and simulators will alleviate the progress of \causalml methods. About a decade ago, similar problems existed in the field of (deep) RL, which shares a few parallels with \causalml, as debated in \Cref{chapter:rl}. In the meantime, RL simulators paved the way to beating world champions in board games \cite{alphago, alphago_2}, achieving super-human performance in Atari \cite{atari} and grandmaster level in StarCraft II \cite{vinyals2019grandmaster}, treating sepsis in intensive care \cite{komorowski2018artificial}, or navigating stratospheric balloons \cite{bellemare2020autonomous}.

\subsection{Untestable Assumptions Are Inevitable}
By making assumptions about the data-generating process in our SCM, we can reason about interventions and counterfactuals. However, making such assumptions can also result in bias amplification \cite{bias_amplification_1} and harming external validity \cite{l2010research} compared to purely statistical models. Using an analogy of Ockham's Razor \cite{ockham_razor}, one may argue that more assumptions easily lead to wrong models.

For example, \citet{bias_amplification_1} illustrates bias amplification in a setting of hidden confounding (\Cref{related_work:unobserved_confounders}). They show that while adjusting for covariates acting like instrumental variables (i.e., variables that are more strongly associated with the treatment assignment than with the outcome), one may reduce confounding bias, but at the same time, residual bias carried by unmeasured confounders can build up at a faster rate. By making the causal model more complex by adding more covariates that should aid backdoor adjustment, the model residual bias of the causal effect increases in harmful ways. A \say{simpler} model that excludes covariates that are predictive of the treatments can work better \footnote{Chapter 18.5 of \cite{hernan2010causal} discusses such issues in more detail.}.

\chapter{Related Work}
\section{Other Surveys}
\citet{schoelkopf2019causality} discusses links between machine learning and graphical causal inference while introducing causality concepts along the way. \citet{scholkopf2021towards} reviews fundamental concepts of causal inference and relates them to open machine learning problems with a particular focus on representation learning. They highlight two issues in current deep learning systems: robustness to distribution shifts and learning reusable and modular mechanisms. 

\citet{feder2021causal} examines the intersection of NLP and causality and argues that causal formalisms can make NLP methods more robust and understandable. They list three problems from the NLP literature that motivate this claim: purely associational models may (i) latch onto spurious associations, failing to generalize in OOD settings (e.g., \cite{mccoy-etal-2019-right}); (ii) exhibit unacceptable performance differences across groups of users (e.g., \cite{zhao-etal-2017-men}); (iii) be too inscrutable to incorporate into high-stakes decisions (e.g., \cite{inscrutable}). 

\citet{survey_social_responsible_AI} discuss how causality may address ethical challenges in socially responsible artificial intelligence. They focus on seven causal inference tools, some of which we have discussed in \Cref{sec:prelim} (e.g., the do-operator or counterfactual analysis), and some of which we have left out (e.g., mediation analysis or propensity scores). 

\citet{causal_reasoning_visual} highlight how causal reasoning can support visual representation learning. Specifically, the authors categorize existing causality-aware visual representation learning work into (i) causal visual comprehension, (ii) causal visual robustness, (iii) causal visual question answering, and (iv) causal datasets. For future work, they enumerate the following potential directions: (i) more reasonable causal relation modeling, (ii) more precise approximations of intervention distributions, (iii) more proper counterfactual synthesizing processes, and (iv) large-scale benchmarks and evaluation pipelines. 

\citet{causal_ml_healthcare} explore how causal inference can be incorporated into clinical decision support systems using modern machine learning techniques. As a running example throughout their paper, they use Alzheimer's disease (AD) to illustrate how \causalml can benefit clinical scenarios. The authors observe that important challenges in healthcare applications are processing high-dimensional and unstructured data, generalization to out-of-distribution samples, and temporal relationships. All these challenges may be addressed through \causalml, which the authors divide into (i) causal representation learning (\Cref{sec:crl}), (ii) causal discovery (\Cref{rw:cd}), and (iii) causal reasoning (estimating interventional distributions, \Cref{sec:cbn_intervention}).

\citet{causality_medical_images} discuss how causality can assist in creating robust and adaptable medical image analysis algorithms. Specifically, the authors point out that many healthcare machine learning approaches fail to translate into clinical practice due to the inability to adapt and be robust to real-world conditions. One cause of such existing models' inabilities is their lack of ability to distinguish between correlations and causation, which may result in deadly mistakes. For example, \cite{degrave2021ai} identified several approaches that claim to have been able to diagnose COVID-19 from chest X-rays but ultimately failed to do so as they instead relied on spurious features like hospital identifiers and the patient's ethnicity.

\section{Machine Learning for Causal Inference} \label{sec:ml_for_causality}
This survey uses causality theory to solve common machine learning problems and provide new perspectives on such. An astute reader may wonder about research in the other direction, using machine learning to answer causal questions. We observe numerous studies using modern representation learning techniques to estimate and answer causal queries, such as causal effects. For completeness, we briefly list recent advances in two common causal inference tasks: \emph{causal effect estimation} and \emph{causal discovery}.  

\subsection{Causal Effect Estimation}
Estimating causal effects from observational data is a fundamental problem in many fields that face challenges in running randomized control trials. We want to answer causal questions in many scientific or commercial setups; hence, it is fallible to argue just from observed associations. Supervised learning methods face two challenges in such settings: (i) missing interventions, i.e., the fact that we only observe one treatment for each individual means models must extrapolate to new treatments without access to ground truth, and (ii) confounding variables affecting both treatment assignment and the outcome, such that extrapolation from observation to intervention requires additional causal assumptions. The literature on \emph{treatment effect estimation} deals with constructing models that address these issues. 

\textbf{Observed Confounders} \label{rw:hte}
Generally speaking, we can identify causal effects if we observe all confounders. However, depending on the structure of the treatment effect (e.g., smoothness or sparsity \cite{kennedy2020optimal}), different estimators behave differently. For example,  \citet{DML} and \citet{x-learner} show that simple regression models trained on all training data points regardless of their treatment can easily lead to biased estimates due to imbalance in the treatment assignment; \citet{shalit2017estimating} and \citet{dragonnet} make similar arguments for neural network models. Hence, the bulk of treatment effect estimation works focuses on model regularization to identify the causal associations in the data.

One line of work for causal effect estimation that utilizes modern ML techniques are \emph{meta-learners} (or \emph{plug-in learners}) \cite{x-learner,caron2020estimating}: they decompose effect estimation into multiple sub-problems (so-called \emph{nuisance components}), each solvable using any modern machine learning technique \cite{DML, x-learner, r-learner, SIN}. 

We highlight a few of these techniques that utilize neural networks. For binary treatments, \citet{curth2021nonparametric} implement multiple meta-learning strategies with neural networks, concluding that theoretically-optimal estimators may not perform best in finite-simple regimes. For scalar-continuous treatments, \citet{nie2021vcnet} propose using splines to preserve continuity over the treatment domain. For higher-dimensional treatments, \citet{SIN} present a meta-learning strategy that learns propensity features. For arbitrary treatments, \citet{transtee} utilize Transformer networks \cite{attention_is_all_you_need} to construct a flexible architecture.

For causal estimand identification, \citet{malek2021asymptotically} use the bandit framework to learn the arm that will produce the best estimator for identifying a causal estimand of interest. They assume an online setting in which the practitioner controls the data collection procedure and aims to find the estimand formula with the lowest asymptotic variance in as few samples as possible. \citet{xia2021causal} develop a necessary and sufficient algorithm that verifies identifiability automatically for differentiable SCMs. Their approach relies on enforcing $L_1$-consistency with the data distribution.

\textbf{Unobserved Confounders} \label{related_work:unobserved_confounders}
There are primarily three strategies to deal with unobserved confounding (\Cref{sec:confounding}): (1) estimating bounds based on additional assumptions, also referred to as \emph{partial identification} \cite{manski1990nonparametric,manski2003partial}, (2) sensitivity analysis of how strong the confounder's effect has to be to make the true estimand substantially different from our estimate \cite{imai2010identification, cinelli2019sensitivity}, and (3) utilizing other observed variables, such as \emph{instrumental} \cite{baiocchi2014instrumental} or \emph{proxy} variables \cite{miao2018identifying}. Specifically for the third category, some recent work explores ML techniques, such as NNs or kernel methods, discussed below.

\textbf{Proxy Variables.}
\emph{Proxy variables} contain relevant auxiliary information on the confounder; ideally, enough to completely recover the confounder \cite{CEVAE, miao2018identifying}. For example, we may be interested in estimating the impact of flight ticket prices on sales \cite{xu2021deep}. A possible confounder is the people's desire to fly, e.g., driven by holiday seasons that affect both the number of ticket sales and the prices customers are willing to pay. A suitable proxy variable for this desire could be the number of views of the ticket reservation page. While simply adjusting on such a proxy variable alone would produce a biased effect estimate, there are recent methods aiming at producing less biased estimates \cite{complex_causes, mastouri2021proximal, minimax_proximal,proximal_survival,proximal_mediator}.

\textbf{Instrumental Variable Regression.}\label{rw:ivr}
Another class of methods rely on an \emph{instrumental variable} $I$ (often just called \emph{instrument}) \cite{iv_1, kiv, iv_2, iv_3, iv_4}. Three conditions have to hold for an instrument to be  \emph{valid}: (i) it is independent of the hidden confounder, $I \Perp C$; (ii) it is not independent of the treatment, $I \not \Perp T$; and (iii) it is independent of the outcome conditional on the treatment and confounder $I \Perp Y \mid \{T, C\}$. 

In the above flight ticket example, we might consider shifting supply factors, such as the oil price, as a valid instrument, as it only affects sales via price, thereby identifying the customers' demand \cite{kiv}.

Sometimes, we observe treatment corrupted by measurement error, e.g., when we do not observe it directly. For instance, in the setting of medical treatments, suppose patients are asked to take a drug at home instead of a hospital with supervision. Some of them may not comply, and a self-reported measurement is erroneous due to the patient lying or being forgetful. \citet{zhu2022causal} propose a method to deal with such measurement error scenarios assuming access to an instrument.

\subsection{Causal Discovery} \label{rw:cd}

\emph{Causal Discovery} is an umbrella term for methods that try to recover the underlying causal structure of a DGP from observational and/or interventional data. Depending on the assumptions one is willing to make, the output ranges from partial node orderings to the full SCM (assuming linear structural equations).

Generally, causal discovery is difficult due to (i) structure identifiability and (ii) computational complexity. First, the identifiability challenge means that, given observational data only, the causal DAG $\gG$ is typically not identifiable, as a set of possible graphs could have generated the data \cite{pearl2009causality}. Secondly, due to the combinatorial nature of the solution space, its size grows super-exponentially with the number of variables, rendering naive methods computationally infeasible \cite{Chickering1996}.

We point keen readers looking for more details on these methods to primarily three classes: (i) combinatorial methods \cite{singh2005finding,NIPS2013_8ce6790c,Cussens11,cd_review,NIPS2015_2b38c2df,aragam2015concave,ramseyGSG17}, (ii) continuous relaxations \cite{notears,yuCGY19,zheng2020learning,unknown_interventions,ngG020,brouillardLLLD20,he0SXLJ21,enco}, and (iii) permutation-based methods \cite{friedman2003being,npvar,cundy2021bcd, zantedeschi2022dag}. Further, \citet{vowels2021d,cd_review} provide excellent surveys. 

Recovering the full causal graph is especially difficult in high dimensions. Hence, one line of work focused on more specialized problems, e.g., more coarse-grained \cite{ancestral_ci} ancestral structures, partial ancestral graphs \cite{mooij2020constraint}, or the causal order among a subgraph of variables known to descend from some set of confounding covariates \cite{watson2022causal}. Another line of work aims to perform causal discovery actively to improve data efficiency and minimize the number of required interventions \cite{eberhardt2010causal,hyttinen2013experiment,von2019optimal,learning_neural_causal_models_actively}.

\subsection{Granger Causality}
In many scientific fields, such as neuroscience, econometrics, and civil engineering, it is important to understand causal relationships found in time series data. In neuroscience, for example, researchers study whether activity in one brain region correlates with activity in another \cite{problems_with_GC}. An econometrician may be interested in which macroeconomic indicators predict one another \cite{lutkepohl2004applied}. Civil engineers are interested in understanding differences in traffic across highways to build well-utilized transportation infrastructure \cite{traffic_prediction}.

Granger causality \cite{granger1969investigating} is a framework for time series structure discovery that quantifies the extent to which the past of one time series aids in predicting the future evolution of another time series. It is based on the simple assumption that causes precede their effects. 
\begin{mydef}{Granger Causality \cite{granger1969investigating, EOCI}}{granger_causality} Whenever past observations of $X$ help in predicting the future of time-series $Y$, then $X \operatorname{Granger-causes} Y$. Formally, we write
\begin{align}
    X \operatorname{Granger-causes} Y : \Longleftrightarrow Y_{t} \not \perp X_{\text {past }(t)} \mid Y_{\text {past }(t)}.
\end{align}
\end{mydef}

If our goal is to detect the underlying causal DAG of multiple variables observed over time, we can use Granger causality only sometimes. Specifically, we can recover the DAG if we observe all relevant variables (implying no unobserved confounding, see \Cref{sec:identifiability}) and no \emph{instantaneous} effects exist, i.e., when no variable causes another at the same time step. This condition and other limitations are explained in more detail by \citet[Chapter~10]{EOCI}. 

There are several interesting extensions of Granger causality. Originally, Granger causality was defined using linear relations. \citet{tank2018neural} extended it to the non-linear case.

\citet{mpir} proposes to combine the benefits of Granger causality with deep learning models. They propose relational learning with \emph{minimum predictive information} (MPI) regularization. This MPI regularizer quantifies the directional predictive strength between each pair of time series. 

\citet{sru} utilize Statistical Recurrent Units (SRUs) to model the network topology of Granger causal relations. Specifically, they explain that the structured sparse estimate of the internal parameters of the SRU networks, trained to predict the stochastic processes' time series measurements, relate to the Granger causal relations. Moreover, by regularizing a low-parameterized variant of SRUs, called \emph{economy-SRU}, towards interpretability of its learned predictive features, this model is less likely to overfit the time series data.

\citet{lowe2020amortized} point out that most causal discovery methods based on Granger causality fit a new model whenever they encounter samples from a new underlying causal graph. However, time series with different underlying causal graphs may share relevant information, e.g., when inferring synaptic brain connections between neurons based on their spiking behavior: test subjects may have varying brain connectivity but the same underlying neurochemistry. Hence, they propose \emph{Amortized Causal Discovery}, an approach that separates the causal relation prediction from modeling the time series dynamics, permitting us to aggregate statistical strength across samples. 
\chapter{Conclusion}
We summarize some of our key findings:
\begin{enumerate}[leftmargin=*]
\item Causal Inference (\Cref{sec:prelim}) encodes causal assumptions about a system of interest. Therefore, as opposed to conventional statistical or probabilistic inference, it allows us to reason about interventional and counterfactual estimands. 
\item We argue that these estimands benefit specific areas of ML research, namely:
    \begin{enumerate}[leftmargin=*]
        \item \textbf{Causal Supervised Learning} (\Cref{chapter:cil}) improves predictive generalization by learning invariant features or mechanisms, both aiming at deconfounding models' reliance w.r.t. reliance on spurious associations. Future work should investigate targeted benchmarks testing for learned invariances, connections to adversarial robustness and meta-learning, and the potential utilization of additional supervision signals. 
        \item \textbf{Causal Generative Modeling}  (\Cref{chapter:cgm}) supports sampling from interventional or counterfactual distributions, naturally performing principled controllable generation or sample editing tasks, respectively. All existing methods learn structural assignments; some infer the causal structure from data. However, it is underexplored what levels of abstractions should be considered for different applications, how to scale assignment learning up to larger graphs, and when counterfactually-generated data augmentations are effective (and when not). 
        \item \textbf{Causal Explanations} (\Cref{chapter:explanations}) explain model predictions while accounting for the causal structure of either the model mechanics or the data-generating process. We divide these methods into (i) feature attributions, which quantify the causal impact of input features, and (ii) contrastive explanations, representing altered instances achieving the desired outcome. So far, it is unclear how to unify both classes of methodology best, scale up explanations, make them robust against distribution shifts, secure and private against attackers, and circumvent the inevitable trade-off of their robustness against recourse sensitivity. 
        \item \textbf{Causal Fairness} (\Cref{chapter:fairness}) paves the way for criteria that assess a model's fairness as well as mitigate harmful disparities w.r.t. causal relationships of the underlying data. However, the criteria rely on either counterfactual or interventional distributions. Future work should shed light on alternatives to equality, fairness outside standard prediction settings, weaker observability assumptions (e.g., hidden confounding), and the validity of interventionist views on social categories.  
        \item \textbf{Causal Reinforcement Learning} (\Cref{chapter:rl}) describes RL methods taking the explicit causal structure of the decision-making environment into account. We categorize these methods into eight categories and observe that their claimed benefits over non-causal methods include deconfounding (resulting in better generalization), intrinsic rewards, and data efficiency. Open problems suggest that some formalisms might be unifiable, deconfounding of offline data has been largely unaddressed in offline RL sections, and agents making decisions based on counterfactuals might offer further benefits. 
        \item \textbf{Modality-Applications}: We review how previously introduced, and modality-specific principles provide opportunities to improve computer vision, natural language processing, and graph representation learning settings.  
    \end{enumerate}
\item We review existing \textbf{Causal Benchmarks} designed for \causalml methods, including ground-truth causal interventions and counterfactuals. 
\item We discussed some of the benefits of using \causalml methodology, as well as issues around untestable assumptions, lack of benchmarks, lack of software, and lack of comparisons to non-causal methods.
\item We touched upon two categories of previous related work; other \causalml surveys and research that utilizes machine learning to infer causal estimands.
\end{enumerate}



\begin{acknowledgements}
In alphabetical order, we would like to thank Damien Teney, David Watson, Jakob Zeitler, Limor Gultchin, Matej Zečević, Phillip Lippe, Stefano Blumberg, and Tom Everitt for helpful feedback on earlier drafts of this paper. 
\end{acknowledgements}

\appendix

\backmatter  

\printbibliography

@article{gardner2015deep,
  title={Deep manifold traversal: Changing labels with convolutional features},
  author={Gardner, Jacob R and Upchurch, Paul and Kusner, Matt J and Li, Yixuan and Weinberger, Kilian Q and Bala, Kavita and Hopcroft, John E},
  journal={arXiv preprint arXiv:1511.06421},
  year={2015}
}

@inproceedings{xian2017zero,
  title={Zero-shot learning-the good, the bad and the ugly},
  author={Xian, Yongqin and Schiele, Bernt and Akata, Zeynep},
  booktitle={Proceedings of the IEEE conference on computer vision and pattern recognition},
  pages={4582--4591},
  year={2017}
}

@article{bricken2023monosemanticity,
       title={Towards Monosemanticity: Decomposing Language Models With Dictionary Learning},
       author={Bricken, Trenton and Templeton, Adly and Batson, Joshua and Chen, Brian and Jermyn, Adam and Conerly, Tom and Turner, Nick and Anil, Cem and Denison, Carson and Askell, Amanda and Lasenby, Robert and Wu, Yifan and Kravec, Shauna and Schiefer, Nicholas and Maxwell, Tim and Joseph, Nicholas and Hatfield-Dodds, Zac and Tamkin, Alex and Nguyen, Karina and McLean, Brayden and Burke, Josiah E and Hume, Tristan and Carter, Shan and Henighan, Tom and Olah, Christopher},
       year={2023},
       journal={Transformer Circuits Thread},
       note={https://transformer-circuits.pub/2023/monosemantic-features/index.html}
    }

@article{gpt2,
  title={Language models are unsupervised multitask learners},
  author={Radford, Alec and Wu, Jeffrey and Child, Rewon and Luan, David and Amodei, Dario and Sutskever, Ilya and others},
  journal={OpenAI blog},
  volume={1},
  number={8},
  pages={9},
  year={2019}
}

@article{meng2022locating,
  title={Locating and editing factual associations in GPT},
  author={Meng, Kevin and Bau, David and Andonian, Alex and Belinkov, Yonatan},
  journal={Advances in Neural Information Processing Systems},
  volume={35},
  pages={17359--17372},
  year={2022}
}

@inproceedings{kusner2019making,
  title={Making decisions that reduce discriminatory impacts},
  author={Kusner, Matt and Russell, Chris and Loftus, Joshua and Silva, Ricardo},
  booktitle={International Conference on Machine Learning},
  pages={3591--3600},
  year={2019},
  organization={PMLR}
}

@article{loftus2018causal,
  title={Causal reasoning for algorithmic fairness},
  author={Loftus, Joshua R and Russell, Chris and Kusner, Matt J and Silva, Ricardo},
  journal={arXiv preprint arXiv:1805.05859},
  year={2018}
}

@inproceedings{kilbertus2020sensitivity,
  title={The sensitivity of counterfactual fairness to unmeasured confounding},
  author={Kilbertus, Niki and Ball, Philip J and Kusner, Matt J and Weller, Adrian and Silva, Ricardo},
  booktitle={Uncertainty in artificial intelligence},
  pages={616--626},
  year={2020},
  organization={PMLR}
}

@inproceedings{counterfactual_fairness,
  author    = {Matt J. Kusner and
               Joshua R. Loftus and
               Chris Russell and
               Ricardo Silva},
  editor    = {Isabelle Guyon and
               Ulrike von Luxburg and
               Samy Bengio and
               Hanna M. Wallach and
               Rob Fergus and
               S. V. N. Vishwanathan and
               Roman Garnett},
  title     = {Counterfactual Fairness},
  booktitle = {Advances in Neural Information Processing Systems 30: Annual Conference
               on Neural Information Processing Systems 2017, December 4-9, 2017,
               Long Beach, CA, {USA}},
  pages     = {4066--4076},
  year      = {2017},
  url       = {https://proceedings.neurips.cc/paper/2017/hash/a486cd07e4ac3d270571622f4f316ec5-Abstract.html},
  bibsource = {dblp computer science bibliography, https://dblp.org}
}

@article{lynch2023spawrious,
  title={Spawrious: A benchmark for fine control of spurious correlation biases},
  author={Lynch, Aengus and Dovonon, Gb{\`e}tondji JS and Kaddour, Jean and Silva, Ricardo},
  journal={arXiv preprint arXiv:2303.05470},
  year={2023}
}

@article{bottou2013counterfactual,
  title={Counterfactual Reasoning and Learning Systems: The Example of Computational Advertising.},
  author={Bottou, L{\'e}on and Peters, Jonas and Qui{\~n}onero-Candela, Joaquin and Charles, Denis X and Chickering, D Max and Portugaly, Elon and Ray, Dipankar and Simard, Patrice and Snelson, Ed},
  journal={Journal of Machine Learning Research},
  volume={14},
  number={11},
  year={2013}
}

@misc{causal_bandits_separating_sets,
  doi = {10.48550/ARXIV.2009.07916},
  
  url = {https://arxiv.org/abs/2009.07916},
  
  author = {de Kroon, Arnoud A. W. M. and Belgrave, Danielle and Mooij, Joris M.},
  
  title = {Causal Discovery for Causal Bandits utilizing Separating Sets},
  
  publisher = {arXiv},
  
  year = {2020},
  
  copyright = {arXiv.org perpetual, non-exclusive license}
}

@inproceedings{
malek2021asymptotically,
title={Asymptotically Best Causal Effect Identification with Multi-Armed Bandits},
author={Alan Malek and Silvia Chiappa},
booktitle={Advances in Neural Information Processing Systems},
editor={A. Beygelzimer and Y. Dauphin and P. Liang and J. Wortman Vaughan},
year={2021},
url={https://openreview.net/forum?id=1dqrBgHYC0d}
}

@inproceedings{gpt3,
	title        = {Language models are few-shot learners},
	author       = {Brown, Tom B and Mann, Benjamin and Ryder, Nick and Subbiah, Melanie and Kaplan, Jared and Dhariwal, Prafulla and Neelakantan, Arvind and Shyam, Pranav and Sastry, Girish and Askell, Amanda and others},
	year         = 2020,
	booktitle    = {NeurIPS}
}

@inproceedings{byol,
	title        = {Bootstrap your own latent: A new approach to self-supervised learning},
	author       = {Grill, Jean-Bastien and Strub, Florian and Altch{\'e}, Florent and Tallec, Corentin and Richemond, Pierre H and Buchatskaya, Elena and Doersch, Carl and Pires, Bernardo Avila and Guo, Zhaohan Daniel and Azar, Mohammad Gheshlaghi and others},
	year         = 2020,
	booktitle    = {NeurIPS}
}

@article{mae,
	title        = {Masked Autoencoders Are Scalable Vision Learners},
	author       = {Kaiming He and Xinlei Chen and Saining Xie and Yanghao Li and Piotr Doll{\'{a}}r and Ross B. Girshick},
	year         = 2021,
	journal      = {arXiv:2111.06377}
}

@inproceedings{simclr,
	title        = {A simple framework for contrastive learning of visual representations},
	author       = {Chen, Ting and Kornblith, Simon and Norouzi, Mohammad and Hinton, Geoffrey},
	year         = 2020,
	booktitle    = {ICML}
}

@article{icp,
  title={Causal inference by using invariant prediction: identification and confidence intervals},
  author={Peters, Jonas and B{\"u}hlmann, Peter and Meinshausen, Nicolai},
  journal={Journal of the Royal Statistical Society: Series B (Statistical Methodology)},
  volume={78},
  number={5},
  pages={947--1012},
  year={2016},
  publisher={Wiley Online Library}
}

@inproceedings{brady,
  title={Introduction to Causal Inference},
  author={Brady Neal},
  year={2020}
}

@inproceedings{PAML,
  author    = {Jean Kaddour and
               Steind{\'{o}}r S{\ae}mundsson and
               Marc Peter Deisenroth},
  editor    = {Hugo Larochelle and
               Marc'Aurelio Ranzato and
               Raia Hadsell and
               Maria{-}Florina Balcan and
               Hsuan{-}Tien Lin},
  title     = {Probabilistic Active Meta-Learning},
  booktitle = {Advances in Neural Information Processing Systems 33: Annual Conference
               on Neural Information Processing Systems 2020, NeurIPS 2020, December
               6-12, 2020, virtual},
  year      = {2020},
  url       = {https://proceedings.neurips.cc/paper/2020/hash/ef0d17b3bdb4ee2aa741ba28c7255c53-Abstract.html},
  bibsource = {dblp computer science bibliography, https://dblp.org}
}

@conference{Jinetal21,
  title = {Causal Direction of Data Collection Matters: Implications of Causal and Anticausal Learning for NLP},
  author = {Jin, Z. and von K{\"u}gelgen, J. and Ni, J. and Vaidhya, T. and Kaushal, A. and Sachan, M. and Sch{\"o}lkopf, B.},
  booktitle = {Proceedings of the 2021 Conference on Empirical Methods in Natural Language Processing (EMNLP)},
  pages = {9499--9513},
  editors = {Marie{-}Francine Moens and Xuanjing Huang and Lucia Specia and Scott Wen{-}tau Yih},
  publisher = {Association for Computational Linguistics},
  month = nov,
  year = {2021},
  note = {*equal contribution},
  doi = {10.18653/v1/2021.emnlp-main.748},
  url = {https://aclanthology.org/2021.emnlp-main.748},
  month_numeric = {11}
}

@misc{schoelkopf2019causality,
      title={Causality for Machine Learning}, 
      author={Bernhard Schölkopf},
      year={2019},
      eprint={1911.10500},
      archivePrefix={arXiv},
      primaryClass={cs.LG}
}

@book{causal_primer,
  title={Causal inference in statistics: A primer},
  author={Glymour, Madelyn and Pearl, Judea and Jewell, Nicholas P},
  year={2016},
  publisher={John Wiley \& Sons}
}

@inproceedings{
mouli2022asymmetry,
title={Asymmetry Learning for Counterfactually-invariant Classification in {OOD} Tasks},
author={S Chandra Mouli and Bruno Ribeiro},
booktitle={International Conference on Learning Representations},
year={2022},
url={https://openreview.net/forum?id=avgclFZ221l}
}

@inproceedings{counterfactual_oppe_gm,
  author    = {Michael Oberst and
               David A. Sontag},
  editor    = {Kamalika Chaudhuri and
               Ruslan Salakhutdinov},
  title     = {Counterfactual Off-Policy Evaluation with Gumbel-Max Structural Causal
               Models},
  booktitle = {Proceedings of the 36th International Conference on Machine Learning,
               {ICML} 2019, 9-15 June 2019, Long Beach, California, {USA}},
  series    = {Proceedings of Machine Learning Research},
  volume    = {97},
  pages     = {4881--4890},
  publisher = {{PMLR}},
  year      = {2019},
  url       = {http://proceedings.mlr.press/v97/oberst19a.html},
  bibsource = {dblp computer science bibliography, https://dblp.org}
}

@misc{generalizing_ope,
 
  
  author = {Parbhoo, Sonali and Joshi, Shalmali and Doshi-Velez, Finale},
  
  title = {Generalizing Off-Policy Evaluation From a Causal Perspective For Sequential Decision-Making},
  
  publisher = {arXiv},
  
  year = {2022},
  
  copyright = {Creative Commons Attribution 4.0 International}
}

@inproceedings{
cf_gps,
title={Woulda, Coulda, Shoulda: Counterfactually-Guided Policy Search},
author={Lars Buesing and Theophane Weber and Yori Zwols and Nicolas Heess and Sebastien Racaniere and Arthur Guez and Jean-Baptiste Lespiau},
booktitle={International Conference on Learning Representations},
year={2019},
url={https://openreview.net/forum?id=BJG0voC9YQ},
}

@article{DBLP:journals/corr/abs-2312-11805,
  author       = {Rohan Anil and
                  Sebastian Borgeaud and
                  Yonghui Wu and
                  Jean{-}Baptiste Alayrac and
                  Jiahui Yu and
                  Radu Soricut and
                  Johan Schalkwyk and
                  Andrew M. Dai and
                  Anja Hauth and
                  Katie Millican and
                  David Silver and
                  Slav Petrov and
                  Melvin Johnson and
                  Ioannis Antonoglou and
                  Julian Schrittwieser and
                  Amelia Glaese and
                  Jilin Chen and
                  Emily Pitler and
                  Timothy P. Lillicrap and
                  Angeliki Lazaridou and
                  Orhan Firat and
                  James Molloy and
                  Michael Isard and
                  Paul Ronald Barham and
                  Tom Hennigan and
                  Benjamin Lee and
                  Fabio Viola and
                  Malcolm Reynolds and
                  Yuanzhong Xu and
                  Ryan Doherty and
                  Eli Collins and
                  Clemens Meyer and
                  Eliza Rutherford and
                  Erica Moreira and
                  Kareem Ayoub and
                  Megha Goel and
                  George Tucker and
                  Enrique Piqueras and
                  Maxim Krikun and
                  Iain Barr and
                  Nikolay Savinov and
                  Ivo Danihelka and
                  Becca Roelofs and
                  Ana{\"{\i}}s White and
                  Anders Andreassen and
                  Tamara von Glehn and
                  Lakshman Yagati and
                  Mehran Kazemi and
                  Lucas Gonzalez and
                  Misha Khalman and
                  Jakub Sygnowski and
                  et al.},
  title        = {Gemini: {A} Family of Highly Capable Multimodal Models},
  journal      = {CoRR},
  volume       = {abs/2312.11805},
  year         = {2023}
}

@article{DBLP:journals/corr/abs-2305-10403,
  author       = {Rohan Anil and
                  Andrew M. Dai and
                  Orhan Firat and
                  Melvin Johnson and
                  Dmitry Lepikhin and
                  Alexandre Passos and
                  Siamak Shakeri and
                  Emanuel Taropa and
                  Paige Bailey and
                  Zhifeng Chen and
                  Eric Chu and
                  Jonathan H. Clark and
                  Laurent El Shafey and
                  Yanping Huang and
                  Kathy Meier{-}Hellstern and
                  Gaurav Mishra and
                  Erica Moreira and
                  Mark Omernick and
                  Kevin Robinson and
                  Sebastian Ruder and
                  Yi Tay and
                  Kefan Xiao and
                  Yuanzhong Xu and
                  Yujing Zhang and
                  Gustavo Hern{\'{a}}ndez {\'{A}}brego and
                  Junwhan Ahn and
                  Jacob Austin and
                  Paul Barham and
                  Jan A. Botha and
                  James Bradbury and
                  Siddhartha Brahma and
                  Kevin Brooks and
                  Michele Catasta and
                  Yong Cheng and
                  Colin Cherry and
                  Christopher A. Choquette{-}Choo and
                  Aakanksha Chowdhery and
                  Cl{\'{e}}ment Crepy and
                  Shachi Dave and
                  Mostafa Dehghani and
                  Sunipa Dev and
                  Jacob Devlin and
                  Mark D{\'{\i}}az and
                  Nan Du and
                  Ethan Dyer and
                  Vladimir Feinberg and
                  Fangxiaoyu Feng and
                  Vlad Fienber and
                  Markus Freitag and
                  Xavier Garcia and
                  Sebastian Gehrmann and
                  Lucas Gonzalez and
                  et al.},
  title        = {PaLM 2 Technical Report},
  journal      = {CoRR},
  volume       = {abs/2305.10403},
  year         = {2023}
}

@article{DBLP:journals/corr/abs-2307-09288,
  author       = {Hugo Touvron and
                  Louis Martin and
                  Kevin Stone and
                  Peter Albert and
                  Amjad Almahairi and
                  Yasmine Babaei and
                  Nikolay Bashlykov and
                  Soumya Batra and
                  Prajjwal Bhargava and
                  Shruti Bhosale and
                  Dan Bikel and
                  Lukas Blecher and
                  Cristian Canton{-}Ferrer and
                  Moya Chen and
                  Guillem Cucurull and
                  David Esiobu and
                  Jude Fernandes and
                  Jeremy Fu and
                  Wenyin Fu and
                  Brian Fuller and
                  Cynthia Gao and
                  Vedanuj Goswami and
                  Naman Goyal and
                  Anthony Hartshorn and
                  Saghar Hosseini and
                  Rui Hou and
                  Hakan Inan and
                  Marcin Kardas and
                  Viktor Kerkez and
                  Madian Khabsa and
                  Isabel Kloumann and
                  Artem Korenev and
                  Punit Singh Koura and
                  Marie{-}Anne Lachaux and
                  Thibaut Lavril and
                  Jenya Lee and
                  Diana Liskovich and
                  Yinghai Lu and
                  Yuning Mao and
                  Xavier Martinet and
                  Todor Mihaylov and
                  Pushkar Mishra and
                  Igor Molybog and
                  Yixin Nie and
                  Andrew Poulton and
                  Jeremy Reizenstein and
                  Rashi Rungta and
                  Kalyan Saladi and
                  Alan Schelten and
                  Ruan Silva and
                  Eric Michael Smith and
                  Ranjan Subramanian and
                  Xiaoqing Ellen Tan and
                  Binh Tang and
                  Ross Taylor and
                  Adina Williams and
                  Jian Xiang Kuan and
                  Puxin Xu and
                  Zheng Yan and
                  Iliyan Zarov and
                  Yuchen Zhang and
                  Angela Fan and
                  Melanie Kambadur and
                  Sharan Narang and
                  Aur{\'{e}}lien Rodriguez and
                  Robert Stojnic and
                  Sergey Edunov and
                  Thomas Scialom},
  title        = {Llama 2: Open Foundation and Fine-Tuned Chat Models},
  journal      = {CoRR},
  volume       = {abs/2307.09288},
  year         = {2023}
}

@article{llama3modelcard,

title={Llama 3 Model Card},

author={AI@Meta},

year={2024},

url = {https://github.com/meta-llama/llama3/blob/main/MODEL_CARD.md}

}

@article{anthropic2024claude,
  title={The claude 3 model family: Opus, sonnet, haiku},
  author={Anthropic, AI},
  journal={Claude-3 Model Card},
  year={2024}
}

@misc{claude2,
    author    = {Anthropic},
    title     = {Anthropic. Model card and evaluations for claude models.},
    year      = {2023},
    url       = {https://www-files.anthropic.com/production/images/Model-Card-Claude-2.pdf},
}

@article{DBLP:journals/corr/abs-2303-08774,
  author       = {OpenAI},
  title        = {{GPT-4} Technical Report},
  journal      = {CoRR},
  volume       = {abs/2303.08774},
  year         = {2023}
}

@article{DBLP:journals/corr/abs-2307-10169,
  author       = {Jean Kaddour and
                  Joshua Harris and
                  Maximilian Mozes and
                  Herbie Bradley and
                  Roberta Raileanu and
                  Robert McHardy},
  title        = {Challenges and Applications of Large Language Models},
  journal      = {CoRR},
  volume       = {abs/2307.10169},
  year         = {2023},
  url          = {https://doi.org/10.48550/arXiv.2307.10169},
  doi          = {10.48550/ARXIV.2307.10169},
  eprinttype    = {arXiv},
  eprint       = {2307.10169},
  bibsource    = {dblp computer science bibliography, https://dblp.org}
}

@article{proximal_RL,
  author    = {Andrew Bennett and
               Nathan Kallus},
  title     = {Proximal Reinforcement Learning: Efficient Off-Policy Evaluation in
               Partially Observed Markov Decision Processes},
  journal   = {CoRR},
  volume    = {abs/2110.15332},
  year      = {2021},
  url       = {https://arxiv.org/abs/2110.15332},
  eprinttype = {arXiv},
  eprint    = {2110.15332},
  bibsource = {dblp computer science bibliography, https://dblp.org}
}

@inproceedings{confounding_robust,
  author    = {Nathan Kallus and
               Angela Zhou},
  editor    = {Samy Bengio and
               Hanna M. Wallach and
               Hugo Larochelle and
               Kristen Grauman and
               Nicol{\`{o}} Cesa{-}Bianchi and
               Roman Garnett},
  title     = {Confounding-Robust Policy Improvement},
  booktitle = {Advances in Neural Information Processing Systems 31: Annual Conference
               on Neural Information Processing Systems 2018, NeurIPS 2018, December
               3-8, 2018, Montr{\'{e}}al, Canada},
  pages     = {9289--9299},
  year      = {2018},
  url       = {https://proceedings.neurips.cc/paper/2018/hash/3a09a524440d44d7f19870070a5ad42f-Abstract.html},
  bibsource = {dblp computer science bibliography, https://dblp.org}
}

@inproceedings{causal_bandits,
  author    = {Finnian Lattimore and
               Tor Lattimore and
               Mark D. Reid},
  editor    = {Daniel D. Lee and
               Masashi Sugiyama and
               Ulrike von Luxburg and
               Isabelle Guyon and
               Roman Garnett},
  title     = {Causal Bandits: Learning Good Interventions via Causal Inference},
  booktitle = {Advances in Neural Information Processing Systems 29: Annual Conference
               on Neural Information Processing Systems 2016, December 5-10, 2016,
               Barcelona, Spain},
  pages     = {1181--1189},
  year      = {2016},
  url       = {https://proceedings.neurips.cc/paper/2016/hash/b4288d9c0ec0a1841b3b3728321e7088-Abstract.html},
  bibsource = {dblp computer science bibliography, https://dblp.org}
}

@article{hussein2017imitation,
  title={Imitation learning: A survey of learning methods},
  author={Hussein, Ahmed and Gaber, Mohamed Medhat and Elyan, Eyad and Jayne, Chrisina},
  journal={ACM Computing Surveys (CSUR)},
  volume={50},
  number={2},
  pages={1--35},
  year={2017},
  publisher={ACM New York, NY, USA}
}

@inproceedings{causal_imitation_learning,
 author = {Zhang, Junzhe and Kumor, Daniel and Bareinboim, Elias},
 booktitle = {Advances in Neural Information Processing Systems},
 editor = {H. Larochelle and M. Ranzato and R. Hadsell and M. F. Balcan and H. Lin},
 pages = {12263--12274},
 publisher = {Curran Associates, Inc.},
 title = {Causal Imitation Learning With Unobserved Confounders},
 url = {https://proceedings.neurips.cc/paper/2020/file/8fdd149fcaa7058caccc9c4ad5b0d89a-Paper.pdf},
 volume = {33},
 year = {2020}
}

@misc{MISA,
 
  author = {Tomar, Manan and Zhang, Amy and Calandra, Roberto and Taylor, Matthew E. and Pineau, Joelle},
  
  title = {Model-Invariant State Abstractions for Model-Based Reinforcement Learning},
  
  publisher = {arXiv},
  
  year = {2021},
  
  copyright = {arXiv.org perpetual, non-exclusive license}
}

@article{causal_world,
  title={Causalworld: A robotic manipulation benchmark for causal structure and transfer learning},
  author={Ahmed, Ossama and Tr{\"a}uble, Frederik and Goyal, Anirudh and Neitz, Alexander and Bengio, Yoshua and Sch{\"o}lkopf, Bernhard and W{\"u}thrich, Manuel and Bauer, Stefan},
  journal={arXiv preprint arXiv:2010.04296},
  year={2020}
}

@inproceedings{maml,
  title={Model-agnostic meta-learning for fast adaptation of deep networks},
  author={Finn, Chelsea and Abbeel, Pieter and Levine, Sergey},
  booktitle={International conference on machine learning},
  pages={1126--1135},
  year={2017},
  organization={PMLR}
}

@inproceedings{asymmetric_shap_values,
  author    = {Christopher Frye and
               Colin Rowat and
               Ilya Feige},
  editor    = {Hugo Larochelle and
               Marc'Aurelio Ranzato and
               Raia Hadsell and
               Maria{-}Florina Balcan and
               Hsuan{-}Tien Lin},
  title     = {Asymmetric Shapley values: incorporating causal knowledge into model-agnostic
               explainability},
  booktitle = {Advances in Neural Information Processing Systems 33: Annual Conference
               on Neural Information Processing Systems 2020, NeurIPS 2020, December
               6-12, 2020, virtual},
  year      = {2020},
  url       = {https://proceedings.neurips.cc/paper/2020/hash/0d770c496aa3da6d2c3f2bd19e7b9d6b-Abstract.html},
  bibsource = {dblp computer science bibliography, https://dblp.org}
}

@inproceedings{anticausal,
  author    = {Bernhard Sch{\"{o}}lkopf and
               Dominik Janzing and
               Jonas Peters and
               Eleni Sgouritsa and
               Kun Zhang and
               Joris M. Mooij},
  title     = {On causal and anticausal learning},
  booktitle = {Proceedings of the 29th International Conference on Machine Learning,
               {ICML} 2012, Edinburgh, Scotland, UK, June 26 - July 1, 2012},
  publisher = {icml.cc / Omnipress},
  year      = {2012},
  url       = {http://icml.cc/2012/papers/625.pdf},
  bibsource = {dblp computer science bibliography, https://dblp.org}
}

@inproceedings{beery2018recognition,
  title={Recognition in terra incognita},
  author={Beery, Sara and Van Horn, Grant and Perona, Pietro},
  booktitle={Proceedings of the European conference on computer vision (ECCV)},
  pages={456--473},
  year={2018}
}

@inproceedings{traffic_prediction,
  title={SATISFy: Towards a self-learning analyzer for time series forecasting in self-improving systems},
  author={Krupitzer, Christian and Pfannem{\"u}ller, Martin and Kaddour, Jean and Becker, Christian},
  booktitle={2018 IEEE 3rd International Workshops on Foundations and Applications of Self* Systems (FAS* W)},
  pages={182--189},
  year={2018},
  organization={IEEE}
}

@inproceedings{imagenet,
  title={Imagenet: A large-scale hierarchical image database},
  author={Deng, Jia and Dong, Wei and Socher, Richard and Li, Li-Jia and Li, Kai and Fei-Fei, Li},
  booktitle={2009 IEEE conference on computer vision and pattern recognition},
  pages={248--255},
  year={2009},
  organization={Ieee}
}

@inproceedings{
singla2022salient,
title={Salient ImageNet: How to discover spurious features in Deep Learning?},
author={Sahil Singla and Soheil Feizi},
booktitle={International Conference on Learning Representations},
year={2022},
url={https://openreview.net/forum?id=XVPqLyNxSyh}
}

@inproceedings{cbo_ricardo,
  author    = {Ricardo Silva},
  editor    = {Daniel D. Lee and
               Masashi Sugiyama and
               Ulrike von Luxburg and
               Isabelle Guyon and
               Roman Garnett},
  title     = {Observational-Interventional Priors for Dose-Response Learning},
  booktitle = {Advances in Neural Information Processing Systems 29: Annual Conference
               on Neural Information Processing Systems 2016, December 5-10, 2016,
               Barcelona, Spain},
  pages     = {1561--1569},
  year      = {2016},
  url       = {https://proceedings.neurips.cc/paper/2016/hash/aff1621254f7c1be92f64550478c56e6-Abstract.html},
  bibsource = {dblp computer science bibliography, https://dblp.org}
}

@article{gershman2017reinforcement,
  title={Reinforcement learning and causal models},
  author={Gershman, Samuel J},
  journal={The Oxford handbook of causal reasoning},
  volume={1},
  pages={295},
  year={2017},
  publisher={Oxford University Press}
}

@inproceedings{
kumor2021sequential,
title={Sequential Causal Imitation Learning with Unobserved Confounders},
author={Daniel Kumor and Junzhe Zhang and Elias Bareinboim},
booktitle={Advances in Neural Information Processing Systems},
editor={A. Beygelzimer and Y. Dauphin and P. Liang and J. Wortman Vaughan},
year={2021},
url={https://openreview.net/forum?id=o6-k168bBD8}
}

@inproceedings{audibert2010best,
  title={Best arm identification in multi-armed bandits.},
  author={Audibert, Jean-Yves and Bubeck, S{\'e}bastien and Munos, R{\'e}mi},
  booktitle={COLT},
  pages={41--53},
  year={2010},
  organization={Citeseer}
}

@inproceedings{structural_causal_bandits,
 author = {Lee, Sanghack and Bareinboim, Elias},
 booktitle = {Advances in Neural Information Processing Systems},
 editor = {S. Bengio and H. Wallach and H. Larochelle and K. Grauman and N. Cesa-Bianchi and R. Garnett},
 pages = {},
 publisher = {Curran Associates, Inc.},
 title = {Structural Causal Bandits: Where to Intervene?},
 url = {https://proceedings.neurips.cc/paper/2018/file/c0a271bc0ecb776a094786474322cb82-Paper.pdf},
 volume = {31},
 year = {2018}
}

@article{bareinboim2015bandits,
  title={Bandits with unobserved confounders: A causal approach},
  author={Bareinboim, Elias and Forney, Andrew and Pearl, Judea},
  journal={Advances in Neural Information Processing Systems},
  volume={28},
  year={2015}
}

@article{learning_by_doing,
  title={Learning by Doing: Controlling a Dynamical System using Causality, Control, and Reinforcement Learning},
  author={Weichwald, Sebastian and Mogensen, S{\o}ren Wengel and Lee, Tabitha Edith and Baumann, Dominik and Kroemer, Oliver and Guyon, Isabelle and Trimpe, Sebastian and Peters, Jonas and Pfister, Niklas},
  journal={arXiv preprint arXiv:2202.06052},
  year={2022}
}

@unpublished{causal_from_RL_perspective,
title= {Causality from the Perspective of Reinforcement Learning},
author = {Csaba Szepesvari},
year = {2018},
note= {Machine Learning for Causal Inference, Counterfactual Prediction, and Autonomous Action (CausalML)
},
URL= {https://sites.google.com/site/faim18wscausalml/invited-talks},
}

@inproceedings{causal_bo,
  author    = {Virginia Aglietti and
               Xiaoyu Lu and
               Andrei Paleyes and
               Javier Gonz{\'{a}}lez},
  editor    = {Silvia Chiappa and
               Roberto Calandra},
  title     = {Causal Bayesian Optimization},
  booktitle = {The 23rd International Conference on Artificial Intelligence and Statistics,
               {AISTATS} 2020, 26-28 August 2020, Online [Palermo, Sicily, Italy]},
  series    = {Proceedings of Machine Learning Research},
  volume    = {108},
  pages     = {3155--3164},
  publisher = {{PMLR}},
  year      = {2020},
  url       = {http://proceedings.mlr.press/v108/aglietti20a.html},
  bibsource = {dblp computer science bibliography, https://dblp.org}
}

@inproceedings{chiappa2019path,
  title={Path-specific counterfactual fairness},
  author={Chiappa, Silvia},
  booktitle={Proceedings of the AAAI Conference on Artificial Intelligence},
  volume={33},
  number={01},
  pages={7801--7808},
  year={2019}
}

@inproceedings{ahuja2020invariant,
  title={Invariant risk minimization games},
  author={Ahuja, Kartik and Shanmugam, Karthikeyan and Varshney, Kush and Dhurandhar, Amit},
  booktitle={International Conference on Machine Learning},
  pages={145--155},
  year={2020},
  organization={PMLR}
}

@article{rezende2020causally,
  title={Causally correct partial models for reinforcement learning},
  author={Rezende, Danilo J and Danihelka, Ivo and Papamakarios, George and Ke, Nan Rosemary and Jiang, Ray and Weber, Theophane and Gregor, Karol and Merzic, Hamza and Viola, Fabio and Wang, Jane and others},
  journal={arXiv preprint arXiv:2002.02836},
  year={2020}
}

@inproceedings{kamath2021does,
  title={Does invariant risk minimization capture invariance?},
  author={Kamath, Pritish and Tangella, Akilesh and Sutherland, Danica and Srebro, Nathan},
  booktitle={International Conference on Artificial Intelligence and Statistics},
  pages={4069--4077},
  year={2021},
  organization={PMLR}
}

@inproceedings{OPPE_POMDP,
  author    = {Guy Tennenholtz and
               Uri Shalit and
               Shie Mannor},
  title     = {Off-Policy Evaluation in Partially Observable Environments},
  booktitle = {The Thirty-Fourth {AAAI} Conference on Artificial Intelligence, {AAAI}
               2020, The Thirty-Second Innovative Applications of Artificial Intelligence
               Conference, {IAAI} 2020, The Tenth {AAAI} Symposium on Educational
               Advances in Artificial Intelligence, {EAAI} 2020, New York, NY, USA,
               February 7-12, 2020},
  pages     = {10276--10283},
  publisher = {{AAAI} Press},
  year      = {2020},
  url       = {https://ojs.aaai.org/index.php/AAAI/article/view/6590},
  bibsource = {dblp computer science bibliography, https://dblp.org}
}

@article{lillicrap2015continuous,
  title={Continuous control with deep reinforcement learning},
  author={Lillicrap, Timothy P and Hunt, Jonathan J and Pritzel, Alexander and Heess, Nicolas and Erez, Tom and Tassa, Yuval and Silver, David and Wierstra, Daan},
  journal={arXiv preprint arXiv:1509.02971},
  year={2015}
}

@article{charpentier2021reinforcement,
  title={Reinforcement learning in economics and finance},
  author={Charpentier, Arthur and Elie, Romuald and Remlinger, Carl},
  journal={Computational Economics},
  pages={1--38},
  year={2021},
  publisher={Springer}
}

@inproceedings{crpe_inf,
 author={Kallus, Nathan and Zhou, Angela},
 booktitle = {Advances in Neural Information Processing Systems},
 editor = {H. Larochelle and M. Ranzato and R. Hadsell and M.F. Balcan and H. Lin},
 pages = {22293--22304},
 publisher = {Curran Associates, Inc.},
 title = {Confounding-Robust Policy Evaluation in Infinite-Horizon Reinforcement Learning},
 url = {https://proceedings.neurips.cc/paper/2020/file/fd4f21f2556dad0ea8b7a5c04eabebda-Paper.pdf},
 volume = {33},
 year = {2020}
}

@inproceedings{bennett2021off,
  title={Off-policy evaluation in infinite-horizon reinforcement learning with latent confounders},
  author={Bennett, Andrew and Kallus, Nathan and Li, Lihong and Mousavi, Ali},
  booktitle={International Conference on Artificial Intelligence and Statistics},
  pages={1999--2007},
  year={2021},
  organization={PMLR}
}

@article{namkoong2020off,
  title={Off-policy policy evaluation for sequential decisions under unobserved confounding},
  author={Namkoong, Hongseok and Keramati, Ramtin and Yadlowsky, Steve and Brunskill, Emma},
  journal={Advances in Neural Information Processing Systems},
  volume={33},
  pages={18819--18831},
  year={2020}
}

@inproceedings{causal_confusion,
  author    = {Pim de Haan and
               Dinesh Jayaraman and
               Sergey Levine},
  editor    = {Hanna M. Wallach and
               Hugo Larochelle and
               Alina Beygelzimer and
               Florence d'Alch{\'{e}}{-}Buc and
               Emily B. Fox and
               Roman Garnett},
  title     = {Causal Confusion in Imitation Learning},
  booktitle = {Advances in Neural Information Processing Systems 32: Annual Conference
               on Neural Information Processing Systems 2019, NeurIPS 2019, December
               8-14, 2019, Vancouver, BC, Canada},
  pages     = {11693--11704},
  year      = {2019},
  url       = {https://proceedings.neurips.cc/paper/2019/hash/947018640bf36a2bb609d3557a285329-Abstract.html},
  bibsource = {dblp computer science bibliography, https://dblp.org}
}

@misc{transtee,
  doi = {10.48550/ARXIV.2202.01336},
  
  url = {https://arxiv.org/abs/2202.01336},
  
  author = {Zhang, Yi-Fan and Zhang, Hanlin and Lipton, Zachary C. and Li, Li Erran and Xing, Eric P.},
  
  title = {Exploring Transformer Backbones for Heterogeneous Treatment Effect Estimation},
  
  publisher = {arXiv},
  
  year = {2022},
  
  copyright = {arXiv.org perpetual, non-exclusive license}
}

@article{bo,
  title={Efficient global optimization of expensive black-box functions},
  author={Jones, Donald R and Schonlau, Matthias and Welch, William J},
  journal={Journal of Global optimization},
  volume={13},
  number={4},
  pages={455--492},
  year={1998},
  publisher={Springer}
}

@article{scholkopf2021towards,
  title={Towards causal representation learning},
  author={Sch{\"o}lkopf, Bernhard and Locatello, Francesco and Bauer, Stefan and Ke, Nan Rosemary and Kalchbrenner, Nal and Goyal, Anirudh and Bengio, Yoshua},
  journal={arXiv preprint arXiv:2102.11107}, 
  year={2021}
}

@article{inscrutable,
author = {Guidotti, Riccardo and Monreale, Anna and Ruggieri, Salvatore and Turini, Franco and Giannotti, Fosca and Pedreschi, Dino},
title = {A Survey of Methods for Explaining Black Box Models},
year = {2018},
journal = {ACM Comput. Surv.}
}

@inproceedings{zhao-etal-2017-men,
    title = "Men Also Like Shopping: Reducing Gender Bias Amplification using Corpus-level Constraints",
    author = "Zhao, Jieyu  and
      Wang, Tianlu  and
      Yatskar, Mark  and
      Ordonez, Vicente  and
      Chang, Kai-Wei",
    booktitle = "Proceedings of the 2017 Conference on Empirical Methods in Natural Language Processing",
    year = "2017"
}

@inproceedings{mccoy-etal-2019-right,
    title = "Right for the Wrong Reasons: Diagnosing Syntactic Heuristics in Natural Language Inference",
    author = "McCoy, Tom  and
      Pavlick, Ellie  and
      Linzen, Tal",
    booktitle = "Proceedings of the 57th Annual Meeting of the Association for Computational Linguistics",
    year = "2019"
}

@article{causaLM,
  author    = {Amir Feder and
               Nadav Oved and
               Uri Shalit and
               Roi Reichart},
  title     = {CausaLM: Causal Model Explanation Through Counterfactual Language
               Models},
  journal   = {Comput. Linguistics},
  volume    = {47},
  number    = {2},
  pages     = {333--386},
  year      = {2021},
  url       = {https://doi.org/10.1162/coli\_a\_00404},
  doi       = {10.1162/coli\_a\_00404},
  bibsource = {dblp computer science bibliography, https://dblp.org}
}

@inproceedings{counterfactual_generator,
  author    = {Xiangji Zeng and
               Yunliang Li and
               Yuchen Zhai and
               Yin Zhang},
  editor    = {Bonnie Webber and
               Trevor Cohn and
               Yulan He and
               Yang Liu},
  title     = {Counterfactual Generator: {A} Weakly-Supervised Method for Named Entity
               Recognition},
  booktitle = {Proceedings of the 2020 Conference on Empirical Methods in Natural
               Language Processing, {EMNLP} 2020, Online, November 16-20, 2020},
  pages     = {7270--7280},
  publisher = {Association for Computational Linguistics},
  year      = {2020},
  url       = {https://doi.org/10.18653/v1/2020.emnlp-main.590},
  doi       = {10.18653/v1/2020.emnlp-main.590},
  bibsource = {dblp computer science bibliography, https://dblp.org}
}

@inproceedings{cad,
  author    = {Divyansh Kaushik and
               Amrith Setlur and
               Eduard H. Hovy and
               Zachary Chase Lipton},
  title     = {Explaining the Efficacy of Counterfactually Augmented Data},
  booktitle = {9th International Conference on Learning Representations, {ICLR} 2021,
               Virtual Event, Austria, May 3-7, 2021},
  publisher = {OpenReview.net},
  year      = {2021},
  url       = {https://openreview.net/forum?id=HHiiQKWsOcV},
  bibsource = {dblp computer science bibliography, https://dblp.org}
}

@inproceedings{gender_bias,
  author    = {Jesse Vig and
               Sebastian Gehrmann and
               Yonatan Belinkov and
               Sharon Qian and
               Daniel Nevo and
               Yaron Singer and
               Stuart M. Shieber},
  editor    = {Hugo Larochelle and
               Marc'Aurelio Ranzato and
               Raia Hadsell and
               Maria{-}Florina Balcan and
               Hsuan{-}Tien Lin},
  title     = {Investigating Gender Bias in Language Models Using Causal Mediation
               Analysis},
  booktitle = {Advances in Neural Information Processing Systems 33: Annual Conference
               on Neural Information Processing Systems 2020, NeurIPS 2020, December
               6-12, 2020, virtual},
  year      = {2020},
  url       = {https://proceedings.neurips.cc/paper/2020/hash/92650b2e92217715fe312e6fa7b90d82-Abstract.html},
  bibsource = {dblp computer science bibliography, https://dblp.org}
}

@inproceedings{cf_fairness_text_classification,
  author    = {Sahaj Garg and
               Vincent Perot and
               Nicole Limtiaco and
               Ankur Taly and
               Ed H. Chi and
               Alex Beutel},
  editor    = {Vincent Conitzer and
               Gillian K. Hadfield and
               Shannon Vallor},
  title     = {Counterfactual Fairness in Text Classification through Robustness},
  booktitle = {Proceedings of the 2019 {AAAI/ACM} Conference on AI, Ethics, and Society,
               {AIES} 2019, Honolulu, HI, USA, January 27-28, 2019},
  pages     = {219--226},
  publisher = {{ACM}},
  year      = {2019},
  url       = {https://doi.org/10.1145/3306618.3317950},
  doi       = {10.1145/3306618.3317950},
  bibsource = {dblp computer science bibliography, https://dblp.org}
}

@inproceedings{
Kaushik2020Learning,
title={Learning The Difference That Makes A Difference With Counterfactually-Augmented Data},
author={Divyansh Kaushik and Eduard Hovy and Zachary Lipton},
booktitle={International Conference on Learning Representations},
year={2020},
url={https://openreview.net/forum?id=Sklgs0NFvr}
}

@inproceedings{generative_interventions,
  author    = {Chengzhi Mao and
               Augustine Cha and
               Amogh Gupta and
               Hao Wang and
               Junfeng Yang and
               Carl Vondrick},
  title     = {Generative Interventions for Causal Learning},
  booktitle = {{IEEE} Conference on Computer Vision and Pattern Recognition, {CVPR}
               2021, virtual, June 19-25, 2021},
  pages     = {3947--3956},
  publisher = {Computer Vision Foundation / {IEEE}},
  year      = {2021},
  url       = {https://openaccess.thecvf.com/content/CVPR2021/html/Mao\_Generative\_Interventions\_for\_Causal\_Learning\_CVPR\_2021\_paper.html},
  bibsource = {dblp computer science bibliography, https://dblp.org}
}

@article{causal_visual_feature_learning,
  author    = {Xin Li and
               Zhizheng Zhang and
               Guoqiang Wei and
               Cuiling Lan and
               Wenjun Zeng and
               Xin Jin and
               Zhibo Chen},
  title     = {Confounder Identification-free Causal Visual Feature Learning},
  journal   = {CoRR},
  volume    = {abs/2111.13420},
  year      = {2021},
  url       = {https://arxiv.org/abs/2111.13420},
  eprinttype = {arXiv},
  eprint    = {2111.13420},
  bibsource = {dblp computer science bibliography, https://dblp.org}
}

@inproceedings{wsss,
  author    = {Dong Zhang and
               Hanwang Zhang and
               Jinhui Tang and
               Xian{-}Sheng Hua and
               Qianru Sun},
  editor    = {Hugo Larochelle and
               Marc'Aurelio Ranzato and
               Raia Hadsell and
               Maria{-}Florina Balcan and
               Hsuan{-}Tien Lin},
  title     = {Causal Intervention for Weakly-Supervised Semantic Segmentation},
  booktitle = {Advances in Neural Information Processing Systems 33: Annual Conference
               on Neural Information Processing Systems 2020, NeurIPS 2020, December
               6-12, 2020, virtual},
  year      = {2020},
  url       = {https://proceedings.neurips.cc/paper/2020/hash/07211688a0869d995947a8fb11b215d6-Abstract.html},
  bibsource = {dblp computer science bibliography, https://dblp.org}
}

@inproceedings{
wu2022discovering,
title={Discovering Invariant Rationales for Graph Neural Networks},
author={Yingxin Wu and Xiang Wang and An Zhang and Xiangnan He and Tat-Seng Chua},
booktitle={International Conference on Learning Representations},
year={2022},
url={https://openreview.net/forum?id=hGXij5rfiHw}
}

@inproceedings{causal_curiosity,
  author    = {Sumedh A. Sontakke and
               Arash Mehrjou and
               Laurent Itti and
               Bernhard Sch{\"{o}}lkopf},
  editor    = {Marina Meila and
               Tong Zhang},
  title     = {Causal Curiosity: {RL} Agents Discovering Self-supervised Experiments
               for Causal Representation Learning},
  booktitle = {Proceedings of the 38th International Conference on Machine Learning,
               {ICML} 2021, 18-24 July 2021, Virtual Event},
  series    = {Proceedings of Machine Learning Research},
  volume    = {139},
  pages     = {9848--9858},
  publisher = {{PMLR}},
  year      = {2021},
  url       = {http://proceedings.mlr.press/v139/sontakke21a.html},
  bibsource = {dblp computer science bibliography, https://dblp.org}
}

@misc{feder2021causal,
      title={Causal Inference in Natural Language Processing: Estimation, Prediction, Interpretation and Beyond}, 
      author={Amir Feder and Katherine A. Keith and Emaad Manzoor and Reid Pryzant and Dhanya Sridhar and Zach Wood-Doughty and Jacob Eisenstein and Justin Grimmer and Roi Reichart and Margaret E. Roberts and Brandon M. Stewart and Victor Veitch and Diyi Yang},
      year={2021},
      eprint={2109.00725},
      archivePrefix={arXiv},
      primaryClass={cs.CL}
}

@inproceedings{
veitch2021counterfactual,
title={Counterfactual Invariance to Spurious Correlations in Text Classification},
author={Victor Veitch and Alexander D'Amour and Steve Yadlowsky and Jacob Eisenstein},
booktitle={Advances in Neural Information Processing Systems},
editor={A. Beygelzimer and Y. Dauphin and P. Liang and J. Wortman Vaughan},
year={2021},
url={https://openreview.net/forum?id=BdKxQp0iBi8}
}

@inproceedings{PEARL,
  author    = {Kate Rakelly and
               Aurick Zhou and
               Chelsea Finn and
               Sergey Levine and
               Deirdre Quillen},
  editor    = {Kamalika Chaudhuri and
               Ruslan Salakhutdinov},
  title     = {Efficient Off-Policy Meta-Reinforcement Learning via Probabilistic
               Context Variables},
  booktitle = {Proceedings of the 36th International Conference on Machine Learning,
               {ICML} 2019, 9-15 June 2019, Long Beach, California, {USA}},
  series    = {Proceedings of Machine Learning Research},
  volume    = {97},
  pages     = {5331--5340},
  publisher = {{PMLR}},
  year      = {2019},
  url       = {http://proceedings.mlr.press/v97/rakelly19a.html},
  bibsource = {dblp computer science bibliography, https://dblp.org}
}

@article{IRM,
  author    = {Mart{\'{\i}}n Arjovsky and
               L{\'{e}}on Bottou and
               Ishaan Gulrajani and
               David Lopez{-}Paz},
  title     = {Invariant Risk Minimization},
  journal   = {CoRR},
  volume    = {abs/1907.02893},
  year      = {2019},
  url       = {http://arxiv.org/abs/1907.02893},
  eprinttype = {arXiv},
  eprint    = {1907.02893},
  bibsource = {dblp computer science bibliography, https://dblp.org}
}

@inproceedings{
sauer2021counterfactual,
title={Counterfactual Generative Networks},
author={Axel Sauer and Andreas Geiger},
booktitle={International Conference on Learning Representations},
year={2021},
url={https://openreview.net/forum?id=BXewfAYMmJw}
}

@article{schrouff2022maintaining,
  title={Maintaining fairness across distribution shift: do we have viable solutions for real-world applications?},
  author={Schrouff, Jessica and Harris, Natalie and Koyejo, Oluwasanmi and Alabdulmohsin, Ibrahim and Schnider, Eva and Opsahl-Ong, Krista and Brown, Alex and Roy, Subhrajit and Mincu, Diana and Chen, Christina and others},
  journal={arXiv preprint arXiv:2202.01034},
  year={2022}
}

@inproceedings{cxplain,
  author    = {Patrick Schwab and
               Walter Karlen},
  editor    = {Hanna M. Wallach and
               Hugo Larochelle and
               Alina Beygelzimer and
               Florence d'Alch{\'{e}}{-}Buc and
               Emily B. Fox and
               Roman Garnett},
  title     = {CXPlain: Causal Explanations for Model Interpretation under Uncertainty},
  booktitle = {Advances in Neural Information Processing Systems 32: Annual Conference
               on Neural Information Processing Systems 2019, NeurIPS 2019, December
               8-14, 2019, Vancouver, BC, Canada},
  pages     = {10220--10230},
  year      = {2019},
  url       = {https://proceedings.neurips.cc/paper/2019/hash/3ab6be46e1d6b21d59a3c3a0b9d0f6ef-Abstract.html},
  bibsource = {dblp computer science bibliography, https://dblp.org}
}

@inproceedings{causal_explanations,
  author    = {Matthew R. O'Shaughnessy and
               Gregory Canal and
               Marissa Connor and
               Christopher Rozell and
               Mark A. Davenport},
  editor    = {Hugo Larochelle and
               Marc'Aurelio Ranzato and
               Raia Hadsell and
               Maria{-}Florina Balcan and
               Hsuan{-}Tien Lin},
  title     = {Generative causal explanations of black-box classifiers},
  booktitle = {Advances in Neural Information Processing Systems 33: Annual Conference
               on Neural Information Processing Systems 2020, NeurIPS 2020, December
               6-12, 2020, virtual},
  year      = {2020},
  url       = {https://proceedings.neurips.cc/paper/2020/hash/3a93a609b97ec0ab0ff5539eb79ef33a-Abstract.html},
  bibsource = {dblp computer science bibliography, https://dblp.org}
}

@article{underspecification,
  title={Underspecification presents challenges for credibility in modern machine learning},
  author={D'Amour, Alexander and Heller, Katherine and Moldovan, Dan and Adlam, Ben and Alipanahi, Babak and Beutel, Alex and Chen, Christina and Deaton, Jonathan and Eisenstein, Jacob and Hoffman, Matthew D and others},
  journal={arXiv preprint arXiv:2011.03395},
  year={2020}
}

@article{atzmon_causal_2020,
	title = {A causal view of compositional zero-shot recognition},
	url = {http://arxiv.org/abs/2006.14610},
	journal = {arXiv:2006.14610 [cs]},
	author = {Atzmon, Yuval and Kreuk, Felix and Shalit, Uri and Chechik, Gal},
	year = {2020}
}

@inproceedings{ng2003autonomous,
  title={Autonomous helicopter flight via reinforcement learning.},
  author={Ng, Andrew Y and Kim, H Jin and Jordan, Michael I and Sastry, Shankar and Ballianda, Shiv},
  booktitle={NIPS},
  volume={16},
  year={2003},
  organization={Citeseer}
}

@book{wright1928tariff,
  title={Tariff on animal and vegetable oils},
  author={Wright, Philip G},
  year={1928},
  publisher={Macmillan Company, New York}
}

@misc{cil_under_temporally_correlated_noise,
  doi = {10.48550/ARXIV.2202.01312},
  
  url = {https://arxiv.org/abs/2202.01312},
  
  author = {Swamy, Gokul and Choudhury, Sanjiban and Bagnell, J. Andrew and Wu, Zhiwei Steven},
  
  title = {Causal Imitation Learning under Temporally Correlated Noise},
  
  publisher = {arXiv},
  
  year = {2022},
  
  copyright = {Creative Commons Attribution Share Alike 4.0 International}
}

@article{berkenkamp2017safe,
  title={Safe model-based reinforcement learning with stability guarantees},
  author={Berkenkamp, Felix and Turchetta, Matteo and Schoellig, Angela and Krause, Andreas},
  journal={Advances in neural information processing systems},
  volume={30},
  year={2017}
}

@InProceedings{causal_fairness_consequences,
  title = 	 {Causal Conceptions of Fairness and their Consequences},
  author =       {Nilforoshan, Hamed and Gaebler, Johann D and Shroff, Ravi and Goel, Sharad},
  booktitle = 	 {Proceedings of the 39th International Conference on Machine Learning},
  year = 	 {2022}
}

@misc{oversmoothing,
  doi = {10.48550/ARXIV.1909.03211},
  
  url = {https://arxiv.org/abs/1909.03211},
  
  author = {Chen, Deli and Lin, Yankai and Li, Wei and Li, Peng and Zhou, Jie and Sun, Xu},
  
  title = {Measuring and Relieving the Over-smoothing Problem for Graph Neural Networks from the Topological View},
  
  publisher = {arXiv},
  
  year = {2019},
  
  copyright = {arXiv.org perpetual, non-exclusive license}
}

@article{user_tampering_recommender_systems,
  author    = {Charles Evans and
               Atoosa Kasirzadeh},
  title     = {User Tampering in Reinforcement Learning Recommender Systems},
  journal   = {CoRR},
  volume    = {abs/2109.04083},
  year      = {2021},
  url       = {https://arxiv.org/abs/2109.04083},
  eprinttype = {arXiv},
  eprint    = {2109.04083},
  bibsource = {dblp computer science bibliography, https://dblp.org}
}

@article{reward_tampering,
  title={Reward tampering problems and solutions in reinforcement learning: A causal influence diagram perspective},
  author={Everitt, Tom and Hutter, Marcus and Kumar, Ramana and Krakovna, Victoria},
  journal={Synthese},
  volume={198},
  number={27},
  pages={6435--6467},
  year={2021},
  publisher={Springer}
}

@inproceedings{human_intervention,
  author    = {William Saunders and
               Girish Sastry and
               Andreas Stuhlm{\"{u}}ller and
               Owain Evans},
  editor    = {Elisabeth Andr{\'{e}} and
               Sven Koenig and
               Mehdi Dastani and
               Gita Sukthankar},
  title     = {Trial without Error: Towards Safe Reinforcement Learning via Human
               Intervention},
  booktitle = {Proceedings of the 17th International Conference on Autonomous Agents
               and MultiAgent Systems, {AAMAS} 2018, Stockholm, Sweden, July 10-15,
               2018},
  pages     = {2067--2069},
  publisher = {International Foundation for Autonomous Agents and Multiagent Systems
               Richland, SC, {USA} / {ACM}},
  year      = {2018},
  url       = {http://dl.acm.org/citation.cfm?id=3238074},
  bibsource = {dblp computer science bibliography, https://dblp.org}
}

@inproceedings{langlois2021rl,
  title={How RL agents behave when their actions are modified},
  author={Langlois, Eric D and Everitt, Tom},
  booktitle={Proceedings of the AAAI Conference on Artificial Intelligence},
  volume={35},
  number={13},
  pages={11586--11594},
  year={2021}
}

@misc{path_specific_objectives,
  doi = {10.48550/ARXIV.2204.10018},
  
  url = {https://arxiv.org/abs/2204.10018},
  
  author = {Farquhar, Sebastian and Carey, Ryan and Everitt, Tom},
  
  title = {Path-Specific Objectives for Safer Agent Incentives},
  
  publisher = {arXiv},
  
  year = {2022},
  
  copyright = {arXiv.org perpetual, non-exclusive license}
}

@article{shachter1986evaluating,
  title={Evaluating influence diagrams},
  author={Shachter, Ross D},
  journal={Operations research},
  volume={34},
  number={6},
  pages={871--882},
  year={1986},
  publisher={INFORMS}
}

@article{fagiuoli1998note,
  title={A note about redundancy in influence diagrams},
  author={Fagiuoli, Enrico and Zaffalon, Marco},
  journal={International Journal of Approximate Reasoning},
  volume={19},
  number={3-4},
  pages={351--365},
  year={1998},
  publisher={Elsevier}
}

@misc{everitt_carey_hammond_fox_langlois_legg_2021, url={https://deepmindsafetyresearch.medium.com/progress-on-causal-influence-diagrams-a7a32180b0d1#4e50}, title={Progress on Causal Influence Diagrams}, publisher={Medium}, author={Everitt, Tom and Carey, Ryan and Hammond, Lewis and Fox, James and Langlois, Eric and Legg, Shane}, year={2021}, month={Jun}}

@misc{ivr_ope,
  doi = {10.48550/ARXIV.2105.10148},
  
  url = {https://arxiv.org/abs/2105.10148},
  
  author = {Chen, Yutian and Xu, Liyuan and Gulcehre, Caglar and Paine, Tom Le and Gretton, Arthur and de Freitas, Nando and Doucet, Arnaud},
  
  title = {On Instrumental Variable Regression for Deep Offline Policy Evaluation},
  
  publisher = {arXiv},
  
  year = {2021},
  
  copyright = {Creative Commons Attribution 4.0 International}
}

@article{desautels2017prediction,
  title={Prediction of early unplanned intensive care unit readmission in a UK tertiary care hospital: a cross-sectional machine learning approach},
  author={Desautels, Thomas and Das, Ritankar and Calvert, Jacob and Trivedi, Monica and Summers, Charlotte and Wales, David J and Ercole, Ari},
  journal={BMJ open},
  volume={7},
  number={9},
  pages={e017199},
  year={2017},
  publisher={British Medical Journal Publishing Group}
}

@incollection{bareinboim2022pearl,
  title={On Pearl’s hierarchy and the foundations of causal inference},
  author={Bareinboim, Elias and Correa, Juan D and Ibeling, Duligur and Icard, Thomas},
  booktitle={Probabilistic and Causal Inference: The Works of Judea Pearl},
  pages={507--556},
  publisher={Association for Computing Machinery},
  year={2022}
}

@article{dawid2002influence,
  title={Influence diagrams for causal modelling and inference},
  author={Dawid, A Philip},
  journal={International Statistical Review},
  volume={70},
  number={2},
  pages={161--189},
  year={2002},
  publisher={Wiley Online Library}
}

@article{howard1966information,
  title={Information value theory},
  author={Howard, Ronald A},
  journal={IEEE Transactions on systems science and cybernetics},
  volume={2},
  number={1},
  pages={22--26},
  year={1966},
  publisher={IEEE}
}

@book{lutkepohl2004applied,
  title={Applied time series econometrics},
  author={L{\"u}tkepohl, Helmut and Kr{\"a}tzig, Markus},
  year={2004},
  publisher={Cambridge university press}
}

@article{granger1969investigating,
  title={Investigating causal relations by econometric models and cross-spectral methods},
  author={Granger, Clive WJ},
  journal={Econometrica: journal of the Econometric Society},
  pages={424--438},
  year={1969},
  publisher={JSTOR}
}

@article{
problems_with_GC,
author = {Patrick A. Stokes  and Patrick L. Purdon },
title = {A study of problems encountered in Granger causality analysis from a neuroscience perspective},
journal = {Proceedings of the National Academy of Sciences},
volume = {114},
number = {34},
pages = {E7063-E7072},
year = {2017},
}

@misc{frohberg2021crass,
      title={CRASS: A Novel Data Set and Benchmark to Test Counterfactual Reasoning of Large Language Models}, 
      author={Jörg Frohberg and Frank Binder},
      year={2021},
      eprint={2112.11941},
      archivePrefix={arXiv},
      primaryClass={cs.CL}
}

@article{yang2022towards,
  title={Towards Fine-grained Causal Reasoning and QA},
  author={Yang, Linyi and Wang, Zhen and Wu, Yuxiang and Yang, Jie and Zhang, Yue},
  journal={arXiv preprint arXiv:2204.07408},
  year={2022}
}

@inproceedings{
goyal2021recurrent,
title={Recurrent Independent Mechanisms},
author={Anirudh Goyal and Alex Lamb and Jordan Hoffmann and Shagun Sodhani and Sergey Levine and Yoshua Bengio and Bernhard Sch{\"o}lkopf},
booktitle={International Conference on Learning Representations},
year={2021},
url={https://openreview.net/forum?id=mLcmdlEUxy-}
}

@article{big_bench,
  title={Beyond the Imitation Game: Quantifying and extrapolating the capabilities of language models},
  author={Aarohi Srivastava and Abhinav Rastogi and Abhishek B Rao and Abu Awal Md Shoeb and Abubakar Abid and Adam Fisch and Adam R. Brown and Adam Santoro and Aditya Gupta and Adri{\`a} Garriga-Alonso and Agnieszka Kluska and Aitor Lewkowycz and Akshat Agarwal and Alethea Power and Alex Ray and Alex Warstadt and Alexander W. Kocurek and Ali Safaya and Ali Tazarv and Alice Xiang and Alicia Parrish and Allen Nie and Aman Hussain and Amanda Askell and Amanda Dsouza and Ameet Annasaheb Rahane and Anantharaman S. Iyer and Anders Johan Andreassen and Andrea Santilli and Andreas Stuhlmuller and Andrew M. Dai and Andrew D. La and Andrew Kyle Lampinen and Andy Zou and Angela Jiang and Angelica Chen and Anh Vuong and Animesh Gupta and Anna Gottardi and Antonio Norelli and Anu Venkatesh and Arash Gholamidavoodi and Arfa Tabassum and Arul Menezes and Arun Kirubarajan and Asher Mullokandov and Ashish Sabharwal and Austin Herrick and Avia Efrat and Aykut Erdem and Ayla Karakacs and Bridget R. Roberts and Bao Sheng Loe and Barret Zoph and Bartlomiej Bojanowski and Batuhan Ozyurt and Behnam Hedayatnia and Behnam Neyshabur and Benjamin Inden and Benno Stein and Berk Ekmekci and Bill Yuchen Lin and Blake Stephen Howald and Cameron Diao and Cameron Dour and Catherine Stinson and Cedrick Argueta and C'esar Ferri Ram'irez and Chandan Singh and Charles Rathkopf and Chenlin Meng and Chitta Baral and Chiyu Wu and Chris Callison-Burch and Chris Waites and Christian Voigt and Christopher D. Manning and Christopher Potts and Cindy Tatiana Ramirez and Clara Rivera and Clemencia Siro and Colin Raffel and Courtney Ashcraft and Cristina Garbacea and Damien Sileo and Daniel H Garrette and Dan Hendrycks and Dan Kilman and Dan Roth and Daniel Freeman and Daniel Khashabi and Daniel Levy and Daniel Gonz'alez and Danny Hernandez and Danqi Chen and Daphne Ippolito and Dar Gilboa and David Dohan and D. Drakard and David Jurgens and Debajyoti Datta and Deep Ganguli and Denis Emelin and Denis Kleyko and Deniz Yuret and Derek Chen and Derek Tam and Dieuwke Hupkes and Diganta Misra and Dilyar Buzan and Dimitri Coelho Mollo and Diyi Yang and Dong-Ho Lee and Ekaterina Shutova and Ekin Dogus Cubuk and Elad Segal and Eleanor Hagerman and Elizabeth Barnes and Elizabeth P. Donoway and Ellie Pavlick and Emanuele Rodol{\`a} and Emma FC Lam and Eric Chu and Eric Tang and Erkut Erdem and Ernie Chang and Ethan A. Chi and Ethan Dyer and Ethan Jerzak and Ethan Kim and Eunice Engefu Manyasi and Evgenii Zheltonozhskii and Fan Xia and Fatemeh Siar and Fernando Mart'inez-Plumed and Francesca Happ'e and François Chollet and Frieda Rong and Gaurav Mishra and Genta Indra Winata and Gerard de Melo and Germ{\'a}n Kruszewski and Giambattista Parascandolo and Giorgio Mariani and Gloria Wang and Gonzalo Jaimovitch-L'opez and Gregor Betz and Guy Gur-Ari and Hana Galijasevic and Han Sol Kim and Hannah Rashkin and Hanna Hajishirzi and Harsh Mehta and Hayden Bogar and Henry Shevlin and Hinrich Sch{\"u}tze and Hiromu Yakura and Hongming Zhang and Hubert Wong and Ian Aik-Soon Ng and Isaac Noble and Jaap Jumelet and Jack Geissinger and John Kernion and Jacob Hilton and Jaehoon Lee and Jaime Fern{\'a}ndez Fisac and J. Brooker Simon and James Koppel and James Zheng and James Zou and Jan Koco'n and Jana Thompson and Jared Kaplan and Jarema Radom and Jascha Sohl-Dickstein and Jason Phang and Jason Wei and Jason Yosinski and Jekaterina Novikova and Jelle Bosscher and Jenni Marsh and Jeremy Kim and Jeroen Taal and Jesse Engel and Jesujoba Oluwadara Alabi and Jiacheng Xu and Jiaming Song and Jillian Tang and Jane W Waweru and John Burden and John Miller and John U. Balis and Jonathan Berant and Jorg Frohberg and Jos Rozen and Jos{\'e} Hern{\'a}ndez-Orallo and Joseph Boudeman and Joseph Jones and Joshua B. Tenenbaum and Joshua S. Rule and Joyce Chua and Kamil Kanclerz and Karen Livescu and Karl Krauth and Karthik Gopalakrishnan and Katerina Ignatyeva and Katja Markert and Kaustubh D. Dhole and Kevin Gimpel and Kevin Ochieng’ Omondi and Kory Wallace Mathewson and Kristen Chiafullo and Ksenia Shkaruta and Kumar Shridhar and Kyle McDonell and Kyle Richardson and Laria Reynolds and Leo Gao and Li Zhang and Liam Dugan and Lianhui Qin and Lidia Contreras-Ochando and Louis-Philippe Morency and Luca Moschella and Luca Lam and Lucy Noble and Ludwig Schmidt and Luheng He and Luis Oliveros Col'on and Luke Metz and Lutfi Kerem cSenel and Maarten Bosma and Maarten Sap and Maartje ter Hoeve and Madotto Andrea and Maheen Saleem Farooqi and Manaal Faruqui and Mantas Mazeika and Marco Baturan and Marco Marelli and Marco Maru and M Quintana and Marie Tolkiehn and Mario Giulianelli and Martha Lewis and Martin Potthast and Matthew Leavitt and Matthias Hagen and M'aty'as Schubert and Medina Baitemirova and Melissa Arnaud and Melvin Andrew McElrath and Michael A. Yee and Michael Cohen and Mi Gu and Michael I. Ivanitskiy and Michael Starritt and Michael Strube and Michal Swkedrowski and Michele Bevilacqua and Michihiro Yasunaga and Mihir Kale and Mike Cain and Mimee Xu and Mirac Suzgun and Monica Tiwari and Mohit Bansal and Moin Aminnaseri and Mor Geva and Mozhdeh Gheini and T MukundVarma and Nanyun Peng and Nathan Chi and Nayeon Lee and Neta Gur-Ari Krakover and Nicholas Cameron and Nicholas S. Roberts and Nicholas Doiron and Nikita Nangia and Niklas Deckers and Niklas Muennighoff and Nitish Shirish Keskar and Niveditha Iyer and Noah Constant and Noah Fiedel and Nuan Wen and Oliver Zhang and Omar Agha and Omar Elbaghdadi and Omer Levy and Owain Evans and Pablo Casares and Parth Doshi and Pascale Fung and Paul Pu Liang and Paul Vicol and Pegah Alipoormolabashi and Peiyuan Liao and Percy Liang and Peter W. Chang and Peter Eckersley and Phu Mon Htut and Pi-Bei Hwang and P. Milkowski and Piyush S. Patil and Pouya Pezeshkpour and Priti Oli and Qiaozhu Mei and QING LYU and Qinlang Chen and Rabin Banjade and Rachel Etta Rudolph and Raefer Gabriel and Rahel Habacker and Ram'on Risco Delgado and Rapha{\"e}l Milli{\`e}re and Rhythm Garg and Richard Barnes and Rif A. Saurous and Riku Arakawa and Robbe Raymaekers and Robert Frank and Rohan Sikand and Roman Novak and Roman Sitelew and Ronan Lebras and Rosanne Liu and Rowan Jacobs and Rui Zhang and Ruslan Salakhutdinov and Ryan Chi and Ryan Lee and Ryan Stovall and Ryan Teehan and Rylan Yang and Sahib J. Singh and Saif M. Mohammad and Sajant Anand and Sam Dillavou and Sam Shleifer and Sam Wiseman and Samuel Gruetter and Sam Bowman and Samuel S. Schoenholz and Sanghyun Han and Sanjeev Kwatra and Sarah A. Rous and Sarik Ghazarian and Sayan Ghosh and Sean Casey and Sebastian Bischoff and Sebastian Gehrmann and Sebastian Schuster and Sepideh Sadeghi and Shadi Sameh Hamdan and Sharon Zhou and Shashank Srivastava and Sherry Shi and Shikhar Singh and Shima Asaadi and Shixiang Shane Gu and Shubh Pachchigar and Shubham Toshniwal and Shyam Upadhyay and Shyamolima Debnath and Siamak Shakeri and Simon Thormeyer and Simone Melzi and Siva Reddy and Sneha Priscilla Makini and Soo-hwan Lee and Spencer Bradley Torene and Sriharsha Hatwar and Stanislas Dehaene and Stefan Divic and Stefano Ermon and Stella Rose Biderman and Stephanie C. Lin and Stephen Prasad and Steven T. Piantadosi and Stuart M. Shieber and Summer Misherghi and Svetlana Kiritchenko and Swaroop Mishra and Tal Linzen and Tal Schuster and Tao Li and Tao Yu and Tariq A. Ali and Tatsuo Hashimoto and Te-Lin Wu and Theo Desbordes and Theodore Rothschild and Thomas Phan and Tianle Wang and Tiberius Nkinyili and Timo Schick and T. N. Kornev and Timothy Telleen-Lawton and Titus Tunduny and Tobias Gerstenberg and Trenton Chang and Trishala Neeraj and Tushar Khot and Tyler O. Shultz and Uri Shaham and Vedant Misra and Vera Demberg and Victoria Nyamai and Vikas Raunak and Vinay V. Ramasesh and Vinay Uday Prabhu and Vishakh Padmakumar and Vivek Srikumar and William Fedus and William Saunders and William Zhang and W Vossen and Xiang Ren and Xiaoyu F Tong and Xinyi Wu and Xudong Shen and Yadollah Yaghoobzadeh and Yair Lakretz and Yang Song and Yasaman Bahri and Ye Ji Choi and Yichi Yang and Yiding Hao and Yifu Chen and Yonatan Belinkov and Yu Hou and Yu Hou and Yushi Bai and Zachary Seid and Zhao Xinran and Zhuoye Zhao and Zi Fu Wang and Zijie J. Wang and Zirui Wang and Ziyi Wu and Sahib Singh and Uri Shaham},
  journal={ArXiv},
  year={2022},
  volume={abs/2206.04615}
}

@article{pearl2019seven,
  title={The seven tools of causal inference, with reflections on machine learning},
  author={Pearl, Judea},
  journal={Communications of the ACM},
  volume={62},
  number={3},
  pages={54--60},
  year={2019},
  publisher={ACM New York, NY, USA}
}

@article{lundberg2017unified,
  title={A unified approach to interpreting model predictions},
  author={Lundberg, Scott M and Lee, Su-In},
  journal={Advances in neural information processing systems},
  volume={30},
  year={2017}
}

@article{tank2018neural,
  title={Neural granger causality},
  author={Tank, Alex and Covert, Ian and Foti, Nicholas and Shojaie, Ali and Fox, Emily},
  journal={arXiv preprint arXiv:1802.05842},
  year={2018}
}

@inproceedings{sru,
  author    = {Saurabh Khanna and
               Vincent Y. F. Tan},
  title     = {Economy Statistical Recurrent Units For Inferring Nonlinear Granger
               Causality},
  booktitle = {8th International Conference on Learning Representations, {ICLR} 2020,
               Addis Ababa, Ethiopia, April 26-30, 2020},
  publisher = {OpenReview.net},
  year      = {2020},
  url       = {https://openreview.net/forum?id=SyxV9ANFDH},
  bibsource = {dblp computer science bibliography, https://dblp.org}
}

@misc{mpir,
  doi = {10.48550/ARXIV.2001.01885},
  
  url = {https://arxiv.org/abs/2001.01885},
  
  author = {Wu, Tailin and Breuel, Thomas and Skuhersky, Michael and Kautz, Jan},
  
  title = {Discovering Nonlinear Relations with Minimum Predictive Information Regularization},
  
  publisher = {arXiv},
  
  year = {2020},
  
  copyright = {arXiv.org perpetual, non-exclusive license}
}

@conference{KilParSch18,
  title = {Generalization in anti-causal learning},
  author = {Kilbertus*, N. and Parascandolo*, G. and Sch{\"o}lkopf*, B.},
  booktitle = {NeurIPS 2018 Workshop on Critiquing and Correcting Trends in Machine Learning},
  month = dec,
  year = {2018},
  note = {*authors are listed in alphabetical order},
  doi = {},
  url = {https://ml-critique-correct.github.io/},
  month_numeric = {12}
}

@InProceedings{learning_icm,
  title = 	 {Learning Independent Causal Mechanisms},
  author =       {Parascandolo, Giambattista and Kilbertus, Niki and Rojas-Carulla, Mateo and Sch{\"o}lkopf, Bernhard},
  booktitle = 	 {Proceedings of the 35th International Conference on Machine Learning},
  pages = 	 {4036--4044},
  year = 	 {2018},
  editor = 	 {Dy, Jennifer and Krause, Andreas},
  volume = 	 {80},
  series = 	 {Proceedings of Machine Learning Research},
  month = 	 {10--15 Jul},
  publisher =    {PMLR},
  url = 	 {https://proceedings.mlr.press/v80/parascandolo18a.html}
}

@article{mnih2015human,
  title={Human-level control through deep reinforcement learning},
  author={Mnih, Volodymyr and Kavukcuoglu, Koray and Silver, David and Rusu, Andrei A and Veness, Joel and Bellemare, Marc G and Graves, Alex and Riedmiller, Martin and Fidjeland, Andreas K and Ostrovski, Georg and others},
  journal={nature},
  volume={518},
  number={7540},
  pages={529--533},
  year={2015},
  publisher={Nature Publishing Group}
}

@misc{
madan2021fast,
title={Fast And Slow Learning Of Recurrent Independent Mechanisms},
author={Kanika Madan and Nan Rosemary Ke and Anirudh Goyal and Bernhard Sch{\"o}lkopf and Yoshua Bengio},
booktitle={International Conference on Learning Representations},
year={2021},
url={https://openreview.net/forum?id=Lc28QAB4ypz}
}

@misc{kansky2017schema,
      title={Schema Networks: Zero-shot Transfer with a Generative Causal Model of Intuitive Physics}, 
      author={Ken Kansky and Tom Silver and David A. Mély and Mohamed Eldawy and Miguel Lázaro-Gredilla and Xinghua Lou and Nimrod Dorfman and Szymon Sidor and Scott Phoenix and Dileep George},
      year={2017},
      eprint={1706.04317},
      archivePrefix={arXiv},
      primaryClass={cs.AI}
}

@misc{relicv2,
  doi = {10.48550/ARXIV.2201.05119},
  
  url = {https://arxiv.org/abs/2201.05119},
  
  author = {Tomasev, Nenad and Bica, Ioana and McWilliams, Brian and Buesing, Lars and Pascanu, Razvan and Blundell, Charles and Mitrovic, Jovana},
  
  title = {Pushing the limits of self-supervised ResNets: Can we outperform supervised learning without labels on ImageNet?},
  
  publisher = {arXiv},
  
  year = {2022},
  
  copyright = {Creative Commons Attribution 4.0 International}
}

@inproceedings{evaluating_model_robustness,
  author    = {Adarsh Subbaswamy and
               Roy Adams and
               Suchi Saria},
  editor    = {Arindam Banerjee and
               Kenji Fukumizu},
  title     = {Evaluating Model Robustness and Stability to Dataset Shift},
  booktitle = {The 24th International Conference on Artificial Intelligence and Statistics,
               {AISTATS} 2021, April 13-15, 2021, Virtual Event},
  series    = {Proceedings of Machine Learning Research},
  volume    = {130},
  pages     = {2611--2619},
  publisher = {{PMLR}},
  year      = {2021},
  url       = {http://proceedings.mlr.press/v130/subbaswamy21a.html},
  bibsource = {dblp computer science bibliography, https://dblp.org}
}

@inproceedings{fair_MAML,
  author    = {Dylan Slack and
               Sorelle A. Friedler and
               Emile Givental},
  editor    = {Mireille Hildebrandt and
               Carlos Castillo and
               L. Elisa Celis and
               Salvatore Ruggieri and
               Linnet Taylor and
               Gabriela Zanfir{-}Fortuna},
  title     = {Fairness warnings and fair-MAML: learning fairly with minimal data},
  booktitle = {FAT* '20: Conference on Fairness, Accountability, and Transparency,
               Barcelona, Spain, January 27-30, 2020},
  pages     = {200--209},
  publisher = {{ACM}},
  year      = {2020},
  url       = {https://doi.org/10.1145/3351095.3372839},
  doi       = {10.1145/3351095.3372839},
  bibsource = {dblp computer science bibliography, https://dblp.org}
}

@inproceedings{nan2021interventional,
  title={Interventional video grounding with dual contrastive learning},
  author={Nan, Guoshun and Qiao, Rui and Xiao, Yao and Liu, Jun and Leng, Sicong and Zhang, Hao and Lu, Wei},
  booktitle={Proceedings of the IEEE/CVF Conference on Computer Vision and Pattern Recognition},
  pages={2765--2775},
  year={2021}
}

@inproceedings{singh2021fairness,
  title={Fairness violations and mitigation under covariate shift},
  author={Singh, Harvineet and Singh, Rina and Mhasawade, Vishwali and Chunara, Rumi},
  booktitle={Proceedings of the 2021 ACM Conference on Fairness, Accountability, and Transparency},
  pages={3--13},
  year={2021}
}

@inproceedings{gradient_supervision,
  author    = {Damien Teney and
               Ehsan Abbasnejad and
               Anton van den Hengel},
  editor    = {Andrea Vedaldi and
               Horst Bischof and
               Thomas Brox and
               Jan{-}Michael Frahm},
  title     = {Learning What Makes a Difference from Counterfactual Examples and
               Gradient Supervision},
  booktitle = {Computer Vision - {ECCV} 2020 - 16th European Conference, Glasgow,
               UK, August 23-28, 2020, Proceedings, Part {X}},
  series    = {Lecture Notes in Computer Science},
  volume    = {12355},
  pages     = {580--599},
  publisher = {Springer},
  year      = {2020},
  url       = {https://doi.org/10.1007/978-3-030-58607-2\_34},
  doi       = {10.1007/978-3-030-58607-2\_34},
  bibsource = {dblp computer science bibliography, https://dblp.org}
}

@article{causal_influence,
  title={Quantifying causal influences},
  author={Janzing, Dominik and Balduzzi, David and Grosse-Wentrup, Moritz and Sch{\"o}lkopf, Bernhard},
  journal={The Annals of Statistics},
  volume={41},
  number={5},
  pages={2324--2358},
  year={2013},
  publisher={Institute of Mathematical Statistics}
}

@article{yudkowsky2016ai,
  title={The AI alignment problem: why it is hard, and where to start},
  author={Yudkowsky, Eliezer},
  journal={Symbolic Systems Distinguished Speaker},
  year={2016}
}

@misc{incentives_are_causal,
  doi = {10.48550/ARXIV.1910.10362},
  
  url = {https://arxiv.org/abs/1910.10362},
  
  author = {Miller, John and Milli, Smitha and Hardt, Moritz},
  
  title = {Strategic Classification is Causal Modeling in Disguise},
  
  publisher = {arXiv},
  
  year = {2019},
  
  copyright = {arXiv.org perpetual, non-exclusive license}
}

@inproceedings{causal_influence_diagram,
  author    = {Tom Everitt and
               Ryan Carey and
               Eric D. Langlois and
               Pedro A. Ortega and
               Shane Legg},
  title     = {Agent Incentives: {A} Causal Perspective},
  booktitle = {Thirty-Fifth {AAAI} Conference on Artificial Intelligence, {AAAI}
               2021, Thirty-Third Conference on Innovative Applications of Artificial
               Intelligence, {IAAI} 2021, The Eleventh Symposium on Educational Advances
               in Artificial Intelligence, {EAAI} 2021, Virtual Event, February 2-9,
               2021},
  pages     = {11487--11495},
  publisher = {{AAAI} Press},
  year      = {2021},
  url       = {https://ojs.aaai.org/index.php/AAAI/article/view/17368},
  bibsource = {dblp computer science bibliography, https://dblp.org}
}

@article{kleinberg2016inherent,
  title={Inherent trade-offs in the fair determination of risk scores},
  author={Kleinberg, Jon and Mullainathan, Sendhil and Raghavan, Manish},
  journal={arXiv preprint arXiv:1609.05807},
  year={2016}
}

@book{christian2021alignment,
  title={The alignment problem: How can machines learn human values?},
  author={Christian, Brian},
  year={2021},
  publisher={Atlantic Books}
}

@book{bostrom_superintelligence,
author = {Bostrom, Nick},
title = {Superintelligence: Paths, Dangers, Strategies},
year = {2014},
isbn = {0199678111},
publisher = {Oxford University Press, Inc.}
}

@inproceedings{everitt2018agi,
  author    = {Tom Everitt and
               Gary Lea and
               Marcus Hutter},
  editor    = {J{\'{e}}r{\^{o}}me Lang},
  title     = {{AGI} Safety Literature Review},
  booktitle = {Proceedings of the Twenty-Seventh International Joint Conference on
               Artificial Intelligence, {IJCAI} 2018, July 13-19, 2018, Stockholm,
               Sweden},
  pages     = {5441--5449},
  publisher = {ijcai.org},
  year      = {2018},
  url       = {https://doi.org/10.24963/ijcai.2018/768},
  doi       = {10.24963/ijcai.2018/768},
  bibsource = {dblp computer science bibliography, https://dblp.org}
}

@InProceedings{james_fox-proc-scipy-2021,
  author    = { {J}ames {F}ox and {T}om {E}veritt and {R}yan {C}arey and {E}ric {L}anglois and {A}lessandro {A}bate and {M}ichael {W}ooldridge },
  title     = { {P}y{C}{I}{D}: {A} {P}ython {L}ibrary for {C}ausal {I}nfluence {D}iagrams },
  booktitle = { {P}roceedings of the 20th {P}ython in {S}cience {C}onference },
  pages     = { 43 - 51 },
  year      = { 2021 },
  editor    = { {M}eghann {A}garwal and {C}hris {C}alloway and {D}illon {N}iederhut and {D}avid {S}hupe },
  doi       = {10.25080/majora-1b6fd038-008}
}

@inproceedings{corbett2017algorithmic,
  title={Algorithmic decision making and the cost of fairness},
  author={Corbett-Davies, Sam and Pierson, Emma and Feller, Avi and Goel, Sharad and Huq, Aziz},
  booktitle={Proceedings of the 23rd acm sigkdd international conference on knowledge discovery and data mining},
  pages={797--806},
  year={2017}
}

@article{chouldechova2017fair,
  title={Fair prediction with disparate impact: A study of bias in recidivism prediction instruments},
  author={Chouldechova, Alexandra},
  journal={Big data},
  volume={5},
  number={2},
  pages={153--163},
  year={2017},
  publisher={Mary Ann Liebert, Inc. 140 Huguenot Street, 3rd Floor New Rochelle, NY 10801 USA}
}

@article{beyond_impossibility,
  author    = {Limor Gultchin and
               Vincent Cohen{-}Addad and
               Sophie Giffard{-}Roisin and
               Varun Kanade and
               Frederik Mallmann{-}Trenn},
  title     = {Beyond Impossibility: Balancing Sufficiency, Separation and Accuracy},
  journal   = {CoRR},
  volume    = {abs/2205.12327},
  year      = {2022},
  url       = {https://doi.org/10.48550/arXiv.2205.12327},
  doi       = {10.48550/arXiv.2205.12327},
  eprinttype = {arXiv},
  eprint    = {2205.12327},
  bibsource = {dblp computer science bibliography, https://dblp.org}
}

@inproceedings{over-smoothing,
  author    = {Deli Chen and
               Yankai Lin and
               Wei Li and
               Peng Li and
               Jie Zhou and
               Xu Sun},
  title     = {Measuring and Relieving the Over-Smoothing Problem for Graph Neural
               Networks from the Topological View},
  booktitle = {The Thirty-Fourth {AAAI} Conference on Artificial Intelligence, {AAAI}
               2020, The Thirty-Second Innovative Applications of Artificial Intelligence
               Conference, {IAAI} 2020, The Tenth {AAAI} Symposium on Educational
               Advances in Artificial Intelligence, {EAAI} 2020, New York, NY, USA,
               February 7-12, 2020},
  pages     = {3438--3445},
  publisher = {{AAAI} Press},
  year      = {2020},
  url       = {https://ojs.aaai.org/index.php/AAAI/article/view/5747},
  bibsource = {dblp computer science bibliography, https://dblp.org}
}

@inproceedings{feng2021should,
  title={Should graph convolution trust neighbors? a simple causal inference method},
  author={Feng, Fuli and Huang, Weiran and He, Xiangnan and Xin, Xin and Wang, Qifan and Chua, Tat-Seng},
  booktitle={Proceedings of the 44th International ACM SIGIR Conference on Research and Development in Information Retrieval},
  pages={1208--1218},
  year={2021}
}

@inproceedings{yang2021deconfounded,
  title={Deconfounded video moment retrieval with causal intervention},
  author={Yang, Xun and Feng, Fuli and Ji, Wei and Wang, Meng and Chua, Tat-Seng},
  booktitle={Proceedings of the 44th International ACM SIGIR Conference on Research and Development in Information Retrieval},
  pages={1--10},
  year={2021}
}

@article{cf_exp_nl,
  author    = {Lisa Anne Hendricks and
               Ronghang Hu and
               Trevor Darrell and
               Zeynep Akata},
  title     = {Generating Counterfactual Explanations with Natural Language},
  journal   = {CoRR},
  volume    = {abs/1806.09809},
  year      = {2018},
  url       = {http://arxiv.org/abs/1806.09809},
  eprinttype = {arXiv},
  eprint    = {1806.09809},
  bibsource = {dblp computer science bibliography, https://dblp.org}
}

@InProceedings{cf_ve,
  title = 	 {Counterfactual Visual Explanations},
  author =       {Goyal, Yash and Wu, Ziyan and Ernst, Jan and Batra, Dhruv and Parikh, Devi and Lee, Stefan},
  booktitle = 	 {Proceedings of the 36th International Conference on Machine Learning},
  pages = 	 {2376--2384},
  year = 	 {2019},
  editor = 	 {Chaudhuri, Kamalika and Salakhutdinov, Ruslan},
  volume = 	 {97},
  series = 	 {Proceedings of Machine Learning Research},
  month = 	 {09--15 Jun},
  publisher =    {PMLR},
  url = 	 {https://proceedings.mlr.press/v97/goyal19a.html}
}

@inproceedings{interventional_fsl,
  author    = {Zhongqi Yue and
               Hanwang Zhang and
               Qianru Sun and
               Xian{-}Sheng Hua},
  editor    = {Hugo Larochelle and
               Marc'Aurelio Ranzato and
               Raia Hadsell and
               Maria{-}Florina Balcan and
               Hsuan{-}Tien Lin},
  title     = {Interventional Few-Shot Learning},
  booktitle = {Advances in Neural Information Processing Systems 33: Annual Conference
               on Neural Information Processing Systems 2020, NeurIPS 2020, December
               6-12, 2020, virtual},
  year      = {2020},
  url       = {https://proceedings.neurips.cc/paper/2020/hash/1cc8a8ea51cd0adddf5dab504a285915-Abstract.html},
  bibsource = {dblp computer science bibliography, https://dblp.org}
}

@article{Sui2021DeconfoundedTF,
  title={Deconfounded Training for Graph Neural Networks},
  author={Yongduo Sui and Xiang Wang and Jiancan Wu and Xiangnan He and Tat-Seng Chua},
  journal={ArXiv},
  year={2021},
  volume={abs/2112.15089}
}

@misc{scm_gnn,
      title={Relating Graph Neural Networks to Structural Causal Models}, 
      author={Matej Zečević and Devendra Singh Dhami and Petar Veličković and Kristian Kersting},
      year={2021},
      eprint={2109.04173},
      archivePrefix={arXiv},
      primaryClass={cs.LG}
}

@inproceedings{zhang2020invariant,
  title={Invariant causal prediction for block mdps},
  author={Zhang, Amy and Lyle, Clare and Sodhani, Shagun and Filos, Angelos and Kwiatkowska, Marta and Pineau, Joelle and Gal, Yarin and Precup, Doina},
  booktitle={International Conference on Machine Learning},
  pages={11214--11224},
  year={2020},
  organization={PMLR}
}

@misc{ortega2021shaking,
      title={Shaking the foundations: delusions in sequence models for interaction and control}, 
      author={Pedro A. Ortega and Markus Kunesch and Grégoire Delétang and Tim Genewein and Jordi Grau-Moya and Joel Veness and Jonas Buchli and Jonas Degrave and Bilal Piot and Julien Perolat and Tom Everitt and Corentin Tallec and Emilio Parisotto and Tom Erez and Yutian Chen and Scott Reed and Marcus Hutter and Nando de Freitas and Shane Legg},
      year={2021},
      eprint={2110.10819},
      archivePrefix={arXiv},
      primaryClass={cs.LG}
}

@inproceedings{abbasnejad_counterfactual_2020,
	address = {Seattle, WA, USA},
	title = {Counterfactual {Vision} and {Language} {Learning}},
	isbn = {978-1-72817-168-5},
	url = {https://ieeexplore.ieee.org/document/9156448/},
	doi = {10.1109/CVPR42600.2020.01006},
	language = {en},
	urldate = {2022-02-03},
	booktitle = {2020 {IEEE}/{CVF} {Conference} on {Computer} {Vision} and {Pattern} {Recognition} ({CVPR})},
	publisher = {IEEE},
	author = {Abbasnejad, Ehsan and Teney, Damien and Parvaneh, Amin and Shi, Javen and van den Hengel, Anton},
	month = jun,
	year = {2020}
}

@article{ipl,
  author    = {Sorawit Saengkyongam and
               Nikolaj Thams and
               Jonas Peters and
               Niklas Pfister},
  title     = {Invariant Policy Learning: {A} Causal Perspective},
  journal   = {CoRR},
  volume    = {abs/2106.00808},
  year      = {2021},
  url       = {https://arxiv.org/abs/2106.00808},
  eprinttype = {arXiv},
  eprint    = {2106.00808},
  bibsource = {dblp computer science bibliography, https://dblp.org}
}

@article{dulac2019challenges,
  title={Challenges of real-world reinforcement learning},
  author={Dulac-Arnold, Gabriel and Mankowitz, Daniel and Hester, Todd},
  journal={arXiv preprint arXiv:1904.12901},
  year={2019}
}

@book{EOCI,
  title = {Elements of Causal Inference - Foundations and Learning Algorithms},
  author = {Peters, J. and Janzing, D. and Sch{\"o}lkopf, B.},
  series = {Adaptive Computation and Machine Learning Series},
  publisher = {The MIT Press},
  address = {Cambridge, MA, USA},
  year = {2017},
  doi = {}
}

@inproceedings{backdoor_discovery,
  author    = {Limor Gultchin and
               Matt J. Kusner and
               Varun Kanade and
               Ricardo Silva},
  editor    = {Silvia Chiappa and
               Roberto Calandra},
  title     = {Differentiable Causal Backdoor Discovery},
  booktitle = {The 23rd International Conference on Artificial Intelligence and Statistics,
               {AISTATS} 2020, 26-28 August 2020, Online [Palermo, Sicily, Italy]},
  series    = {Proceedings of Machine Learning Research},
  volume    = {108},
  pages     = {3970--3979},
  publisher = {{PMLR}},
  year      = {2020},
  url       = {http://proceedings.mlr.press/v108/gultchin20a.html},
  bibsource = {dblp computer science bibliography, https://dblp.org}
}

@article{yue_counterfactual_2021,
	title = {Counterfactual {Zero}-{Shot} and {Open}-{Set} {Visual} {Recognition}},
	url = {http://arxiv.org/abs/2103.00887},
	language = {en},
	urldate = {2022-02-03},
	journal = {arXiv:2103.00887 [cs]},
	author = {Yue, Zhongqi and Wang, Tan and Zhang, Hanwang and Sun, Qianru and Hua, Xian-Sheng},
	month = mar,
	year = {2021}
}

@article{consistency_rule,
 ISSN = {10443983},
 URL = {http://www.jstor.org/stable/20788241},
 author = {Judea Pearl},
 journal = {Epidemiology},
 number = {6},
 pages = {872--875},
 publisher = {Lippincott Williams & Wilkins},
 title = {Brief Report: On the Consistency Rule in Causal Inference: "Axiom, Definition, Assumption, or Theorem?"},
 volume = {21},
 year = {2010}
}

@book{pearl2009causality,
  title={Causality},
  author={Pearl, Judea},
  year={2009},
  publisher={Cambridge university press}
}

@inproceedings{wang2021causal,
  title={Causal Attention for Unbiased Visual Recognition},
  author={Wang, Tan and Zhou, Chang and Sun, Qianru and Zhang, Hanwang},
  booktitle={Proceedings of the IEEE/CVF International Conference on Computer Vision (ICCV)},
  year={2021}
}

@inproceedings{DBLP:conf/fat/KarimiSV21,
  author    = {Amir{-}Hossein Karimi and
               Bernhard Sch{\"{o}}lkopf and
               Isabel Valera},
  editor    = {Madeleine Clare Elish and
               William Isaac and
               Richard S. Zemel},
  title     = {Algorithmic Recourse: from Counterfactual Explanations to Interventions},
  booktitle = {FAccT '21: 2021 {ACM} Conference on Fairness, Accountability, and
               Transparency, Virtual Event / Toronto, Canada, March 3-10, 2021},
  pages     = {353--362},
  publisher = {{ACM}},
  year      = {2021},
  url       = {https://doi.org/10.1145/3442188.3445899},
  doi       = {10.1145/3442188.3445899},
  bibsource = {dblp computer science bibliography, https://dblp.org}
}

@inproceedings{complex_causes,
  author    = {Limor Gultchin and
               David S. Watson and
               Matt J. Kusner and
               Ricardo Silva},
  editor    = {Marina Meila and
               Tong Zhang},
  title     = {Operationalizing Complex Causes: {A} Pragmatic View of Mediation},
  booktitle = {Proceedings of the 38th International Conference on Machine Learning,
               {ICML} 2021, 18-24 July 2021, Virtual Event},
  series    = {Proceedings of Machine Learning Research},
  volume    = {139},
  pages     = {3875--3885},
  publisher = {{PMLR}},
  year      = {2021},
  url       = {http://proceedings.mlr.press/v139/gultchin21a.html},
  bibsource = {dblp computer science bibliography, https://dblp.org}
}

@misc{equalizing_recourse,
  doi = {10.48550/ARXIV.1909.03166},
  
  url = {https://arxiv.org/abs/1909.03166},
  
  author = {Gupta, Vivek and Nokhiz, Pegah and Roy, Chitradeep Dutta and Venkatasubramanian, Suresh},
  
  title = {Equalizing Recourse across Groups},
  
  publisher = {arXiv},
  
  year = {2019},
  
  copyright = {Creative Commons Attribution 4.0 International}
}

@article{von2020fairness,
  title={On the fairness of causal algorithmic recourse},
  author={von K{\"u}gelgen, Julius and Karimi, Amir-Hossein and Bhatt, Umang and Valera, Isabel and Weller, Adrian and Sch{\"o}lkopf, Bernhard},
  journal={arXiv preprint arXiv:2010.06529},
  year={2020}
}

@inproceedings{DBLP:conf/nips/KarimiKSV20,
  author    = {Amir{-}Hossein Karimi and
               Bodo Julius von K{\"{u}}gelgen and
               Bernhard Sch{\"{o}}lkopf and
               Isabel Valera},
  editor    = {Hugo Larochelle and
               Marc'Aurelio Ranzato and
               Raia Hadsell and
               Maria{-}Florina Balcan and
               Hsuan{-}Tien Lin},
  title     = {Algorithmic recourse under imperfect causal knowledge: a probabilistic
               approach},
  booktitle = {Advances in Neural Information Processing Systems 33: Annual Conference
               on Neural Information Processing Systems 2020, NeurIPS 2020, December
               6-12, 2020, virtual},
  year      = {2020},
  url       = {https://proceedings.neurips.cc/paper/2020/hash/02a3c7fb3f489288ae6942498498db20-Abstract.html},
  bibsource = {dblp computer science bibliography, https://dblp.org}
}

@inproceedings{bender2021dangers,
  title={On the Dangers of Stochastic Parrots: Can Language Models Be Too Big?},
  author={Bender, Emily M and Gebru, Timnit and McMillan-Major, Angelina and Shmitchell, Shmargaret},
  booktitle={Proceedings of the 2021 ACM Conference on Fairness, Accountability, and Transparency},
  pages={610--623},
  year={2021}
}

@article{survey_ac,
  author    = {Amir{-}Hossein Karimi and
               Gilles Barthe and
               Bernhard Sch{\"{o}}lkopf and
               Isabel Valera},
  title     = {A survey of algorithmic recourse: definitions, formulations, solutions,
               and prospects},
  journal   = {CoRR},
  volume    = {abs/2010.04050},
  year      = {2020},
  url       = {https://arxiv.org/abs/2010.04050},
  eprinttype = {arXiv},
  eprint    = {2010.04050},
  bibsource = {dblp computer science bibliography, https://dblp.org}
}

@article{DBLP:journals/corr/abs-1907-09615,
  author    = {Shalmali Joshi and
               Oluwasanmi Koyejo and
               Warut Vijitbenjaronk and
               Been Kim and
               Joydeep Ghosh},
  title     = {Towards Realistic Individual Recourse and Actionable Explanations
               in Black-Box Decision Making Systems},
  journal   = {CoRR},
  volume    = {abs/1907.09615},
  year      = {2019},
  url       = {http://arxiv.org/abs/1907.09615},
  eprinttype = {arXiv},
  eprint    = {1907.09615},
  bibsource = {dblp computer science bibliography, https://dblp.org}
}

@article{counterfactual_explanations,
  author    = {Sandra Wachter and
               Brent D. Mittelstadt and
               Chris Russell},
  title     = {Counterfactual Explanations without Opening the Black Box: Automated
               Decisions and the {GDPR}},
  journal   = {CoRR},
  volume    = {abs/1711.00399},
  year      = {2017},
  url       = {http://arxiv.org/abs/1711.00399},
  eprinttype = {arXiv},
  eprint    = {1711.00399},
  bibsource = {dblp computer science bibliography, https://dblp.org}
}

@inproceedings{ribeiro2016should,
  title={" Why should i trust you?" Explaining the predictions of any classifier},
  author={Ribeiro, Marco Tulio and Singh, Sameer and Guestrin, Carlos},
  booktitle={Proceedings of the 22nd ACM SIGKDD international conference on knowledge discovery and data mining},
  pages={1135--1144},
  year={2016}
}

@article{baiocchi2014instrumental,
  title={Instrumental variable methods for causal inference},
  author={Baiocchi, Michael and Cheng, Jing and Small, Dylan S},
  journal={Statistics in medicine},
  volume={33},
  number={13},
  pages={2297--2340},
  year={2014},
  publisher={Wiley Online Library}
}

@article{mehrabi2021survey,
  title={A survey on bias and fairness in machine learning},
  author={Mehrabi, Ninareh and Morstatter, Fred and Saxena, Nripsuta and Lerman, Kristina and Galstyan, Aram},
  journal={ACM Computing Surveys (CSUR)},
  volume={54},
  number={6},
  pages={1--35},
  year={2021},
  publisher={ACM New York, NY, USA}
}

@article{cheng2022evaluation,
  title={Evaluation methods and measures for causal learning algorithms},
  author={Cheng, Lu and Guo, Ruocheng and Moraffah, Raha and Sheth, Paras and Candan, Kasim Selcuk and Liu, Huan},
  journal={IEEE Transactions on Artificial Intelligence},
  year={2022},
  publisher={IEEE}
}

@inproceedings{
invariant_cil,
title={Invariant Causal Imitation Learning for Generalizable Policies},
author={Ioana Bica and Daniel Jarrett and Mihaela van der Schaar},
booktitle={Advances in Neural Information Processing Systems},
editor={A. Beygelzimer and Y. Dauphin and P. Liang and J. Wortman Vaughan},
year={2021},
url={https://openreview.net/forum?id=715E7e6j4gU}
}

@inproceedings{pixel_cnn,
  author    = {A{\"{a}}ron van den Oord and
               Nal Kalchbrenner and
               Lasse Espeholt and
               Koray Kavukcuoglu and
               Oriol Vinyals and
               Alex Graves},
  editor    = {Daniel D. Lee and
               Masashi Sugiyama and
               Ulrike von Luxburg and
               Isabelle Guyon and
               Roman Garnett},
  title     = {Conditional Image Generation with PixelCNN Decoders},
  booktitle = {Advances in Neural Information Processing Systems 29: Annual Conference
               on Neural Information Processing Systems 2016, December 5-10, 2016,
               Barcelona, Spain},
  pages     = {4790--4798},
  year      = {2016},
  url       = {https://proceedings.neurips.cc/paper/2016/hash/b1301141feffabac455e1f90a7de2054-Abstract.html},
  bibsource = {dblp computer science bibliography, https://dblp.org}
}

@article{imai2010identification,
  title={Identification, inference and sensitivity analysis for causal mediation effects},
  author={Imai, Kosuke and Keele, Luke and Yamamoto, Teppei},
  journal={Statistical science},
  volume={25},
  number={1},
  pages={51--71},
  year={2010},
  publisher={Institute of Mathematical Statistics}
}

@article{lowe2020amortized,
  title={Amortized causal discovery: Learning to infer causal graphs from time-series data},
  author={L{\"o}we, Sindy and Madras, David and Zemel, Richard and Welling, Max},
  journal={arXiv preprint arXiv:2006.10833},
  year={2020}
}

@inproceedings{cinelli2019sensitivity,
  title={Sensitivity analysis of linear structural causal models},
  author={Cinelli, Carlos and Kumor, Daniel and Chen, Bryant and Pearl, Judea and Bareinboim, Elias},
  booktitle={International conference on machine learning},
  pages={1252--1261},
  year={2019},
  organization={PMLR}
}

@book{manski2003partial,
  title={Partial identification of probability distributions},
  author={Manski, Charles},
  volume={5},
  year={2003},
  publisher={Springer}
}

@inproceedings{
zhu2022causal,
title={Causal Inference with Treatment Measurement Error: A Nonparametric Instrumental Variable Approach},
author={Yuchen Zhu and Limor Gultchin and Arthur Gretton and Matt Kusner and Ricardo Silva},
booktitle={The 38th Conference on Uncertainty in Artificial Intelligence},
year={2022},
url={https://openreview.net/forum?id=SLcxbOUi9gq}
}

@article{miao2018identifying,
  title={Identifying causal effects with proxy variables of an unmeasured confounder},
  author={Miao, Wang and Geng, Zhi and Tchetgen Tchetgen, Eric J},
  journal={Biometrika},
  volume={105},
  number={4},
  pages={987--993},
  year={2018},
  publisher={Oxford University Press}
}

@inproceedings{actionable_recourse,
author = {Ustun, Berk and Spangher, Alexander and Liu, Yang},
title = {Actionable Recourse in Linear Classification},
year = {2019},
isbn = {9781450361255},
publisher = {Association for Computing Machinery},
address = {New York, NY, USA},
url = {https://doi.org/10.1145/3287560.3287566},
doi = {10.1145/3287560.3287566},
booktitle = {Proceedings of the Conference on Fairness, Accountability, and Transparency}
}

@inproceedings{local_explanations_necessity_sufficiency,
  author    = {David S. Watson and
               Limor Gultchin and
               Ankur Taly and
               Luciano Floridi},
  editor    = {Cassio P. de Campos and
               Marloes H. Maathuis and
               Erik Quaeghebeur},
  title     = {Local explanations via necessity and sufficiency: unifying theory
               and practice},
  booktitle = {Proceedings of the Thirty-Seventh Conference on Uncertainty in Artificial
               Intelligence, {UAI} 2021, Virtual Event, 27-30 July 2021},
  series    = {Proceedings of Machine Learning Research},
  volume    = {161},
  pages     = {1382--1392},
  publisher = {{AUAI} Press},
  year      = {2021},
  url       = {https://proceedings.mlr.press/v161/watson21a.html},
  bibsource = {dblp computer science bibliography, https://dblp.org}
}

@article{manski1990nonparametric,
  title={Nonparametric bounds on treatment effects},
  author={Manski, Charles F},
  journal={The American Economic Review},
  volume={80},
  number={2},
  pages={319--323},
  year={1990},
  publisher={JSTOR}
}

@inproceedings{nabi2018fair,
  title={Fair inference on outcomes},
  author={Nabi, Razieh and Shpitser, Ilya},
  booktitle={Proceedings of the AAAI Conference on Artificial Intelligence},
  volume={32},
  number={1},
  year={2018}
}

@article{wu2019pc,
  title={Pc-fairness: A unified framework for measuring causality-based fairness},
  author={Wu, Yongkai and Zhang, Lu and Wu, Xintao and Tong, Hanghang},
  journal={Advances in Neural Information Processing Systems},
  volume={32},
  year={2019}
}

@article{bannon2020causality,
  title={Causality and Batch Reinforcement Learning: Complementary Approaches To Planning In Unknown Domains},
  author={Bannon, James and Windsor, Brad and Song, Wenbo and Li, Tao},
  journal={arXiv preprint arXiv:2006.02579},
  year={2020}
}

@article{gabillon2012best,
  title={Best arm identification: A unified approach to fixed budget and fixed confidence},
  author={Gabillon, Victor and Ghavamzadeh, Mohammad and Lazaric, Alessandro},
  journal={Advances in Neural Information Processing Systems},
  volume={25},
  year={2012}
}

@inproceedings{introduced_unfairness,
  author    = {Carolyn Ashurst and
               Ryan Carey and
               Silvia Chiappa and
               Tom Everitt},
  title     = {Why Fair Labels Can Yield Unfair Predictions: Graphical Conditions
               for Introduced Unfairness},
  booktitle = {Thirty-Sixth {AAAI} Conference on Artificial Intelligence, {AAAI}
               2022, Thirty-Fourth Conference on Innovative Applications of Artificial
               Intelligence, {IAAI} 2022, The Twelveth Symposium on Educational Advances
               in Artificial Intelligence, {EAAI} 2022 Virtual Event, February 22
               - March 1, 2022},
  pages     = {9494--9503},
  publisher = {{AAAI} Press},
  year      = {2022},
  url       = {https://ojs.aaai.org/index.php/AAAI/article/view/21182},
  bibsource = {dblp computer science bibliography, https://dblp.org}
}

@inproceedings{bubeck2009pure,
  title={Pure exploration in multi-armed bandits problems},
  author={Bubeck, S{\'e}bastien and Munos, R{\'e}mi and Stoltz, Gilles},
  booktitle={International conference on Algorithmic learning theory},
  pages={23--37},
  year={2009},
  organization={Springer}
}

@inproceedings{wu2019counterfactual,
  title={Counterfactual fairness: Unidentification, bound and algorithm},
  author={Wu, Yongkai and Zhang, Lu and Wu, Xintao},
  booktitle={Proceedings of the Twenty-Eighth International Joint Conference on Artificial Intelligence},
  year={2019}
}

@inproceedings{zhang2018fairness,
  title={Fairness in decision-making—the causal explanation formula},
  author={Zhang, Junzhe and Bareinboim, Elias},
  booktitle={Proceedings of the AAAI Conference on Artificial Intelligence},
  volume={32},
  number={1},
  year={2018}
}

@article{race_and_social_equity,
author = {McGinnis Johnson, Jasmine},
year = {2015},
month = {03},
pages = {262-264},
title = {Race and Social Equity: A Nervous Area of Government},
volume = {34},
journal = {Equality, Diversity and Inclusion: An International Journal},
doi = {10.1108/EDI-12-2014-0084}
}

@article{guy2012social,
  title={Social equity: Its legacy, its promise},
  author={Guy, Mary E and McCandless, Sean A},
  journal={Public Administration Review},
  volume={72},
  number={s1},
  pages={S5--S13},
  year={2012},
  publisher={Wiley Online Library}
}

@Article{bellemare13arcade,
    author = {{Bellemare}, M.~G. and {Naddaf}, Y. and {Veness}, J. and {Bowling}, M.},
    title = {The Arcade Learning Environment: An Evaluation Platform for General Agents},
    journal = {Journal of Artificial Intelligence Research},
    year = "2013",
    month = "jun",
    volume = "47",
    pages = "253--279",
}

@article{zhu2019causal,
  title={Causal discovery with reinforcement learning},
  author={Zhu, Shengyu and Ng, Ignavier and Chen, Zhitang},
  journal={arXiv preprint arXiv:1906.04477},
  year={2019}
}

@inproceedings{todorov2012mujoco,
  title={MuJoCo: A physics engine for model-based control},
  author={Todorov, Emanuel and Erez, Tom and Tassa, Yuval},
  booktitle={2012 IEEE/RSJ International Conference on Intelligent Robots and Systems},
  pages={5026--5033},
  year={2012},
  organization={IEEE},
  doi={10.1109/IROS.2012.6386109}
}

@techreport{zhang2016markov,
  title={Markov decision processes with unobserved confounders: A causal approach},
  author={Zhang, Junzhe and Bareinboim, Elias},
  year={2016},
  institution={Technical report, Technical Report R-23, Purdue AI Lab}
}

@misc{ riw,
   author = "{Wikipedia contributors}",
   title = "Reinventing the wheel",
   year = "2022",
   url = "https://en.wikipedia.org/wiki/Reinventing_the_wheel",
   note = "[Online; accessed 10-May-2022]"
 }

@misc{ openai_shutting_down_robotics,
   author = "Katyanna Quach",
   title = "OpenAI shuts down robotics team because it doesn't have enough data yet",
   year = "2021",
   url = " https://www.theregister.com/2021/07/18/in_brief_ai/
",
   note = "[Online; accessed 30-May-2022]"
 }

@inproceedings{contrastive_explanations_for_model_interpretability,
  author    = {Alon Jacovi and
               Swabha Swayamdipta and
               Shauli Ravfogel and
               Yanai Elazar and
               Yejin Choi and
               Yoav Goldberg},
  editor    = {Marie{-}Francine Moens and
               Xuanjing Huang and
               Lucia Specia and
               Scott Wen{-}tau Yih},
  title     = {Contrastive Explanations for Model Interpretability},
  booktitle = {Proceedings of the 2021 Conference on Empirical Methods in Natural
               Language Processing, {EMNLP} 2021, Virtual Event / Punta Cana, Dominican
               Republic, 7-11 November, 2021},
  pages     = {1597--1611},
  publisher = {Association for Computational Linguistics},
  year      = {2021},
  url       = {https://doi.org/10.18653/v1/2021.emnlp-main.120},
  doi       = {10.18653/v1/2021.emnlp-main.120},
  bibsource = {dblp computer science bibliography, https://dblp.org}
}

@article{lipton1990contrastive,
  title={Contrastive explanation},
  author={Lipton, Peter},
  journal={Royal Institute of Philosophy Supplements},
  volume={27},
  pages={247--266},
  year={1990},
  publisher={Cambridge University Press}
}

@inproceedings{salimi2019interventional,
  title={Interventional fairness: Causal database repair for algorithmic fairness},
  author={Salimi, Babak and Rodriguez, Luke and Howe, Bill and Suciu, Dan},
  booktitle={Proceedings of the 2019 International Conference on Management of Data},
  pages={793--810},
  year={2019}
}

@article{DML,
    author = {Chernozhukov, Victor and Chetverikov, Denis and Demirer, Mert and Duflo, Esther and Hansen, Christian and Newey, Whitney and Robins, James},
    title = "{Double/debiased machine learning for treatment and structural parameters}",
    journal = {The Econometrics Journal},
    volume = {21},
    number = {1},
    pages = {C1-C68},
    year = {2018},
    month = {01},
    issn = {1368-4221},
}

@inproceedings{interpretable_cf_exps,
  author    = {Lisa Schut and
               Oscar Key and
               Rory McGrath and
               Luca Costabello and
               Bogdan Sacaleanu and
               Medb Corcoran and
               Yarin Gal},
  editor    = {Arindam Banerjee and
               Kenji Fukumizu},
  title     = {Generating Interpretable Counterfactual Explanations By Implicit Minimisation
               of Epistemic and Aleatoric Uncertainties},
  booktitle = {The 24th International Conference on Artificial Intelligence and Statistics,
               {AISTATS} 2021, April 13-15, 2021, Virtual Event},
  series    = {Proceedings of Machine Learning Research},
  volume    = {130},
  pages     = {1756--1764},
  publisher = {{PMLR}},
  year      = {2021},
  url       = {http://proceedings.mlr.press/v130/schut21a.html},
  bibsource = {dblp computer science bibliography, https://dblp.org}
}

@book{barocas-hardt-narayanan,
  title = {Fairness and Machine Learning},
  author = {Solon Barocas and Moritz Hardt and Arvind Narayanan},
  publisher = {fairmlbook.org},
  note = {\url{http://www.fairmlbook.org}},
  year = {2019}
}

@inproceedings{
relic,
title={Representation Learning via Invariant Causal Mechanisms},
author={Jovana Mitrovic and Brian McWilliams and Jacob C Walker and Lars Holger Buesing and Charles Blundell},
booktitle={International Conference on Learning Representations},
year={2021},

}

@inproceedings{dragonnet,
 author = {Shi, Claudia and Blei, David and Veitch, Victor},
 booktitle = {Advances in Neural Information Processing Systems},
 editor = {H. Wallach and H. Larochelle and A. Beygelzimer and F. d\textquotesingle Alch\'{e}-Buc and E. Fox and R. Garnett},
 pages = {},
 publisher = {Curran Associates, Inc.},
 title = {Adapting Neural Networks for the Estimation of Treatment Effects},
 volume = {32},
 year = {2019}
}

@inproceedings{causal_vae,
  author    = {Mengyue Yang and
               Furui Liu and
               Zhitang Chen and
               Xinwei Shen and
               Jianye Hao and
               Jun Wang},
  title     = {CausalVAE: Disentangled Representation Learning via Neural Structural
               Causal Models},
  booktitle = {{IEEE} Conference on Computer Vision and Pattern Recognition, {CVPR}
               2021, virtual, June 19-25, 2021},
  pages     = {9593--9602},
  publisher = {Computer Vision Foundation / {IEEE}},
  year      = {2021},
  url       = {https://openaccess.thecvf.com/content/CVPR2021/html/Yang\_CausalVAE\_Disentangled\_Representation\_Learning\_via\_Neural\_Structural\_Causal\_Models\_CVPR\_2021\_paper.html},
  bibsource = {dblp computer science bibliography, https://dblp.org}
}

@inproceedings{
cundy2021bcd,
title={{BCD} Nets: Scalable Variational Approaches for Bayesian Causal Discovery},
author={Chris Cundy and Aditya Grover and Stefano Ermon},
booktitle={Advances in Neural Information Processing Systems},
editor={A. Beygelzimer and Y. Dauphin and P. Liang and J. Wortman Vaughan},
year={2021},

}

@inproceedings{npvar,
  author    = {Ming Gao and
               Yi Ding and
               Bryon Aragam},
  title     = {A polynomial-time algorithm for learning nonparametric causal graphs},
  booktitle = {Advances in Neural Information Processing Systems},
  year      = {2020},
}

@article{friedman2003being,
  title={Being Bayesian about network structure. A Bayesian approach to structure discovery in Bayesian networks},
  author={Friedman, Nir and Koller, Daphne},
  journal={Machine learning},
  volume={50},
  year={2003},
}

@article{cubuk2018autoaugment,
  title={Autoaugment: Learning augmentation policies from data},
  author={Cubuk, Ekin D and Zoph, Barret and Mane, Dandelion and Vasudevan, Vijay and Le, Quoc V},
  journal={arXiv preprint arXiv:1805.09501},
  year={2018}
}

@software{haiku2020github,
  author = {Tom Hennigan and Trevor Cai and Tamara Norman and Igor Babuschkin},
  title = {{H}aiku: {S}onnet for {JAX}},
  url = {http://github.com/deepmind/dm-haiku},
  version = {0.0.3},
  year = {2020},
}

@software{flax2020github,
  author = {Jonathan Heek and Anselm Levskaya and Avital Oliver and Marvin Ritter and Bertrand Rondepierre and Andreas Steiner and Marc van {Z}ee},
  title = {{F}lax: A neural network library and ecosystem for {JAX}},
  url = {http://github.com/google/flax},
  version = {0.5.2},
  year = {2020},
}

@inproceedings{abadi2016tensorflow,
  title={{TensorFlow}: a system for {Large-Scale} machine learning},
  author={Abadi, Mart{\'\i}n and Barham, Paul and Chen, Jianmin and Chen, Zhifeng and Davis, Andy and Dean, Jeffrey and Devin, Matthieu and Ghemawat, Sanjay and Irving, Geoffrey and Isard, Michael and others},
  booktitle={12th USENIX symposium on operating systems design and implementation (OSDI 16)},
  pages={265--283},
  year={2016}
}

@software{jax2018github,
  author = {James Bradbury and Roy Frostig and Peter Hawkins and Matthew James Johnson and Chris Leary and Dougal Maclaurin and George Necula and Adam Paszke and Jake Vander{P}las and Skye Wanderman-{M}ilne and Qiao Zhang},
  title = {{JAX}: composable transformations of {P}ython+{N}um{P}y programs},
  url = {http://github.com/google/jax},
  version = {0.3.13},
  year = {2018},
}

@article{paszke2017automatic,
  title={Automatic differentiation in pytorch},
  author={Paszke, Adam and Gross, Sam and Chintala, Soumith and Chanan, Gregory and Yang, Edward and DeVito, Zachary and Lin, Zeming and Desmaison, Alban and Antiga, Luca and Lerer, Adam},
  year={2017}
}

@inproceedings{
zantedeschi2022dag,
title={{DAG} Learning on the Permutahedron},
author={Valentina Zantedeschi and Jean Kaddour and Luca Franceschi and Matt Kusner and Vlad Niculae},
booktitle={ICLR2022 Workshop on the Elements of Reasoning: Objects, Structure and Causality},
year={2022},

}

@article{pawlowski2020deep,
  title={Deep Structural Causal Models for Tractable Counterfactual Inference},
  author={Pawlowski, Nick and Coelho de Castro, Daniel and Glocker, Ben},
  journal={Advances in Neural Information Processing Systems},
  volume={33},
  year={2020}
}

@misc{harm,
  doi = {10.48550/ARXIV.2204.12993},
  
  url = {https://arxiv.org/abs/2204.12993},
  
  author = {Richens, Jonathan G. and Beard, Rory and Thompson, Daniel H.},
  
  title = {First do no harm: counterfactual objective functions for safe \& ethical AI},
  
  publisher = {arXiv},
  
  year = {2022},
  
  copyright = {Creative Commons Attribution 4.0 International}
}

@article{komorowski2018artificial,
  title={The artificial intelligence clinician learns optimal treatment strategies for sepsis in intensive care},
  author={Komorowski, Matthieu and Celi, Leo A and Badawi, Omar and Gordon, Anthony C and Faisal, A Aldo},
  journal={Nature medicine},
  volume={24},
  number={11},
  pages={1716--1720},
  year={2018},
  publisher={Nature Publishing Group}
}

@article{bellemare2020autonomous,
  title={Autonomous navigation of stratospheric balloons using reinforcement learning},
  author={Bellemare, Marc G and Candido, Salvatore and Castro, Pablo Samuel and Gong, Jun and Machado, Marlos C and Moitra, Subhodeep and Ponda, Sameera S and Wang, Ziyu},
  journal={Nature},
  volume={588},
  number={7836},
  pages={77--82},
  year={2020},
  publisher={Nature Publishing Group}
}

@article{alphago_2,
  author    = {David Silver and
               Julian Schrittwieser and
               Karen Simonyan and
               Ioannis Antonoglou and
               Aja Huang and
               Arthur Guez and
               Thomas Hubert and
               Lucas Baker and
               Matthew Lai and
               Adrian Bolton and
               Yutian Chen and
               Timothy P. Lillicrap and
               Fan Hui and
               Laurent Sifre and
               George van den Driessche and
               Thore Graepel and
               Demis Hassabis},
  title     = {Mastering the game of Go without human knowledge},
  journal   = {Nat.},
  volume    = {550},
  number    = {7676},
  pages     = {354--359},
  year      = {2017},
  url       = {https://doi.org/10.1038/nature24270},
  doi       = {10.1038/nature24270},
  bibsource = {dblp computer science bibliography, https://dblp.org}
}

@article{atari,
  author    = {Volodymyr Mnih and
               Koray Kavukcuoglu and
               David Silver and
               Alex Graves and
               Ioannis Antonoglou and
               Daan Wierstra and
               Martin A. Riedmiller},
  title     = {Playing Atari with Deep Reinforcement Learning},
  journal   = {CoRR},
  volume    = {abs/1312.5602},
  year      = {2013},
  url       = {http://arxiv.org/abs/1312.5602},
  eprinttype = {arXiv},
  eprint    = {1312.5602},
  bibsource = {dblp computer science bibliography, https://dblp.org}
}

@article{alphago,
  author    = {David Silver and
               Aja Huang and
               Chris J. Maddison and
               Arthur Guez and
               Laurent Sifre and
               George van den Driessche and
               Julian Schrittwieser and
               Ioannis Antonoglou and
               Vedavyas Panneershelvam and
               Marc Lanctot and
               Sander Dieleman and
               Dominik Grewe and
               John Nham and
               Nal Kalchbrenner and
               Ilya Sutskever and
               Timothy P. Lillicrap and
               Madeleine Leach and
               Koray Kavukcuoglu and
               Thore Graepel and
               Demis Hassabis},
  title     = {Mastering the game of Go with deep neural networks and tree search},
  journal   = {Nat.},
  volume    = {529},
  number    = {7587},
  pages     = {484--489},
  year      = {2016},
  url       = {https://doi.org/10.1038/nature16961},
  doi       = {10.1038/nature16961},
  bibsource = {dblp computer science bibliography, https://dblp.org}
}

@article{rubin1974estimating,
  title={Estimating causal effects of treatments in randomized and nonrandomized studies.},
  author={Rubin, Donald B},
  journal={Journal of educational Psychology},
  volume={66},
  number={5},
  pages={688},
  year={1974},
  publisher={American Psychological Association}
}

@inproceedings{shalit2017estimating,
  title={Estimating individual treatment effect: generalization bounds and algorithms},
  author={Shalit, Uri and Johansson, Fredrik D and Sontag, David},
  booktitle={International Conference on Machine Learning},
  pages={3076--3085},
  year={2017},
  organization={PMLR}
}

@inproceedings{curth2021nonparametric,
  author    = {Alicia Curth and
               Mihaela van der Schaar},
  editor    = {Arindam Banerjee and
               Kenji Fukumizu},
  title     = {Nonparametric Estimation of Heterogeneous Treatment Effects: From
               Theory to Learning Algorithms},
  booktitle = {The 24th International Conference on Artificial Intelligence and Statistics,
               {AISTATS} 2021, April 13-15, 2021, Virtual Event},
  series    = {Proceedings of Machine Learning Research},
  volume    = {130},
  pages     = {1810--1818},
  publisher = {{PMLR}},
  year      = {2021}
}

@article{r-learner,
    author = {Nie, X and Wager, S},
    title = "{Quasi-oracle estimation of heterogeneous treatment effects}",
    journal = {Biometrika},
    year = {2020},
    month = {09},
    issn = {0006-3444},
    doi = {10.1093/biomet/asaa076},
}

@article{kennedy2020optimal,
  title={Optimal doubly robust estimation of heterogeneous causal effects},
  author={Kennedy, Edward H},
  journal={arXiv preprint arXiv:2004.14497},
  year={2020}
}

@inproceedings{CEVAE,
 author = {Louizos, Christos and Shalit, Uri and Mooij, Joris M and Sontag, David and Zemel, Richard and Welling, Max},
 booktitle = {Advances in Neural Information Processing Systems},
 editor = {I. Guyon and U. V. Luxburg and S. Bengio and H. Wallach and R. Fergus and S. Vishwanathan and R. Garnett},
 pages = {},
 publisher = {Curran Associates, Inc.},
 title = {Causal Effect Inference with Deep Latent-Variable Models},
  volume = {30},
 year = {2017}
}

@inproceedings{
nie2021vcnet,
title={{\{}VCN{\}}et and Functional Targeted Regularization For Learning Causal Effects of Continuous Treatments},
author={Lizhen Nie and Mao Ye and qiang liu and Dan Nicolae},
booktitle={International Conference on Learning Representations},
year={2021},
}

@article{caron2020estimating,
      title={Estimating Individual Treatment Effects using Non-Parametric Regression Models: a Review}, 
      author={Alberto Caron and Ioanna Manolopoulou and Gianluca Baio},
      year={2020},
      eprint={2009.06472},
      archivePrefix={arXiv},
      primaryClass={stat.ME},
        journal={arXiv preprint arXiv:2009.06472},
}

@article {x-learner,
	author = {K{\"u}nzel, S{\"o}ren R. and Sekhon, Jasjeet S. and Bickel, Peter J. and Yu, Bin},
	title = {Metalearners for estimating heterogeneous treatment effects using machine learning},
	volume = {116},
	number = {10},
	pages = {4156--4165},
	year = {2019},
	doi = {10.1073/pnas.1804597116},
	publisher = {National Academy of Sciences},
	issn = {0027-8424},
	journal = {Proceedings of the National Academy of Sciences}
}

@inproceedings{SIN,
  author    = {Jean Kaddour and
               Yuchen Zhu and
               Qi Liu and
               Matt J. Kusner and
               Ricardo Silva},
  editor    = {Marc'Aurelio Ranzato and
               Alina Beygelzimer and
               Yann N. Dauphin and
               Percy Liang and
               Jennifer Wortman Vaughan},
  title     = {Causal Effect Inference for Structured Treatments},
  booktitle = {Advances in Neural Information Processing Systems 34: Annual Conference
               on Neural Information Processing Systems 2021, NeurIPS 2021, December
               6-14, 2021, virtual},
  pages     = {24841--24854},
  year      = {2021},
  url       = {https://proceedings.neurips.cc/paper/2021/hash/d02e9bdc27a894e882fa0c9055c99722-Abstract.html},
  bibsource = {dblp computer science bibliography, https://dblp.org}
}

@inproceedings{
enco,
title={Efficient Neural Causal Discovery without Acyclicity Constraints},
author={Phillip Lippe and Taco Cohen and Efstratios Gavves},
booktitle={International Conference on Learning Representations},
year={2022},
url={https://openreview.net/forum?id=eYciPrLuUhG}
}

@inproceedings{zheng2020learning,
  title={Learning sparse nonparametric dags},
  author={Zheng, Xun and Dan, Chen and Aragam, Bryon and Ravikumar, Pradeep and Xing, Eric},
  booktitle={International Conference on Artificial Intelligence and Statistics},
  pages={3414--3425},
  year={2020},
  organization={PMLR}
}

@inproceedings{watson2022rational,
  title={Rational Shapley Values},
  author={Watson, David},
  booktitle={2022 ACM Conference on Fairness, Accountability, and Transparency},
  pages={1083--1094},
  year={2022}
}

@article{goyal2020inductive,
  title={Inductive biases for deep learning of higher-level cognition},
  author={Goyal, Anirudh and Bengio, Yoshua},
  journal={arXiv preprint arXiv:2011.15091},
  year={2020}
}

@inproceedings{notears,
  author    = {Xun Zheng and
               Bryon Aragam and
               Pradeep Ravikumar and
               Eric P. Xing},
  title     = {DAGs with {NO} {TEARS:} Continuous Optimization for Structure Learning},
  booktitle = {Advances in Neural Information Processing Systems},
  year      = {2018},
}

@inproceedings{
steerability,
title={On the "steerability" of generative adversarial networks},
author={Ali Jahanian* and Lucy Chai* and Phillip Isola},
booktitle={International Conference on Learning Representations},
year={2020},
url={https://openreview.net/forum?id=HylsTT4FvB}
}

@article{lipton2018mythos,
  title={The Mythos of Model Interpretability: In machine learning, the concept of interpretability is both important and slippery.},
  author={Lipton, Zachary C},
  journal={Queue},
  volume={16},
  number={3},
  pages={31--57},
  year={2018},
  publisher={ACM New York, NY, USA}
}

@article{miller2019explanation,
  title={Explanation in artificial intelligence: Insights from the social sciences},
  author={Miller, Tim},
  journal={Artificial intelligence},
  volume={267},
  pages={1--38},
  year={2019},
  publisher={Elsevier}
}

@misc{cd_review,
  doi = {10.48550/ARXIV.2206.01152},
  
  url = {https://arxiv.org/abs/2206.01152},
  
  author = {Squires, Chandler and Uhler, Caroline},
  
  title = {Causal Structure Learning: a Combinatorial Perspective},
  
  publisher = {arXiv},
  
  year = {2022},
  
  copyright = {arXiv.org perpetual, non-exclusive license}
}

@book{l2010research,
  title={Research design explained},
  author={L Mitchell, Mark and M Jolley, Janina},
  year={2010}
}

@article{bias_amplification_1,
  title={Invited commentary: understanding bias amplification},
  author={Pearl, Judea},
  journal={American journal of epidemiology},
  volume={174},
  number={11},
  pages={1223--1227},
  year={2011},
  publisher={Oxford University Press}
}

@article{ockham_razor,
  title={Ockham's razor and Bayesian analysis},
  author={Jefferys, William H and Berger, James O},
  journal={American scientist},
  volume={80},
  number={1},
  pages={64--72},
  year={1992},
  publisher={JSTOR}
}

@book{lattimore2020bandit,
  title={Bandit algorithms},
  author={Lattimore, Tor and Szepesv{\'a}ri, Csaba},
  year={2020},
  publisher={Cambridge University Press}
}

@misc{recourse_cannot_be_robust,
  doi = {10.48550/ARXIV.2205.15834},
  
  url = {https://arxiv.org/abs/2205.15834},
  
  author = {Fokkema, Hidde and de Heide, Rianne and van Erven, Tim},
  
  title = {Attribution-based Explanations that Provide Recourse Cannot be Robust},
  
  publisher = {arXiv},
  
  year = {2022},
  
  copyright = {arXiv.org perpetual, non-exclusive license}
}

@article{aragam2015concave,
  title={Concave penalized estimation of sparse Gaussian Bayesian networks},
  author={Aragam, Bryon and Zhou, Qing},
  journal={The Journal of Machine Learning Research},
  volume={16},
  year={2015},
}

@book{molnar2022,
  title      = {Interpretable Machine Learning},
  author     = {Christoph Molnar},
  year       = {2022},
  subtitle   = {A Guide for Making Black Box Models Explainable},
  edition    = {2},
  url = {christophm.github.io/interpretable-ml-book/}
}

@article{li2020causal,
  title={Causal World Models by Unsupervised Deconfounding of Physical Dynamics},
  author={Li, Minne and Yang, Mengyue and Liu, Furui and Chen, Xu and Chen, Zhitang and Wang, Jun},
  journal={arXiv preprint arXiv:2012.14228},
  year={2020}
}

@article{ramseyGSG17,
  author    = {Joseph D. Ramsey and
               Madelyn Glymour and
               Ruben Sanchez{-}Romero and
               Clark Glymour},
  title     = {A million variables and more: the Fast Greedy Equivalence Search algorithm
               for learning high-dimensional graphical causal models, with an application
               to functional magnetic resonance images},
  journal   = {International Journal of Data Science and Analytics},
  volume    = {3},
  year      = {2017}
}

@inproceedings{NIPS2015_2b38c2df,
 author = {Scanagatta, Mauro and de Campos, Cassio P and Corani, Giorgio and Zaffalon, Marco},
 booktitle = {Advances in Neural Information Processing Systems},
 title = {Learning Bayesian Networks with Thousands of Variables},
 volume = {28},
 year = {2015}
}

@inproceedings{NIPS2013_8ce6790c,
 author = {Xiang, Jing and Kim, Seyoung},
 booktitle = {Advances in Neural Information Processing Systems},
 title = {{A}$^{\ast}$ Lasso for Learning a Sparse Bayesian Network Structure for Continuous Variables},
 volume = {26},
 year = {2013}
}

@inproceedings{Cussens11,
  author    = {James Cussens},
  title     = {Bayesian network learning with cutting planes},
  booktitle = {{Uncertainty in Artificial Intelligence}},
  year      = {2011}
}

@book{singh2005finding,
  title={Finding optimal Bayesian networks by dynamic programming},
  author={Singh, Ajit P and Moore, Andrew W},
  year={2005},
  publisher={Citeseer}
}

@inproceedings{yuCGY19,
  author    = {Yue Yu and
               Jie Chen and
               Tian Gao and
               Mo Yu},
  title     = {{DAG-GNN:} {DAG} Structure Learning with Graph Neural Networks},
  booktitle = {{ICML}},
  volume    = {97},
  year      = {2019}
}

@inproceedings{ngG020,
  author    = {Ignavier Ng and
               AmirEmad Ghassami and
               Kun Zhang},
  title     = {On the Role of Sparsity and {DAG} Constraints for Learning Linear
               DAGs},
  booktitle = {NeurIPS},
  year      = {2020}
}

@article{vowels2021d,
  title={D'ya like dags? a survey on structure learning and causal discovery},
  author={Vowels, Matthew J and Camgoz, Necati Cihan and Bowden, Richard},
  journal={arXiv preprint arXiv:2103.02582},
  year={2021}
}

@inproceedings{he0SXLJ21,
  author    = {Yue He and
               Peng Cui and
               Zheyan Shen and
               Renzhe Xu and
               Furui Liu and
               Yong Jiang},
  title     = {{DARING:} Differentiable Causal Discovery with Residual Independence},
  booktitle = {{KDD}},
  year      = {2021}
}

@inproceedings{brouillardLLLD20,
  author    = {Philippe Brouillard and
               S{\'{e}}bastien Lachapelle and
               Alexandre Lacoste and
               Simon Lacoste{-}Julien and
               Alexandre Drouin},
  title     = {Differentiable Causal Discovery from Interventional Data},
  booktitle = {NeurIPS},
  year      = {2020}
}

@misc{concept_counterfactuals,
  doi = {10.48550/ARXIV.2106.12723},
  
  url = {https://arxiv.org/abs/2106.12723},
  
  author = {Abid, Abubakar and Yuksekgonul, Mert and Zou, James},
  
  title = {Meaningfully Explaining Model Mistakes Using Conceptual Counterfactuals},
  
  publisher = {arXiv},
  
  year = {2021},
  
  copyright = {Creative Commons Attribution 4.0 International}
}

@inproceedings{
risks_of_irm,
title={The Risks of Invariant Risk Minimization},
author={Elan Rosenfeld and Pradeep Kumar Ravikumar and Andrej Risteski},
booktitle={International Conference on Learning Representations},
year={2021},
url={https://openreview.net/forum?id=BbNIbVPJ-42}
}

@inproceedings{caus_match,
  title = 	 {Domain Generalization using Causal Matching},
  author =       {Mahajan, Divyat and Tople, Shruti and Sharma, Amit},
  booktitle = 	 {Proceedings of the 38th International Conference on Machine Learning},
  year = 	 {2021}
}

@inproceedings{csg,
author = {Liu, Chang and Sun, Xinwei and Wang, Jindong and Li, Tao and Qin, Tao and Chen, Wei and Liu, Tie-Yan},
year = {2020},
month = {11},
title = {Learning Causal Semantic Representation for Out-of-Distribution Prediction}
}

@inproceedings{
LaCIM,
title={Recovering Latent Causal Factor for Generalization to Distributional Shifts},
author={Xinwei Sun and Botong Wu and Xiangyu Zheng and Chang Liu and Wei Chen and Tao Qin and Tie-Yan Liu},
booktitle={Advances in Neural Information Processing Systems},
editor={A. Beygelzimer and Y. Dauphin and P. Liang and J. Wortman Vaughan},
year={2021},
url={https://openreview.net/forum?id=go3GvM7aFD}
}

@article{ahuja2021invariance,
  title={Invariance principle meets information bottleneck for out-of-distribution generalization},
  author={Ahuja, Kartik and Caballero, Ethan and Zhang, Dinghuai and Gagnon-Audet, Jean-Christophe and Bengio, Yoshua and Mitliagkas, Ioannis and Rish, Irina},
  journal={Advances in Neural Information Processing Systems},
  volume={34},
  year={2021}
}

@article{chen2022invariance,
  title={Invariance Principle Meets Out-of-Distribution Generalization on Graphs},
  author={Chen, Yongqiang and Zhang, Yonggang and Yang, Han and Ma, Kaili and Xie, Binghui and Liu, Tongliang and Han, Bo and Cheng, James},
  journal={arXiv preprint arXiv:2202.05441},
  year={2022}
}

@inproceedings{cca_mfrl,
  author    = {Thomas Mesnard and
               Theophane Weber and
               Fabio Viola and
               Shantanu Thakoor and
               Alaa Saade and
               Anna Harutyunyan and
               Will Dabney and
               Thomas S. Stepleton and
               Nicolas Heess and
               Arthur Guez and
               Eric Moulines and
               Marcus Hutter and
               Lars Buesing and
               R{\'{e}}mi Munos},
  editor    = {Marina Meila and
               Tong Zhang},
  title     = {Counterfactual Credit Assignment in Model-Free Reinforcement Learning},
  booktitle = {Proceedings of the 38th International Conference on Machine Learning,
               {ICML} 2021, 18-24 July 2021, Virtual Event},
  series    = {Proceedings of Machine Learning Research},
  volume    = {139},
  pages     = {7654--7664},
  publisher = {{PMLR}},
  year      = {2021},
  url       = {http://proceedings.mlr.press/v139/mesnard21a.html},
  bibsource = {dblp computer science bibliography, https://dblp.org}
}

@inproceedings{social_influence,
  author    = {Natasha Jaques and
               Angeliki Lazaridou and
               Edward Hughes and
               {\c{C}}aglar G{\"{u}}l{\c{c}}ehre and
               Pedro A. Ortega and
               DJ Strouse and
               Joel Z. Leibo and
               Nando de Freitas},
  editor    = {Kamalika Chaudhuri and
               Ruslan Salakhutdinov},
  title     = {Social Influence as Intrinsic Motivation for Multi-Agent Deep Reinforcement
               Learning},
  booktitle = {Proceedings of the 36th International Conference on Machine Learning,
               {ICML} 2019, 9-15 June 2019, Long Beach, California, {USA}},
  series    = {Proceedings of Machine Learning Research},
  volume    = {97},
  pages     = {3040--3049},
  publisher = {{PMLR}},
  year      = {2019},
  url       = {http://proceedings.mlr.press/v97/jaques19a.html},
  bibsource = {dblp computer science bibliography, https://dblp.org}
}

@article{galhotra2021feature,
  title={Feature Attribution and Recourse via Probabilistic Contrastive Counterfactuals},
  author={Galhotra, Sainyam and Pradhan, Romila and Salimi, Babak},
  year={2021}
}

@article{albini2021counterfactual,
  title={Counterfactual Shapley Additive Explanations},
  author={Albini, Emanuele and Long, Jason and Dervovic, Danial and Magazzeni, Daniele},
  journal={arXiv preprint arXiv:2110.14270},
  year={2021}
}

@inproceedings{kommiya2021towards,
  title={Towards unifying feature attribution and counterfactual explanations: Different means to the same end},
  author={Kommiya Mothilal, Ramaravind and Mahajan, Divyat and Tan, Chenhao and Sharma, Amit},
  booktitle={Proceedings of the 2021 AAAI/ACM Conference on AI, Ethics, and Society},
  pages={652--663},
  year={2021}
}

@inproceedings{janzing_fairness,
  author    = {Dominik Janzing and
               Lenon Minorics and
               Patrick Bl{\"{o}}baum},
  editor    = {Silvia Chiappa and
               Roberto Calandra},
  title     = {Feature relevance quantification in explainable {AI:} {A} causal problem},
  booktitle = {The 23rd International Conference on Artificial Intelligence and Statistics,
               {AISTATS} 2020, 26-28 August 2020, Online [Palermo, Sicily, Italy]},
  series    = {Proceedings of Machine Learning Research},
  volume    = {108},
  pages     = {2907--2916},
  publisher = {{PMLR}},
  year      = {2020},
  url       = {http://proceedings.mlr.press/v108/janzing20a.html},
  bibsource = {dblp computer science bibliography, https://dblp.org}
}

@article{he2008active,
  title={Active learning of causal networks with intervention experiments and optimal designs},
  author={He, Yang-Bo and Geng, Zhi},
  journal={Journal of Machine Learning Research},
  volume={9},
  number={Nov},
  pages={2523--2547},
  year={2008}
}

@article{naidu2021differential,
  title={When differential privacy meets interpretability: A case study},
  author={Naidu, Rakshit and Priyanshu, Aman and Kumar, Aadith and Kotti, Sasikanth and Wang, Haofan and Mireshghallah, Fatemehsadat},
  journal={arXiv preprint arXiv:2106.13203},
  year={2021}
}

@article{aivodji2020model,
  title={Model extraction from counterfactual explanations},
  author={A{\"\i}vodji, Ulrich and Bolot, Alexandre and Gambs, S{\'e}bastien},
  journal={arXiv preprint arXiv:2009.01884},
  year={2020}
}

@inproceedings{shokri2021privacy,
  title={On the privacy risks of model explanations},
  author={Shokri, Reza and Strobel, Martin and Zick, Yair},
  booktitle={Proceedings of the 2021 AAAI/ACM Conference on AI, Ethics, and Society},
  pages={231--241},
  year={2021}
}

@article{slack2021counterfactual,
  title={Counterfactual explanations can be manipulated},
  author={Slack, Dylan and Hilgard, Anna and Lakkaraju, Himabindu and Singh, Sameer},
  journal={Advances in Neural Information Processing Systems},
  volume={34},
  year={2021}
}

@article{simonyan2013deep,
  title={Deep inside convolutional networks: Visualising image classification models and saliency maps},
  author={Simonyan, Karen and Vedaldi, Andrea and Zisserman, Andrew},
  journal={arXiv preprint arXiv:1312.6034},
  year={2013}
}

@article{adebayo2018sanity,
  title={Sanity checks for saliency maps},
  author={Adebayo, Julius and Gilmer, Justin and Muelly, Michael and Goodfellow, Ian and Hardt, Moritz and Kim, Been},
  journal={Advances in neural information processing systems},
  volume={31},
  year={2018}
}

@article{burkart2021survey,
  title={A survey on the explainability of supervised machine learning},
  author={Burkart, Nadia and Huber, Marco F},
  journal={Journal of Artificial Intelligence Research},
  volume={70},
  pages={245--317},
  year={2021}
}

@inproceedings{dovsilovic2018explainable,
  title={Explainable artificial intelligence: A survey},
  author={Do{\v{s}}ilovi{\'c}, Filip Karlo and Br{\v{c}}i{\'c}, Mario and Hlupi{\'c}, Nikica},
  booktitle={2018 41st International convention on information and communication technology, electronics and microelectronics (MIPRO)},
  pages={0210--0215},
  year={2018},
  organization={IEEE}
}

@article{artelt2019computation,
  title={On the computation of counterfactual explanations--A survey},
  author={Artelt, Andr{\'e} and Hammer, Barbara},
  journal={arXiv preprint arXiv:1911.07749},
  year={2019}
}

@article{chen2020causalml,
  title={Causalml: Python package for causal machine learning},
  author={Chen, Huigang and Harinen, Totte and Lee, Jeong-Yoon and Yung, Mike and Zhao, Zhenyu},
  journal={arXiv preprint arXiv:2002.11631},
  year={2020}
}

@article{sharma2020dowhy,
  title={DoWhy: An end-to-end library for causal inference},
  author={Sharma, Amit and Kiciman, Emre},
  journal={arXiv preprint arXiv:2011.04216},
  year={2020}
}

@misc{true_to_model_or_data,

  
  url = {https://arxiv.org/abs/2006.16234},
  
  author = {Chen, Hugh and Janizek, Joseph D. and Lundberg, Scott and Lee, Su-In},
  
  title = {True to the Model or True to the Data?},
  
  publisher = {arXiv},
  
  year = {2020},
  
  copyright = {arXiv.org perpetual, non-exclusive license}
}

@article{bach2021doubleml,
  title={DoubleML--An Object-Oriented Implementation of Double Machine Learning in R},
  author={Bach, Philipp and Chernozhukov, Victor and Kurz, Malte S and Spindler, Martin},
  journal={arXiv preprint arXiv:2103.09603},
  year={2021}
}

@article{mahajan2019preserving,
  title={Preserving causal constraints in counterfactual explanations for machine learning classifiers},
  author={Mahajan, Divyat and Tan, Chenhao and Sharma, Amit},
  journal={arXiv preprint arXiv:1912.03277},
  year={2019}
}

\end{document}